    \RequirePackage{fix-cm}
   \documentclass[smallcondensed]{svjour3}     
    \smartqed  
    \usepackage{graphicx}
    \usepackage[utf8]{inputenc} 
    \usepackage[T1]{fontenc}
    \usepackage[dvipsnames]{xcolor}
    \usepackage{multirow}
    \usepackage{array}
    \usepackage{caption}
    \usepackage{float}
    \usepackage{lipsum}
    \usepackage{pifont}
    \usepackage{array}
    \usepackage{cite}
    \usepackage{amsmath}
    \usepackage{amssymb}
    \usepackage{marvosym}
    \usepackage{enumerate}
    \usepackage{makecell}
    
    \usepackage{hyperref}
    \hypersetup{
    colorlinks=true,
    linkcolor=blue,
    filecolor=magenta,      
    urlcolor=cyan,
}

    \usepackage[final]{changes}

    \usepackage{tabu}
    \usepackage{array,booktabs}
    \newcolumntype{L}{@{}>{\kern\tabcolsep}l<{\kern\tabcolsep}}
    \usepackage{colortbl}
    \usepackage{multirow}

    \newcommand\crule[3][black]{\textcolor{#1}{\rule[3pt]{#2}{#3}}}
    \newcommand\mybox[2]{\fcolorbox{#1}{#2}{\crule[#2]{.4cm}{0.04cm}}}
    \newcommand\mysmallbox[2]{\fcolorbox{#1}{#2}{\crule[#2]{.1cm}{0.04cm}}}

    \newcommand{\LL}{\mathcal{L}}
    
    \newcommand{\tmf}{tinyMiniFrance}
    \newcommand{\MF}{MiniFrance}

    \definecolor{Gray}{gray}{0.85}
    \newcolumntype{a}{>{\columncolor{Gray}}c}

   \definecolor{Caen}{RGB}{81, 173, 81}
   \definecolor{Nice}{RGB}{138, 138, 255}
   \definecolor{Cherbourg}{RGB}{255.,  25.,  25.}
   \definecolor{Lille}{RGB}{255.,  50.,  50.}
   \definecolor{Clermont}{RGB}{255.,  77.,  77.}
   \definecolor{Vannes}{RGB}{255., 102., 102.}
   \definecolor{Quimper}{RGB}{255., 128., 128.}
   \definecolor{Angers}{RGB}{255., 153., 153.}
   \definecolor{Rennes}{RGB}{255., 179., 179.}
   \definecolor{Martigues}{RGB}{255, 204, 204}
   \definecolor{Saint-Brieuc}{RGB}{0, 128, 0}
   \definecolor{Dunkerque}{RGB}{41, 150, 41}
   \definecolor{Brest}{RGB}{162, 218, 162}
   \definecolor{Lorient}{RGB}{123, 195, 123}
   \definecolor{LeMans}{RGB}{203, 240, 203}
   \definecolor{Nantes}{RGB}{46,  46, 255}

    \definecolor{urbanfabric}{RGB}{219, 95, 87}
    \definecolor{industrial}{RGB}{219, 151, 87}
    \definecolor{mine}{RGB}{219, 208, 87}
    \definecolor{artificial}{RGB}{173,219,87}
    \definecolor{arable}{RGB}{117, 219, 87}
    \definecolor{permanent}{RGB}{123, 196., 123}
    \definecolor{pastures}{RGB}{88, 177,  88}
    \definecolor{complexcultiv}{RGB}{212, 246, 212}
    \definecolor{orchards}{RGB}{176, 226, 176}
    \definecolor{forests}{RGB}{0, 128, 0}
    \definecolor{herbacious}{RGB}{88, 176, 167}
    \definecolor{openspaces}{RGB}{153,  93,  19}
    \definecolor{wetlands}{RGB}{87, 155, 219}
    \definecolor{water}{RGB}{0, 98, 255}

    \definecolor{mostaza}{RGB}{234, 157, 15}
    \definecolor{burdeos}{RGB}{109, 7, 26}
    \definecolor{azulmarino}{RGB}{6, 66, 115}

    \definecolor{darkcolor}{RGB}{0, 0, 128}
    \definecolor{lightcolor}{RGB}{230, 230, 255}

    \definecolor{cmap063}{RGB}{64,144,197}
    \definecolor{cmap064}{RGB}{62,142,196}
    \definecolor{cmap087}{RGB}{9,82,157}
    \definecolor{cmap093}{RGB}{8,66,133}

    \definecolor{ggray}{gray}{0.65}

\begin{document}

   \title{Semi-Supervised Semantic Segmentation\\[2pt] in Earth Observation:}

   \subtitle{The MiniFrance suite, dataset analysis\\[2pt] and multi-task network study.}
    
   \titlerunning{Semi-Supervised Semantic Segmentation in Earth Observation}        
    
    \author{Javiera Castillo-Navarro \and
            Bertrand Le Saux \and 
            Alexandre Boulch \and \\[2pt]
            Nicolas Audebert \and
            Sébastien Lef\`{e}vre
    }
    
    
    \institute{J. Castillo Navarro \at
               ONERA, Université Paris-Saclay, F-91123 Palaiseau, France.\\
               Université Bretagne Sud, IRISA UMR 6074, F-56000 Vannes, France.\\
                  \Letter{\quad \href{mailto:javiera.castillo_navarro@onera.fr}{\nolinkurl{javiera.castillo_navarro@onera.fr}}}          
               \and
               B. Le Saux \at
             European Space Agency, ESRIN, I-00044 Frascati (Rome), Italy\\
            \Letter{\quad \href{mailto:bertrand.le.saux@esa.int}{\nolinkurl{bertrand.le.saux@esa.int}}}   
               \and
               A. Boulch \at
               valeo.ai, F-75008 Paris, France\\
               \Letter{\quad \href{mailto:alexandre.boulch@valeo.com}{\nolinkurl{alexandre.boulch@valeo.com}}}
               \and
               N. Audebert \at
               Conservatoire national des arts et métiers, CEDRIC EA 4629, F-75003, Paris, France.\\
               \Letter{\quad \href{mailto:nicolas.audebert@cnam.fr}{\nolinkurl{nicolas.audebert@cnam.fr}}}
               \and
               S. Lefèvre \at
               Université Bretagne Sud, IRISA UMR 6074, F-56000 Vannes, France.\\
               \Letter{\quad \href{mailto:sebastien.lefevre@irisa.fr}{\nolinkurl{sebastien.lefevre@irisa.fr}}}
    }
    
       \date{}

    \maketitle
    
    \begin{abstract}
    The development of semi-supervised learning techniques is essential to enhance the generalization capacities of machine learning algorithms. Indeed, raw image data are abundant while labels are scarce, therefore it is crucial to leverage unlabeled inputs to build better models. The availability of large databases have been key for the development of learning algorithms with high level performance. 

    Despite the major role of machine learning in Earth Observation to derive products such as land cover maps,
 datasets in the field are still limited, either because of modest surface coverage, lack of variety of scenes or restricted classes to identify. We introduce a novel large-scale dataset for semi-supervised semantic segmentation in Earth Observation, the \MF\ suite. MiniFrance has several unprecedented properties: it is large-scale, containing over 2000 very high resolution aerial images, accounting for more than 200 billions samples (pixels); it is varied, covering 16 conurbations in France, with various climates, different landscapes, and urban as well as countryside scenes; and it is challenging, considering land use classes with high-level semantics. Nevertheless, the most distinctive quality of MiniFrance is being the only dataset in the field especially designed for semi-supervised learning: it contains labeled and unlabeled images in its training partition, which reproduces a life-like scenario. Along with this dataset, we present tools for data representativeness analysis in terms of  appearance similarity and a thorough study of MiniFrance data, demonstrating that it is suitable for learning and generalizes well in a semi-supervised setting. Finally, we present semi-supervised deep architectures based on multi-task learning and the first experiments on MiniFrance. These results will serve as baselines for future work on semi-supervised learning over the MiniFrance dataset. The Minifrance suite and related semi-supervised networks will be publicly available to promote semi-supervised works in Earth Observation.


    \end{abstract}
    
    \section{Introduction}
    \label{intro}

Earth Observation (EO) data analysis plays a major role on the way we understand our planet and its dynamics. Indeed, the ever-growing amount of remote sensing imagery data in the last decades has allowed new developments in the fields of ecology, urban planning or natural disaster response~\cite{runting2020opportunities}, and will certainly be crucial on the battle against climate change.

In recent years, deep learning techniques -- and the significant growth of computing power jointly with massive amounts of (labeled) data available -- have transformed the fields of machine learning and computer vision. Moreover, remote sensing imagery has not been the exception since several state-of-the-art methods for classification, object detection and image segmentation have proved to be most effective in this kind of data too \cite{audebert-18isprsj-beyond-RGB, maggiori_can_2017, zhu2017deep}.

 Unfortunately, most of the machine learning algorithms -- and particularly, deep learning methods -- developed to date rely heavily on the availability of annotated image databases. Labeled data is hard to obtain, requiring too much effort and time, while raw data -- without labels -- is abundant, especially in remote sensing where satellites generate data continuously (e.g., Copernicus Sentinels provide up to 5 day coverage of the Earth). Because of this, we are convinced that semi-supervised methods -- which leverage unlabeled data to help on the learning process -- will be essential to push further the generalization capacities of the models.

 To this end, we propose the first large-scale dataset for semi-supervised semantic segmentation in the field: the MiniFrance dataset. It will encourage research on semi-supervised methods and will provide a common and reliable benchmark to new algorithms, just as ImageNet~\cite{deng2009imagenet} did on traditional computer vision a decade ago. Along with the \MF\ suite, we conduct a thorough analysis of data in terms of representativeness to define a convenient partition for semi-supervision and we present semi-supervised methods for semantic segmentation, based on multi-task learning, that show the effectiveness of semi-supervised learning and will serve as baselines for future work on this dataset. For this reason, the \MF\ suite and related semi-supervised networks will be made publicly available.

 Thus, our contributions are three-fold:
 \vspace{-\topsep}
 \begin{itemize}
   \item We introduce \MF\ a new large scale dataset for semi-supervised semantic segmentation in Earth Observation\footnote{Preliminary work on this dataset have been published in~\cite{castillo2019what-data}, where the limitations of existing EO datasets are shown and one can understand the interest of varied and rich datasets as \MF.}.
   \item We define techniques for prior analysis of the representativeness of datasets for training and deploying models which help evaluate the need for domain adaptation.
  \item We show the benefits of semi-supervised learning strategies to improve semantic segmentation:
  \begin{itemize}
      \item In particular, we propose a new loss function for unsupervised or semi-supervised image segmentation;
      \item we report an extensive study of semi-supervised learning with different losses and multi-task architectures.
   \end{itemize}
\end{itemize}

On account of this, we start by exploring some related work in Section~\ref{sec:related-work}.
Section~\ref{sec: MF-dataset-description} describes the \MF\ suite in details, while Section~\ref{sec: statistical-analysis} introduces some tools to analyze data representativeness and appearance similarity in multi-location datasets. This allows us to get meaningful insight about the \MF\ dataset and to define a suitable partition  -- labeled training, unlabeled training and testing -- to perform semi-supervised learning.
We introduce our semi-supervised strategies in Section~\ref{sec: deep-semi-supervised-learning}, including neural network architectures and unsupervised losses to consider in a multi-task learning scheme. We then present in Section~\ref{sec: experiments-analysis} the analysis and experimental study of semi-supervised learning over the \MF\ dataset. They provide deeper understanding about semi-supervised learning and show the interest of the development of these techniques, that use unlabeled data to enhance the learning process, improving the generalization capacities of the models. These results will also serve as baselines for future work on semi-supervised learning over the \MF\ dataset.

\section{Related work}\label{sec:related-work}

Since we aim to perform semantic segmentation on remote sensing data using deep semi-supervised neural networks, we discuss here the related work on semantic segmentation, EO datasets, and semi-supervised learning. 

\subsection{Semantic Segmentation}\label{sec: related-work-semantic-segmentation}

Semantic segmentation consists in the process of assigning a class label to every pixel on an image. It is a relevant task in computer vision because it implies understanding the context of a scene or an image which might be crucial for some applications, like autonomous driving or medical image diagnostics.

If in the last decade Convolutional Neural Networks (CNNs) became the state-of-the-art to perform image classification and object detection, the breakthrough of Fully Convolutional Networks~\cite{long2015fully}  (FCNs) revolutionized the way of obtaining dense pixel-wise predictions. This kind of architectures takes advantage of CNNs replacing the last fully connected layers by convolutional ones, obtaining dense prediction maps. Today, state-of-the-art semantic segmentation networks, from SegNet~\cite{badrinarayanan2015segnet} and U-Net~\cite{ronneberger2018unet} to PSPNet~\cite{zhao2017pspnet} or DeepLab~\cite{chen2017deeplab}, all inherit from the FCN paradigm. A comprehensive review can be found in~\cite{minaee2020image-segm-survey}.

Processing of EO data has also greatly benefited from these techniques which now define the state-of-the-art in the field. Semantic segmentation is one of the main tasks in remote sensing since it provide pixel-wise classification that corresponds to land cover or land use maps (i.e. the most popular EO products). After seminal works for road detection~\cite{mnih_hinton_eccv2010}, generic multi-class segmentation was soon tackled with CNNs and FCNs~\cite{Paisitkriangkrai_2015_CVPR_Workshops, iadf-16jstars-dfc15, audebert-18isprsj-beyond-RGB, rey2017detecting}, until latest developments which result in global cover maps of a continent or the entire planet~\cite{demuzere-bechtel-PONE2019-europe-global-LCZ}. With respect to these approaches, our work aims at leveraging also unlabeled data for estimating the classification model.

\subsection{Datasets for Earth Observation}\label{sec:EO-datasets}

The tremendous progress of computer vision -- where machine learning is applied on images -- in the last decades would not have been possible without the development of large public datasets, such as ImageNet~\cite{deng2009imagenet}, COCO~\cite{lin2014microsoft} or Cityscapes~\cite{cordts2016cityscapes} for learning on visual data. These datasets provide the means to compare models, and to test their scalability and reliability. They are the key to improve performance of algorithms and push research limits further.

In view of the above, the remote sensing community has also published several datasets for different tasks in order to encourage the research in the field. Table~\ref{tab:EO-datasets} describes the main initiatives. 

\begin{table}[h!]
     \begin{center}
        \caption{Earth Observation datasets summary.\\ {\small Abbreviations: SS = Semantic Segmentation; LC = Land Cover; Road = Road Extraction; Build. = Building Extraction; OD = Object Detection; CD = Change Detection; IC = Image Classification; Urb. = Urban; Ctry = Countryside.} \label{tab:EO-datasets}}
       \vspace{-.5em}
        \setlength{\tabcolsep}{4pt}
        \begin{tabular}{c>{\centering\arraybackslash}m{1.8cm}>{\centering\arraybackslash}m{1.5cm}>{\centering\arraybackslash}m{1cm}>{\centering\arraybackslash}m{.8cm}>{\centering\arraybackslash}m{1.2cm}>{\centering\arraybackslash}m{1cm}}
         \toprule
           \emph{Dataset} & \emph{Task} & \emph{Location} & \emph{Zone type}&\emph{Surface (km$^2$)} & \emph{Resolution (cm/px)} & \emph{Number classes} \\\midrule
           Vaihingen~\cite{Haala-Cramer-Jacobsen-DGPF-ISPRS, object-detection-JPRS-2014} & \makecell{Semantic \\segmentation} & \makecell{Vaihingen\\(Germany)} & Urban & 1 & 9 & 6 \\ \midrule
           
           Potsdam~\cite{Haala-Cramer-Jacobsen-DGPF-ISPRS, object-detection-JPRS-2014} & \makecell{Semantic \\segmentation}&\makecell{Potsdam\\(Germany)}& Urban & 3.5 & 5 & 6 \\ \midrule
           
           Inria~\cite{maggiori_can_2017} & \makecell{Semantic \\segmentation} & \makecell{USA, Austria\\(10 cities)}& Urban & 810 & 10 - 30 & 2 \\ \midrule
           
           DOTA~\cite{xia2018dota} & Object detection& Worlwide & Urban & - & Variable & 15\\ \midrule

           xView~\cite{lam-kuzma-mccord-xview-arxiv2018}& Object detection & Worlwide &Urban & 1415 & 30 & 60\\ \midrule
           
           DeepGlobe~\cite{demir-koperski-DeepGlobe18-cvprw18} & Road,  Build., LC & Worlwide &  Urb., Ctry & 2,220/ 984/ 1,717& 50/ 31/ 50 & 2/ 2/ 7 \\ \midrule 

           BigEarthNet~\cite{sumbul2019bigearthnet}& Multi-label classification & \makecell{Europe \\ (10 countries)}& Urb., Ctry & 850,000 &1,000 & $\sim$ 40\\\midrule 
           
           xBD~\cite{Gupta-xBD-CVPRW2019} & CD, Build. & 15 countries & Urban & 45,362 & 30 & - \\\midrule 
 
            HRSCD~\cite{daudt-19cviu-semanticCD} & CD, SS & \makecell{France \\ (2 areas)} & Urb., Ctry & 14,550 & 50 & 5$^2$
           \\\bottomrule 
           MiniFrance & \makecell{Semantic \\segmentation} & \makecell{France \\ (16 areas)} & Urb., Ctry & 53,000 & 50 & 12
           \\\bottomrule 
        \end{tabular}         
     \end{center}
   \end{table}

If some of the datasets mentioned above already take into account multiple locations, most are limited to urban scenes only and they are devoted to a single class (such as buildings) or to land cover (and not land use) classes. Land cover refers to the ground surface coverage: vegetation, urban infrastructure, water, etc; while land use indicates the purpose the land serves: urban, industrial buildings, agriculture, etc. The second is more interesting to analyze, because it provides further information about human activity in a given area, however extracting this information from images only remains a major challenge~\cite{fisher2005land-use-land-cover}. 
MiniFrance, however, offers scenes from urban and countryside zones, with land-use, high semantic level of classes and covers a vast surface (larger than other datasets at Very High Resolution -- VHR), with aerial images at a sub-meter resolution, including $\sim 150$ GB of data.

Furthermore, all the aforementioned datasets were designed for fully supervised learning, which does not correspond to the real practical case where huge amounts of imagery are available, but only a few images come with some labeled regions. MiniFrance is the first dataset that includes labeled and unlabeled data that can be used in training phases, thus recreating a realistic scenario.

\subsection{Semi-supervised Learning}\label{sec: related-work-semisup}

Semi-supervised Learning~\cite{chapelle-scholkopf-semisup-book} refers to all the techniques that are halfway between supervised and unsupervised learning. In these settings, available data can be divided in two parts: a labeled set where raw data and its corresponding target are provided, and an unlabeled set for which only raw data are available. The key idea behind semi-supervised learning is to learn a representation function (that maps a data point to its target) from labeled data as in the supervised approach, but using the  available unlabeled data to leverage information about structure of these data to help the learning process. This is a much realistic and compelling approach than supervised learning, since in real-life applications annotated data is difficult to procure -- even harder in the context of semantic segmentation, since one needs pixel-wise labels -- while raw data is plentiful.

Semi-supervised methods for semantic segmentation in deep learning have been developed in the last years, but mostly in the form of weakly supervision: from scribbles~\cite{improvingmaps2018, durand2017wildcat} bounding boxes~\cite{khoreva2017simple, papandreou2015weakly} and image-level annotations~\cite{papandreou2015weakly} to obtain dense, pixel-wise predictions.  Pseudo-labels~\cite{lee2013pseudo} can also be used to address the semi-supervised problem~\cite{chen2020digging}, propagating labels from annotated examples through non-annotated ones, according to a confidence criterion, to artificially enlarge available training data. Other works include unlabeled data during training in a generative adversarial network framework~\cite{souly2017semi, hung2018adversarial}. The method in \cite{kalluri2019universal} is similar to our settings in the way unlabeled images are exploited, but targets a domain adaptation task and requires an alignment of the features from multiple domains through an entropy module.

Semi-supervised methods for remote sensing applications have also been studied in the last years. ~\cite{xia2013semi-supervised-probabilistic-pca} presents a feature extraction method based on principal component analysis that uses labeled and unlabeled data, \cite{tuia2014semisupervised-manifold-alignment,hong2019learnable-manifold-alignment} leverage unlabeled examples to achieve manifold alignment of data coming from different modalities. More recently, deep learning based approaches have leveraged weakly labeled data for different purposes. \cite{nivaggioli2019weakly, schmitt2020weakly} use weak supervision for land cover classification. \cite{zhang2020map} uses open and incomplete available data (OpenStreetMap -- OSM) to generate maps of a large-scale zone. \cite{bonafilia2019building} also uses OSM data as weak labels for building extraction. \cite{Le2019weakly} performs weakly supervised semantic segmentation to detect penguin colonies on the Antarctic.

Fewer are the works that, like us, exploit completely unlabeled examples during the training process. \cite{tao2017unsupervised} uses labeled and unlabeled data in an alternating training process to perform semi-supervised semantic segmentation of remote sensing images, while \cite{zhu2019semi} leverages unlabeled data for domain adaptation purposes using an adversarial training strategy. 

Conversely to previous works, we aim here to leverage completely unlabeled images and fully annotated ones to jointly train deep neural networks with an adapted loss and architecture, in a multi-task learning framework, training one unique model end-to-end, for semi-supervised semantic segmentation of aerial images.

\vfill

\section{The MiniFrance Suite}\label{sec: MF-dataset-description}

Considering the limitations of current Earth Observation (EO) datasets emphasized in Section~\ref{sec:EO-datasets}, we propose a new large-scale benchmark suite for semi-supervised semantic segmentation: MiniFrance.
As in real life EO applications, it comprises both labeled and unlabeled imagery for developing and training algorithms. To our knowledge, this is the first dataset designed for benchmarking semi-supervised learning in the field. Moreover, it consists of a variety of classes

\begin{minipage}{\linewidth}
       \vspace{-1em}
       \begin{figure}[H]
           \centering
           \includegraphics[width=.6\linewidth]{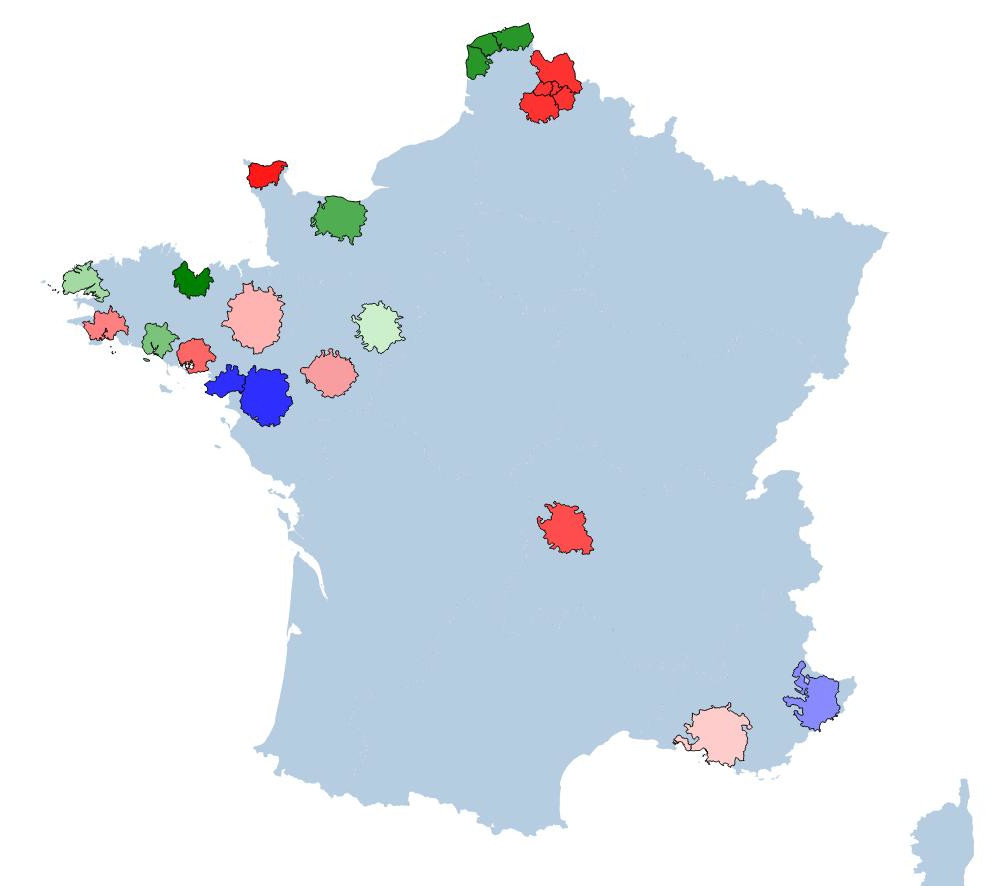}
           \caption{Dataset overview.} 
           \label{fig:minifrance-overview}  
       \end{figure}
       \vspace{-3em}
       \begin{table}[H]
          \begin{center}
          \caption{List of cities in MiniFrance and split details.} \label{tab:minifrance-cities-split}
         \vspace{-.5em}
          \begin{tabular}{ m{1em} m{1em} c c @{\hskip 8pt} r  @{\hskip 8pt} p{1.7em}}    \toprule
          & & \emph{Conurbation} &  \emph{Tiles} &  \emph{\% pixels} & \emph{Color} \\\midrule \vspace{2pt}
          \multirow{8}{*}{\rotatebox{90}{\emph{Training}}} & \multirow{2}{*}{\rotatebox{90}{\emph{Labeled}}} & Nice & 170 & 8.01 \% & \mybox{Nice}{Nice} \\[3pt]
          & & Nantes, Saint-Nazaire  & 226 & 10.65 \% & \mybox{Nantes}{Nantes} \\[4pt] \cmidrule{2-6}
          &  \multirow{6}{*}{\rotatebox{90}{\emph{Unlabeled}}} & Le Mans & 107 & 5.04 \% & \mybox{LeMans}{LeMans}  \\
          & & Brest & 88  & 4.14 \% &  \mybox{Brest}{Brest}\\   
          & & Lorient & 68 & 3.20 \%  & \mybox{Lorient}{Lorient} \\
          & & Caen & 126 & 5.94 \%  & \mybox{Caen}{Caen}  \\ 
          & & Dunkerque, Calais, Boulogne-sur-Mer & 150 & 7.07 \% &  \mybox{Dunkerque}{Dunkerque} \\
          & & Saint-Brieuc & 71 &  3.34 \% &  \mybox{Saint-Brieuc}{Saint-Brieuc}\\ \midrule
          \multirow{8}{*}{\rotatebox{90}{\emph{Test}}} & & Marseille, Martigues & 162 & 7.63 \% &  \mybox{Martigues}{Martigues}\\ 
          & & Rennes & 196 & 9.24 \% & \mybox{Rennes}{Rennes} \\
          & & Angers & 123  & 5.79 \% & \mybox{Angers}{Angers}\\
          & & Quimper & 79 & 3.72 \% & \mybox{Quimper}{Quimper}\\
          & & Vannes & 73 & 3.44 \% & \mybox{Vannes}{Vannes}\\
          & & Clermont-Ferrand & 150 & 7.07 \% & \mybox{Clermont}{Clermont}\\
          & & Lille, Arras, Lens, Douai, H\'enins & 275 & 12.96 \% & \mybox{Lille}{Lille}\\
          & & Cherbourg & 57 & 2.68 \% & \mybox{Cherbourg}{Cherbourg}\\ \bottomrule
          \end{tabular} 
          \end{center}     
       \end{table} 
  
   \end{minipage}

 \noindent on several locations with different appearances: this allows to push further the generalization capacities of the models.

\subsection{MiniFrance}

It consists of data corresponding to 16 conurbations and their surroundings from different regions in France (see Figure~\ref{fig:minifrance-overview} and Table~\ref{tab:minifrance-cities-split}). It includes urban and countryside scenes: residential areas, industrial and commercial zones but also fields, forests, sea-shore or low mountains.

\begin{minipage}{\linewidth}
      \begin{table}[H]
           \begin{center}
              \captionof{table}{Land Use classes available in MiniFrance.\label{tab:minifrance-classes}}
              \vspace{-.5em}
              \setlength{\tabcolsep}{3pt}
              \begin{tabular}{>{\centering\arraybackslash}p{6cm} r c }
               \toprule
                 \emph{Class} & \emph{\% pixels} & \emph{Color} \\\midrule
                 Urban fabric & 9.6 \% & \mybox{urbanfabric}{urbanfabric} \\
                 Industrial, commercial, public, military, private and transport units & 6.4 \% & \mybox{industrial}{industrial} \\
                 Mine, dump ans construction sites & 0.7 \% & \mybox{mine}{mine}  \\
                 Artificial non-agricultural vegetated areas& 1.1 \%& \mybox{artificial}{artificial} \\
                 Arable land (annual crops) & 29.5 \% & \mybox{arable}{arable}\\
                 Permanent crops & 1.0 \% & \mybox{permanent}{permanent}  \\ 
                 Pastures & 29.0 \% & \mybox{pastures}{pastures} \\
                 \textcolor{ggray}{Complex and mixed cultivation patterns} & \textcolor{ggray}{0.0 \%} & \mybox{complexcultiv}{complexcultiv}\\
                 \textcolor{ggray}{Orchards at the fringe of urban classes} & \textcolor{ggray}{0.0 \%} & \mybox{orchards}{orchards}\\
                 Forests & 15.9 \% & \mybox{forests}{forests} \\
                 Herbaceous vegetation associations & 4.6 \% & \mybox{herbacious}{herbacious}\\
                 Open spaces with little or no vegetation & 0.4 \% & \mybox{openspaces}{openspaces}\\
                 Wetlands & 0.7 \% & \mybox{wetlands}{wetlands}\\
                 Water & 1.0 \% & \mybox{water}{water}\\
                 Clouds, shadows or no data & 0.1 \% & \mybox{Black}{Black}\\\bottomrule
              \end{tabular}         
           \end{center}
         \end{table}
         \vspace{-5em}

         \begin{figure}[H]
           \begin{center}
              \vspace{1em}
              \setlength{\tabcolsep}{2pt}
              \begin{tabular}{ccc}
                 \includegraphics[width=.24\linewidth]{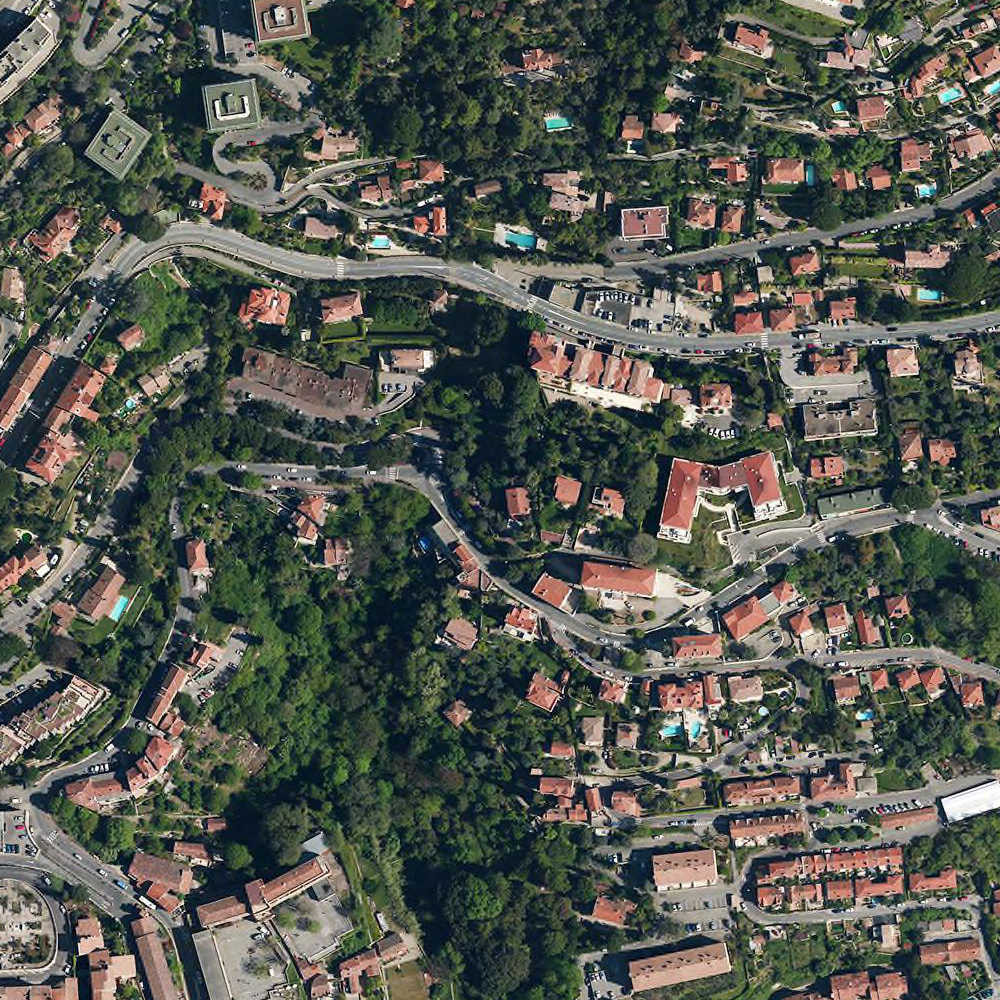} &
                 \includegraphics[width=.24\linewidth]{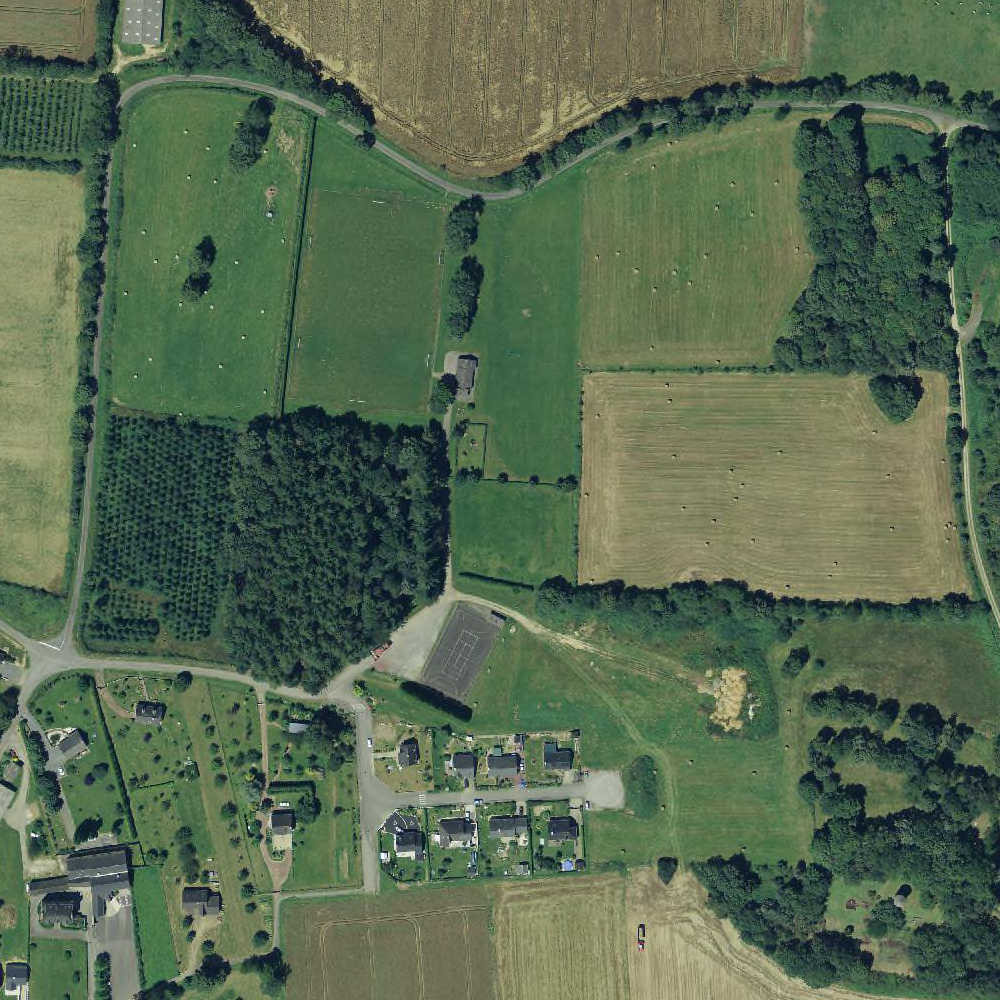} &
                 \includegraphics[width=.24\linewidth]{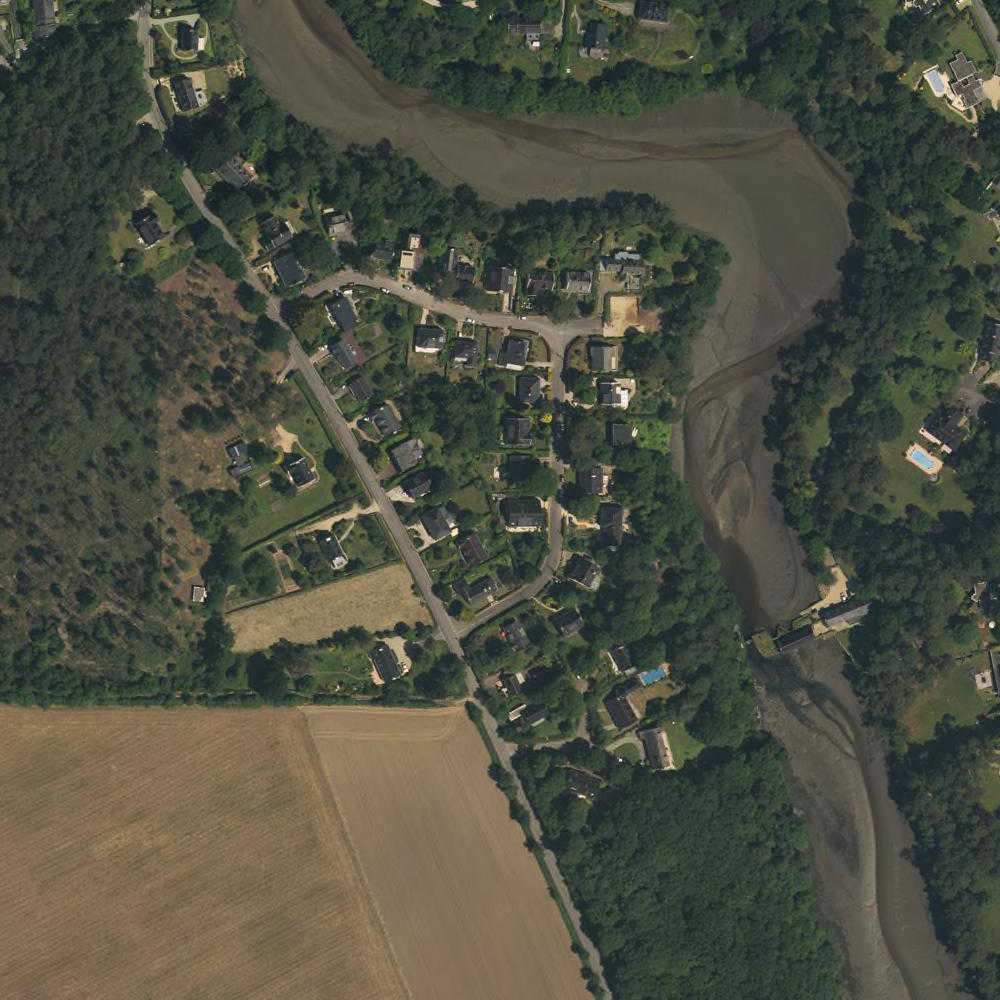} \\
                 \includegraphics[width=.24\linewidth]{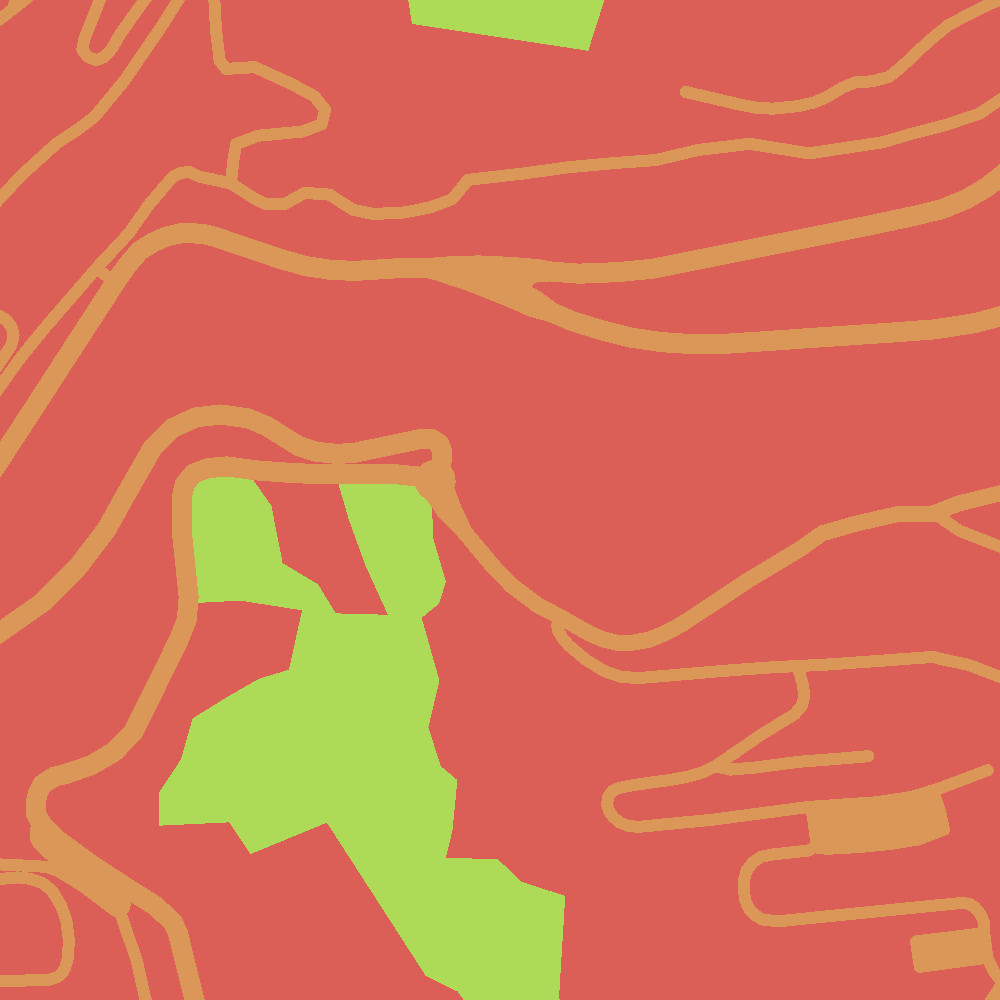} &
                 \includegraphics[width=.24\linewidth]{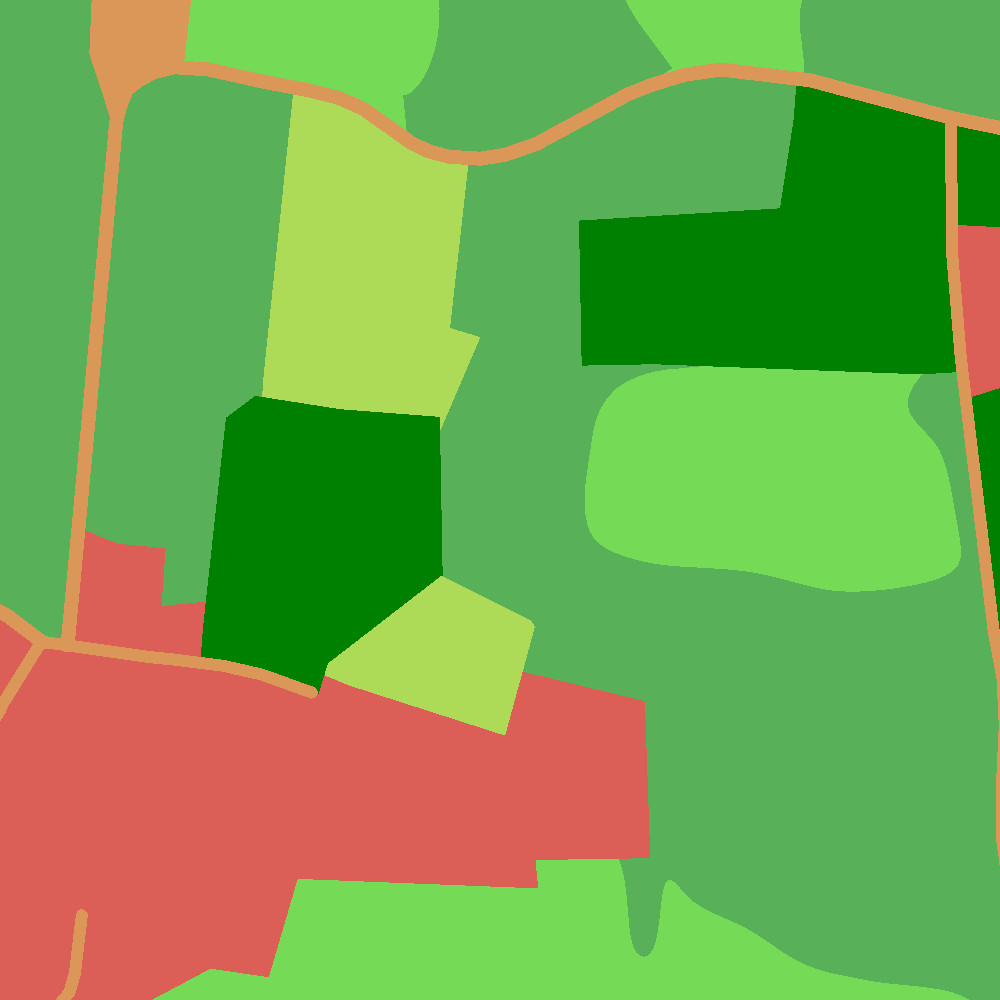} &
                 \includegraphics[width=.24\linewidth]{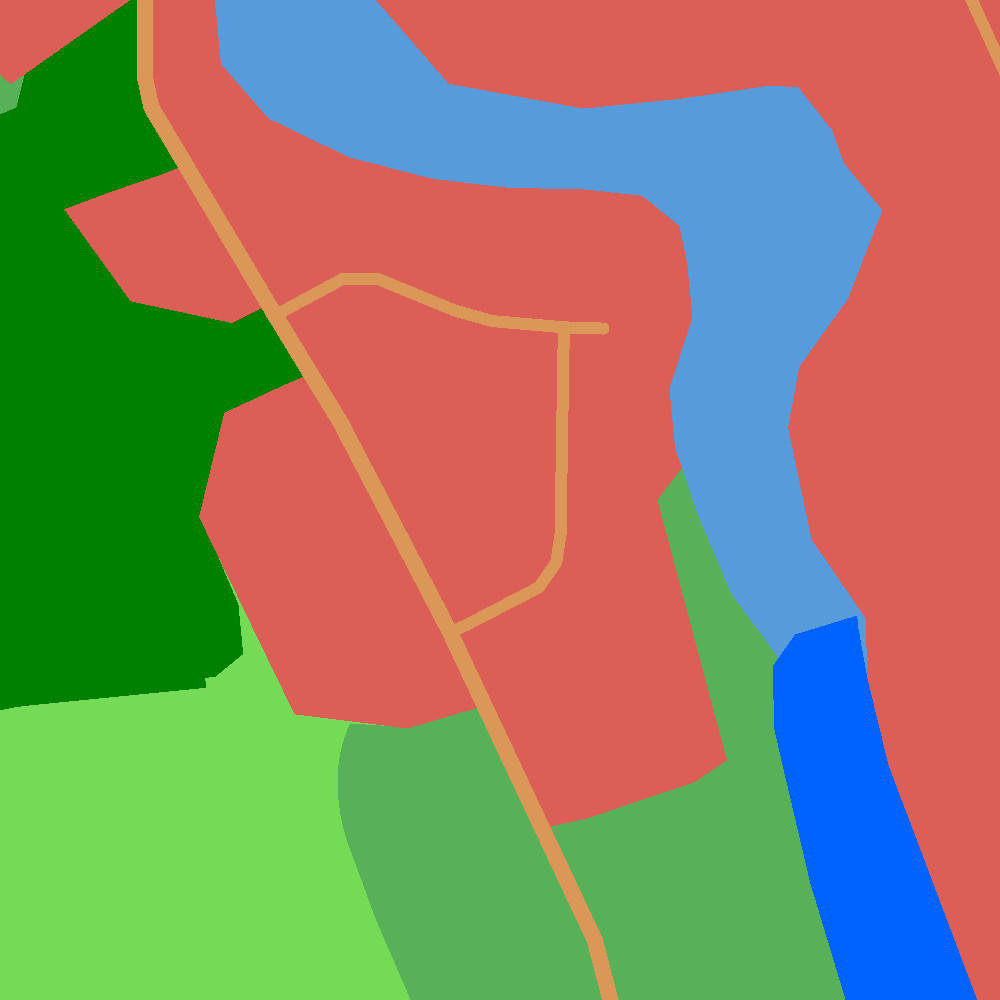}        
              \end{tabular}
           \end{center}
           \vspace{-1.2em}
           \caption{Some samples of \MF\ dataset on different localizations. Images~(up) and their associated ground-truth~(down). From left to right: Nice, Rennes and Vannes.} \label{fig: example-images-mf}
         \end{figure}    

   \end{minipage}

\medskip
MiniFrance gathers data from two sources:

\begin{itemize}
   \item Open data VHR aerial images from the French National Institute of Geographical and Forest Information (IGN) BD~ORTHO database\footnote{\url{https://geoservices.ign.fr/documentation/diffusion/index.html}}.\\   
   They are provided as RGB tiles of size 10,000~px $\times$ 10,000~px at a resolution of 50~cm/px, namely 25~km$^2$ per tile. Images included in this dataset were acquired between 2012 and 2014.
   \item Labeled class-reference from the UrbanAtlas~2012 database. Original data are openly available as vector images (i.e. containing polygon annotations) at the European Copernicus program website\footnote{\url{https://land.copernicus.eu/local/urban-atlas/urban-atlas-2012/view}}. Using the georeferenced data available in the BD~ORTHO, we have made rasters of these images that geographically match the VHR tiles from the BD~ORTHO. We consider 14 land-use classes (see Table~\ref{tab:minifrance-classes}), corresponding to the second level of the semantic hierarchy defined by UrbanAtlas~\cite{montero2014european-land-use-land-cover}. For this reason, some of them might not be present in the regions considered for \MF\ and they are colored in \textcolor{ggray}{gray} in Table~\ref{tab:minifrance-classes}.

\end{itemize}

Collecting data from different sources brings some burden that must be considered. Land use maps from UrbanAtlas are obtained through a semi-automatic process and thus they are not 100\% accurate~\cite{lefebvre2016monitoring-urban-areas}, besides polygon annotations might not match 50 cm/px resolution images precisely. Moreover, additional errors might come from the fact that image and ground-truth may not correspond to the same year.
Nonetheless, MiniFrance has several peculiar, unprecedented properties that we detail now.
\vskip2pt

{\parindent0pt
\textbf{Large-scale.} MiniFrance is a very large-scale dataset. It contains a total of 2,121 aerial images of size 10,000px~$\times $~10,000px at 50cm/px resolution. In terms of ground coverage, with 53,000~km$^2$ it is 12 times larger than DeepGlobe and larger than xBD, among the datasets of similar resolution.

\vskip3pt
\textbf{Rich and varied.} MiniFrance includes aerial images of 16 conurbations and their surroundings from different regions with various climates and landscapes (Mediterranean, oceanic and mountainous) in France. Introducing various locations leads to various appearances for the same class (buildings look different, vegetation is not the same and so on). 
Moreover, it combines urban centers, rural areas and large forest scenes. With respect to remote sensing datasets like ISPRS Vaihingen and Potsdam, it offers much more variety, as already observed in Section~\ref{sec:EO-datasets}. We propose an experimental comparison between \MF\ and Vaihingen in Section~\ref{sec: limits-of-EO-datasets}.

\vskip3pt

\textbf{High semantic level of classes.} %
MiniFrance considers 14 land-use classes, which is more than most of the datasets exposed in Section~\ref{sec:EO-datasets}. However, these classes have higher semantics: to identify an ``urban area'' an algorithm must be able to find several houses or buildings together, same to classify a forest. It is much easier to only consider classes at an object level (cars, buildings, trees, etc). Moreover, land-use classes are hard to learn, even for humans: how to distinguish \emph{pastures}~\mysmallbox{pastures}{pastures} from \emph{artificial non-agricultural vegetated areas}~\mysmallbox{artificial}{artificial} in Figure~\ref{fig: example-images-mf}?

\vskip3pt

\textbf{Underlying domain adaptation problem.}
Since train and test sets were split by city -- instead of excluding random tiles from all the zones -- algorithms developed on MiniFrance must address the underlying problem of domain adaptation. The appearance of classes might vary considerably from one city to another. Architecture is not the same, agriculture may change, etc. In Figure~\ref{fig: example-images-mf} we observe that \emph{urban fabric}~\mysmallbox{urbanfabric}{urbanfabric} does not look alike between the three exposed images.

\vskip3pt

\textbf{Designed for semi-supervised semantic segmentation.} To our knowledge, this is the first dataset specifically designed for semi-supervised learning strategies. Indeed, our training split includes labeled (two cities) and unlabeled images (six ones) while algorithms can be tested on the eight remaining cities.
With such a proportion of unlabeled examples, this fosters the development of new methods to leverage them. Moreover, these methods are likely to be easily transferred to lifelike scenarios and to have better generalization properties by design.
Table~\ref{tab:minifrance-cities-split} presents our training -- labeled and unlabeled images -- and testing splits.
}

\vspace{-1em}

\subsection{Tiny MiniFrance} 
To allow prototyping new algorithms with fast processing and validation times we also introduce \emph{\tmf} (tMF), a small, computationally tractable version of the MiniFrance dataset.

\tmf\ consists in a subsample of the original data: it contains 3,500 images of size 1,000px~$\times $~1,000px. Containing around 1,7\% of the original data, it preserves the variety and richness of MiniFrance. 

\begin{figure}[htbp]
   \centering
   \includegraphics[width=.6\linewidth]{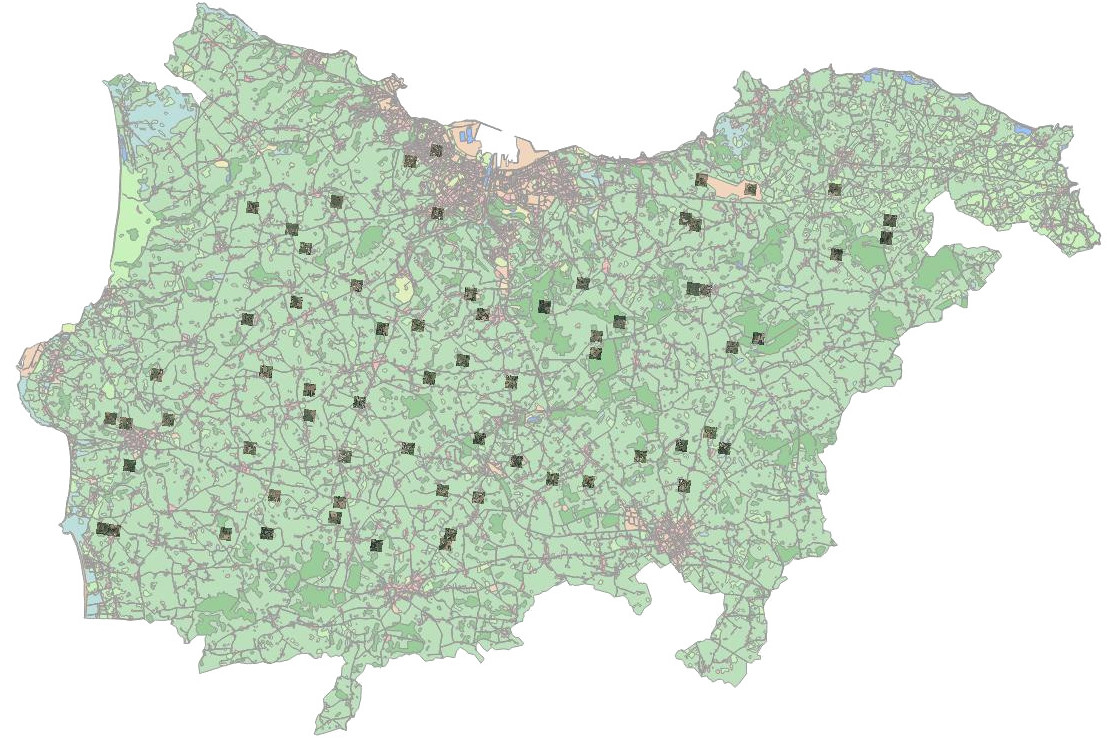}
   \vspace{-.5em}
   \caption{Subsample for \tmf\ over Cherbourg region.}
   \label{fig:tmfrance-sampling-cherbourg}
   \vspace{-1em}
\end{figure}

Sampling is uniform over each region. To preserve the same balance between classes, \added{it is performed by randomly selecting sub-tiles from original tiles in the dataset and verifying that there is at least one sub-tile from each tile in \MF.  Figure~\ref{fig:tmfrance-sampling-cherbourg} illustrates the result of sampling over the region of Cherbourg.} Moreover, we keep the original proportion of images per region on the dataset (e.g. the region of Nice contains more data than Brest, as in table~\ref{tab:minifrance-cities-split}). Training -- labeled and unlabeled -- and testing splits remain unchanged with respect to the original dataset.

Table~\ref{tab:tmf-classes-distribution} shows the classes distribution over \tmf. When compared with  Table~\ref{tab:minifrance-classes}, the original proportions of classes of MiniFrance are well preserved. Thus, we can expect that algorithms developed on \tmf\ will scale up similarly to MiniFrance. For this reason and for computing capacities, all the following analysis and experiments will be performed over \tmf, with the exception of Section~\ref{sec: experiments-MF}. However, and for the sake of simplicity, we will mostly employ the term \MF.

\begin{table}[htb]
   \small
   \begin{center}
   \caption{Classes distribution on \tmf.\label{tab:tmf-classes-distribution}}
   \setlength{\tabcolsep}{4pt}
      \begin{tabular}{cc|cc|cc}\toprule
         \emph{Class} & \emph{\% px} & \emph{Class} & \emph{\% px} & \emph{Class} & \emph{\% px} \\ \midrule
         Urban & 9.9 \% & Permanent & 1.3 \% & Herbaceous & 4.5 \% \\
         Industrial & 6.5 \% &  Pastures & 27.3 \%  &   Open & 0.1 \% \\
         Mine & 0.7 \% & Complex & 0.0 \% & Wetlands & 0.7 \% \\
         Artificial & 1.2 \% & Orchards & 0.0 \% & Water & 1.0 \% \\
         Arable & 30.7 \% & Forest & 16.0 \% & Clouds & 0.1 \%\\\bottomrule
      \end{tabular}
      \vspace{-1em}
\end{center}
\end{table}

\section{Statistical analysis of the representativeness of training and test datasets}\label{sec: statistical-analysis}

\vspace{-1.1mm}
\begin{figure}[H]
   \centering
   \includegraphics[width=.8\linewidth]{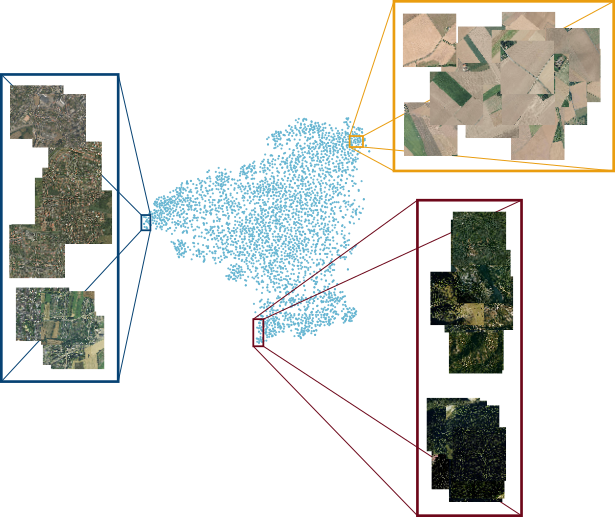}
   \caption{2D representation of images by t-SNE after ResNet34 encoding. Similar projections are close, while different visual features are separated. In~\mysmallbox{azulmarino}{azulmarino}, mostly urban scenes; in~\mysmallbox{mostaza}{mostaza} fields images and in~\mysmallbox{burdeos}{burdeos} mostly forest scenes.}\label{fig:tmfrance-tsne-zoom} 
   
\end{figure}

This section introduces two concepts that are required to have adequate learning conditions to achieve satisfying results and that explain our choice for labeled training data, unlabeled training data and test data for \MF: class representativeness and appearance.

On the one hand, class representativeness refers to the fact that to properly learn a certain class, any learning algorithm needs to see at least some examples of this class during training. Otherwise, it will not be able to identify it successfully at inference time. Hence, the labeled training split should contain examples of all possible classes in the dataset.

On the other hand, in a standard supervised setting, appearance features in the training set should have the same distribution as those on the test set to achieve good inference results. However, in a semi-supervised learning setting, unlabeled training data relax such a strong constraint. Indeed, by providing more information on the possible visual features, they help learning a wider appearance of each class. This is appealing since it favors generalization, but also brings more robustness against distribution shift (i.e. it is more unlikely that the test set contains very new appearances w.r.t. the test set).

\vskip2pt
According to this, we consider that a good training split should satisfy two conditions:
\vspace{-2.5pt}
\begin{enumerate}[($i$)]
\item Labeled training data must contain a good representation of all classes in the dataset, ideally with the same distribution than the testing data.
\item Training data (labeled and unlabeled) must cover all the range of appearances of different visual features in the dataset.
\end{enumerate}

In what follows we present a statistical analysis of the \MF\ dataset to show that our chosen split (in Table~\ref{tab:minifrance-cities-split}) satisfies these two requirements.

\subsection{Appearance analysis.}

To study the appearance similarity between the training split and testing split of \MF\ data, we rely mainly on two tools. First, we use pre-trained Convolutional Neural Networks (CNNs) as image feature extractors. Indeed, thanks to their shared-weight architecture and translation invariance, CNNs are reliable encoding tools for images. Furthermore models pretrained on ImageNet -- a very large database for visual recognition -- have seen a wide variety of representations that allow them to output a vector encoding the image's appearance. Secondly, we apply the t-SNE~\cite{maaten2008visualizing} algorithm to reduce the dimension of the high-dimensional feature vectors and visualize them in a 2D space\footnote{\added{t-SNE is a non-linear dimensionality reduction technique that allows visualization of high-dimensional data. In brief, the algorithm starts by converting the euclidean distances between high dimensional objects into conditional probabilities that represent similarities. Then, it defines a Student t-distribution with one degree of freedom over the low-dimensional points. Finally, it minimizes the Kullback-Leibler divergence between the high and low-dimensional distributions with respect to the locations of the low-dimensional points. At the end, if two high-dimensional objects are similar, then their representations at the low-dimensional t-SNE visualization are close and vice-versa. }}. Given the assumption that CNNs encode for image appearance, look-alike images should be close in the 2D representation space, while images with different visual features should be apart.

\medskip

\begin{figure}[!htbp]
   \centering
   \includegraphics[width=.9\linewidth]{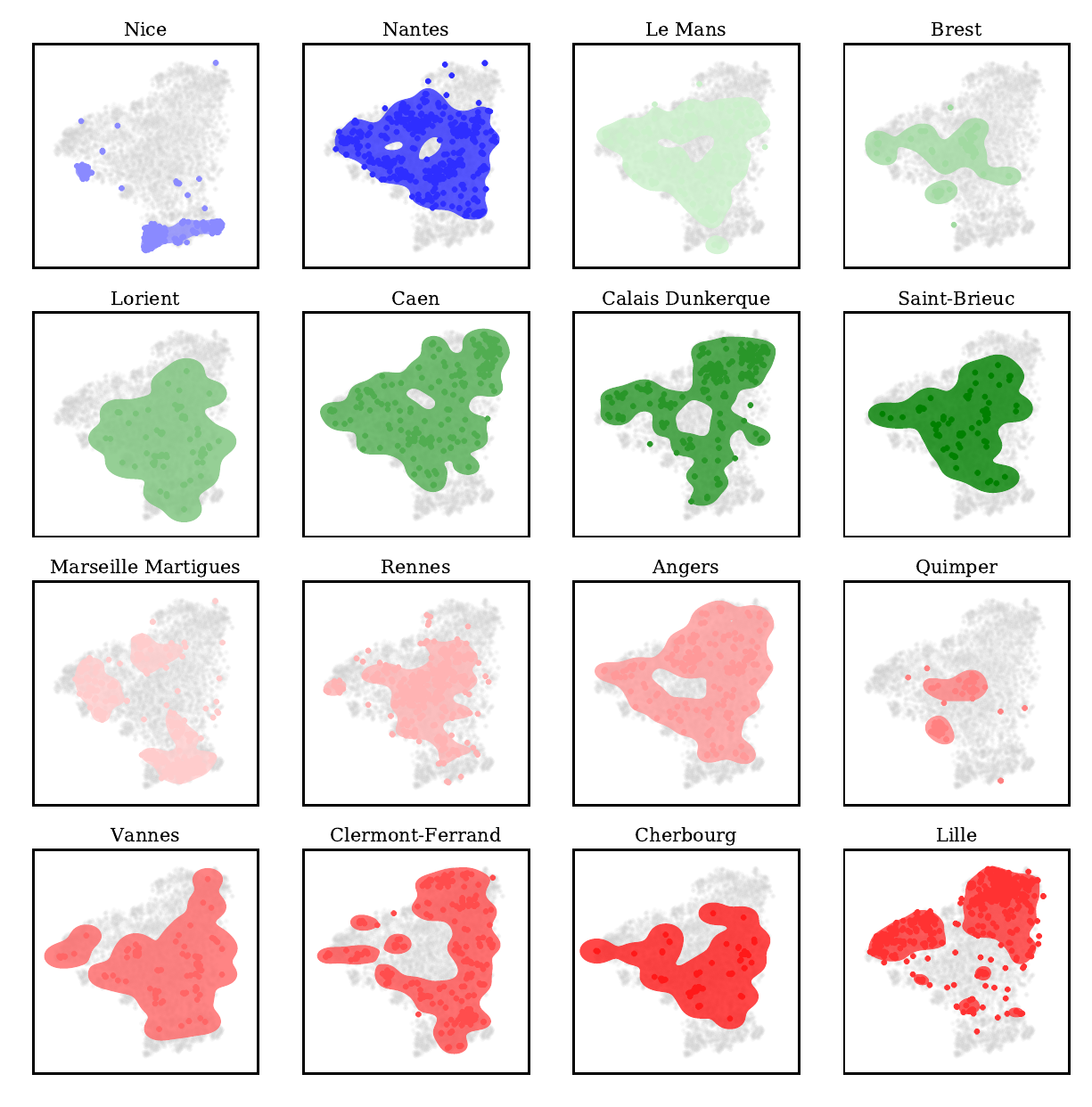}
   \caption{Distributions of cities in the 2D appearance space.}\label{fig:tmfrance-resnet-svmdensity}   
\end{figure}

\begin{figure}[!htbp]
   \centering
   \includegraphics[width=.9\linewidth]{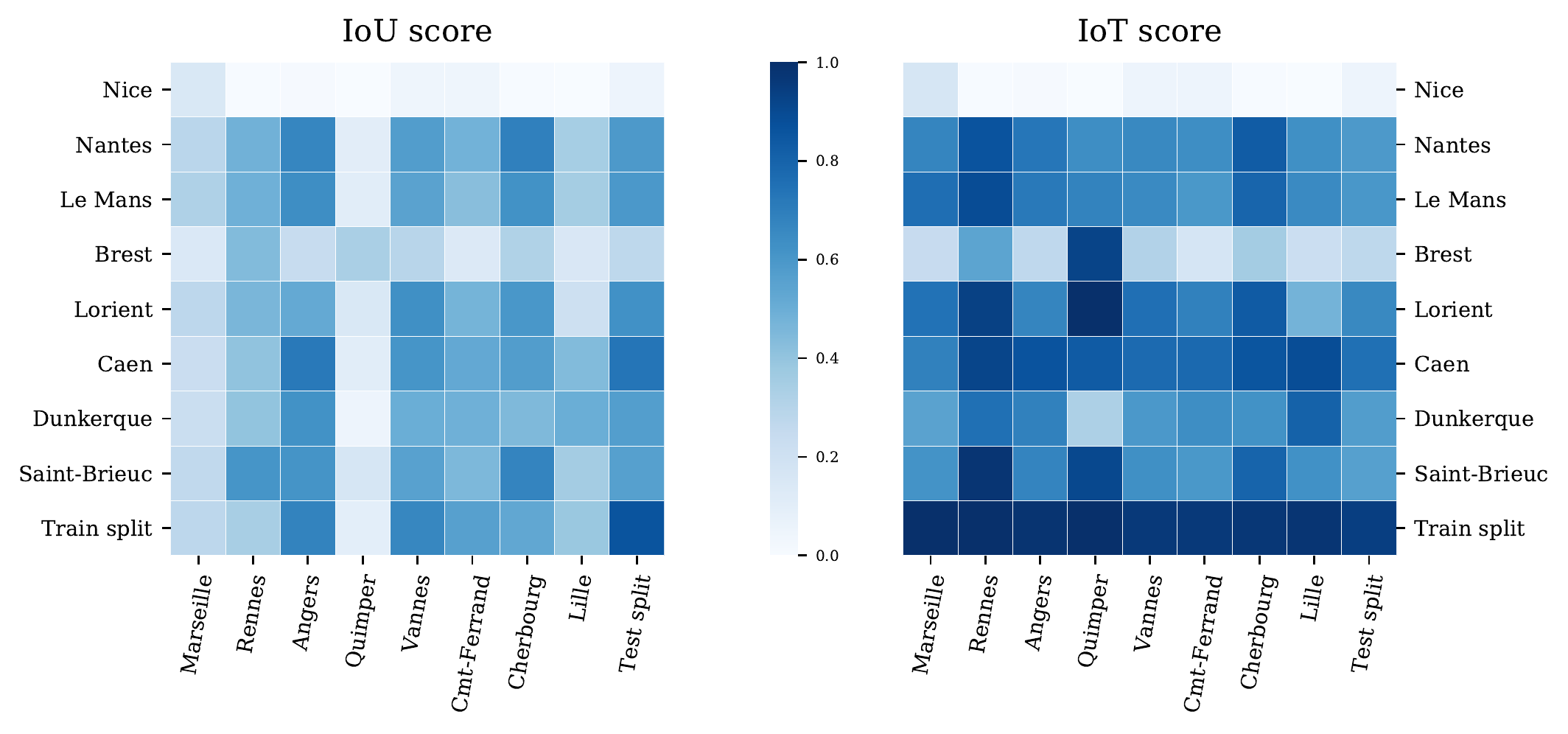}
   \caption{IoU and IoT (Intersection over Test) scores between the 2D distributions of cities in the training split and the testing split, represented as heatmaps. Last column represents the scores between a training city and the union of surfaces of the testing split. Similarly, last row corresponds to the scores between the union of surfaces in the training split and every city in the test. The dark last row of the IoT score indicates that the train split covers well every city in the test partition.}\label{fig:tmfrance-iou-scores}   
\end{figure}

Thus, our algorithm for appearance coverage assessment between datasets is summarized as follows:
\vspace{-4pt}
\begin{itemize}
   \item For each image in the dataset we obtain an encoded feature vector through a CNN (in particular, we use a VGG16~\cite{simonyan2015VGG} and a ResNet34~\cite{he2016deep} \footnote{In what follows, we present only images of the results with ResNet34 encoding. However, VGG16 encoding shows similar results.}).
   \item  Then, we apply a t-SNE to this set of high-dimensional feature vectors to obtain a 2D representation of the dataset images which preserves the original similarity of visual features.  Figure~\ref{fig:tmfrance-tsne-zoom} shows the mapping result and validates that similar images are close while different appearances are put apart. 
   \item Each point in the 2D space can be traced back to the original tile and so to the city it comes from. Then, we use a one-class SVM~\cite{scholkopf2001estimating} to estimate the distribution of the city images in the 2D space. It results in appearance maps which are shown in Figure~\ref{fig:tmfrance-resnet-svmdensity}. 
   \item Finally, we evaluate the appearance similarity and coverage between cities using two metrics:
   \begin{enumerate}[($i$)]
   \item We use the intersection over union score (IoU, the standard metric for object detection) between the surfaces defined by the distributions, or appearance maps, to assess appearance similarity. Let $S_1$ and $S_2$ be two sets, the IoU score between them is defined as $IoU(S_1, S_2) = \tfrac{|S_1 \cap S_2|}{|S_1 \cup S_2|}$. In our context, higher IoU scores relate to  resemblance between the appearance maps of cities. 

   \item We also introduce the Intersection over Test area score (IoT). Let $S_1$ and $S_2$ be two sets, the IoT score between them is defined as $IoT(S_1, S_2) = \tfrac{|S_1 \cap S_2|}{|S_2|}$. This score measures the area covered by the intersection of the two surfaces normalized over the second area, which is the objective. We compute IoT considering $S_1 \in T$ and $S_2 \in E$, where $T$ and $E$ are the set of training cities and the set of testing cities, respectively. Thereby IoT measures how well the objective appearance map is covered by appearances of the training data.
   \end{enumerate}

\end{itemize}

Figure~\ref{fig:tmfrance-iou-scores} shows these scores as two heatmaps between cities in the training set and the ones in the test set. Results are consistent with reality, to name a few examples: Nice exhibits low similarity scores with all cities, except Marseille, because those are the only cities from Mediterranean coast. Quimper has its higher IoU score with Brest, which is coherent because of their geographic proximity; in terms of IoT Quimper is well covered by Lorient and Saint-Brieuc, which are also geographically close (all these cities are located in Brittany). High IoU score between Angers and Caen is justified by the fact that both are agricultural localities, with similar landscapes.

To summarize, we propose a method to assess representativeness in terms of appearance similarity between cities in the \MF\ training split and the ones in the testing split. IoU scores show that, even if there are similarities between cities, no locality in the training set is identical to another one in the test set. However, IoT proves that testing cities are well covered by the ensemble of training cities, which is confirmed by the last dark row of this score in Figure~\ref{fig:tmfrance-iou-scores} (right).

\subsection{Class representativeness analysis.}

A class cannot be learnt if no example of it has been seen at training time. In other words, the labeled training partition has to contain all the existing classes on the dataset. If possible, the distribution of the classes during training should be similar to the one of test data. 

\begin{figure}[htb]
   \centering
   \includegraphics[width=\linewidth]{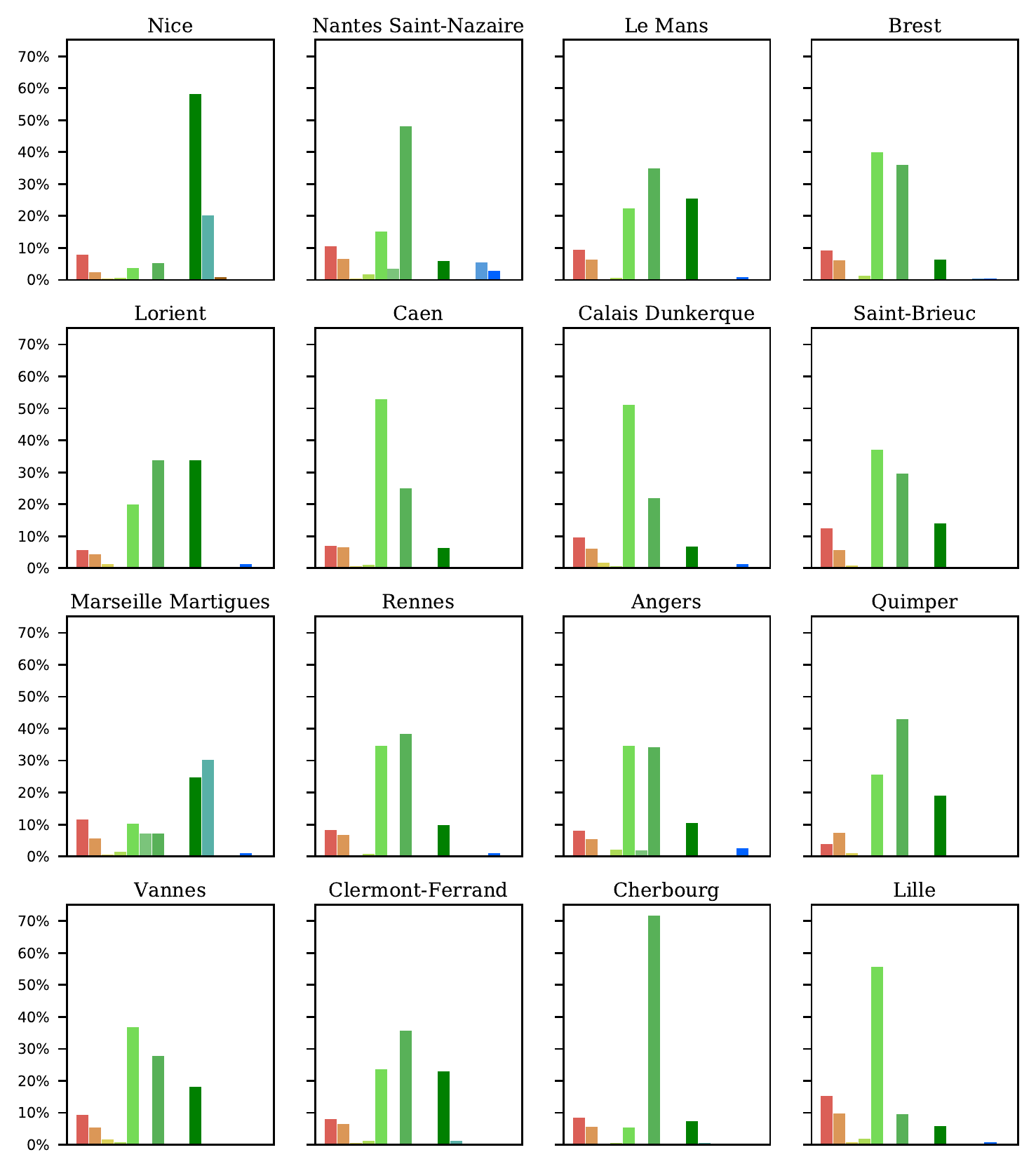}
   \vspace{-1em}
   \caption{Histograms of class distributions by city. $x$ axis represents the classes with colors as in table~\ref{tab:minifrance-classes}. $y$ axis presents the percentage of each class by city.}
   \label{fig:tmfrance-class_distributions}
   
\end{figure}

To fulfill this condition, we study the classes distribution on the dataset. We compute class histograms of each geographic area and present them in Figure~\ref{fig:tmfrance-class_distributions}. We observe that they vary significantly from one city to another. Besides, among the 12 classes that we consider in this analysis -- we do not consider \emph{complex and mixed cultivation patterns}, \emph{orchards at the fringe of urban classes} nor \emph{clouds and shadows}\footnote{\emph{clouds and shadows} is not a land use class and thus it is not interesting to our problem.}, see Table~\ref{tab:minifrance-classes} --, no city contains all of them. The best coverage of classes is given by the \emph{Nantes, Saint-Nazaire} or \emph{Marseille, Martigues} conurbations that exhibit 10 of the classes. However, most of the regions contain only 7 or 8 categories in total. 

Another problem is the heterogeneous proportions of classes in each region. The most striking example is \emph{Cherbourg} where 6 classes are represented and one of them -- \emph{pastures} -- covers 70\% of the total pixels, while the other categories count for less than 10\% each.

Therefore, defining a labeled training split that represents all the classes in a good proportion is not straightforward. %

\begin{figure}[htb]
   \centering
   \includegraphics[width=\linewidth]{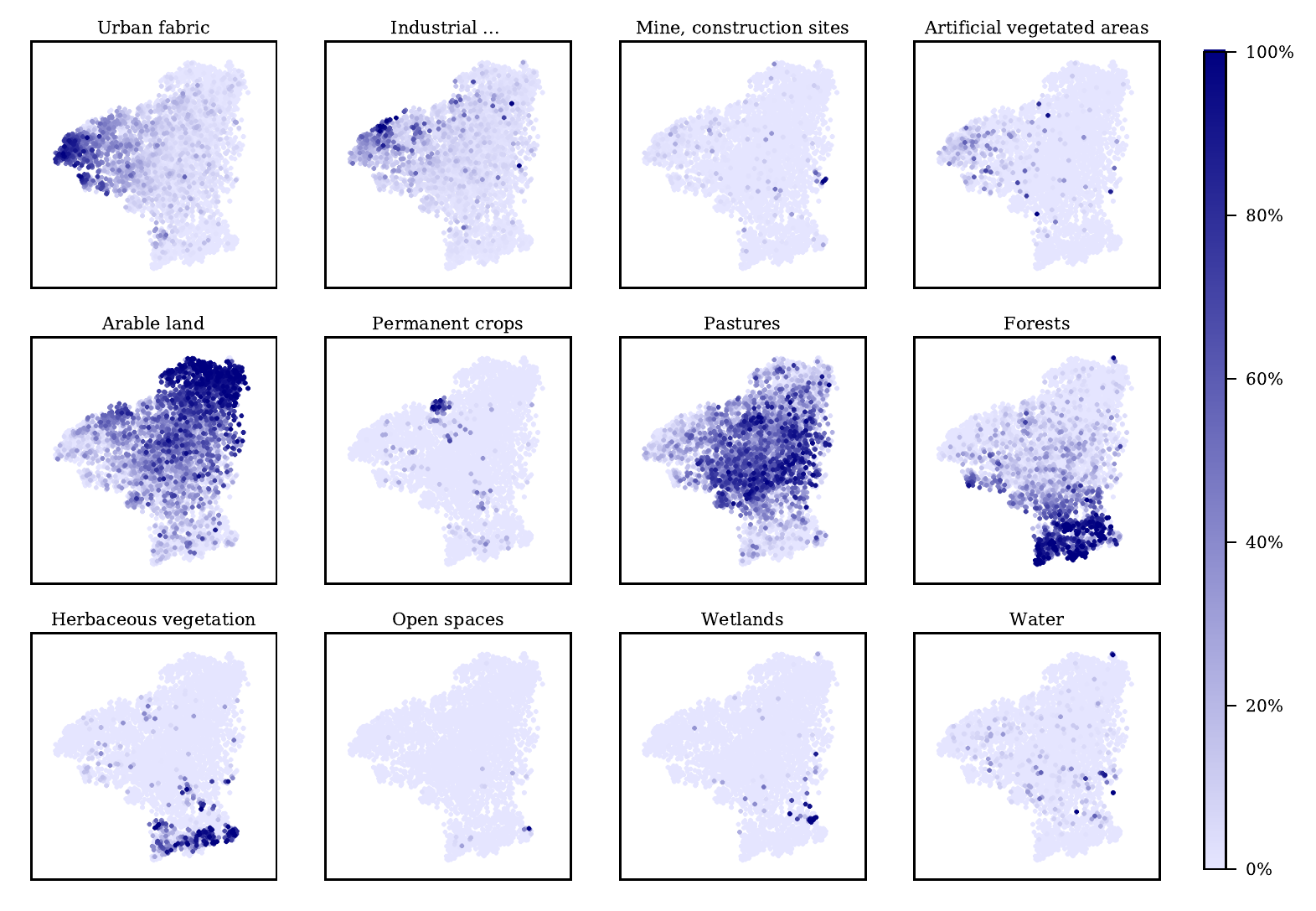}

   \caption{Class distributions in the 2D appearance space. One subplot represents one class. Each point is colored as the proportion occupied by a given class over the corresponding image.}
   \label{fig:tmfrance-resnet-class-presence-proportions}
   
\end{figure}

Along with the histograms, we make use of our precedent analysis to understand the distribution of the classes in the images in terms of appearance. Each subplot in Figure~\ref{fig:tmfrance-resnet-class-presence-proportions} presents a class in the dataset and contains all the images in the 2D appearance representation space.  
Each point is colored according to the proportion occupied by the class over the image. That is, the darker the point in the figure~\mysmallbox{darkcolor}{darkcolor}, the more pixels corresponding to the class are in the image. On the contrary, a light point~\mysmallbox{lightcolor}{lightcolor} indicates that there are very few pixels representing the class.
We observe that some classes (such as \emph{pastures} or \emph{arable land}) are well-spread over the whole appearance space, with high proportions in many tiles. This means that they are represented by diverse images and that they are likely to have a lot of examples (as confirmed by the histograms of Fig.~\ref{fig:tmfrance-class_distributions}). These classes should be easier to learn. Others -- like \emph{urban fabric} or \emph{industrial, commercial, public, military, private and transport units} -- are widespread, but do not reach majority in most of the images in which they are present. This means that these classes have a large variance in their appearance but not so many examples per appearance mode, which could make them more difficult to learn.
Moreover, other categories (like \emph{artificial non-agricultural vegetated areas} or \emph{herbaceous vegetation associations}) are mostly concentrated over one zone -- that could correspond to only one geographic region --, that is, they are present in images of specific appearances, which makes them even harder to learn. Finally, we see classes that are extremely rare (e.g. \emph{wetlands} and \emph{open spaces with little or no vegetation}), they are present in a few images only, and thus they should be the more difficult to learn.


All of the above shows that we can combine class distribution and visual appearance mapping to get further insight on the data. These tools help us to define a suitable partition of the \MF\ dataset -- labeled, unlabeled and test data -- that satisfies the class distribution and appearance conditions as we will show in section~\ref{sec: results-split-definition}. 

\section{Semi-supervised Semantic Segmentation with Deep Neural Networks}\label{sec: deep-semi-supervised-learning}

In this section, we introduce multi-task deep neural networks for semi-supervised semantic segmentation which will serve as baselines on the MiniFrance dataset. We aim to use unlabeled data to help generalization for semantic segmentation of aerial images. The challenge is two-fold: designing network architectures able to deal with both labeled and unlabeled images, and selecting unsupervised tasks to perform along with the appropriate auxiliary loss function.

Let $\phi_s(\cdot)$ be the function learned by a supervised segmentation network (for the sake of simplicity, the corresponding network will also be referred as $\phi_s$).
Such a network can be optimized through supervised learning using stochastic gradient descent and a classification loss $\LL_s$ (cross entropy loss is a standard choice). We denote $x$ the input image and $y$ the target label, then:
\begin{equation}
    (x,y) \mapsto \LL_s( \phi_s(x), y )
\end{equation}

From a general point of view, using unlabeled data to help the previous optimization can be seen as a second task optimized with a loss function $\LL_u$ and a transfer function through the network denoted by $\phi_u$. Without labels, unsupervised losses usually rely on comparing in some way the output to the input image:
\begin{equation}
    x \mapsto \LL_u( \phi_u(x), x )
\end{equation}

In order to improve the genericity of $\phi_s$, one has to relate $\phi_s$ and $\phi_u$. This is generally done by partially sharing parameters between both networks.
Finally, the semi-supervised loss is a weighted sum of the losses for each individual task:
\begin{equation}\label{eq: semisupervised-loss}
    \LL(x) = \LL_s(\phi_s(x), y) + \lambda \LL_u(\phi_u(x),x)
\end{equation}

\subsection{Neural Network Architectures}\label{sec: semisup-neural-networks}

\begin{figure*}[ht!]
    \centering
    \begin{tabular}{>{\centering\arraybackslash}m{.23\linewidth}@{\hspace{1.2cm}}>{\centering\arraybackslash}m{.22\linewidth}@{\hspace{1.2cm}}>{\centering\arraybackslash}m{.3\linewidth}}

     \includegraphics[width=\linewidth]{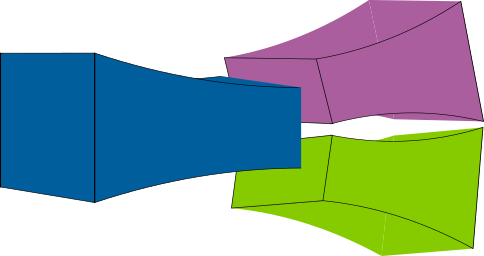} &

     \includegraphics[width=\linewidth]{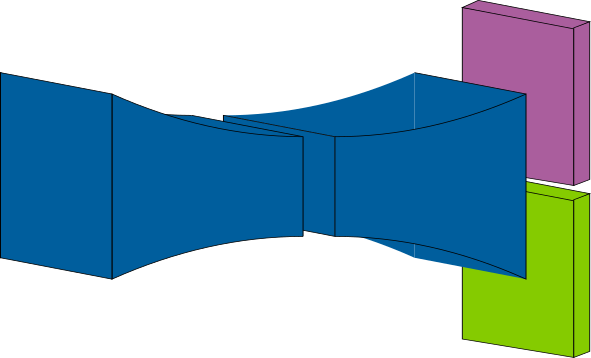}  &

    \includegraphics[width=\linewidth]{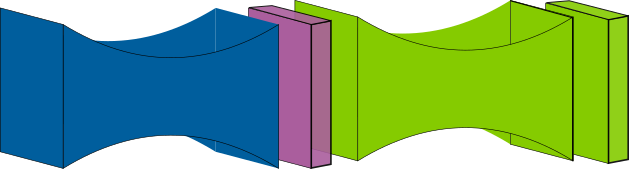}   \\
         (a) BerundaNet-early & (b) BerundaNet-late & (c) W-Net \\
    \end{tabular}
    \caption{Proposed neural network architectures for semi-supervised learning. Shared layers are depicted in blue, supervised layers are in purple, and unsupervised layers are shown in green. \label{fig:semisup-archis}}
 \end{figure*}

We propose here two types of semi-supervised networks which process the multi-task optimization -- semantic segmentation as the supervised task, along with an unsupervised task -- either as parallel streams or as sequential objectives (Figure~\ref{fig:semisup-archis}).

{\parindent0pt

\medskip
\textbf{BerundaNet (with early and late task splitting)}\\
Standard encoder-decoder networks for semantic segmentation \added{-- such as SegNet~\cite{badrinarayanan2015segnet} or U-Net~\cite{ronneberger2018unet} --} can easily be extended for multiple task learning by adding a new head with a loss for the new, unsupervised task~\cite{daudt-19cviu-semanticCD, carvalho-2019grsl-mtl3D}. With such an architecture (thereafter named BerundaNet after the mythological two-headed bird), both tasks have shared parameters until the data streams are split. We distinguish two variants depending on the splitting layer. Early splitting networks have one encoder and two decoders, one for each task (Fig.~\ref{fig:semisup-archis} (a)). On the contrary, with late-splitting task specialization occurs at the very end. It has an almost-all shared decoder with only a single separate convolutional layer for each task (Fig.~\ref{fig:semisup-archis} (b)).

Eventually, all architectures optimize the global loss defined in Eq.~\eqref{eq: semisupervised-loss}. $\LL_s$ can be any supervised loss for semantic segmentation, and in the following we consider the cross-entropy loss. $\LL_u$ is an unsupervised loss. In the experiments we will consider reconstruction losses (such as $\LL_1$ or $\LL_2$) and unsupervised image segmentation losses that will be presented in Section~\ref{sec: unsupervised-losses}.

\medskip
\textbf{W-Net}~\cite{xia2017wnet, chen2018wnet}\\
Multiple task learning can also be processed sequentially, as in W-Net~\cite{xia2017wnet} which combines two unsupervised objectives: segmentation and reconstruction. W-Net consists of two stacked U-Net~\cite{ronneberger2018unet}, hence its name. We adapt the original design to semi-supervised learning by specializing the first U-Net block on the semantic segmentation task and focusing the second one on the unsupervised objective (Fig.~\ref{fig:semisup-archis} (c)). With respect to previous notations, in this case the network $\phi_s$ shares all parameters with $\phi_u$. 
At the end of the first U-Net block, a soft-max layer is included to achieve the supervised classification.

The loss function for our semi-supervised W-Net architecture is then more precisely decomposed as follows:
\begin{equation}\label{eq: semisupervised-loss-wnet}
      \LL(x) = \LL_s(\phi_s(x), y) + \lambda \LL_u(\phi_u(\phi_s(x)),x)
\end{equation}
where $x$ is the input image, $y$ its corresponding ground truth, $\phi_s(\cdot)$  represents the first U-Net block and $\phi_u(\cdot)$ represents the second U-Net block. As before, $\LL_s$ can be any supervised loss for semantic segmentation and $\LL_u$ is an unsupervised loss.

} 

\bigskip
This kind of architectures -- BerundaNet and W-Net -- allows us to deal with both labeled and unlabeled data during training. When a labeled example is processed the gradient is backpropagated trough the whole network, whereas if an unlabeled example is processed gradients are only backpropagated through the unsupervised part and shared parameters of the network (green and blue blocks in Figure \ref{fig:semisup-archis}). However, the main objective is still the semantic segmentation task. Thus, even if unsupervised parts are helpful during the training process, evaluation can be performed without them, which yields in standard-size inference networks.

\subsection{Unsupervised Losses for Image Segmentation}\label{sec: unsupervised-losses}

We now present some unsupervised losses $\LL_u$ which can leverage the information brought by images with no label.  Two task objectives are usually considered, image reconstruction and image segmentation, leading to the following general formulation:
\begin{equation}\label{eq:unsupervised-loss-decomposition}
   \LL_u(\cdot) = \alpha^{(rec)} \LL^{(rec)}(\cdot) + \alpha^{(reg)} \LL^{(reg)}(\cdot)
\end{equation} 
where $\LL^{(rec)}$ is a reconstruction loss, $\LL^{(reg)}$ is a regularization loss and $\alpha^{(rec)}, \alpha^{(reg)}$ are balance coefficients.

In the following, we adapt some existing losses to semi-supervised semantic segmentation, and also propose a novel implementation of a \emph{relaxed K-means} loss for unsupervised image segmentation.

{ 

\parindent0pt

\medskip
\textbf{Image reconstruction losses}\\
Image reconstruction losses can be simply defined using solely standard reconstruction losses such as the classical $\LL_1$ and $\LL_2$, as in equations \eqref{eq: l1-loss} and \eqref{eq: l2-loss}. They enforce the encoding power of internal representations built by the network $\phi_s$ by closing the loop from it to the original input, the image itself. This kind of self-supervision is for example used in~\cite{xia2017wnet}.

\begin{align}\label{eq: l1-loss}
    \LL_1 (x) =& \frac{1}{N}\sum_{i=1}^N |x_i - \hat{x}_i|, \\
    \label{eq: l2-loss}
    \LL_2(x) =& \frac{1}{N}\sum_{i=1}^N (x_i - \hat{x}_i)^2
\end{align}
where $x_i$ denotes the $i^\text{th}$ pixel of the image, $\hat{x}_i$ its reconstructed version and $N$ the number of pixels in the image.

\medskip
\textbf{Relaxed K-means} \\
We propose a new loss for unsupervised image segmentation, which combines the old intuitions behind the $k$-means algorithm with the expressive power of neural network's non-linear modeling. In a standard manner, it is cast as a color image quantization problem, where the objective is to find an optimal, reduced set of $K$ colors for encoding the image. Formally, it minimizes the reconstruction loss $\LL^{(rec)}(x, x_c)$ where $x_c$ is the quantized image.

We still denote $x$ the input image and $x_i$ its value at pixel $i$. $k$-means alternatively optimizes centroids of color clusters $c_k$ ($k \in \{1,K\}$) and membership matrices $\hat{y}^{(k)}$ of $x$ to cluster $k$. It follows:
\begin{equation}\label{eq:cluster-center}
   c_k =\frac{\sum_{i} x_i \hat{y}_i^{(k)}}{\sum_{i} \hat{y}_i^{(k)}}
\end{equation}
and 
\begin{equation}\label{eq: clustered-image}
   x_c = \sum_{k=1}^K c_k\cdot \hat{y}^{(k)}
\end{equation}

In standard $k$-means, memberships $\hat{y}_i^{(k)} \in \{0,1\}$ are then determined such that $|| x_i - c_k ||^2$ is minimum. Instead, we relax the hard constraint so that $\hat{y}_i^{(k)} \in [0,1]$ and estimate memberships as the output $\hat{y}=\phi(x)$ of a network which minimizes $\LL^{(rec)}(x, x_c)$. In our experiments we will use:
\begin{equation}\label{eq:reconstruction-alex}
   \LL_{km}^{(rec)}(x) = \LL_1(x,x_c)
\end{equation}

Eventually, to compensate for the relaxation we add a regularization term which ensures memberships are peaked to a one-cluster-per-pixel distribution:
\begin{equation}\label{eq:regularization-alex}
   \LL_{km}^{(reg)}(x) = \sum_{k=1}^K  \sum_{i} \hat{y}_i^{(k)}\cdot(1-\hat{y}_i^{(k)})
\end{equation}
The whole unsupervised loss is then in the form of Eq.~\eqref{eq:unsupervised-loss-decomposition}.

\medskip
\textbf{Mumford-Shah Loss}\\
Recent works on unsupervised image segmentation have brought the power of level set methods based on minimization of the Mumford-Shah functional~\cite{mumford-shah-min-func-CVPR1985} in CNNs~\cite{kim2019mumford}. 

The unsupervised segmentation loss is then expressed as:
\begin{equation}\label{eq: mumford-shah-loss}
   \LL_{MS}(x) = \sum_{k=1}^K \sum_{i} |x_i - c_k|^2 \hat{y}_i^{(k)} + \alpha^{(reg)} \sum_{k=1}^K \sum_{i} |\nabla \hat{y}_i^{(k)}|
\end{equation}
where we kept the same notations as before. 

In Eq.~\eqref{eq: mumford-shah-loss}, the first term corresponds to the reconstruction loss, while the regularization term penalizes gradient variations in the resulting segmentation, thus leading to more homogeneous regions. 
} 

\section{Experiments with \MF\ and Analysis}\label{sec: experiments-analysis}

This section intends to evaluate two aspects of our work: first, the contributions of the \MF\ dataset with respect to existing EO datasets, and secondly, the potential of semi-supervised learning on a realistic scenario.

Furthermore, the experiments and results presented in the following will serve as baselines for future works on semi-supervised learning on the \MF\ suite.

{ 
\parindent0pt

\medskip
\textbf{Implementation details}
\medskip

For all the following experiments, networks are trained using Adam optimizer~\cite{diedrik2015adam} with learning rate of $10^{-4}$, during 150 pseudo-epochs, \added{where we observed convergence of models}. Each pseudo-epoch consists of 5000 annotated samples and, in the case of semi-supervised methods, 5000 additional unlabeled samples. One sample is a 512 $\times$ 512 tile randomly chosen from training data, \added{this patch size allows to observe enough context on one image to identify the important elements on it}.
In terms of losses hyperparameters for semi-supervised methods, $\lambda$ in Eq.~\eqref{eq: semisupervised-loss} is set to $\lambda= 2.0$ for reconstruction losses ($\LL_1$ and $\LL_2$) and to $\lambda= 5.0$ for unsupervised segmentation losses ($\LL_{km}$ and $\LL_{MS}$) to get comparable values with respect to cross-entropy. $\alpha^{(rec)}$ and $\alpha^{(reg)}$ in Eq.~\eqref{eq:unsupervised-loss-decomposition} are set to $\alpha^{(rec)} = \alpha^{(reg)} = 1$, \added{for simplicity}.
\added{SegNet and U-Net encoders and decoders are implemented using the architectures defined on the original papers (the last layer of the decoder being adapted to the reconstruction or unsupervised segmentation loss under consideration). }

PyTorch~\cite{pytorch2019} is used for all implementations. Experiments over the \tmf\ dataset are executed using a Nvidia GeForce RTX 2080, while experiments over \MF\ run on a Nvidia Tesla V100 32GB.
} 

\subsection{Limits of Supervised and Semi-supervised Learning on Standard EO datasets}\label{sec: limits-of-EO-datasets}

The ISPRS Vaihingen dataset is a very popular dataset for semantic segmentation in EO data. It has served to benchmark several methods over the years. However, as we mentioned in section~\ref{sec:EO-datasets}, it is constrained to  urban scenes over one city only and thus lacks variety.

The following results show the limits of a dataset such as Vaihingen, and prove the necessity of a more realistic dataset -- that includes various locations and scenes -- as offered by \MF.

\begin{figure}[!htbp]
   \centering
   \includegraphics[width=0.6\linewidth]{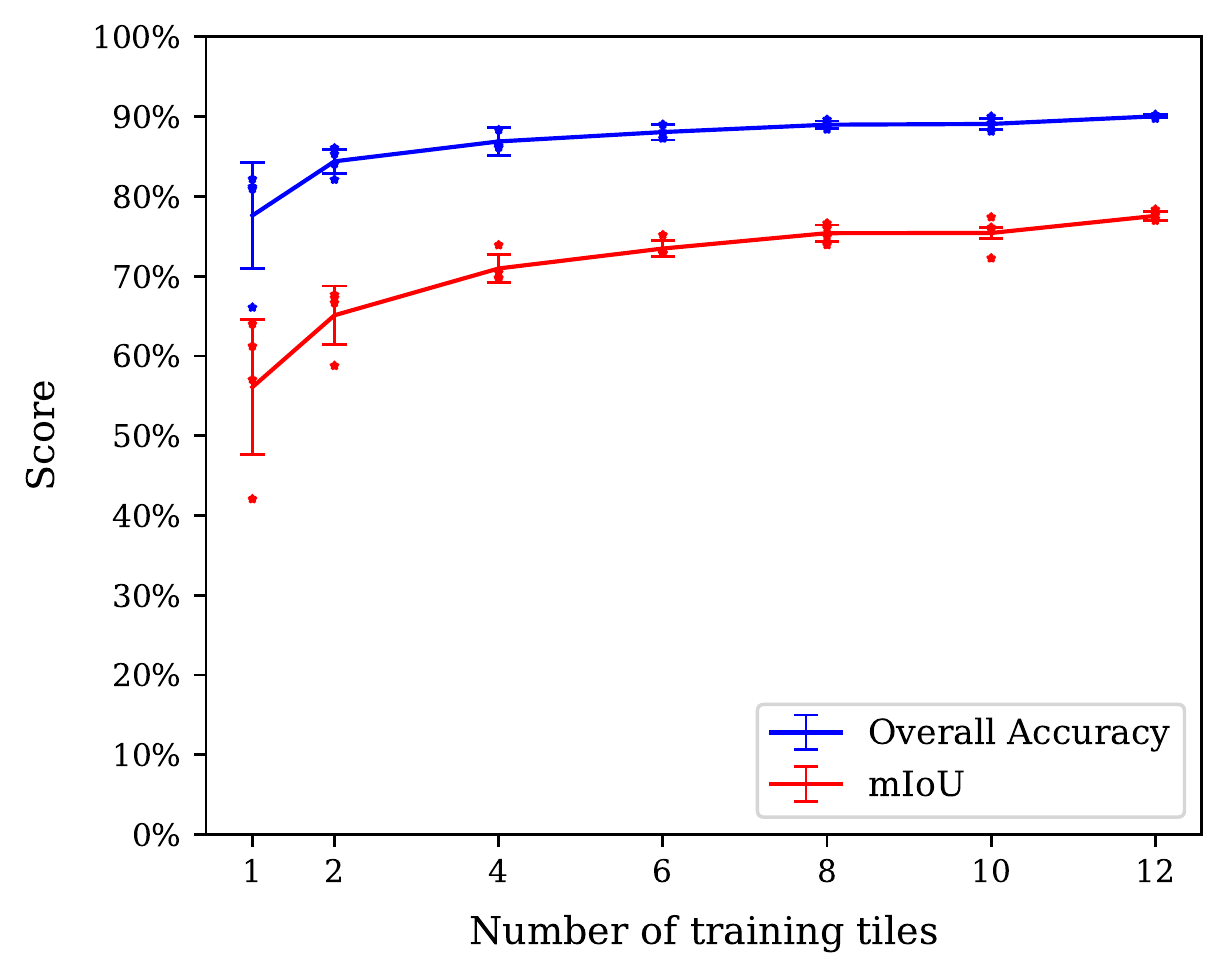}
   \caption{
   Influence of the training set size (number of tiles) on the network performances, in terms of overall accuracy and mean Intersection over union (mIoU). The curves show the mean and the standard deviation for each score and $\star$ shows raw results.}
   \label{fig:results-robustness-vaihingen}

\end{figure}

In our first experiment, we aim to test the sensitivity of
a classic supervised learning framework to the amount of available training data. We train a SegNet model, which has already shown remarkable results on this dataset~\cite{audebert-18isprsj-beyond-RGB}. The experiment consists in reducing the amount of
annotated images used for training, from 12 tiles to only one,
while the validation set remains unchanged (4 tiles). We repeat the experiment four times to get more  statistically significant curves. Results are shown in Figure~\ref{fig:results-robustness-vaihingen}.

The outcomes of this experiment are somehow surprising. When reducing the number of training tiles from 12 to 1 (only 8\% of original data!), we report a decrease of \emph{only} 12\% of overall accuracy (from 90\% to 78\%) and 21\% of mIoU (from 77\% to 56\%), i.e. much less than one would expect. Indeed, we supposed that reducing the number of training tiles would seriously impact the performance of the network. One possible reason is that all the images in the Vaihingen dataset are alike, thus, to generalize on them is a relatively easy task. However, one can note that training with more data is nevertheless preferable in terms of reliability: the variance increases as the number of tiles decreases.

To investigate this explanation, we apply our tool for appearance coverage assessment presented in section~\ref{sec: statistical-analysis} to the union of both datasets, \tmf\ and Vaihingen. To get a fair comparison, images from the Vaihingen dataset were downsampled to the \tmf\ resolution (from 9~cm/px to 50~cm/px) before being encoded by the CNN. 
 
\begin{figure}[!htbp]
   \centering
   \includegraphics[width=0.5\linewidth]{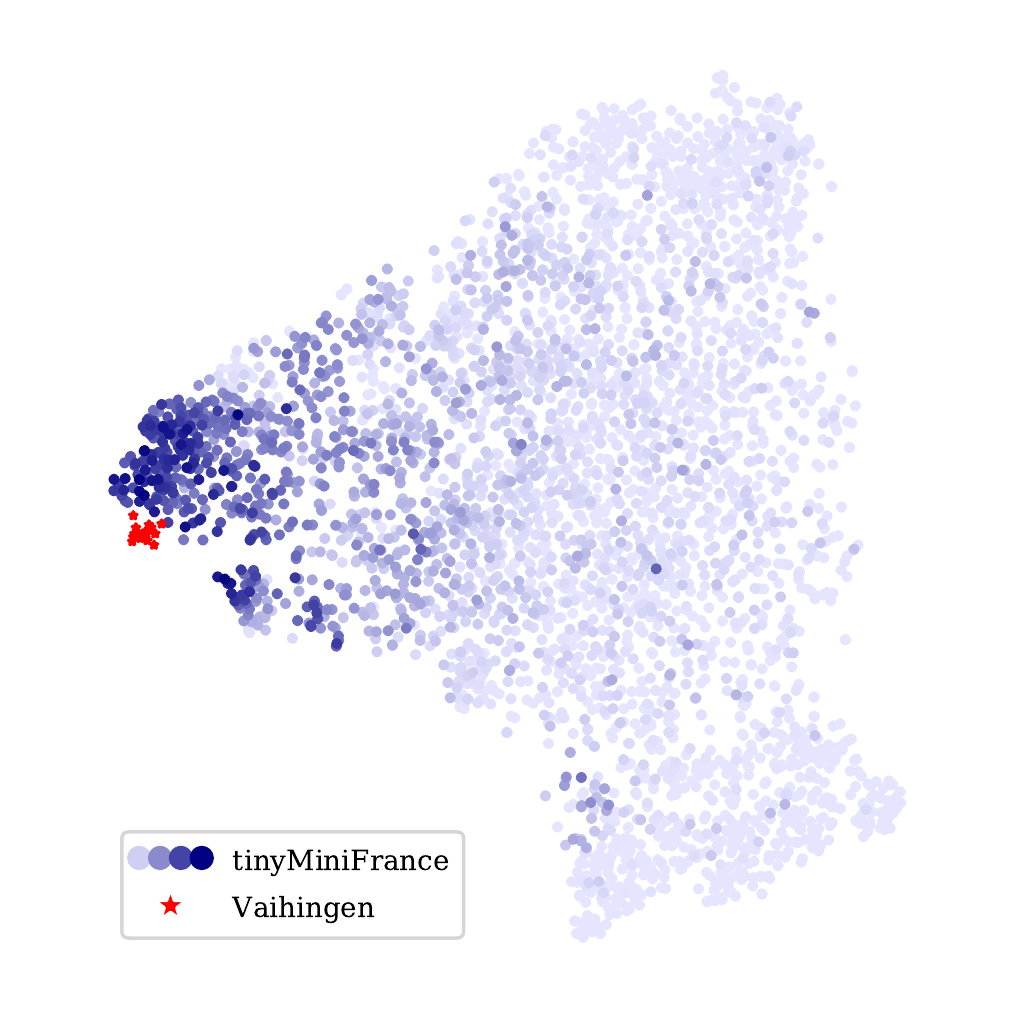}
   \caption{
   2D representation of images by t-SNE, applied to \tmf\ and Vaihingen together, after ResNet34 encoding. Points from \tmf\ are colored according to the proportion occupied by the class \emph{urban fabric}.}
   \label{fig:tsne-tmfrance-vaihingen}
\end{figure}

Due to the stochastic nature of the t-SNE algorithm, it is important to note that subsequent runs can lead to different embeddings. However since \tmf\ is much larger than the 16 Vaihingen tiles, the projection is not noticeably perturbed up to rotation and reflection. We chose the embedding which resulted in the same visualization as Section~\ref{sec: statistical-analysis}.
Results are shown in Figure~\ref{fig:tsne-tmfrance-vaihingen}. Red stars ({\color{red} $\star$}) represent Vaihingen tiles, while shading blue circles ({\color{lightcolor} $\bullet$}$\cdots${\color{darkcolor} $\bullet$}) are \tmf\ tiles, colored according to the proportion occupied by \emph{urban fabric} (as in Figure~\ref{fig:tmfrance-resnet-class-presence-proportions}, darker points contain a higher proportion of urban pixels). We consider specifically the \emph{urban fabric} class since it is the most related to the Vaihingen urban dataset.

The previous visualization is insightful. On the one hand, we realize how small the Vaihingen dataset is compared to \tmf\ (and even more to the entire \MF), in terms of number of available tiles. On the other hand, the t-SNE algorithm places Vaihingen as a very small cluster next to the urban scenes of \tmf, which means that: ($i$) Vaihingen is slightly different from \tmf\ (may be due to the IRRG encoding vs. RGB); ($ii$) at the same time, it remains visually close to the urban images from \tmf\ (confirming our choice to consider here the \emph{urban fabric} class); and ($iii$) the wide surface covered by \tmf\ on the 2D appearance projection space w.r.t. Vaihingen shows that our dataset presents a much larger variety of appearances in terms of urban scenes; furthermore, these urban scenes form only a small part of the appearance space, thus proving the very wide diversity of \tmf, and to a larger extent of \MF.

\subsection{Defining the Labeled/Unlabeled/Test split for \MF} \label{sec: results-split-definition}

\begin{figure}[!htbp]
   \centering
   \includegraphics[width=\linewidth]{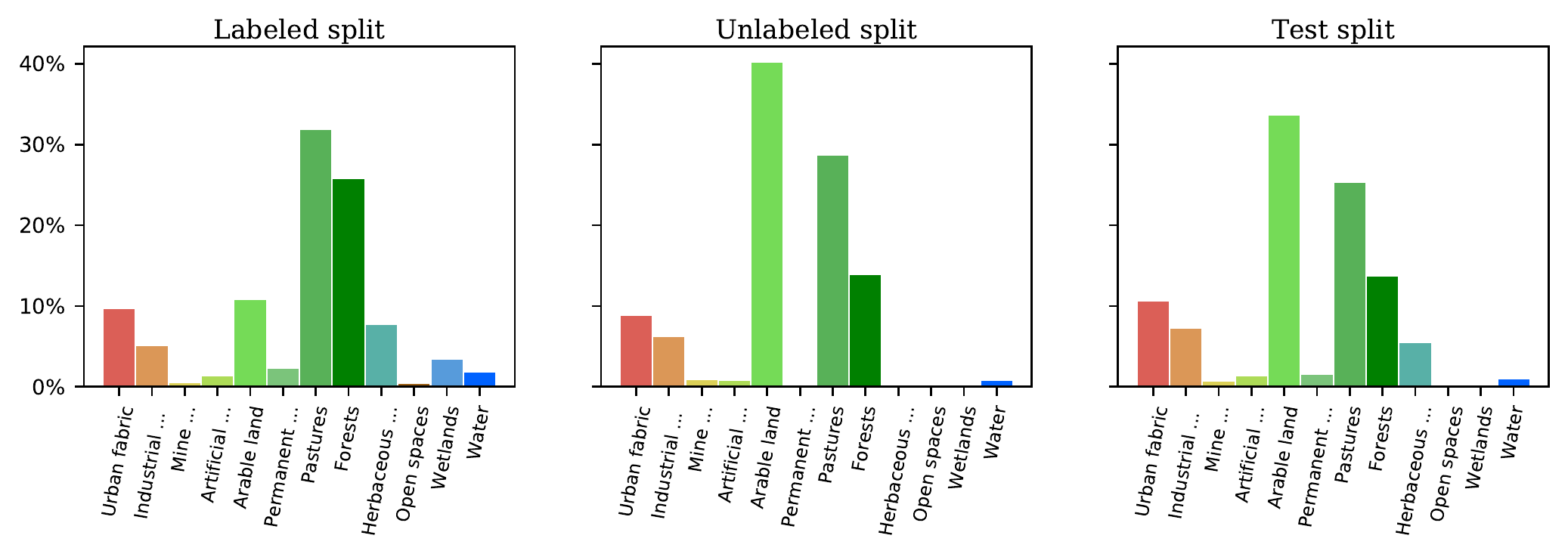}
   \vspace{-1em}
   \caption{Class distributions aggregated by split as defined in Table~\ref{tab:minifrance-cities-split}}
   \label{fig:tmfrance-class_distributions_traintest}
   
\end{figure}

Using all the tools and information presented in section~\ref{sec: statistical-analysis}, \MF\ has been carefully designed to satisfy the conditions of appearance and class representativeness. Indeed, the split proposed in Table~\ref{tab:minifrance-cities-split} allows to represent all the classes with a proper distribution, as shown in the histograms of Figure~\ref{fig:tmfrance-class_distributions_traintest}. Hence, all classes present in the test set have training examples in the labeled split.

\begin{figure}[!htb]
   \centering
   \includegraphics[width=.9\linewidth]{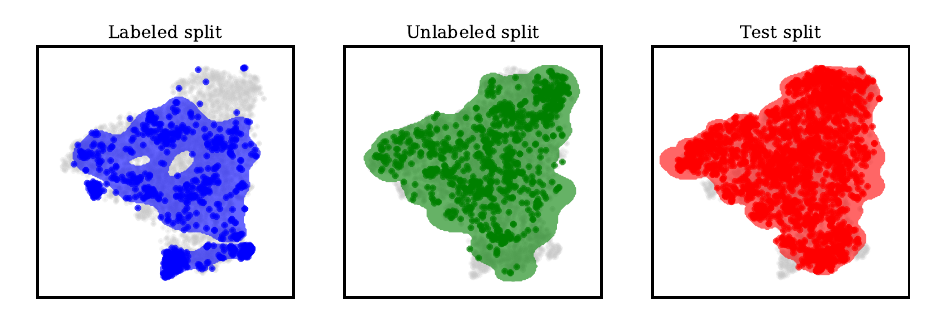}
   \vspace{-1em}
   \caption{Appearance representation aggregated by split as defined in Table~\ref{tab:minifrance-cities-split}}
   \label{fig:tmfrance-appearance_traintest}
   
\end{figure}

On the appearance side as shown in Figure~\ref{fig:tmfrance-appearance_traintest}, even if labeled cities do not cover the whole appearance space of test images, the union of labeled and unlabeled does. This should ensure that all appearances are seen in a semi-supervised setup. Moreover, in terms of IoU scores of appearance shown in Figure~\ref{fig:tmfrance-iou-scores}, the labeled split comprises one region with a high score (Nantes) and one with a low score (Nice) which should help to learn different appearances of classes. In addition, in the unlabeled split most of the cities have a high score with respect to the test set, so they should help to extract the implicit information from images. 

\begin{table}[!htbp]
   \begin{center}
      \caption{IoU and IoT scores between training data -- labeled and unlabeled -- and test data. Scores are presented in numerical form as well as color code for comparison with Figure~\ref{fig:tmfrance-iou-scores}.} \label{tab:iou-iot-labeled-unlabeled-test}
      \begin{tabular}{ccc} \toprule\noalign{\smallskip}
      $S_1 \text{ - } S_2$ & $IoU(S_1, S_2)$ & $IoT(S_1, S_2)$ \\\noalign{\smallskip}\midrule[.8\heavyrulewidth]\noalign{\smallskip}
      \emph{Labeled - Test} & 0.63 \mybox{cmap063}{cmap063} & 0.64 \mybox{cmap064}{cmap064}  \\[2pt]
      \emph{Unlabeled - Test} & 0.87 \mybox{cmap087}{cmap087}& 0.93 \mybox{cmap093}{cmap093} \\[2pt]
      \bottomrule
      \end{tabular}
   \end{center}   
\end{table}

Table~\ref{tab:iou-iot-labeled-unlabeled-test} presents the IoU and IoT scores between the surfaces in Figure~\ref{fig:tmfrance-appearance_traintest} and confirms the information above. Thus, even if the labeled training split contains all classes of the test split, 64\% of IoT means it is far from covering all the possible appearances. However, with 93\% of IoT score with the test area, the unlabeled training split offers wider information about the visual features present in the \MF\ dataset that should be exploited to achieve good quality classification and generalization.

In brief, \MF\ is a very challenging dataset for semantic segmentation that promotes new solutions in a semi-supervised manner as some appearances can only be extracted from the unlabeled data. 
However, train and test adequacy was carefully controlled to avoid domain shift and such disentangle semi-supervised learning from domain adaptation and transfer learning.

\subsection{Supervised and Semi-supervised Learning on \MF}

The purpose of this section is to show that we can benefit from semi-supervised learning -- using unlabeled data during the learning process -- to achieve better results and generalization than vanilla supervised approaches. 

To this end, we perform experiments to compare a semi-supervised setting with an equivalent supervised approach, using different backbone architectures. First, we train supervised networks (SegNet and U-Net) in a classical way, using the cross-entropy loss, over the labeled training split of \tmf. Secondly, we train a BerundaNet-late architecture (with SegNet and U-Net backbone) over \tmf\ -- using both, labeled and unlabeled data --, which is the equivalent semi-supervised strategy. We train BerundaNet-late with a reconstruction task ($\LL_1$ as auxiliary loss) and with an unsupervised segmentation task ($\LL_{km}$ as auxiliary loss) and show that in both cases, semi-supervised learning can improve the results obtained by the supervised network. 

Results of these experiments are summarized in Table~\ref{tab: tmf-supervised-vs-semisupervised}. The \emph{oracle} corresponds to the hypothetical case where annotations are available for all training cities (i.e, we can access the ground-truth for all the images of the 8 regions in the training split) during the training phase. The oracle results might be seen as an upper bound for semi-supervised learning strategies and they are brought out here just for comparison and not as a result of this work. 

\begin{table}[H]
   \begin{center}
      \caption{Supervised vs. Semi-supervised experiments over \tmf\ using different backbone architectures. We refer to the hypothetical case where annotations are available for all 8 training regions as \emph{oracle}. Semi-supervised denotes results for BerundaNet-late with the corresponding backbone.
      \label{tab: tmf-supervised-vs-semisupervised}}

      \begin{tabular}{c|aa|cc|cccc} \toprule
      & \multicolumn{2}{a|}{Oracle} &\multicolumn{2}{c|}{Supervised} & \multicolumn{4}{c}{Semi-supervised (BerundaNet-late)}\\
      Backbone & \multicolumn{2}{a|}{$\LL_{ce}$}  & \multicolumn{2}{c|}{$\LL_{ce}$} & \multicolumn{2}{c}{$\LL_{ce} + \lambda \mathcal{L}_1$}& \multicolumn{2}{c}{$\LL_{ce} + \lambda \mathcal{L}_{km}$}\\
      & OA & mIoU & OA & mIoU & OA & mIoU & OA & mIoU \\ \midrule \midrule
      SegNet & 59.06 & 23.95 & 36.76 & 14.03 &  \bf{45.52} & 14.43 &  42.26 & \bf{15.75} \\
      U-Net  & 57.71 & 25.25 & 46.30 & 18.18& \bf{47.90} & \bf{18.70} & 46.92 & 18.26 \\ 
      \bottomrule
      
      \end{tabular}

   \end{center}
\end{table}

\begin{figure}[!htbp]

   \begin{center}
      \setlength{\tabcolsep}{1.6pt}
      \begin{tabular}{c@{\hspace{6pt}}|@{\hspace{6pt}}cc@{\hspace{6pt}}|@{\hspace{6pt}}ccc}
         \includegraphics[width=.15\linewidth]{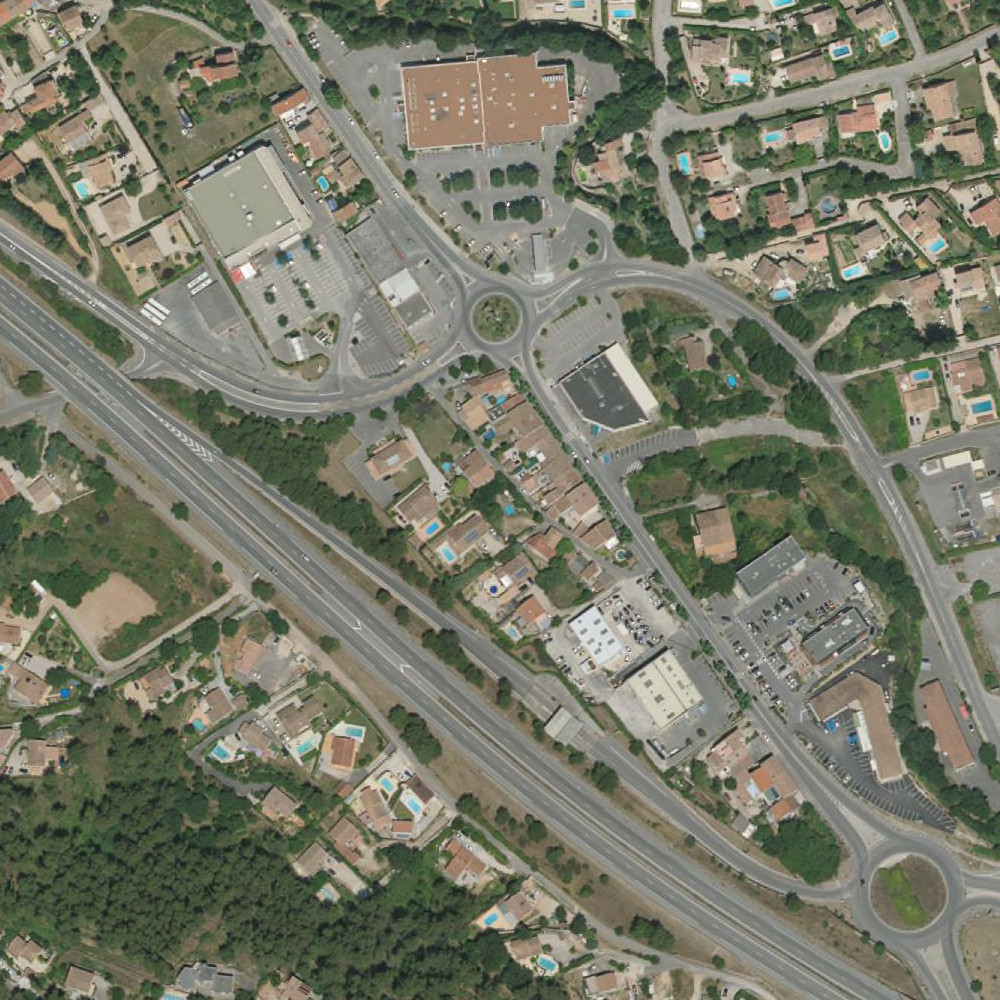} & \includegraphics[width=.15\linewidth]{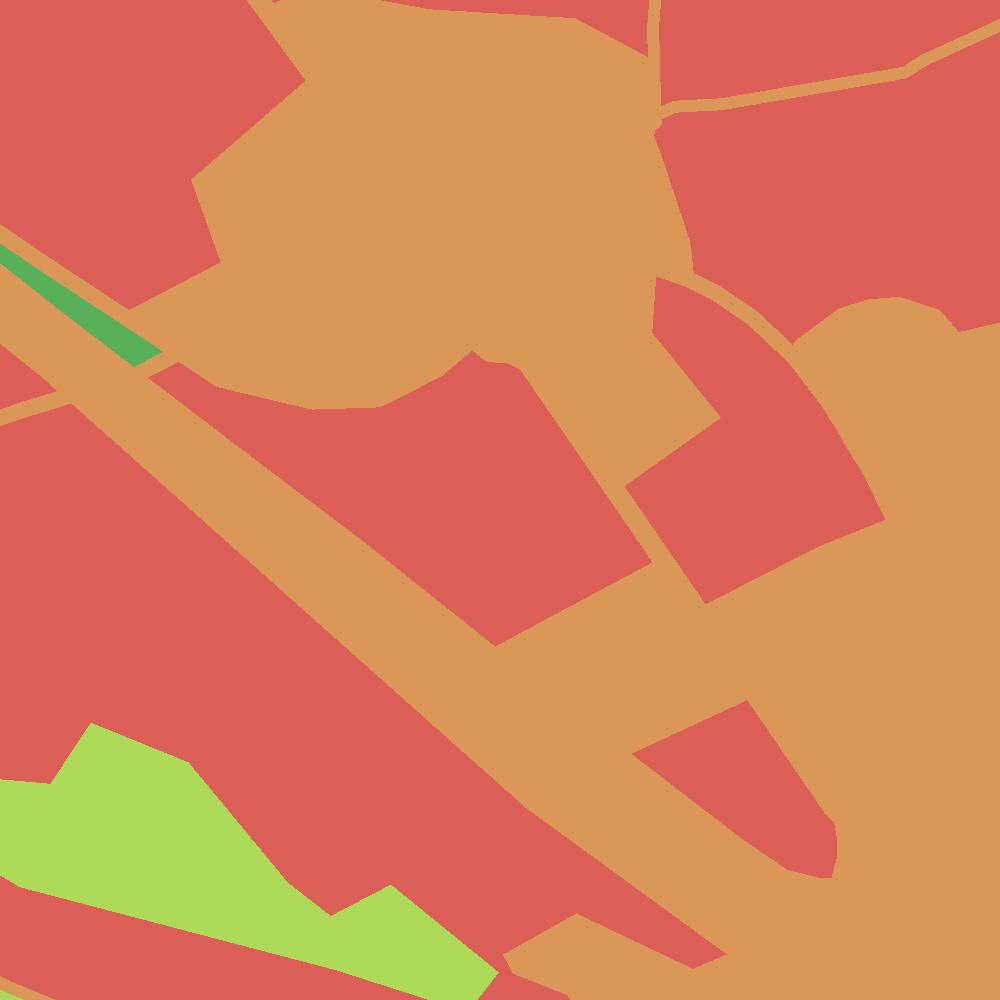} & \includegraphics[width=.15\linewidth]{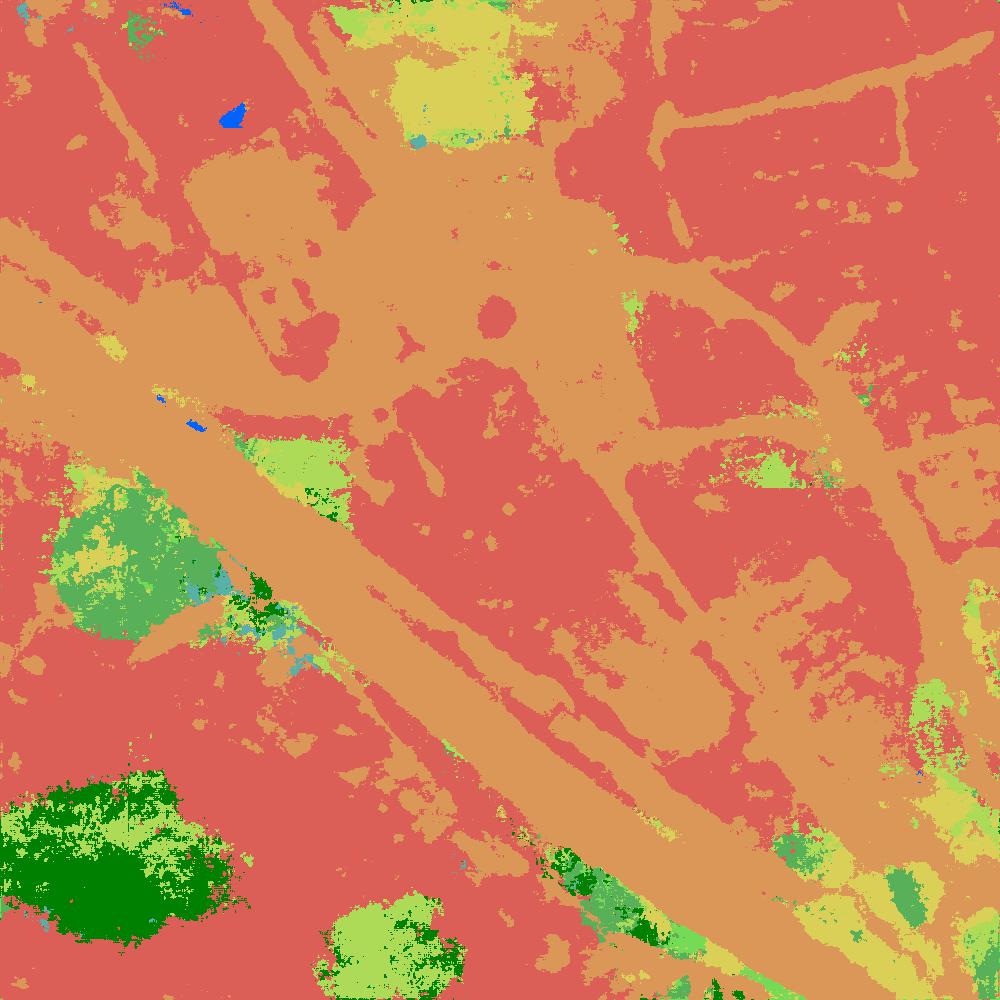} & \includegraphics[width=.15\linewidth]{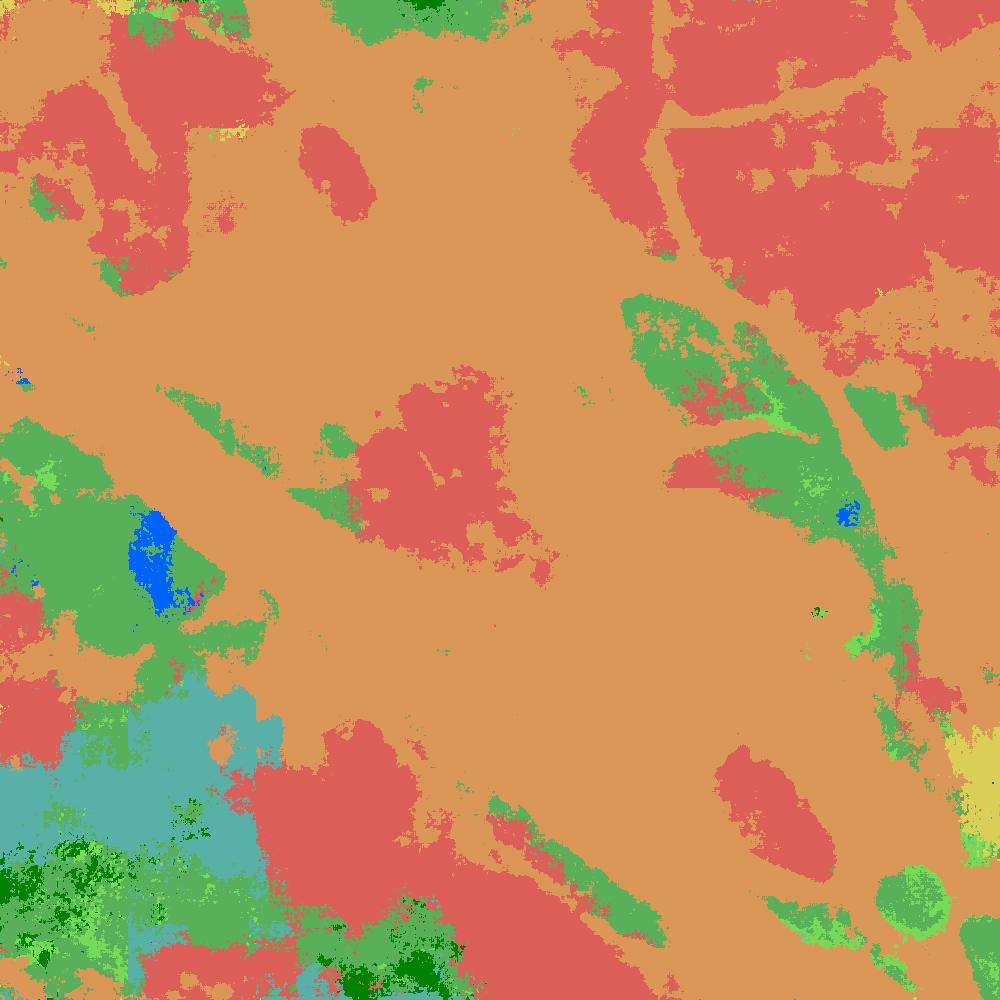} & \includegraphics[width=.15\linewidth]{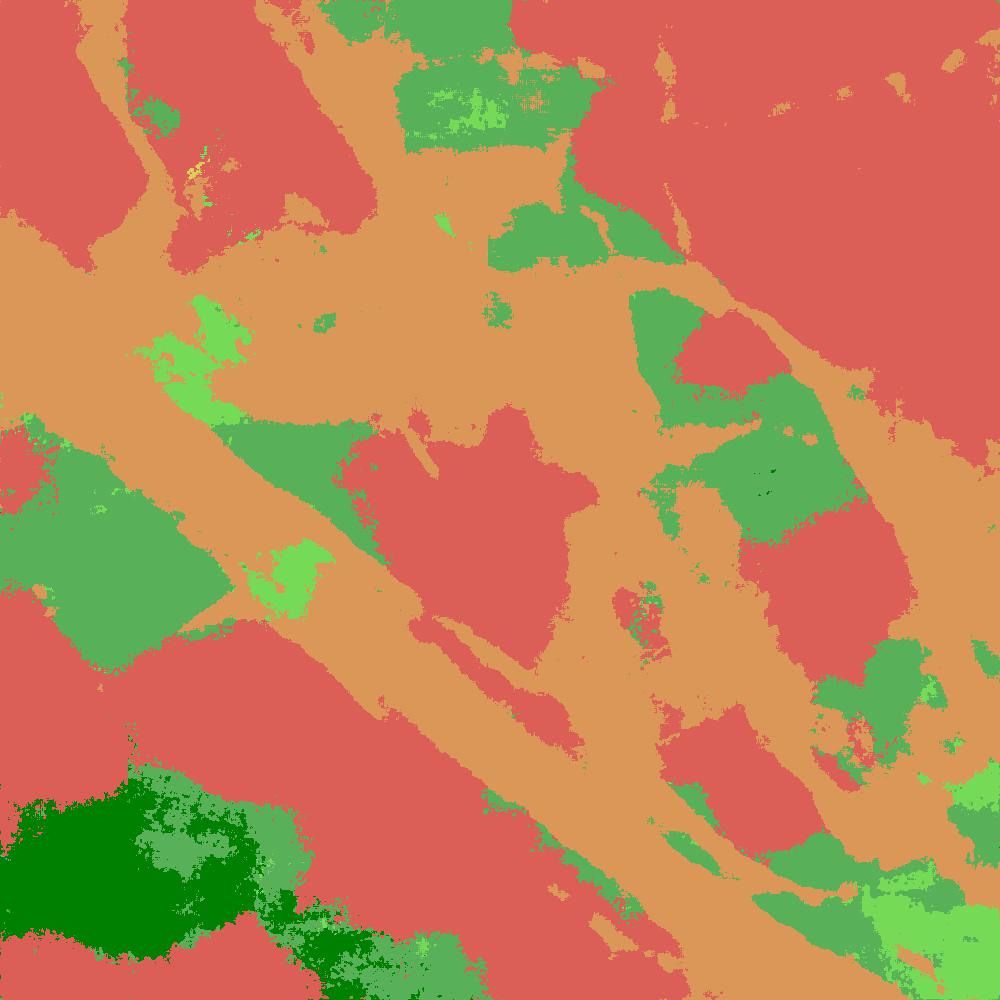} & \includegraphics[width=.15\linewidth]{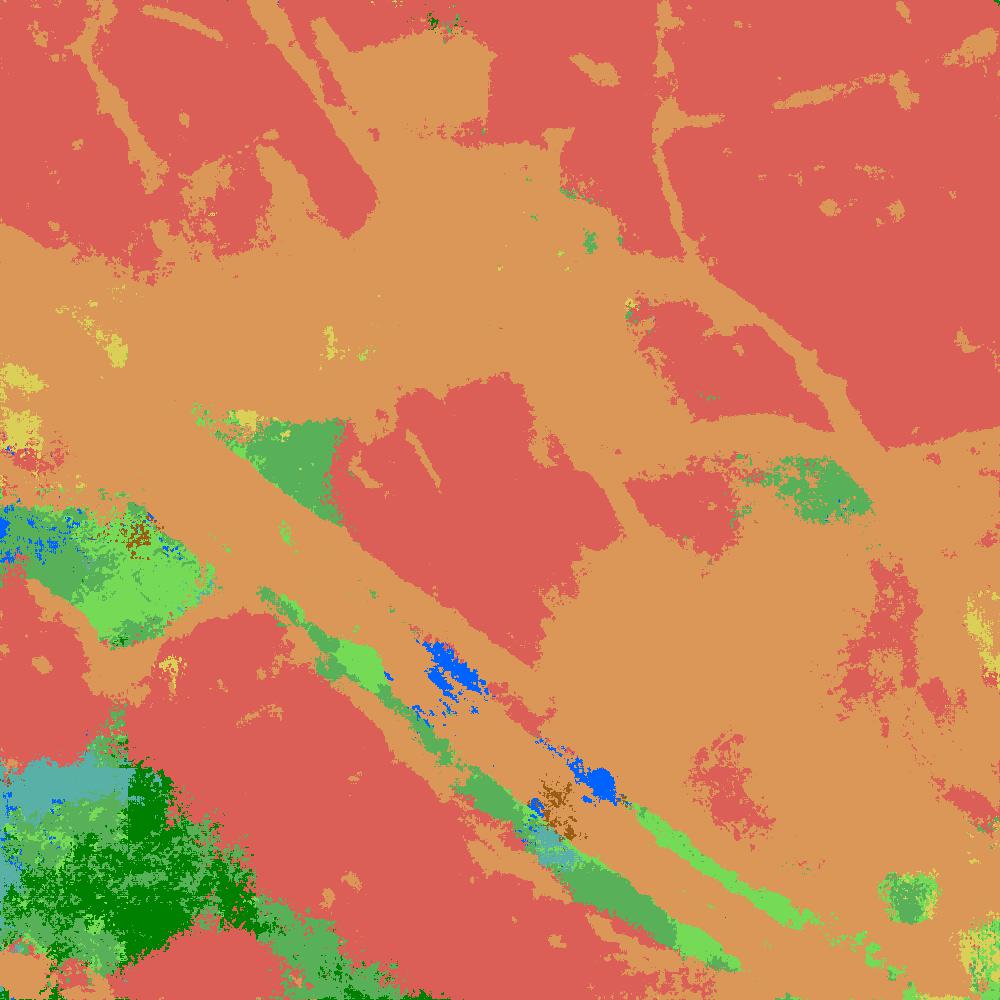}\\
         
         \includegraphics[width=.15\linewidth]{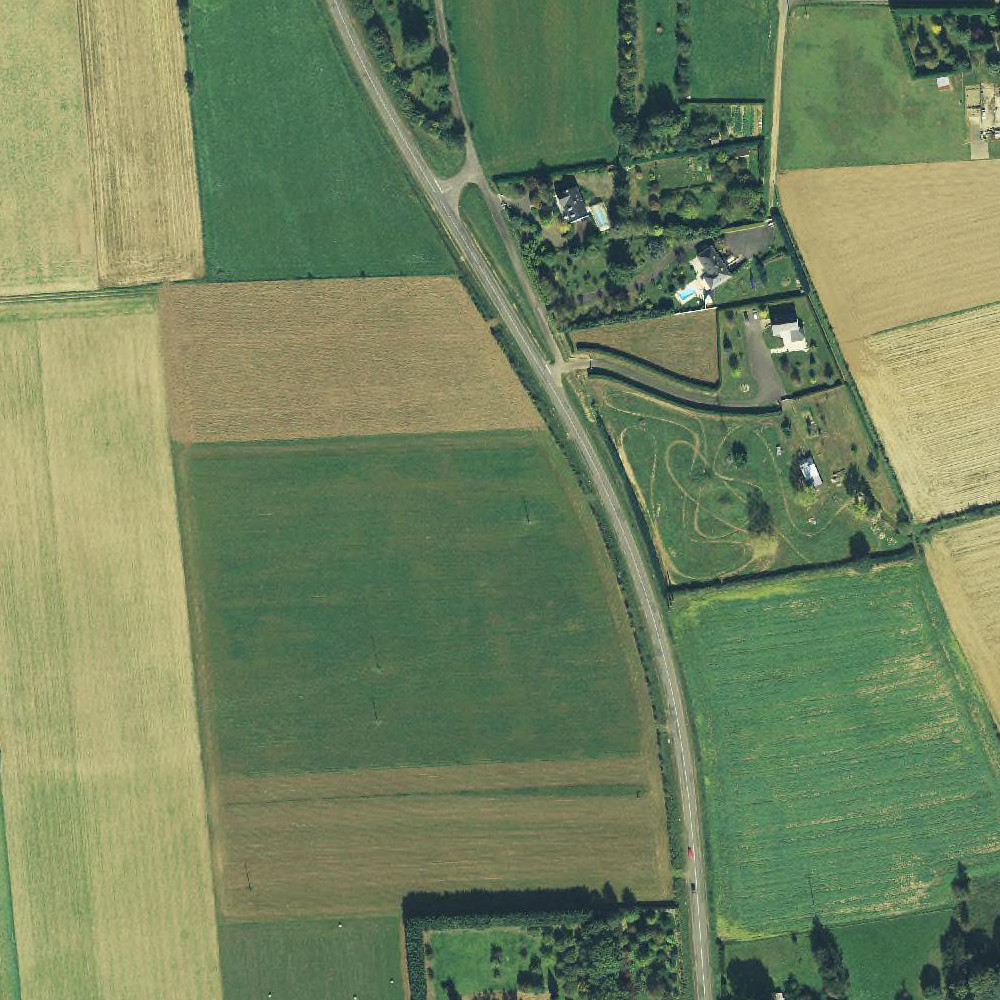} & \includegraphics[width=.15\linewidth]{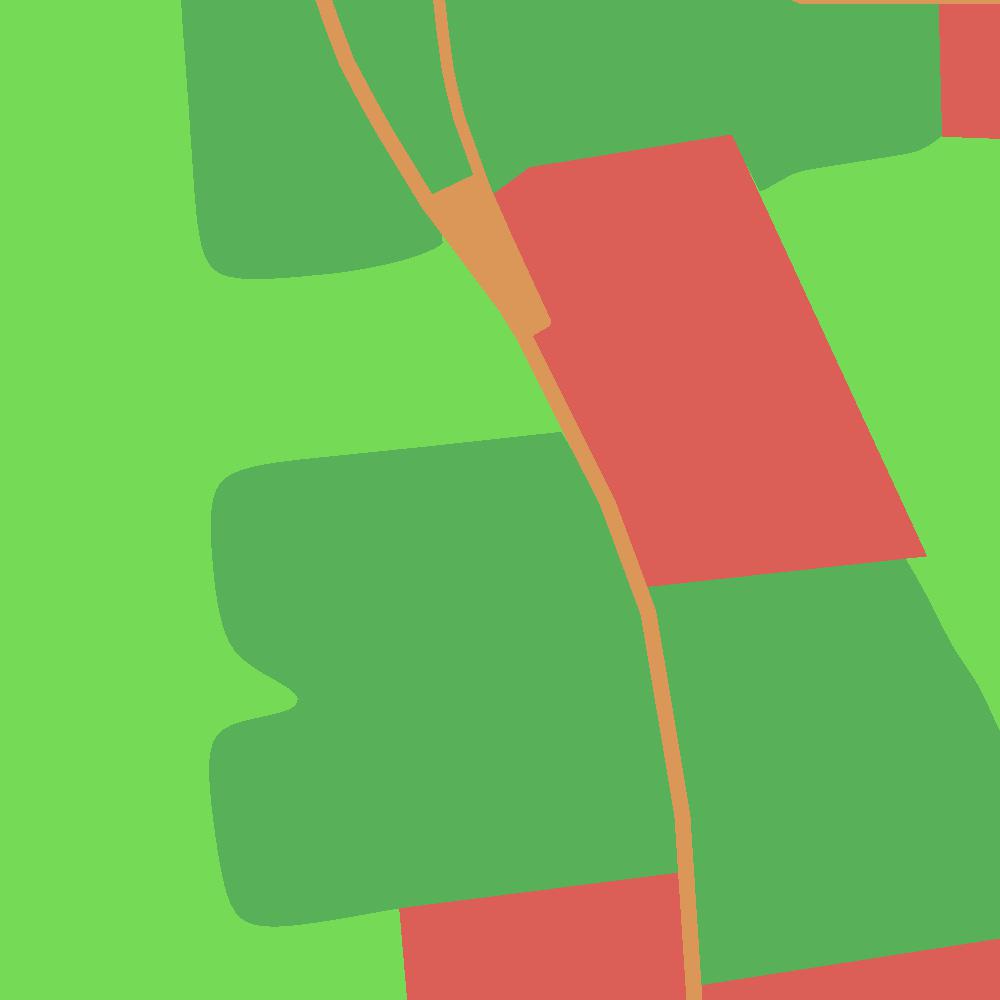} & \includegraphics[width=.15\linewidth]{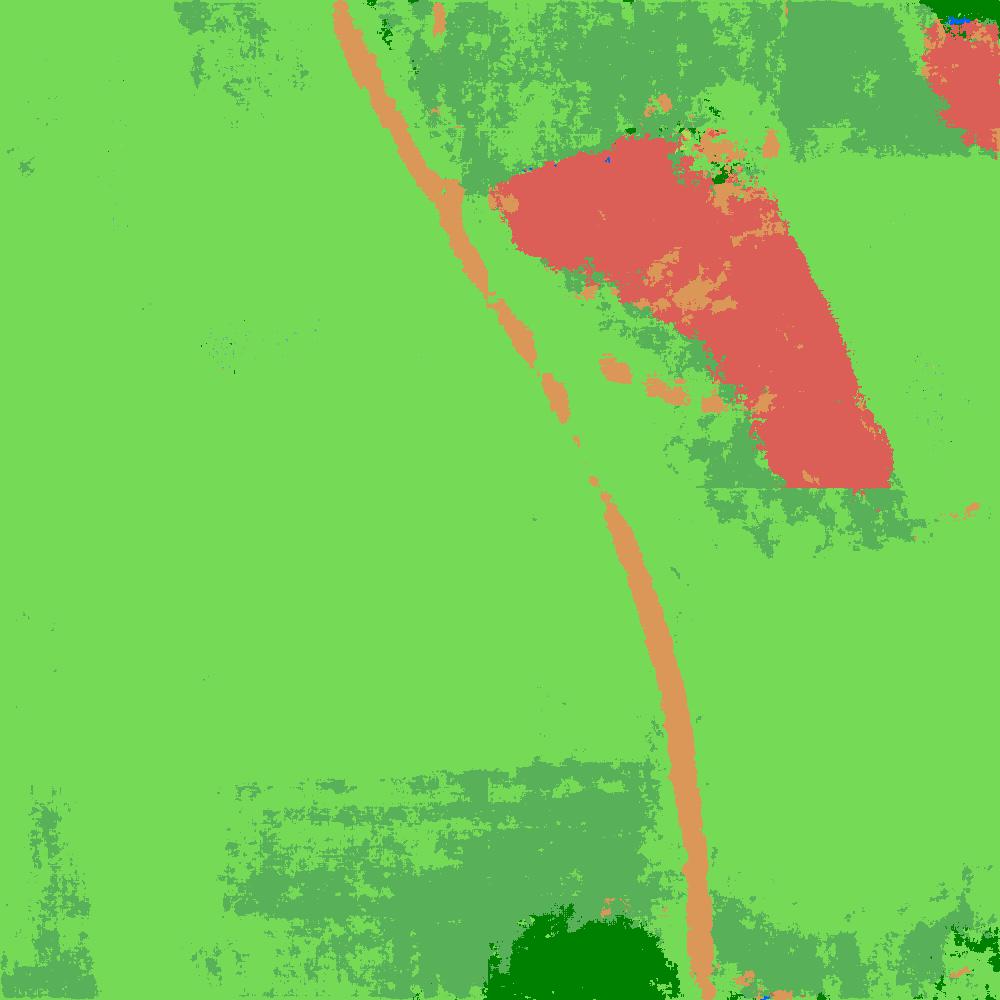} & \includegraphics[width=.15\linewidth]{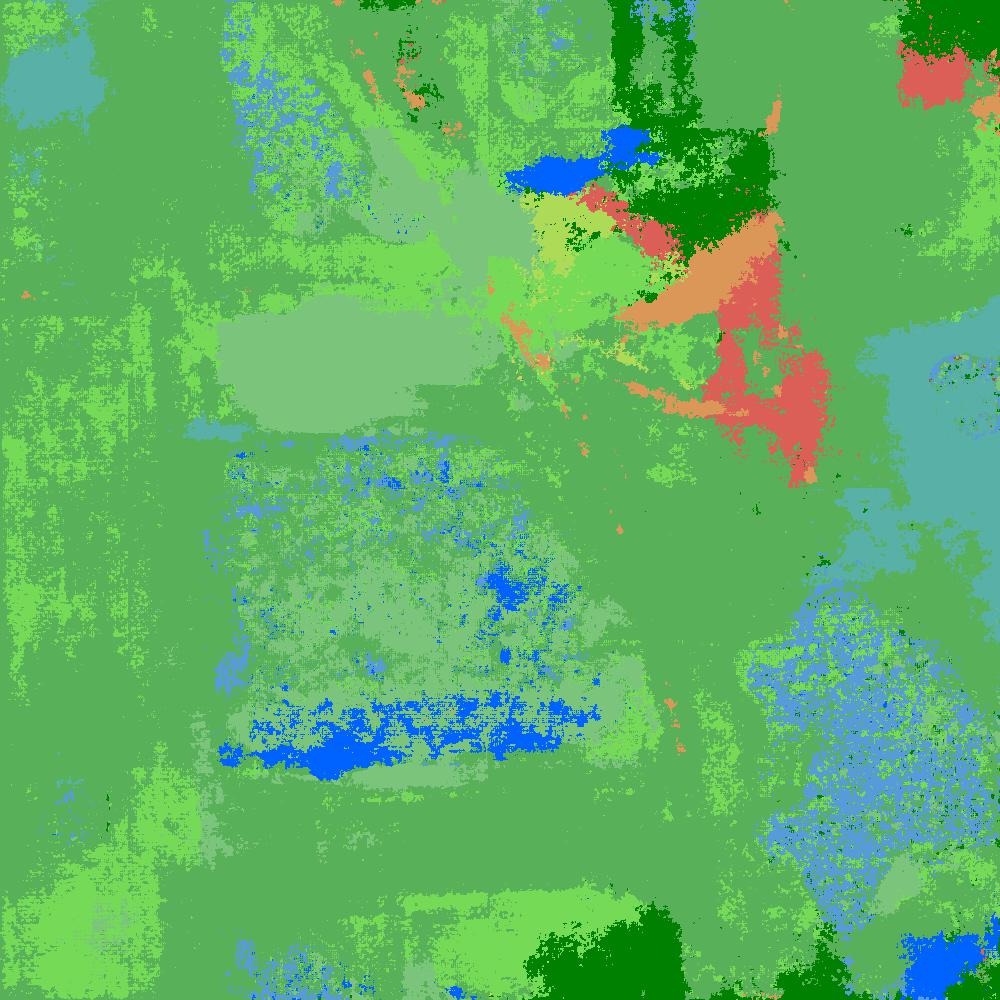} & \includegraphics[width=.15\linewidth]{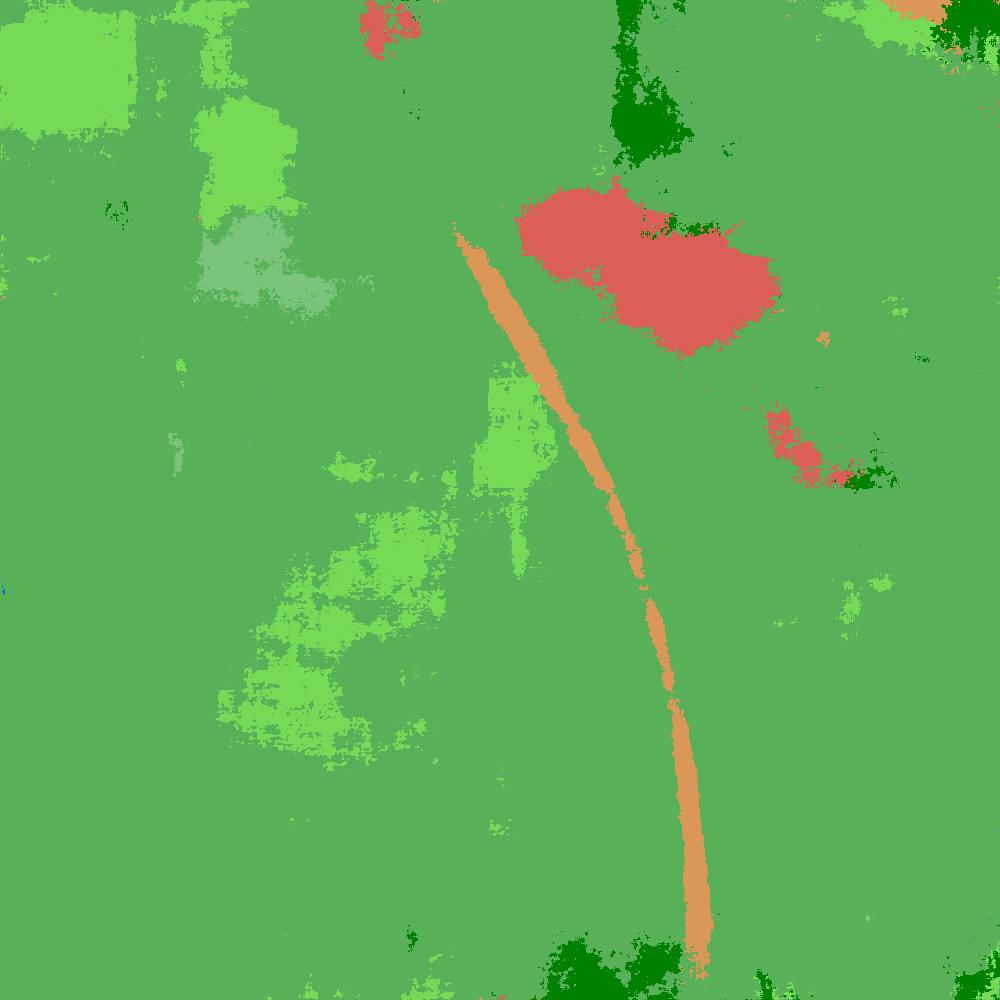} & \includegraphics[width=.15\linewidth]{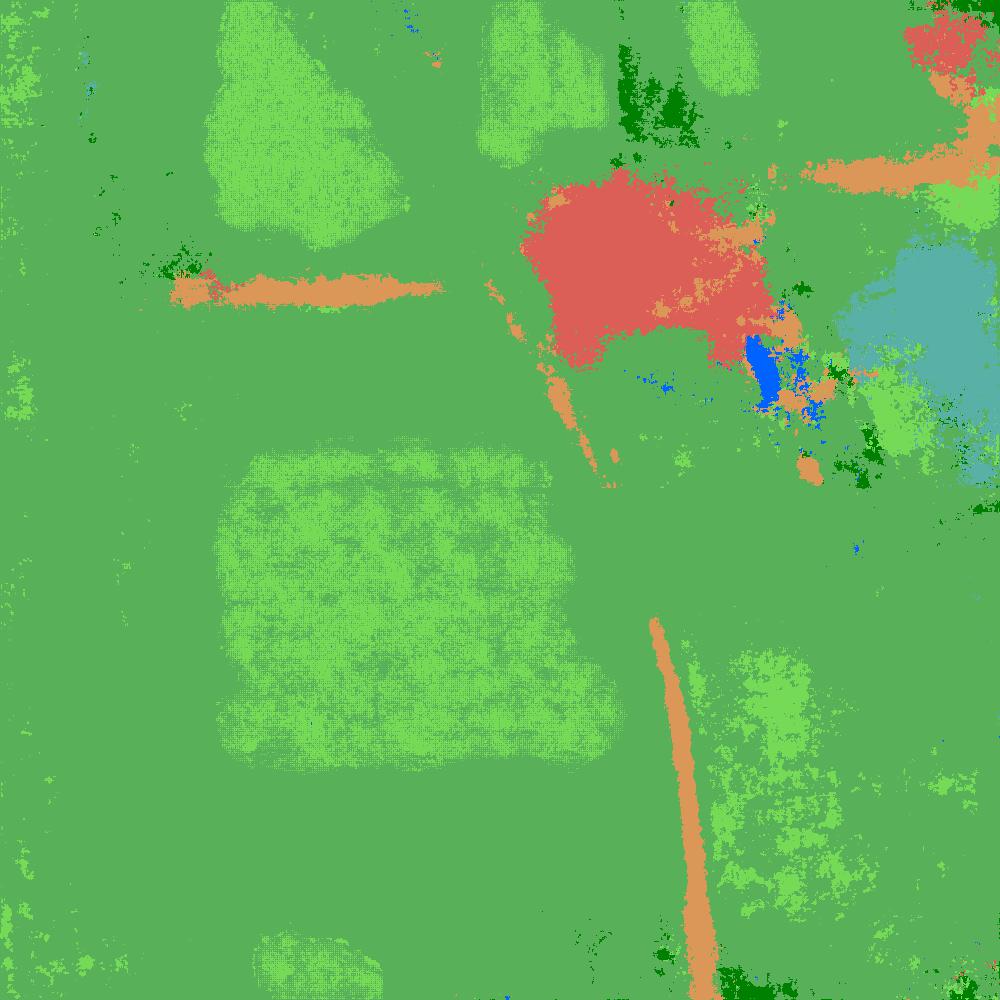}\\

         \includegraphics[width=.15\linewidth]{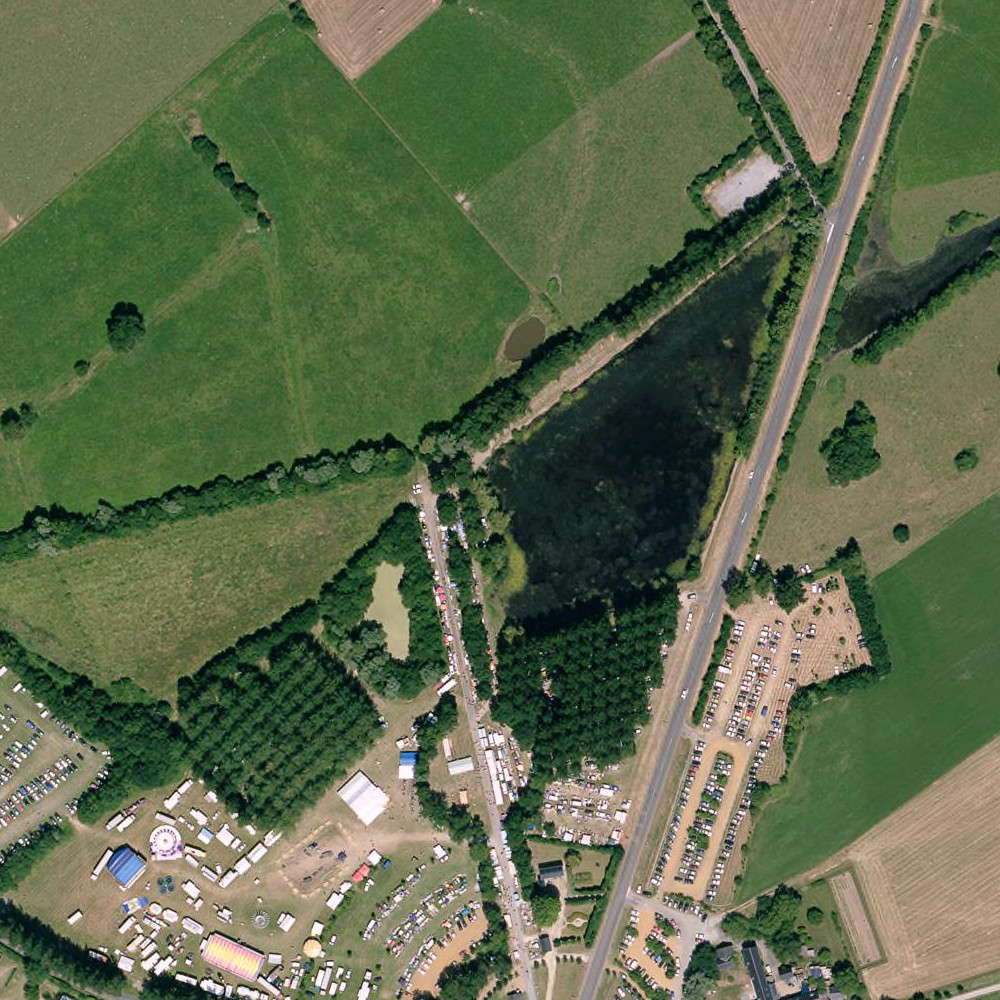} & \includegraphics[width=.15\linewidth]{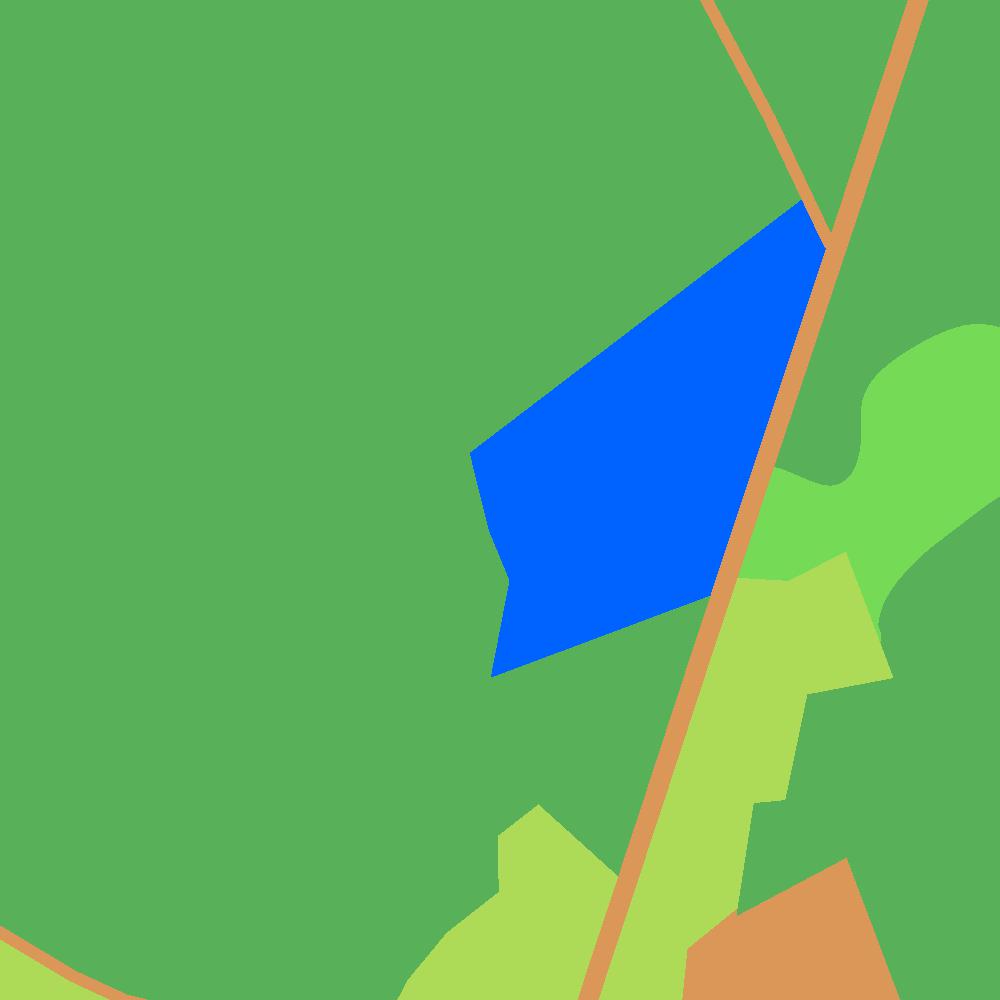} & \includegraphics[width=.15\linewidth]{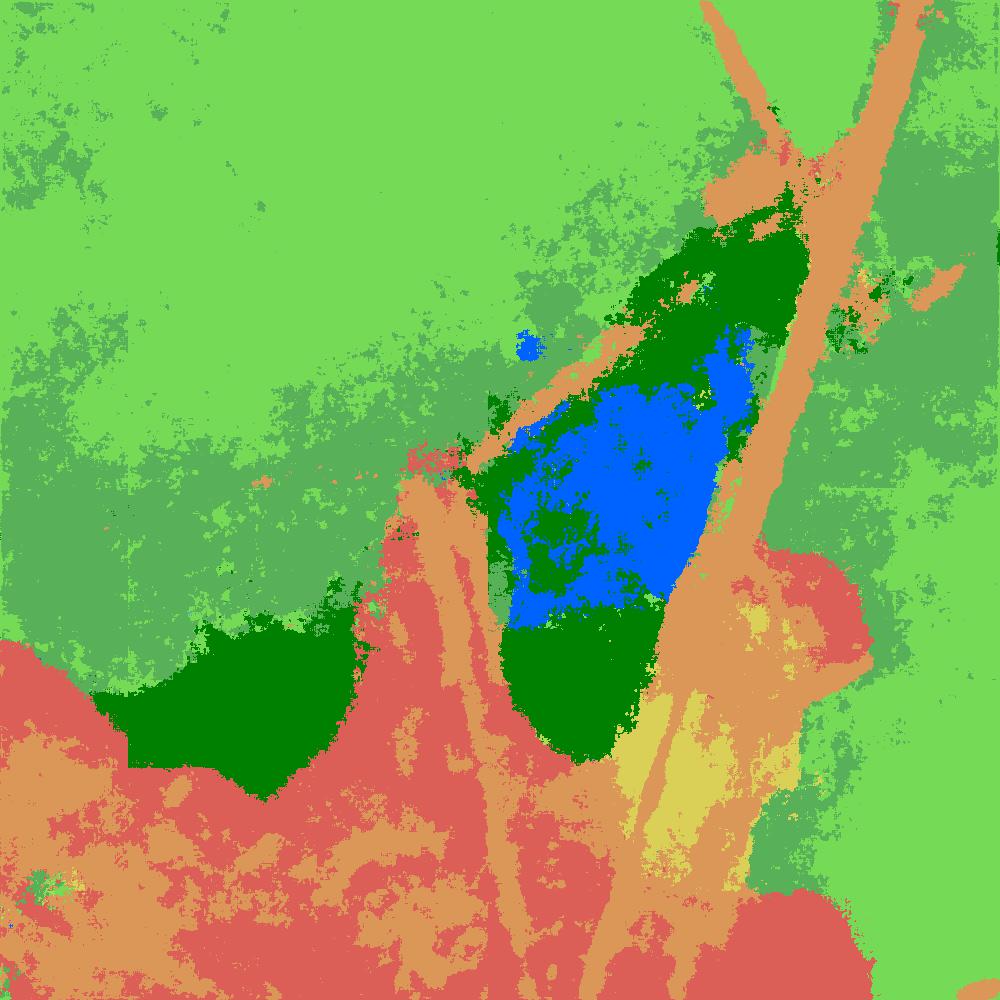} & \includegraphics[width=.15\linewidth]{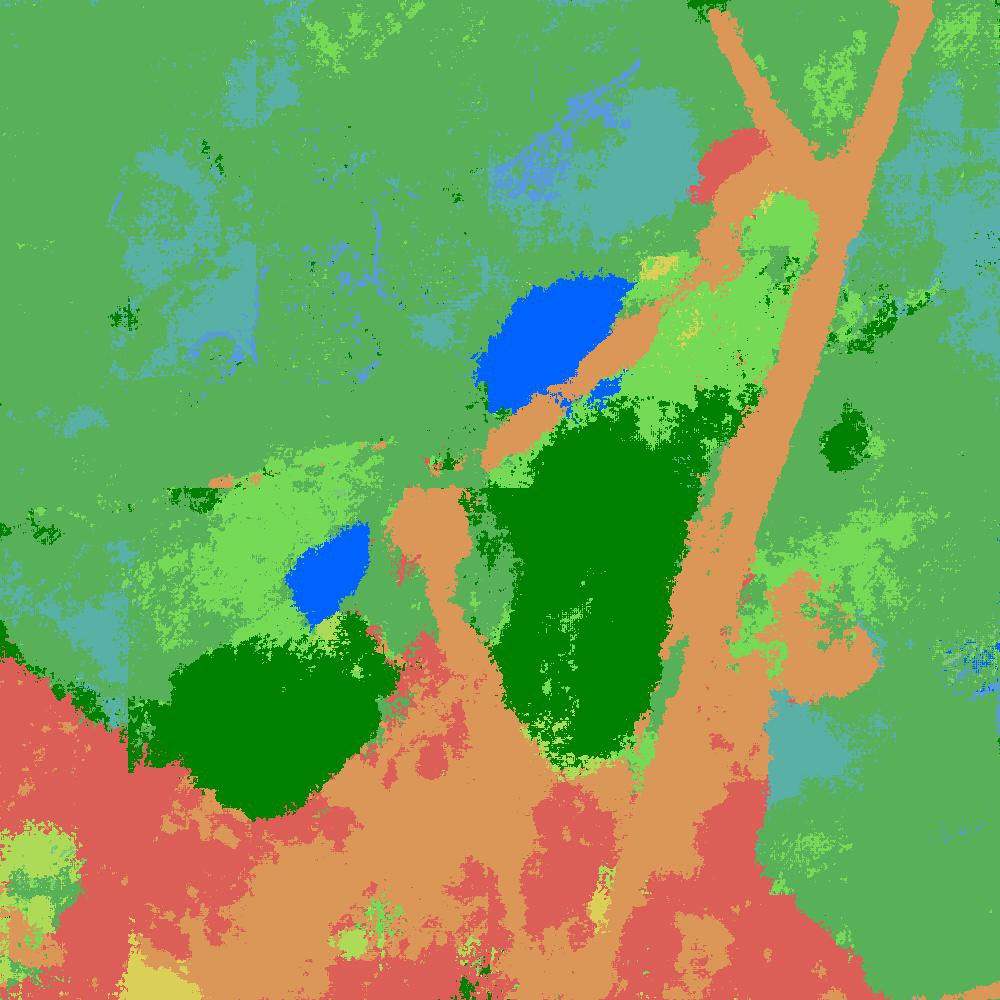} & \includegraphics[width=.15\linewidth]{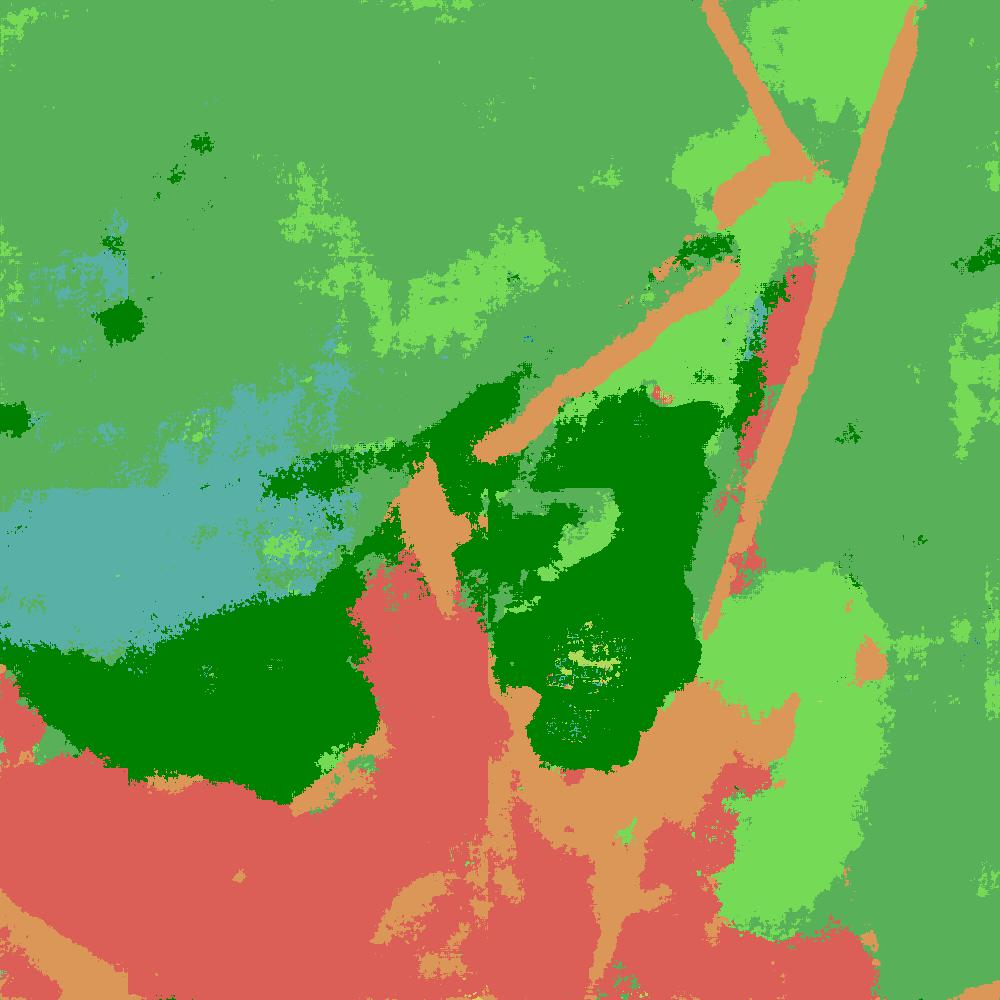} & \includegraphics[width=.15\linewidth]{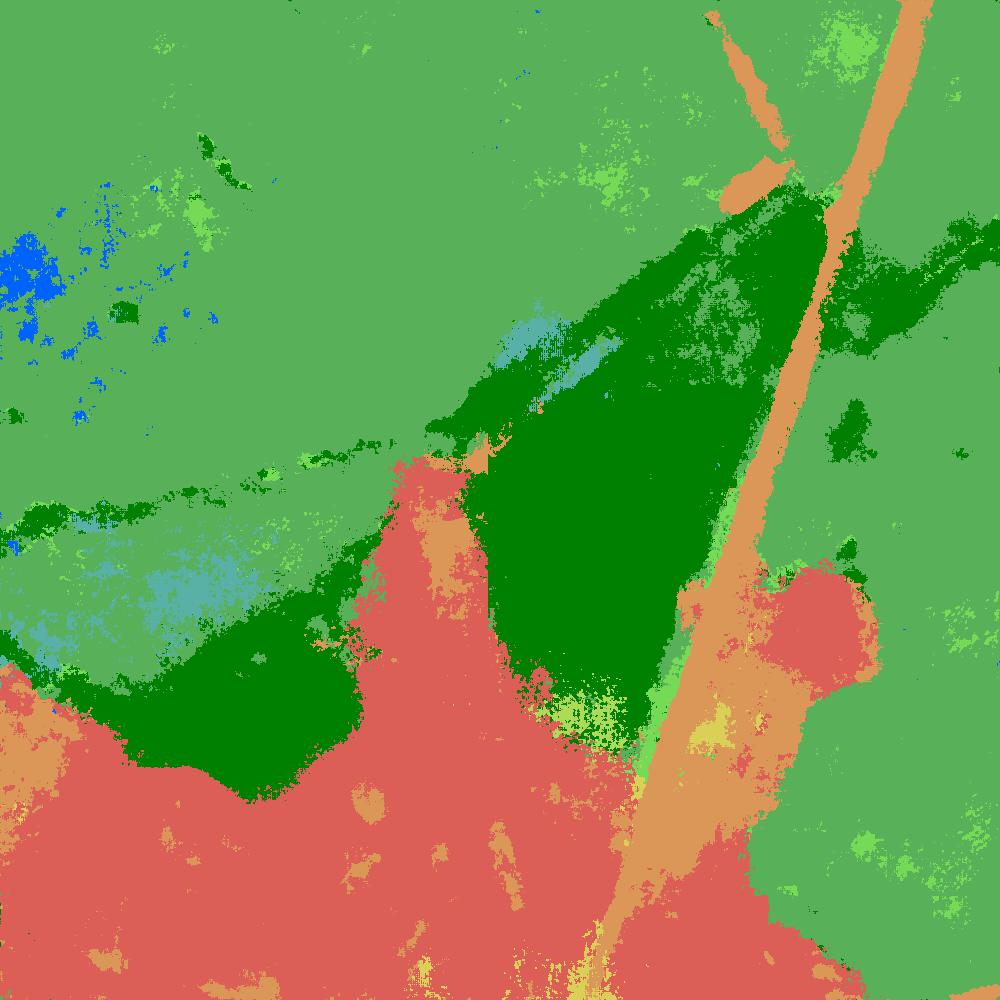}\\

         \includegraphics[width=.15\linewidth]{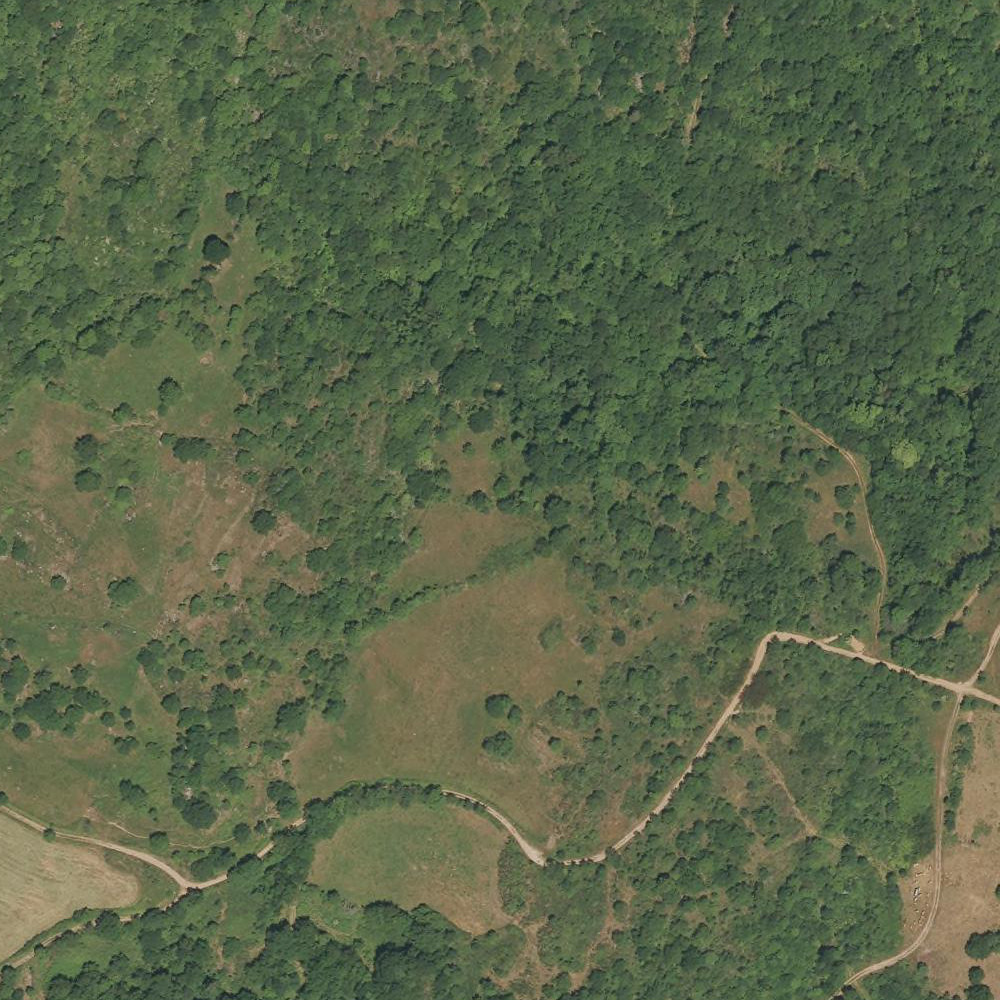} & \includegraphics[width=.15\linewidth]{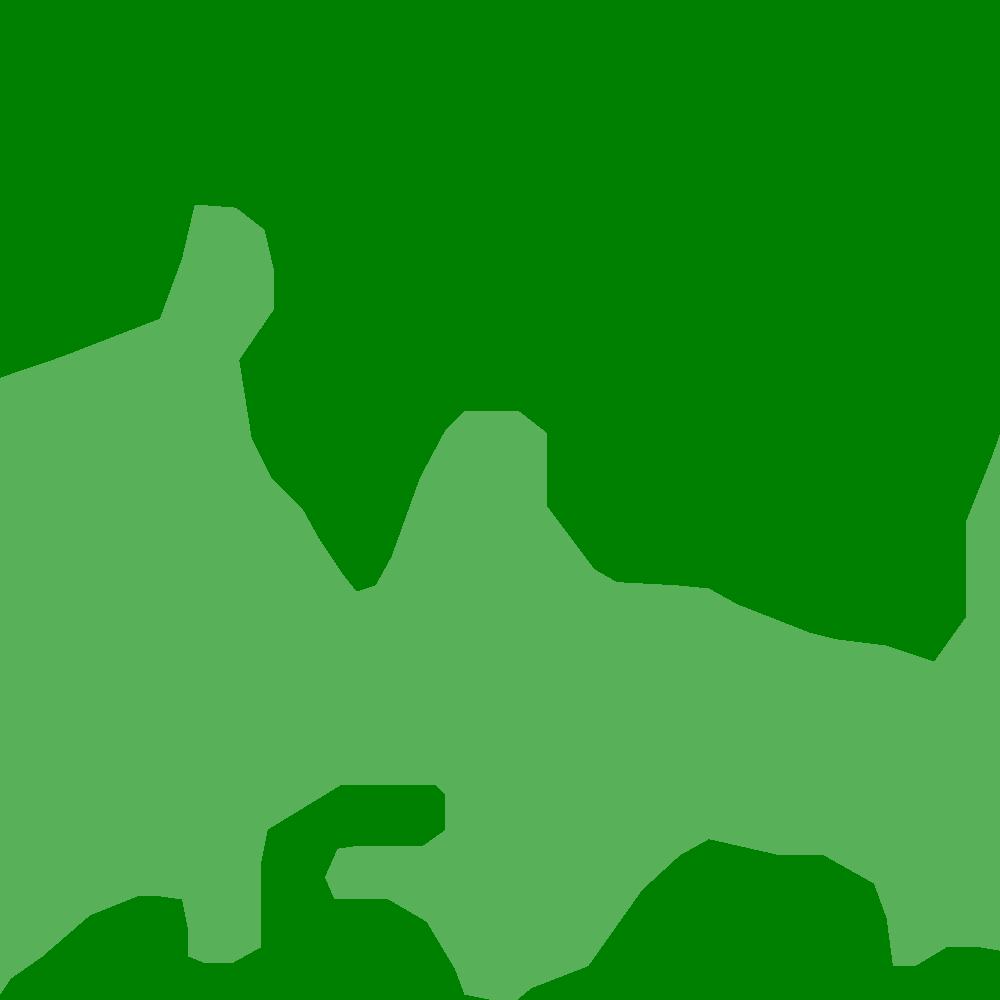} & \includegraphics[width=.15\linewidth]{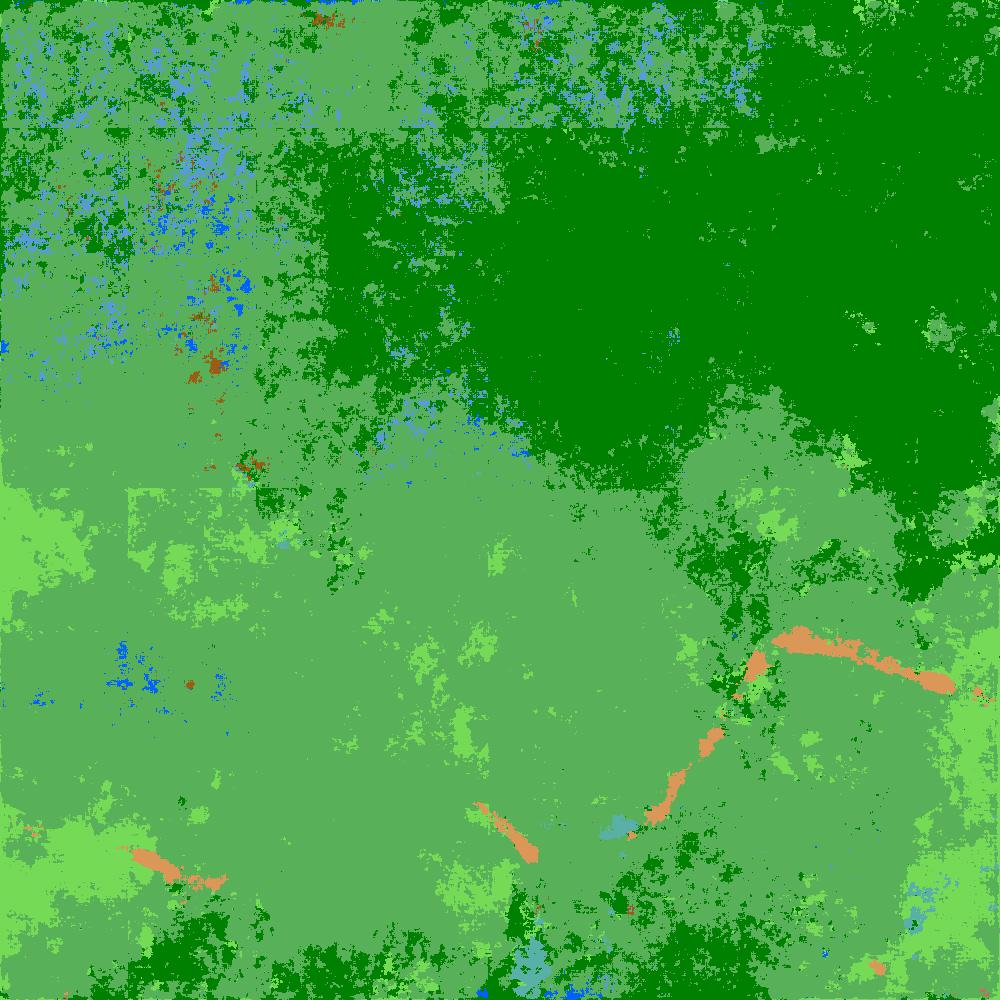} &\includegraphics[width=.15\linewidth]{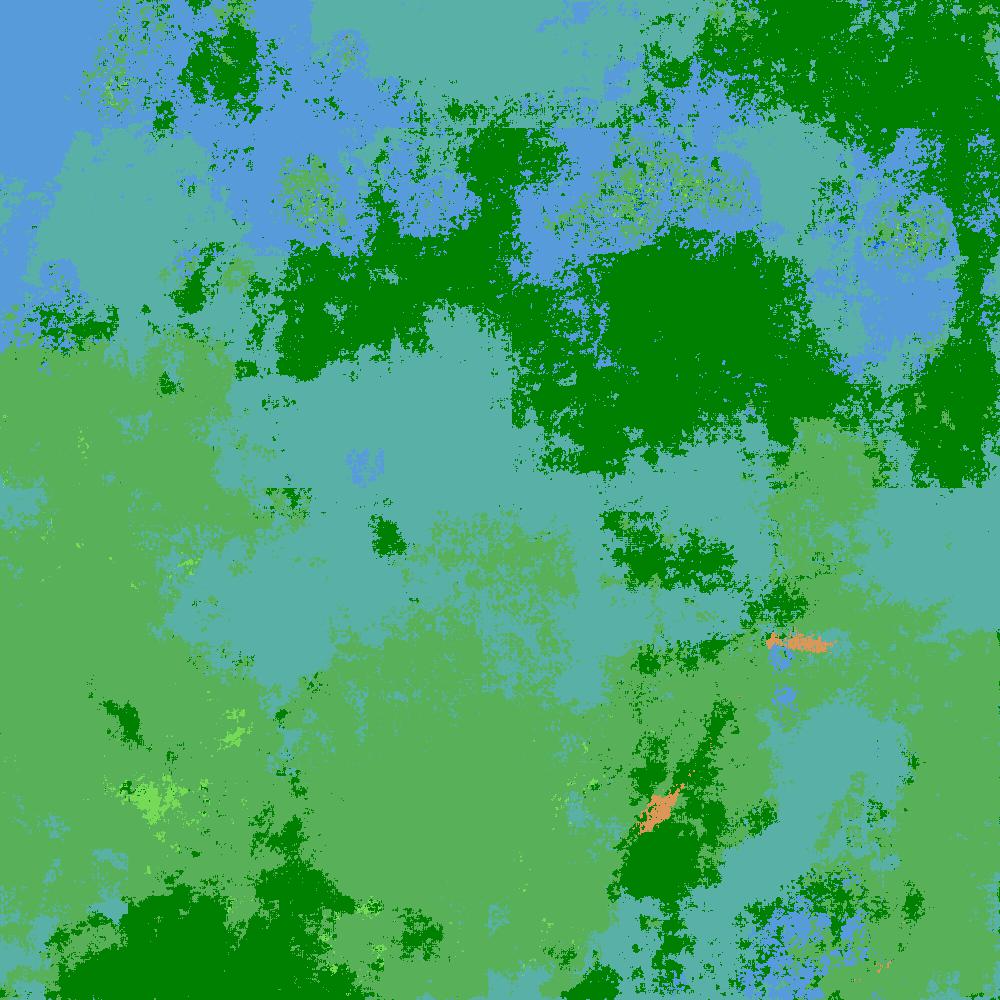} & \includegraphics[width=.15\linewidth]{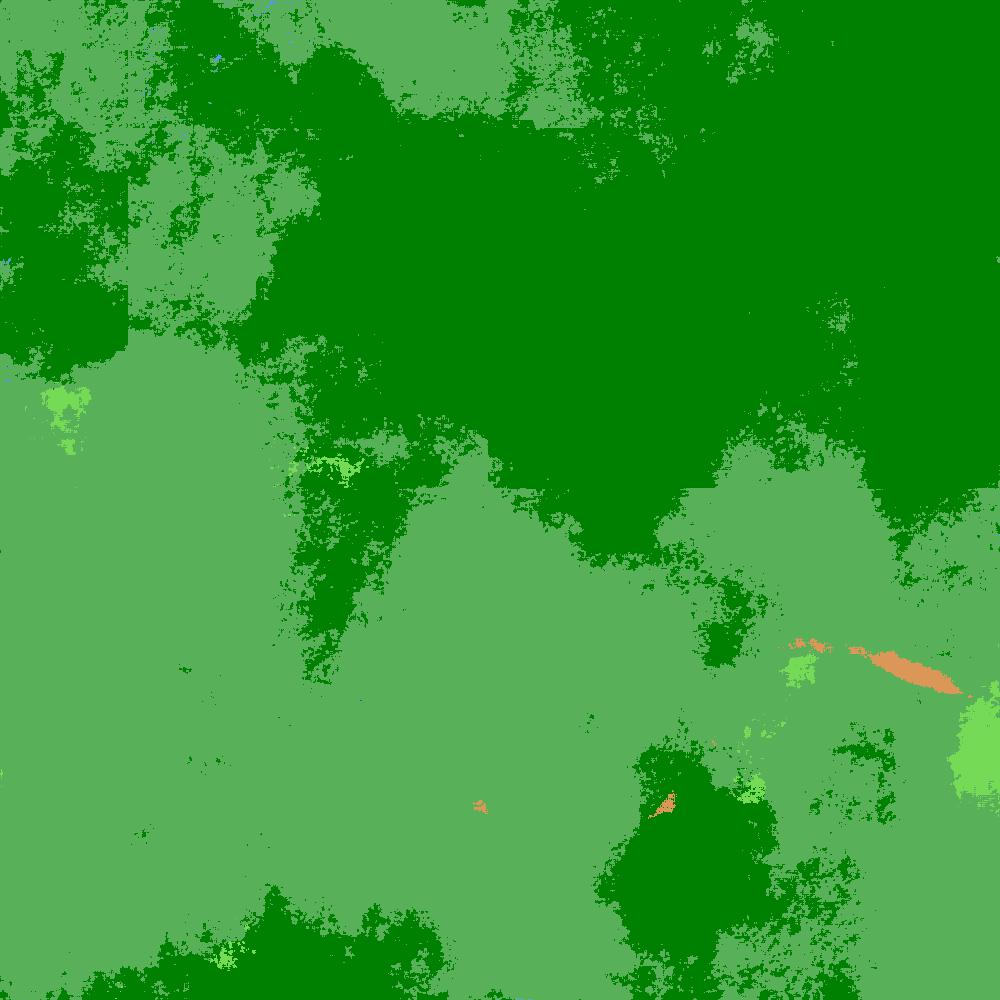} & \includegraphics[width=.15\linewidth]{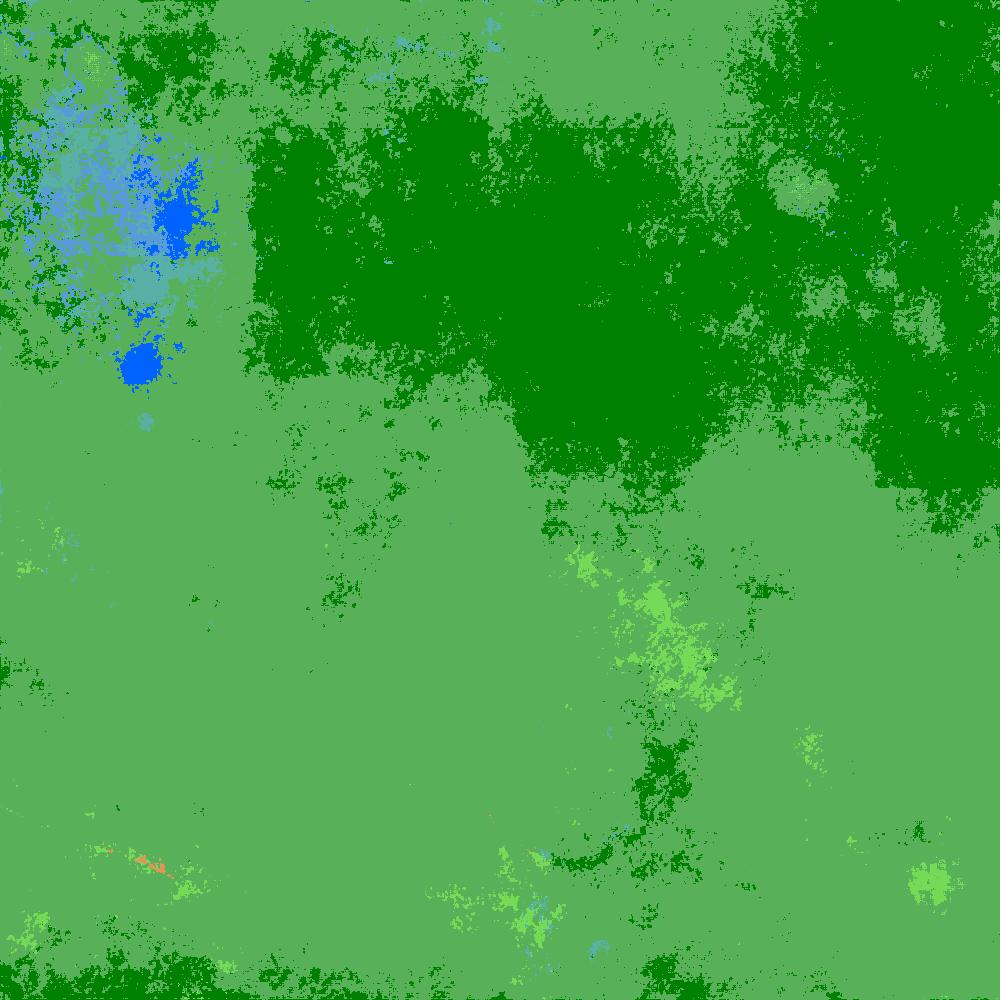}\\

         \includegraphics[width=.15\linewidth]{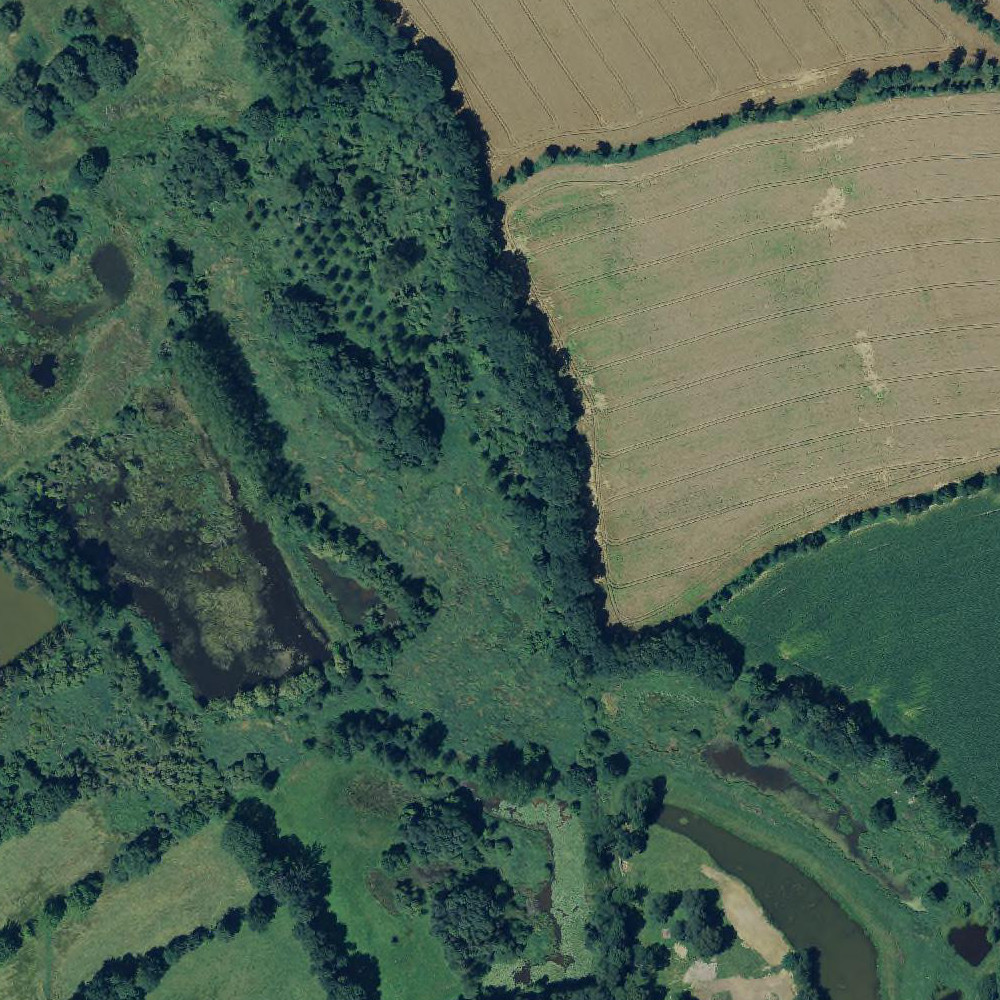} & \includegraphics[width=.15\linewidth]{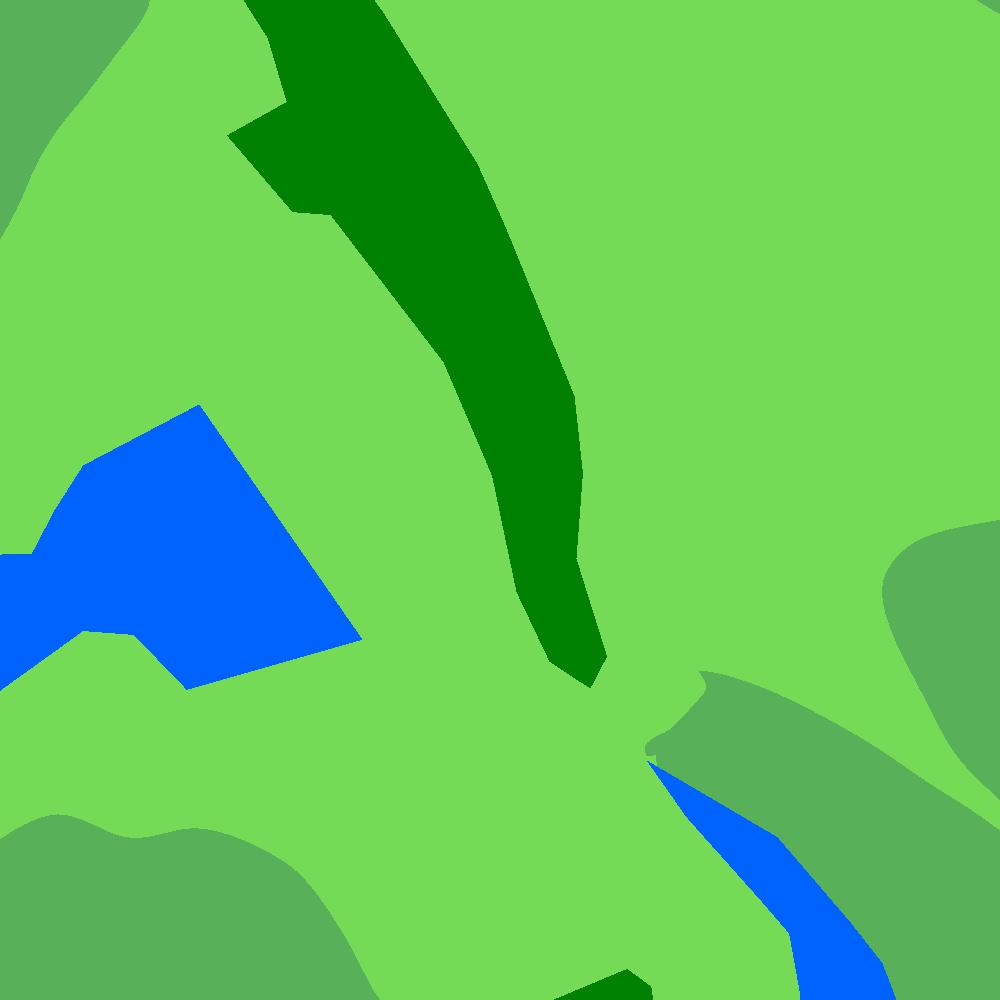} & \includegraphics[width=.15\linewidth]{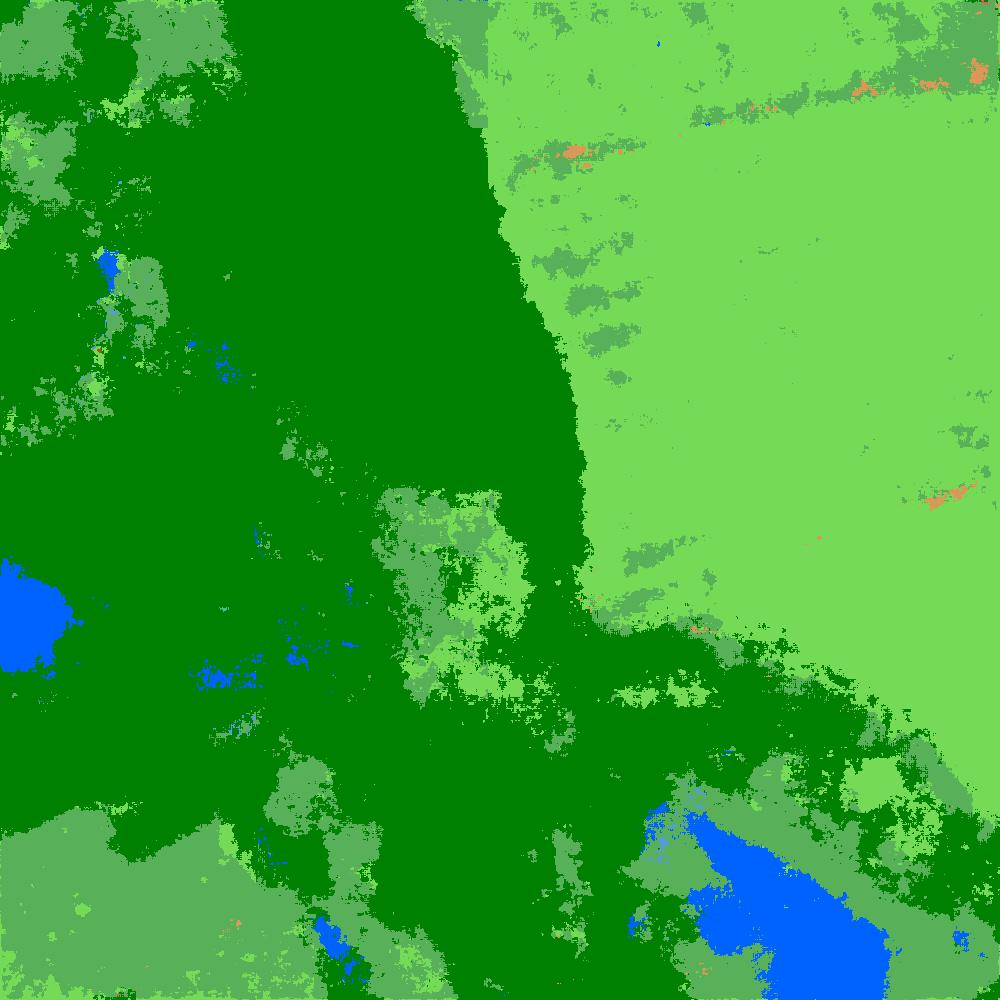} & \includegraphics[width=.15\linewidth]{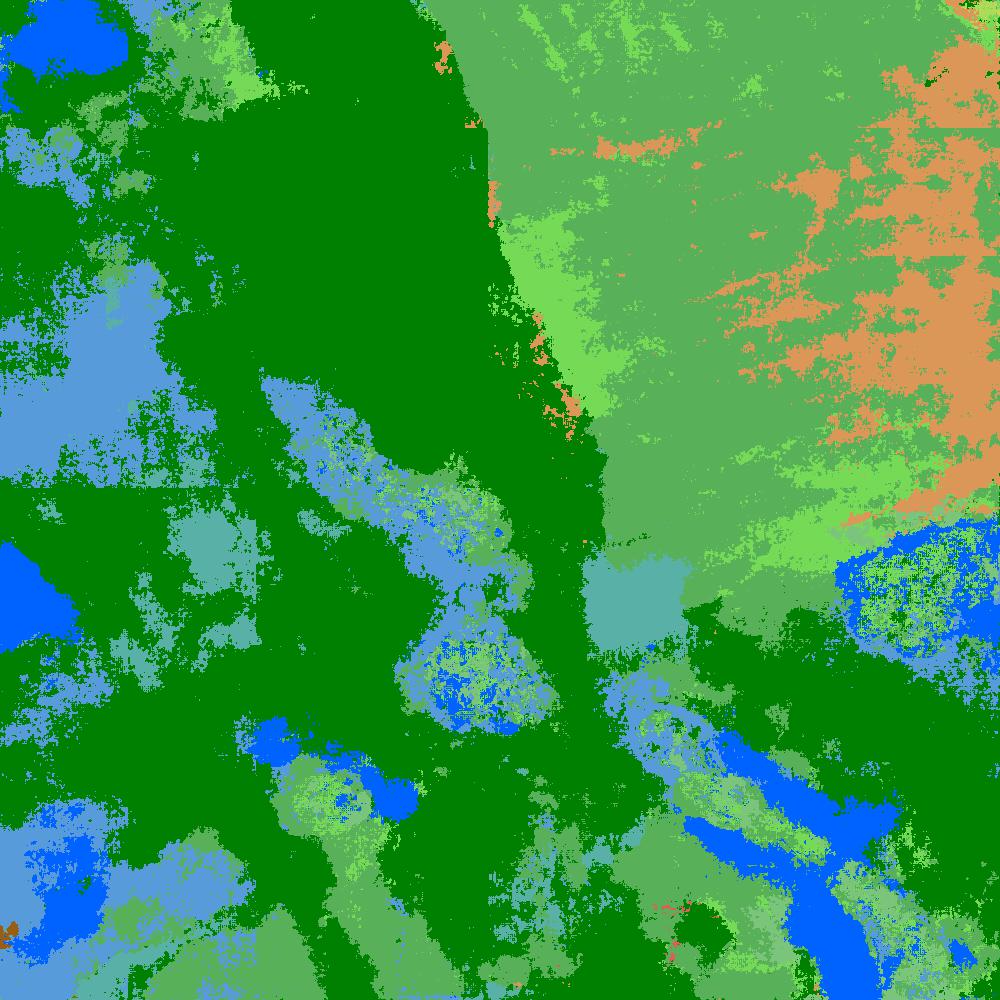} & \includegraphics[width=.15\linewidth]{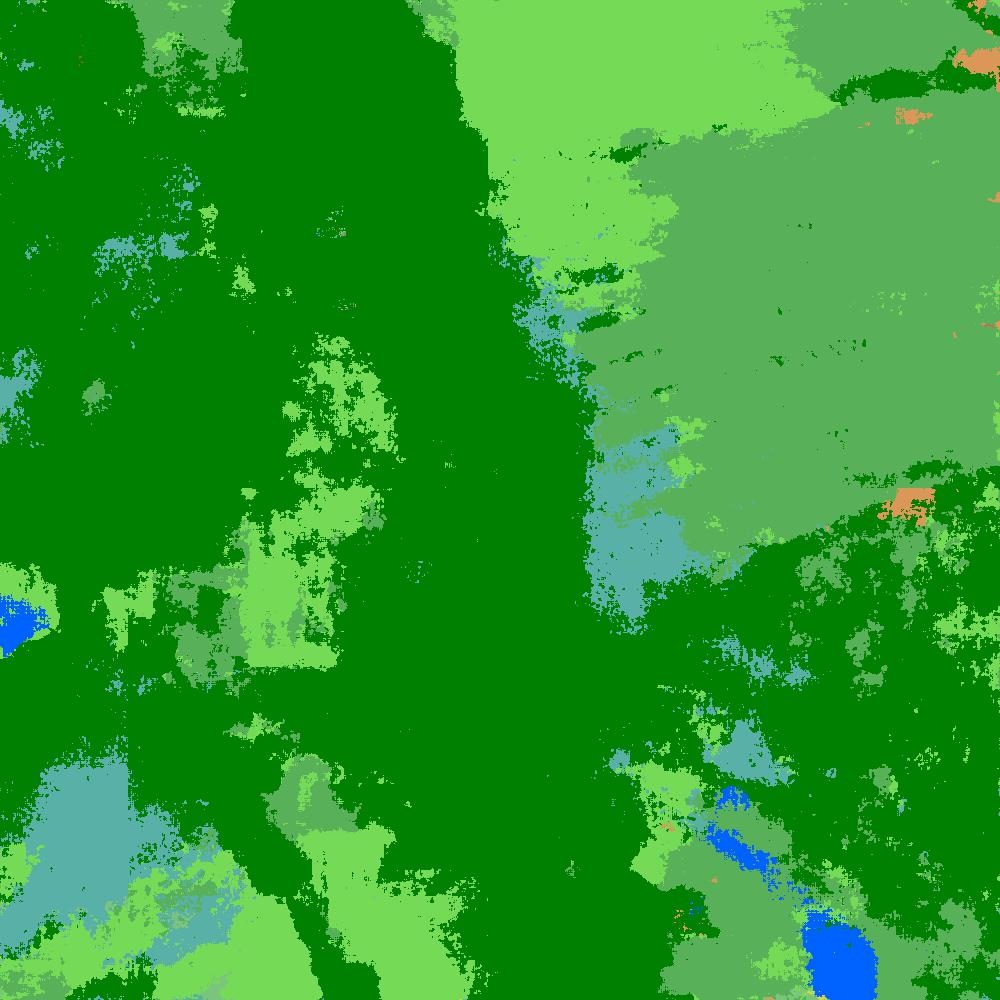} & \includegraphics[width=.15\linewidth]{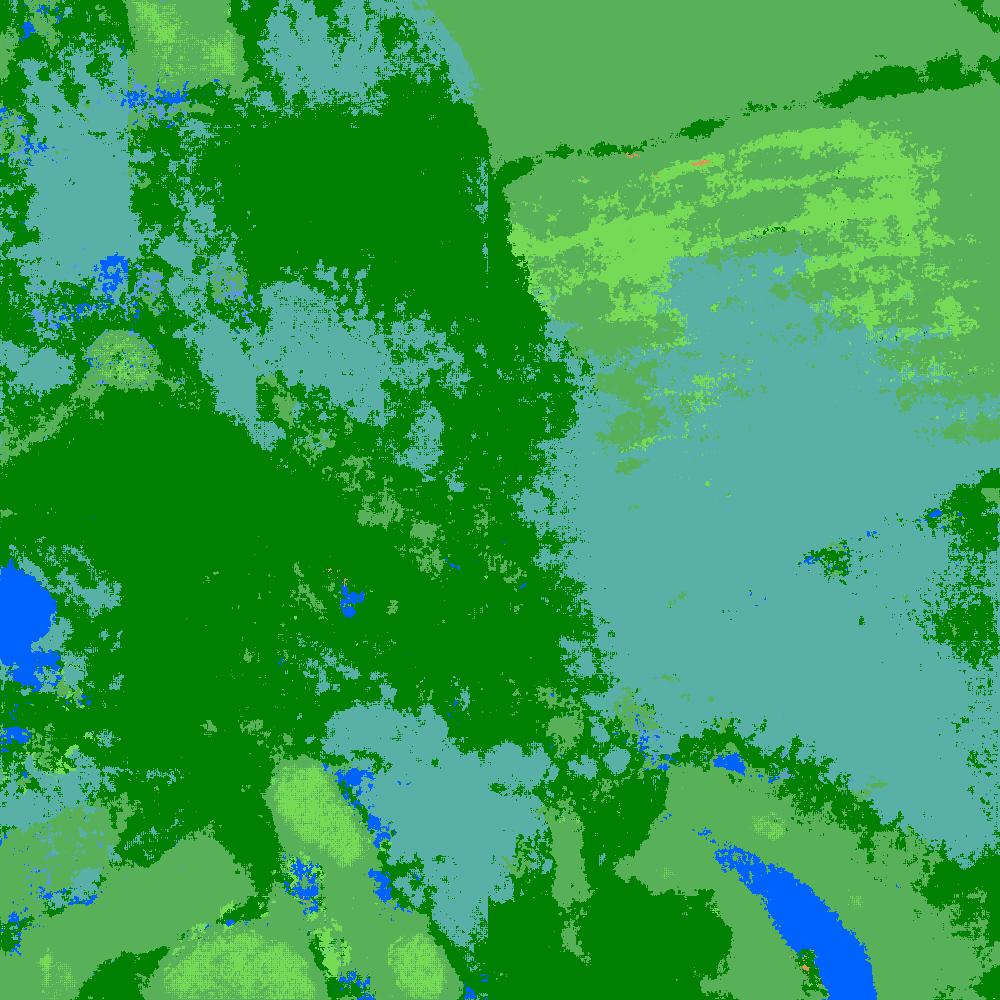}\\

         \includegraphics[width=.15\linewidth]{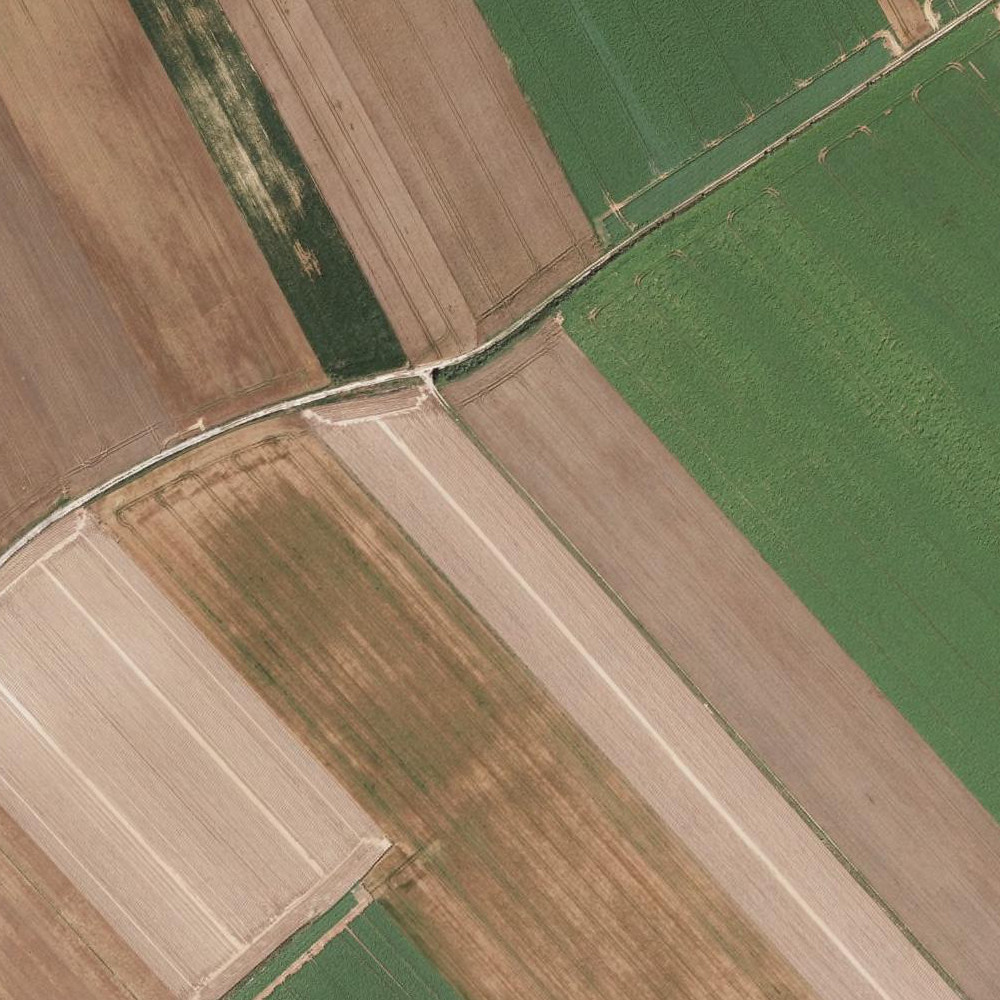} & \includegraphics[width=.15\linewidth]{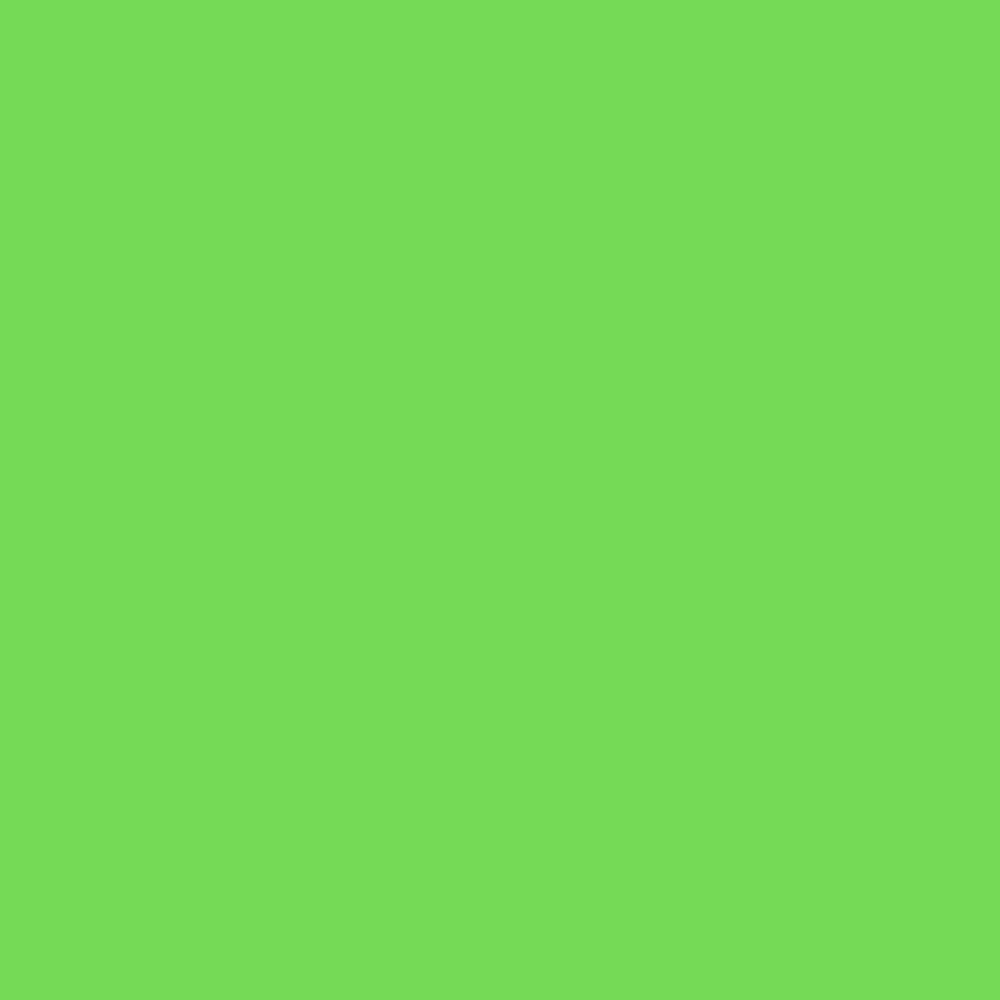} & \includegraphics[width=.15\linewidth]{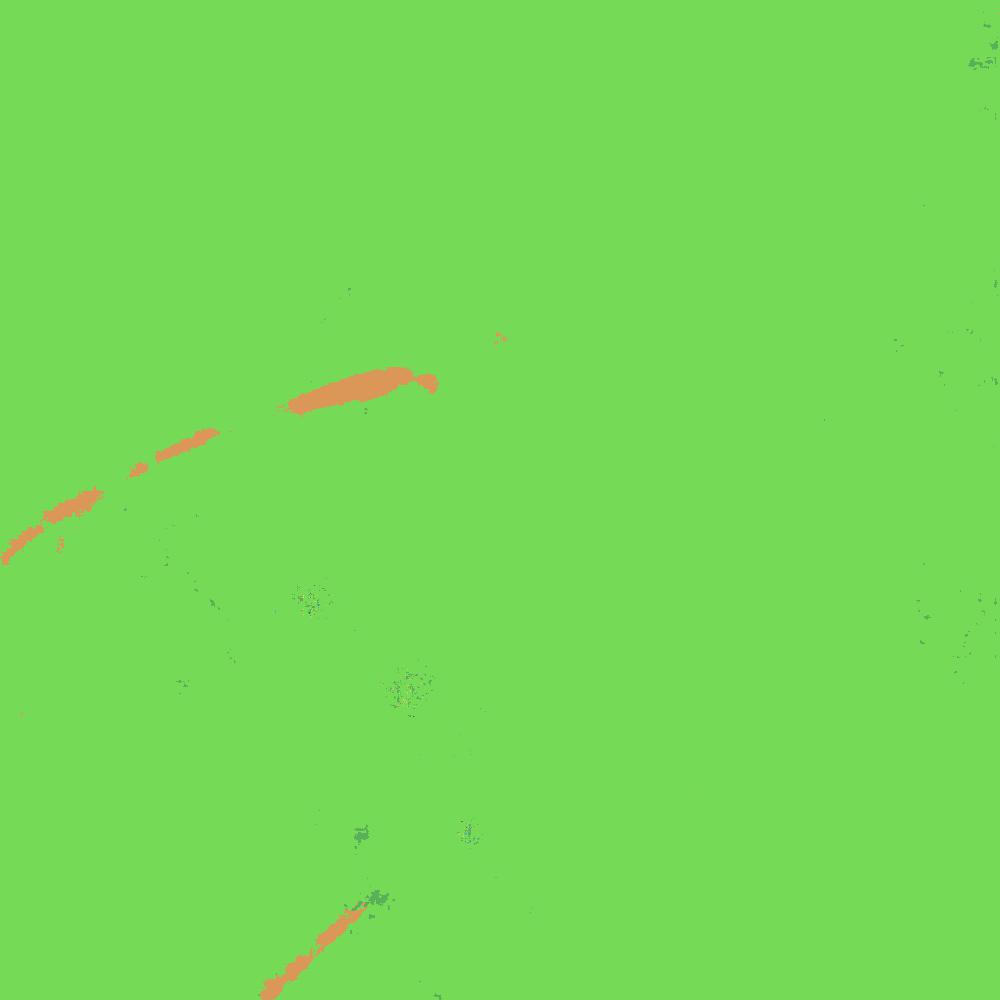} & \includegraphics[width=.15\linewidth]{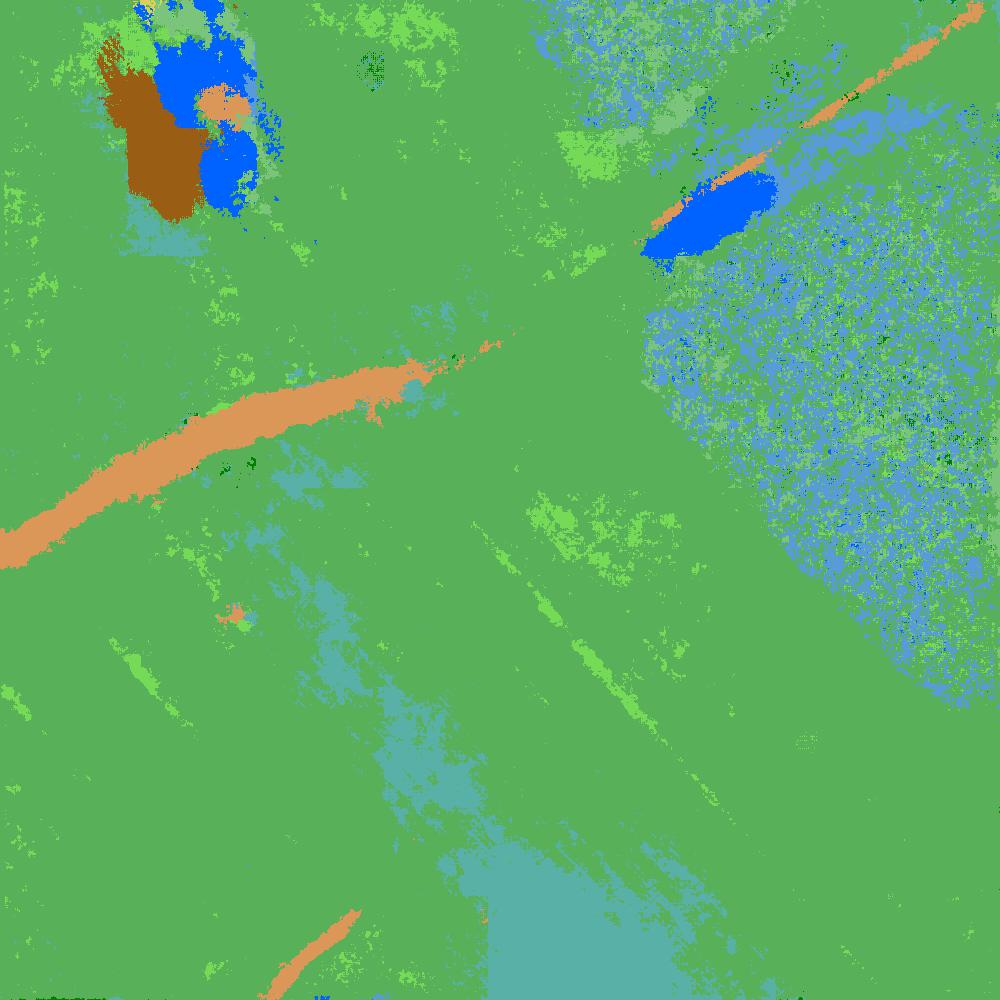} & \includegraphics[width=.15\linewidth]{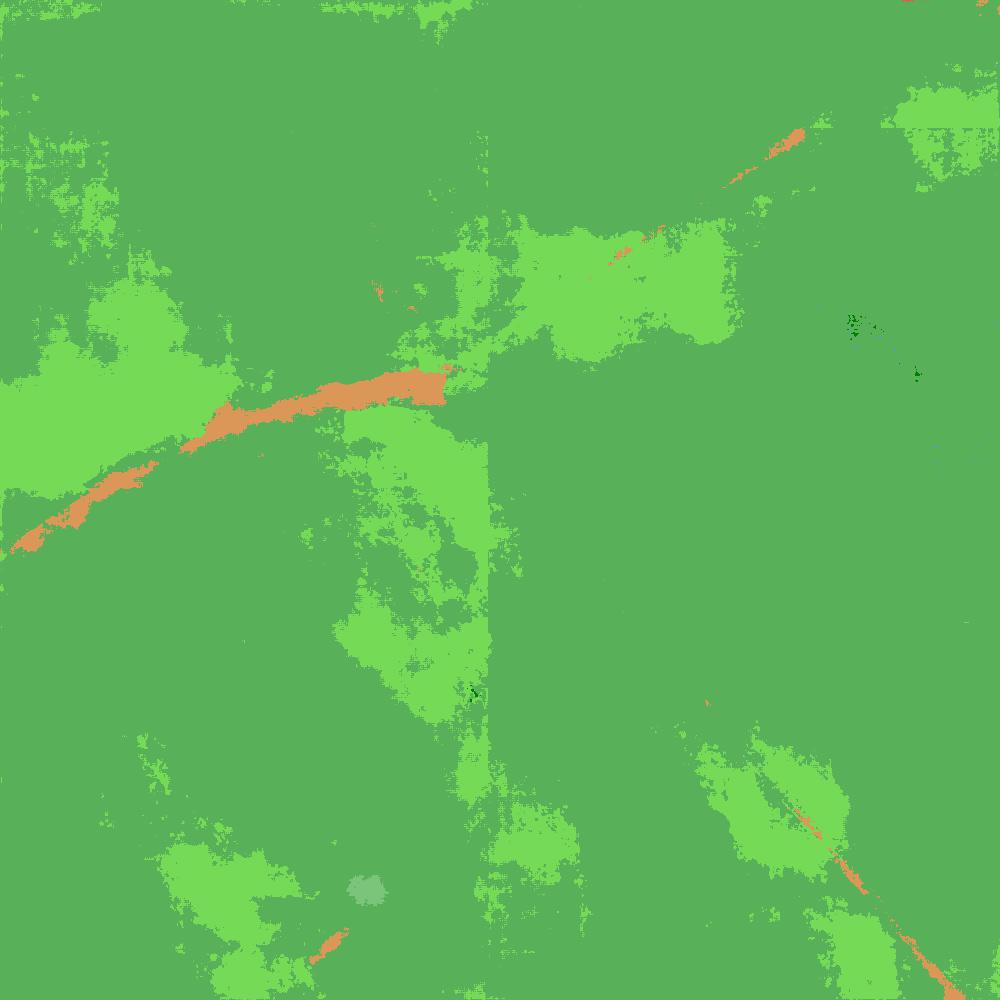} & \includegraphics[width=.15\linewidth]{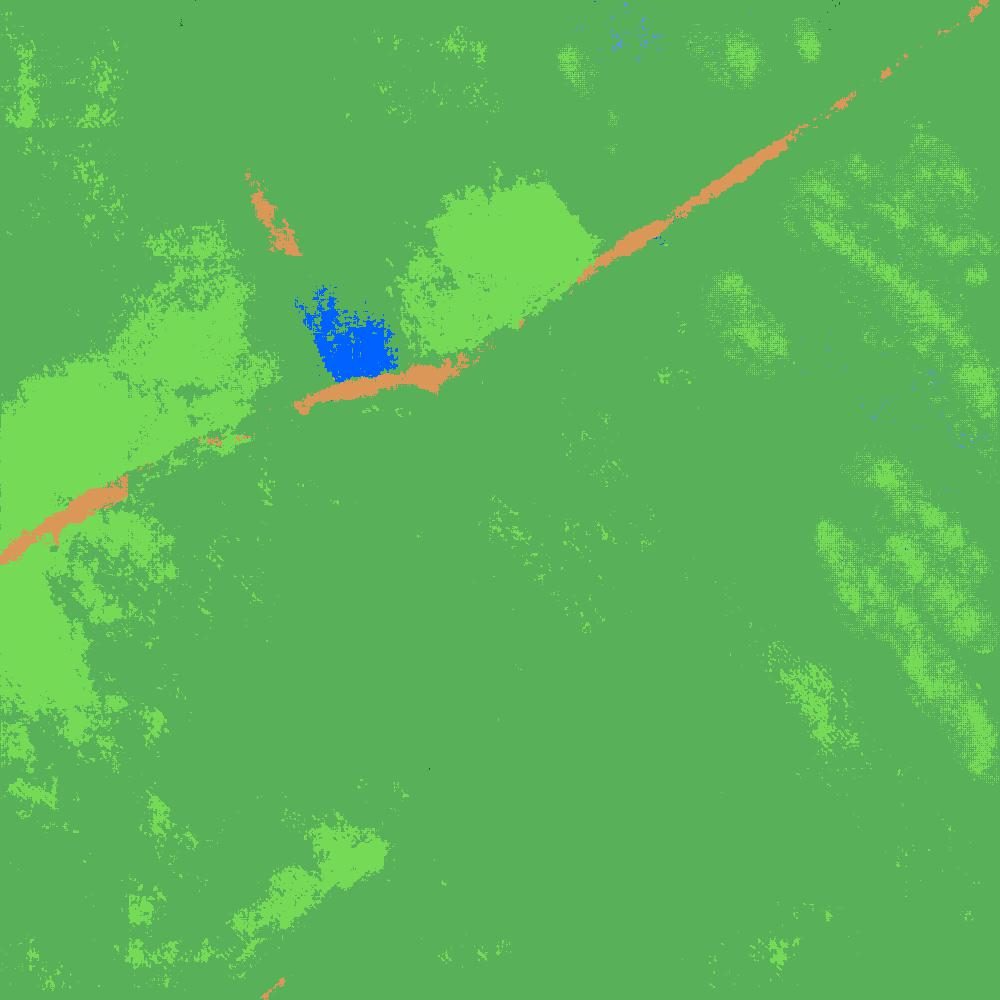}\\
         
         \includegraphics[width=.15\linewidth]{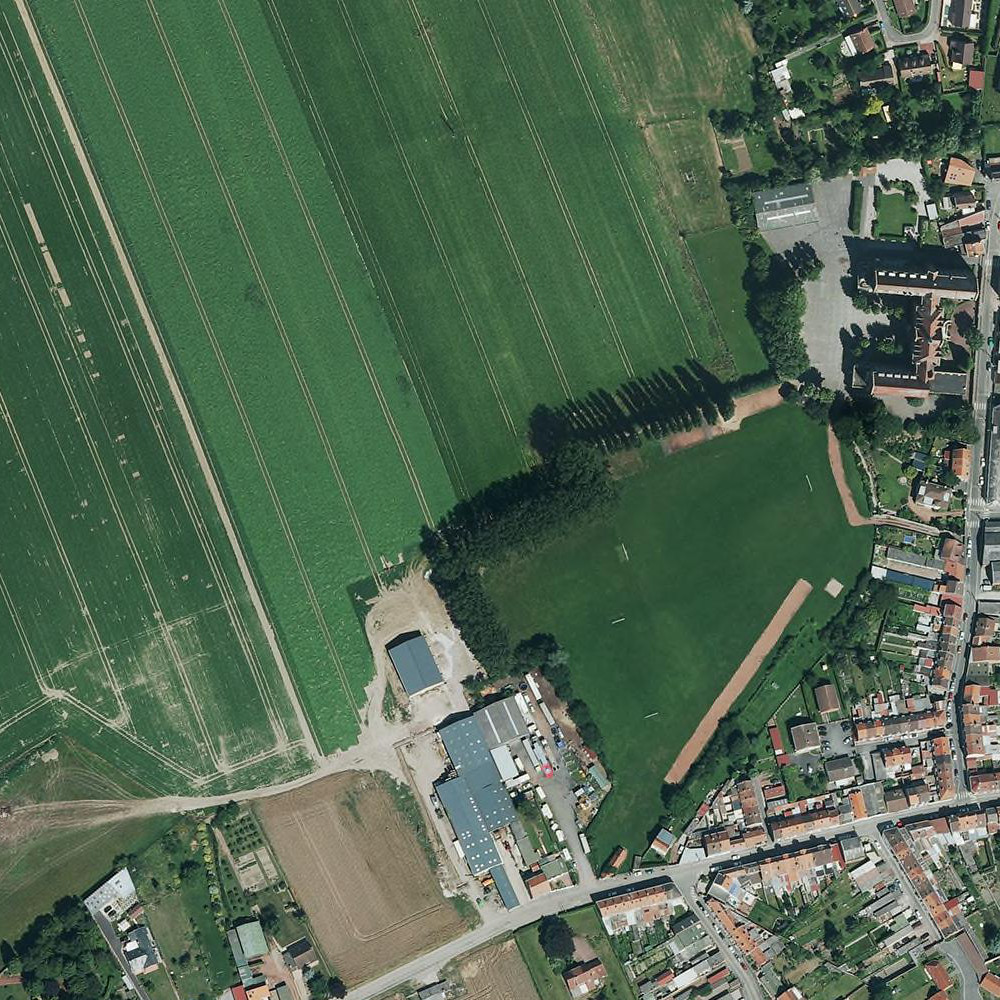} & \includegraphics[width=.15\linewidth]{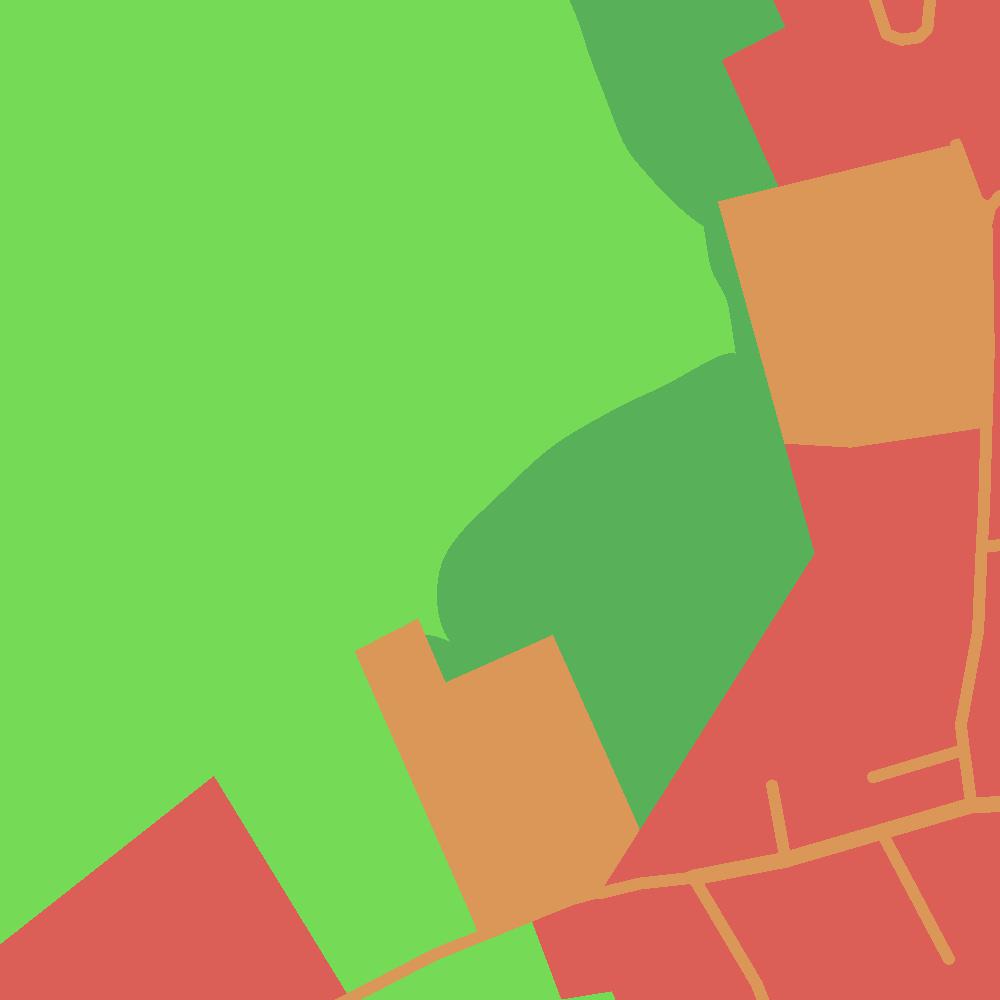} & \includegraphics[width=.15\linewidth]{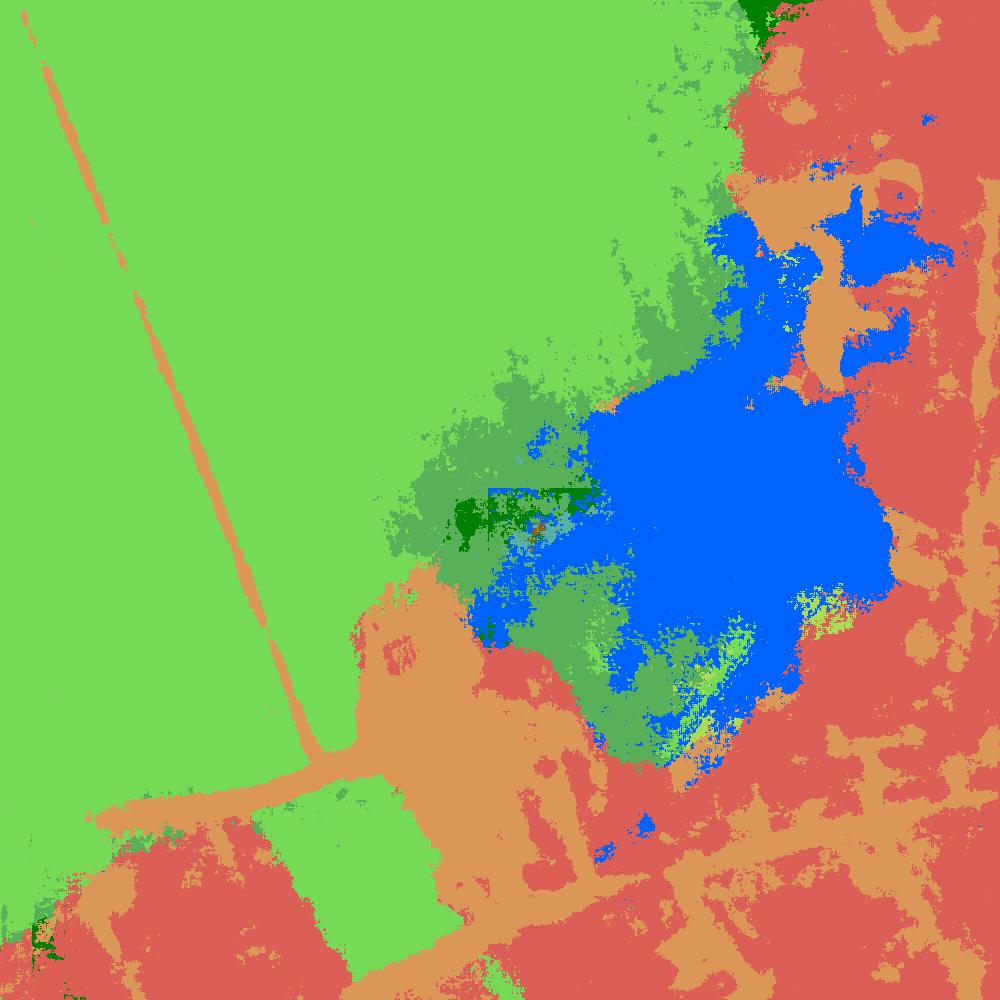} & \includegraphics[width=.15\linewidth]{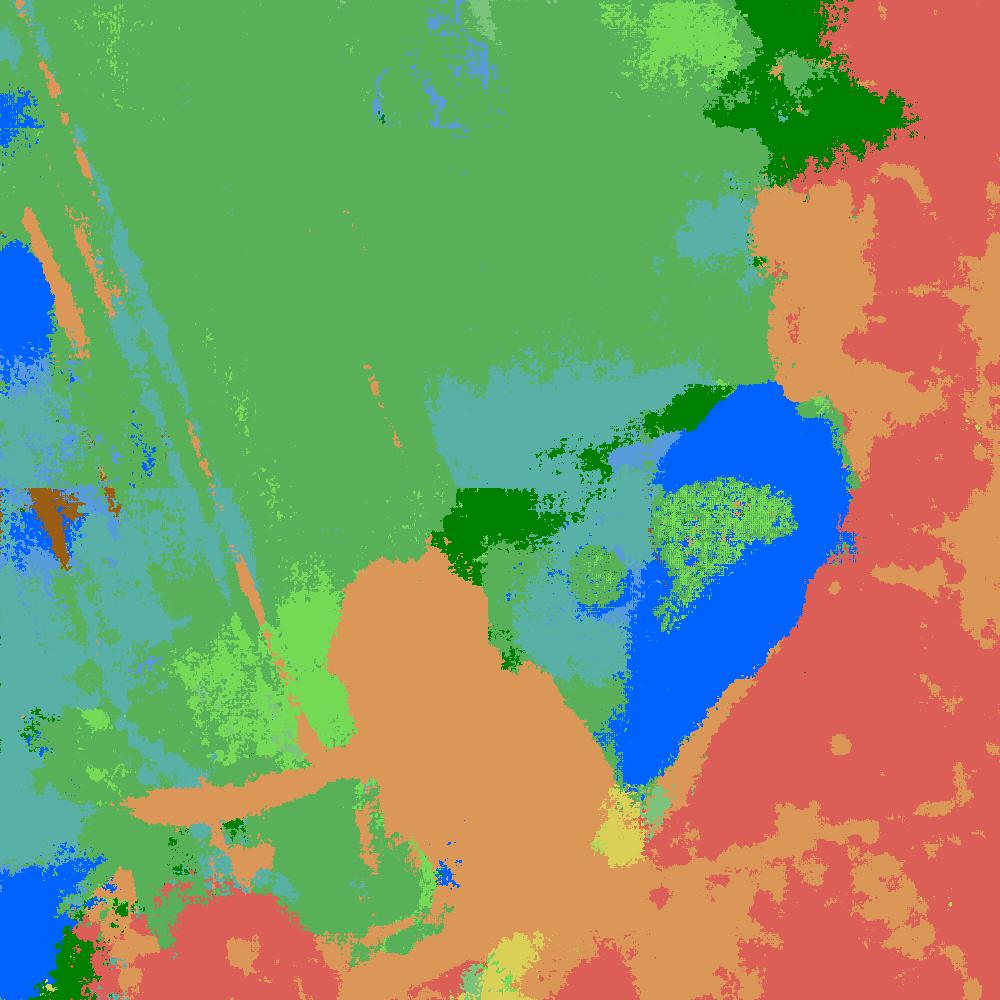} & \includegraphics[width=.15\linewidth]{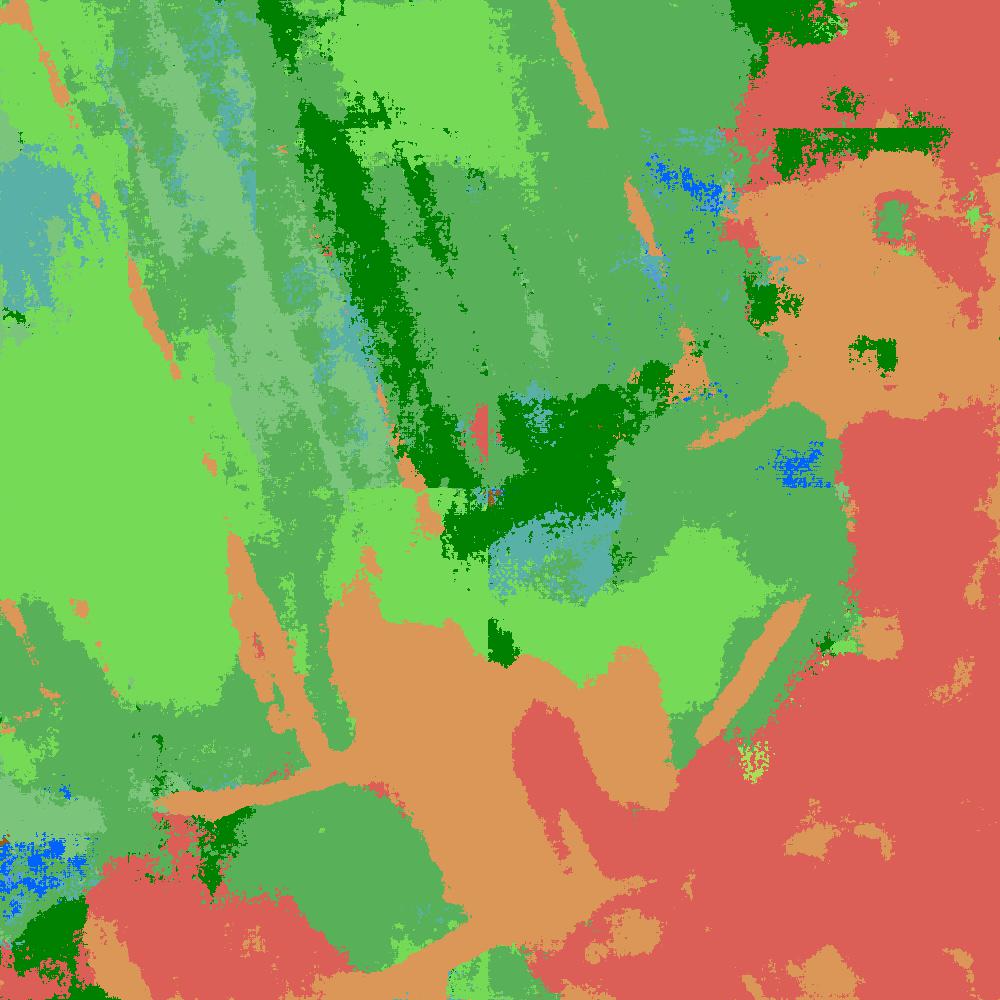} & \includegraphics[width=.15\linewidth]{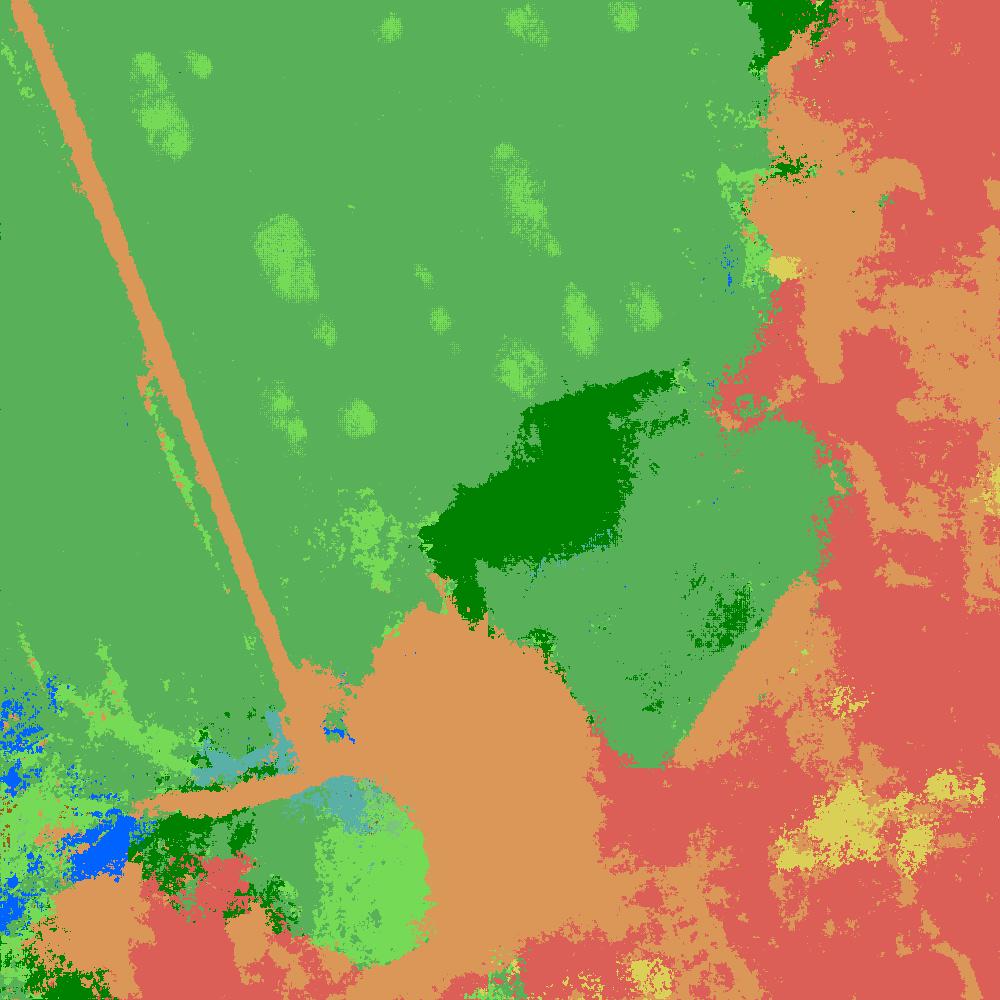}\\[3pt]

         Image & Ground truth & Oracle & Supervised & Semisup ($\LL_1$) & Semisup ($\LL_{km}$) \\[4pt]
         \multicolumn{1}{c}{}& \multicolumn{2}{c}{Undisclosed} & \multicolumn{3}{c}{Results} 
      \end{tabular}
      \caption{Classification examples of different methods. Oracle refers to the hypothetical case where all ground-truths are available for training regions (8 annotated training cities). Supervised refers to the results of a network trained only on the labeled training split of \tmf, while semi-supervised corresponds to  the BerundaNet-late network trained over all available training data (labeled and unlabeled). SegNet architecture is used as backbone. } \label{fig: interest-ssl-tmfrance}
   \end{center}
\end{figure}

Along with Table~\ref{tab: tmf-supervised-vs-semisupervised}, Figure~\ref{fig: interest-ssl-tmfrance} shows segmentation maps obtained during the testing phase for the previous experiments with a SegNet backbone. We refer as \emph{undisclosed} to the entries that are not publicly available but that are shown here as a reference and comparison to our results: ground-truth and oracle. At a global scale, we observe that semi-supervised methods -- whether with reconstruction or with segmentation auxiliary task -- present more homogeneous and finer segmentation maps than their supervised counterpart. This is noticeable in particular in clear roads and less noisy regions. Adding unlabeled data during the learning process helps to regularize and generalize better, especially in the case of \MF\ data, where labels are often approximate. In some cases, semi-supervised methods can even beat the oracle predictions, as in the last row example where the oracle mistook a pasture section for a water section. 

\medskip

Several remarks can be raised from these results:
\vspace{-\topsep}
\begin{itemize}
   \item First, \MF\ is challenging. The oracle shows that even if we could access all images labels (of the 8 cities in the training split) during training, we would only get 59\% overall accuracy with a fully supervised approach (see Table~\ref{tab: tmf-supervised-vs-semisupervised}, oracle column). This is far below  the accuracy that can be achieved with other datasets.
   \item The amount of labeled data influences a lot the performance of supervised settings. Focusing on the results of the oracle and the supervised experiment (second and third columns on Table~\ref{tab: tmf-supervised-vs-semisupervised}), we see that for a SegNet architecture going from 8 to 2 training labeled cities implies a 22\%  loss in accuracy and 10\% less of mIoU. And even if the U-Net seems more robust to the amount of labeled data, reducing annotated data diminishes network performances notoriously. From a visual perspective, prediction quality is noticeably worse for the supervised approach with respect to the oracle (third and forth columns in Figure~\ref{fig: interest-ssl-tmfrance}).

   \item Semi-supervised strategies exhibit promising results. In both cases, whether we use a SegNet or a U-Net backbone, the benefits of semi-supervised learning are clear, regardless of the chosen auxiliary task there is a gain of accuracy and mIoU with respect to the supervised method.
   \item Finally, from a visual perspective, semi-supervised methods (fifth and sixth columns in Figure~\ref{fig: interest-ssl-tmfrance}) are superior to the supervised one (fourth column). Indeed, semi-supervised segmentation maps are more homogeneous than the supervised ones (see the second, fourth and sixth row examples). Besides, urban cartography is better delineated in the semi-supervised semantic maps and seems more appropriated with respect to the original image.

\end{itemize}

Those are encouraging results for future works on semi-supervised learning for semantic segmentation.  

\subsection{Analysis of Semi-supervised Learning on \tmf}

We have seen that semi-supervised learning can be beneficial to improve segmentation results. Next sections intend to explore some possibilities to approach semi-supervised learning, in terms of neural network architectures or losses to use in a multi-task learning strategy.

For this purpose, we present several experiments performed over \tmf\ to analyze the contributions of the neural network architectures in section~\ref{sec: experiments-choice-architecture}. We also study the effect of the choice of auxiliary task to perform and the unsupervised loss function in section~\ref{sec: experiments-choice-loss}.

\subsubsection{Influence of the choice of architecture on semi-supervision} \label{sec: experiments-choice-architecture}

In the following, we compare the architectures presented in section~\ref{sec: semisup-neural-networks} with respect to both auxiliary tasks, reconstruction (using $\LL_1$ loss) and unsupervised segmentation (with $\LL_{km}$ loss). For the BerundaNet-early architecture a SegNet backbone is used. Results of these experiments are reported in Table~\ref{tab:tmf-architectures-comparison}.

\begin{table}[H]
   \begin{center}
      \caption{Neural networks for semi-supervised semantic segmentation comparison.}\label{tab:tmf-architectures-comparison}
 
      \begin{tabular}{ccccc} \toprule
         \emph{Auxiliary Loss} & \emph{Architecture} & \emph{Backbone} & \emph{OA (\%)} &\emph{ mIoU (\%)} \\ \midrule
         \multirow{3}{*}{$\LL_1$} & BerundaNet-early & SegNet & 35.94 & 9.51 \\
         & BerundaNet-late  & SegNet & 45.52 & 14.43 \\
         & BerundaNet-late & U-Net & \bf{47.90} & \bf{18.70} \\
         & W-Net~\cite{xia2017wnet} & U-Net  & 40.72 & 13.79 \\ \midrule
         \multirow{3}{*}{ $\LL_{km}$} & BerundaNet-early & SegNet & 38.20 & 10.26 \\
         & BerundaNet-late & SegNet & 42.26 & 15.75 \\
         & BerundaNet-late & U-Net & \bf{46.92} & \bf{18.26} \\
         & W-Net~\cite{xia2017wnet} & U-Net & 45.20 & 16.13\\       
         \bottomrule
      \end{tabular}

   \end{center}   
\end{table}

Whatever the chosen auxiliary task, BerundaNet-late with U-Net backbone is the architecture that achieves the best scores, followed by W-Net and BerundaNet-late with SegNet backbone. BerundaNet-early is just slightly better than a supervised approach with same backbone. This indicates that, in terms of network architecture, it might be better to split the supervised and unsupervised tasks rather late, enabling more shared parameters. Thus, the image statistics learned through optimization of the auxiliary task are better harnessed for the main objective.

\begin{figure}[!htbp]
   \begin{center}
      \setlength{\tabcolsep}{1pt}
      \begin{tabular}{cc@{\hspace{8pt}}cccc@{\hspace{8pt}}cccc}
         \includegraphics[width=.09\linewidth]{35-2012-0345-6780-LA93-0M50-E080_2493_3283.jpg} & \includegraphics[width=.09\linewidth]{35-2012-0345-6780-LA93-0M50-E080_2493_3283_gt.jpg} & \includegraphics[width=.09\linewidth]{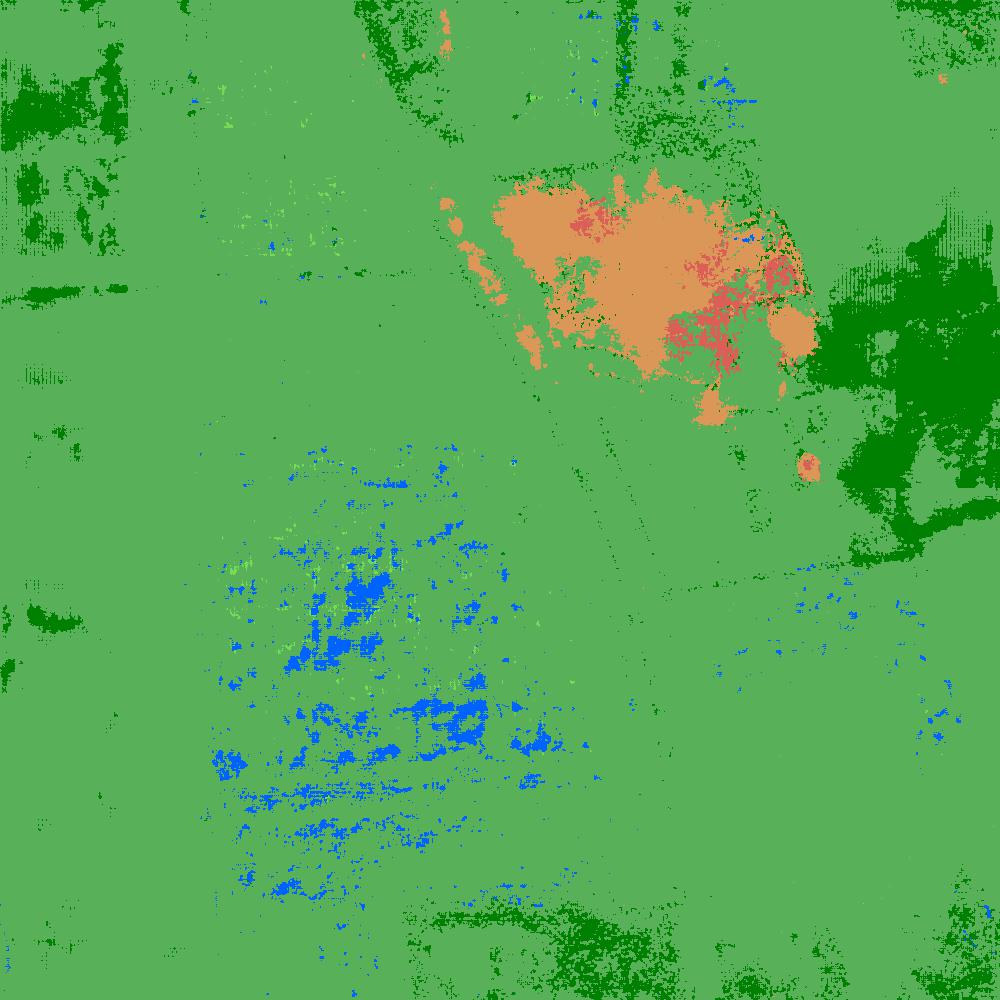} & \includegraphics[width=.09\linewidth]{35-2012-0345-6780-LA93-0M50-E080_2493_3283_segnet_l1_pred.jpg} & \includegraphics[width=.09\linewidth]{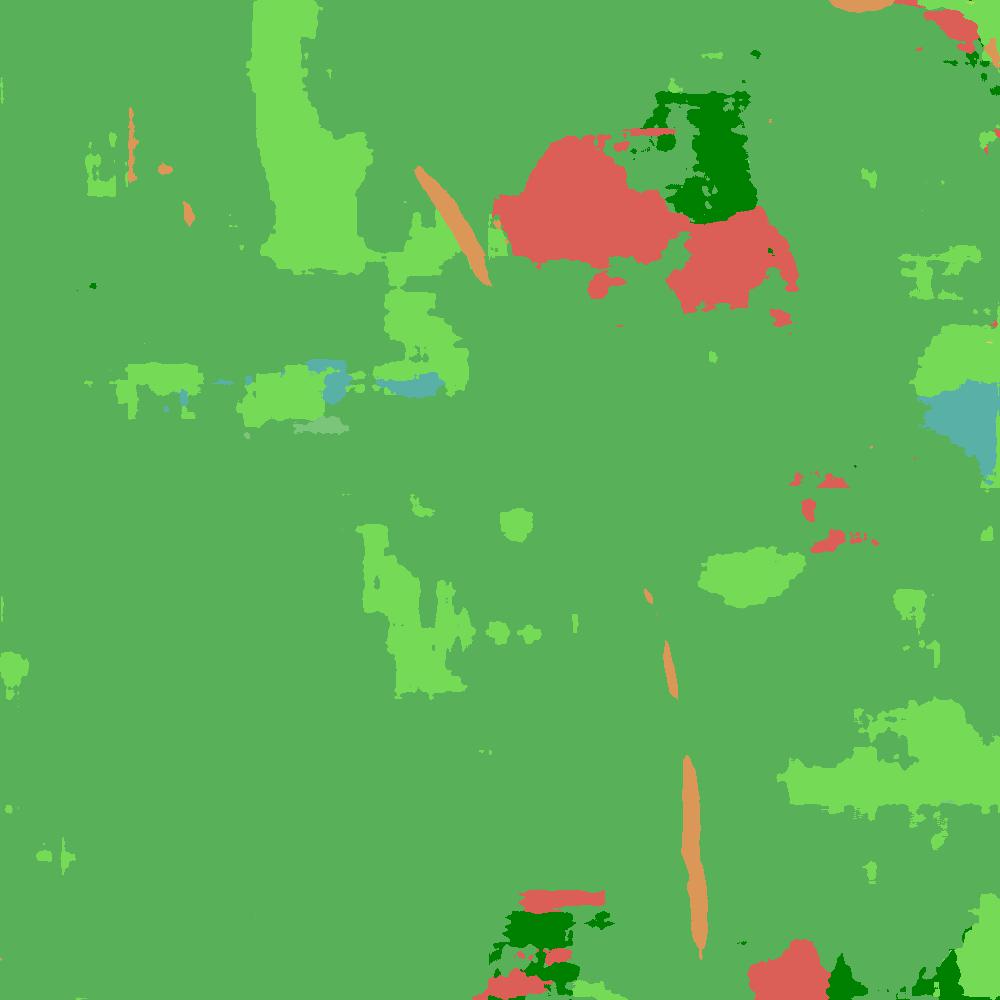} & \includegraphics[width=.09\linewidth]{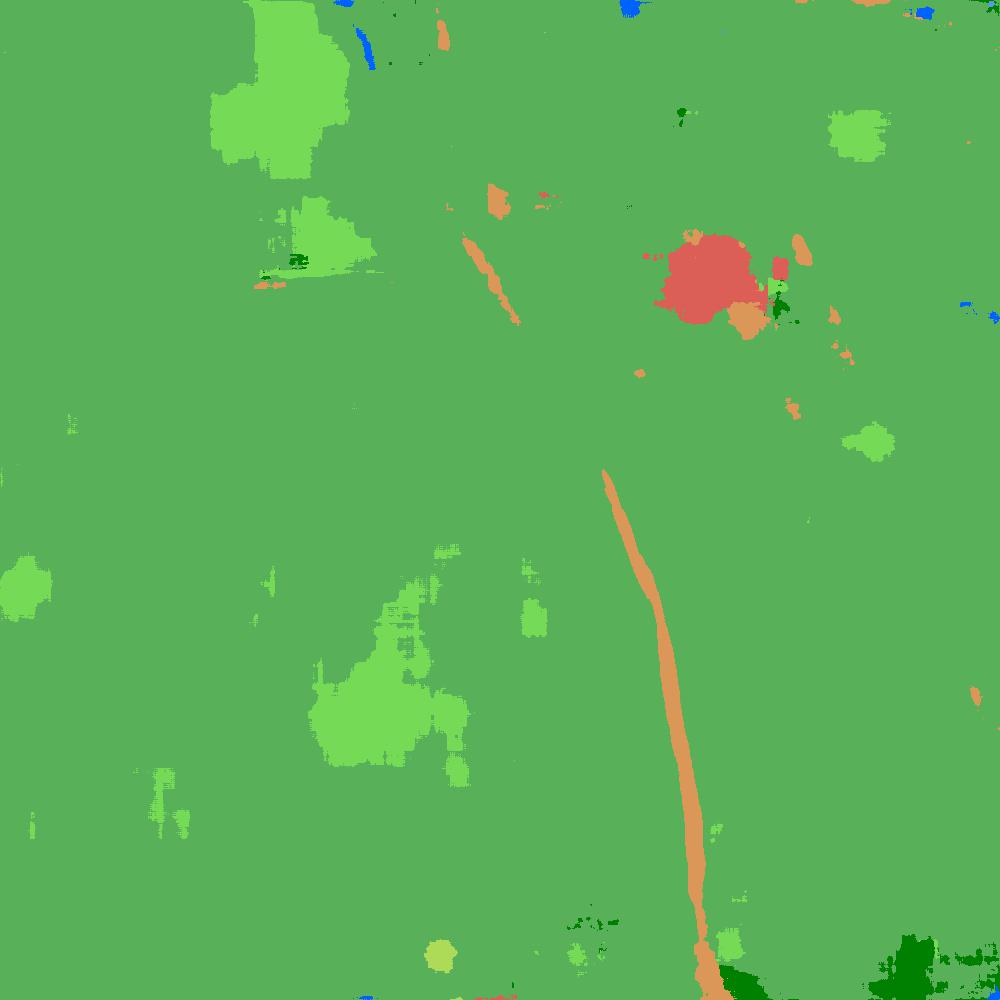} & \includegraphics[width=.09\linewidth]{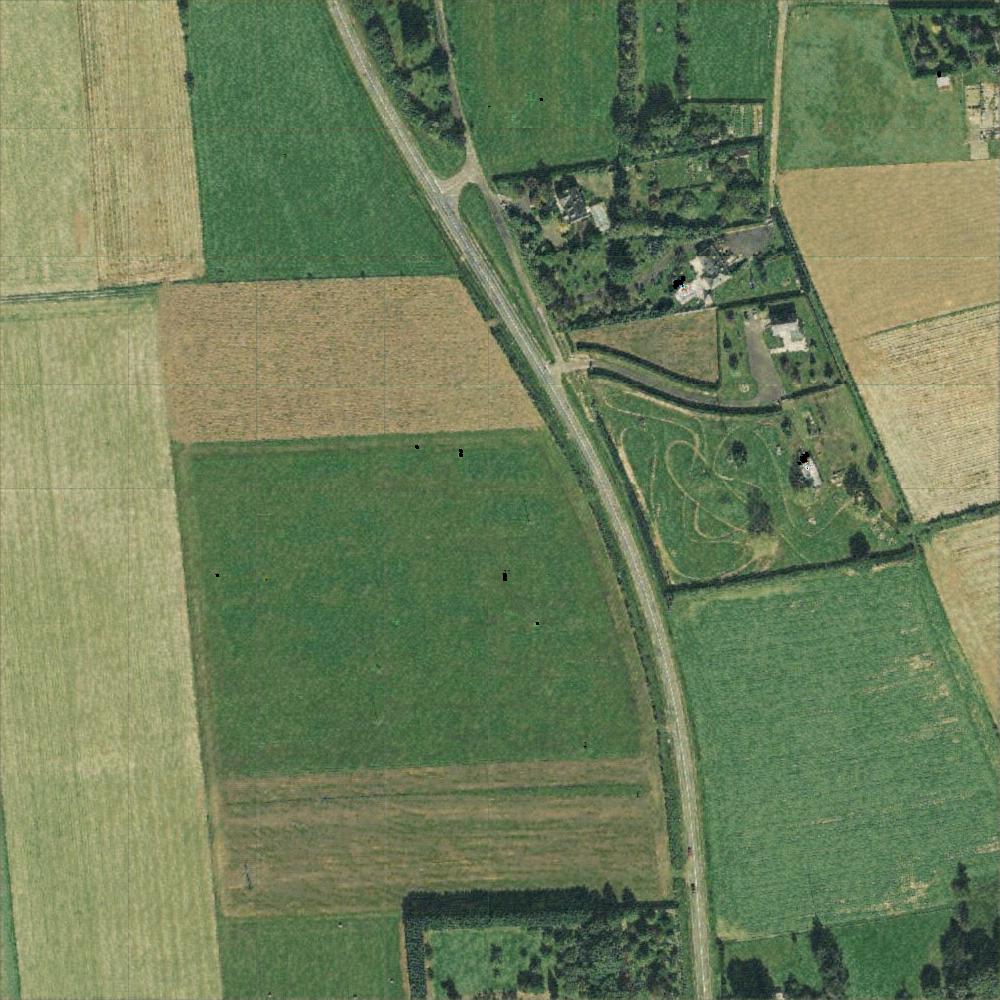} & \includegraphics[width=.09\linewidth]{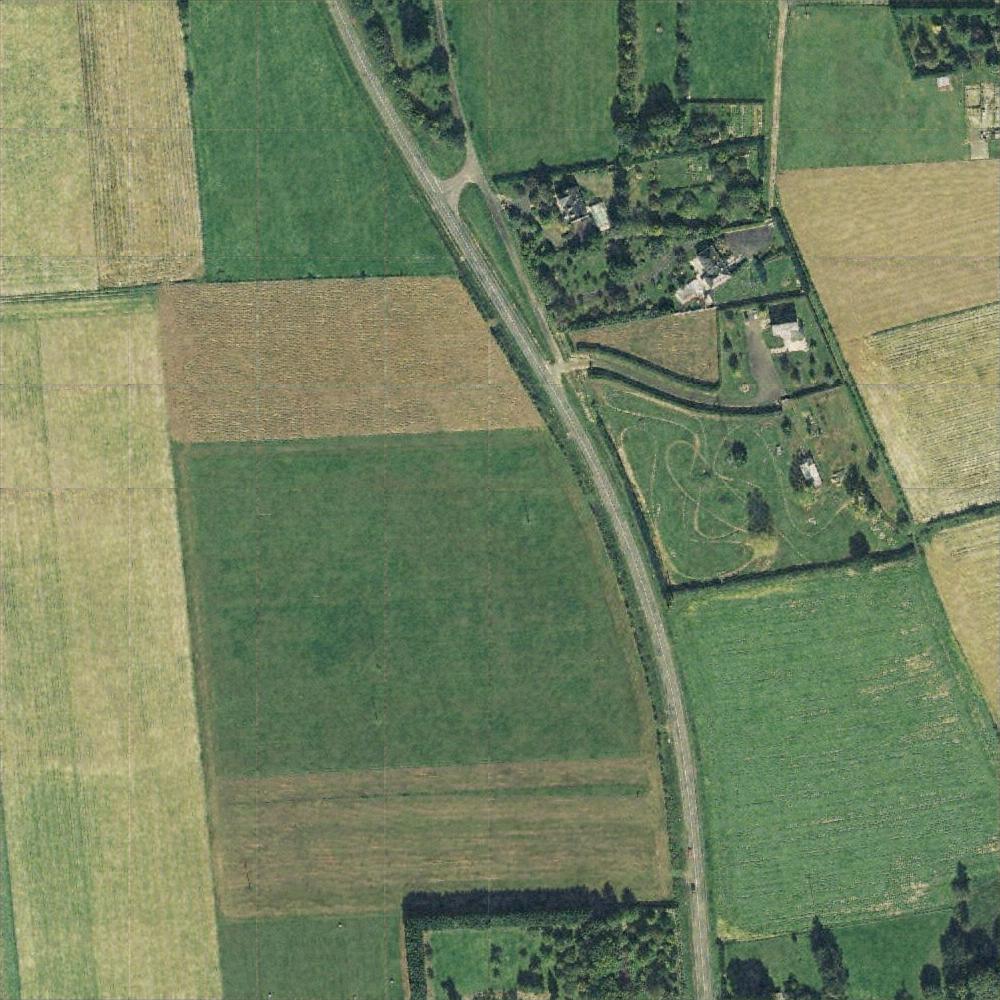} & \includegraphics[width=.09\linewidth]{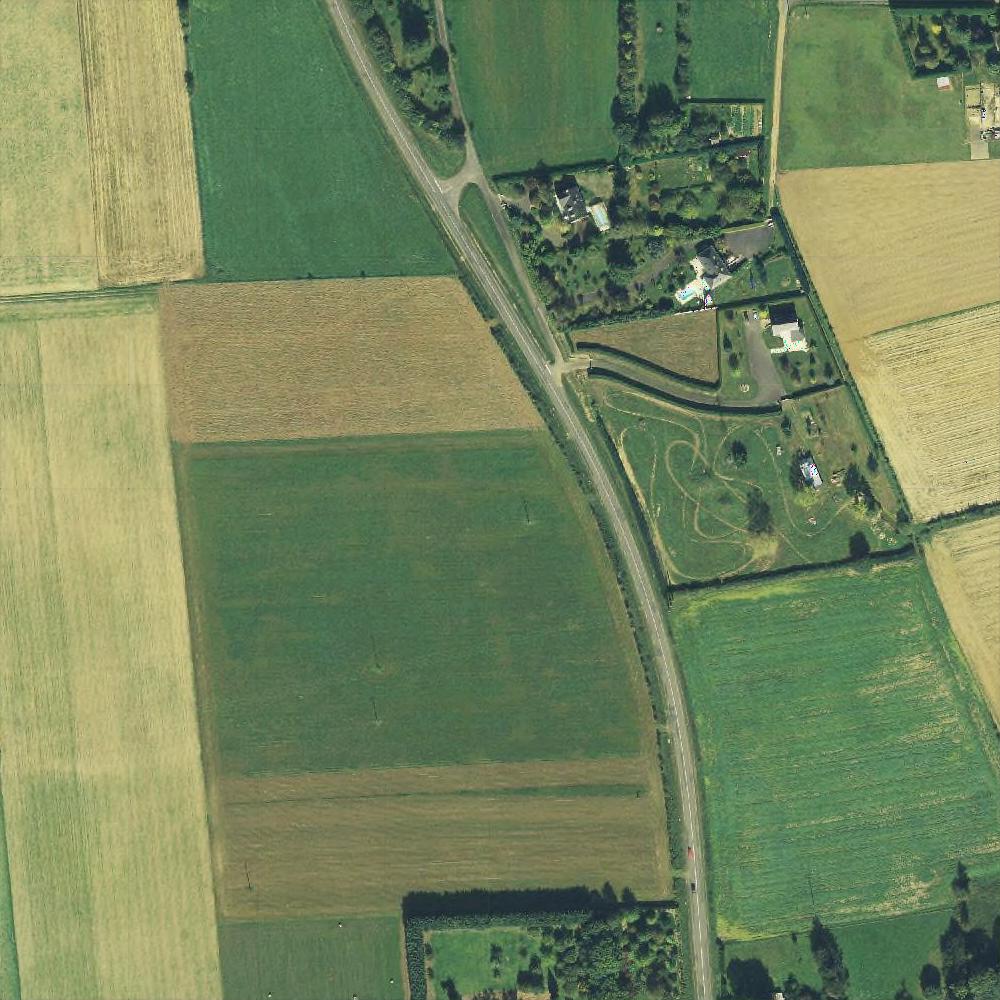} & \includegraphics[width=.09\linewidth]{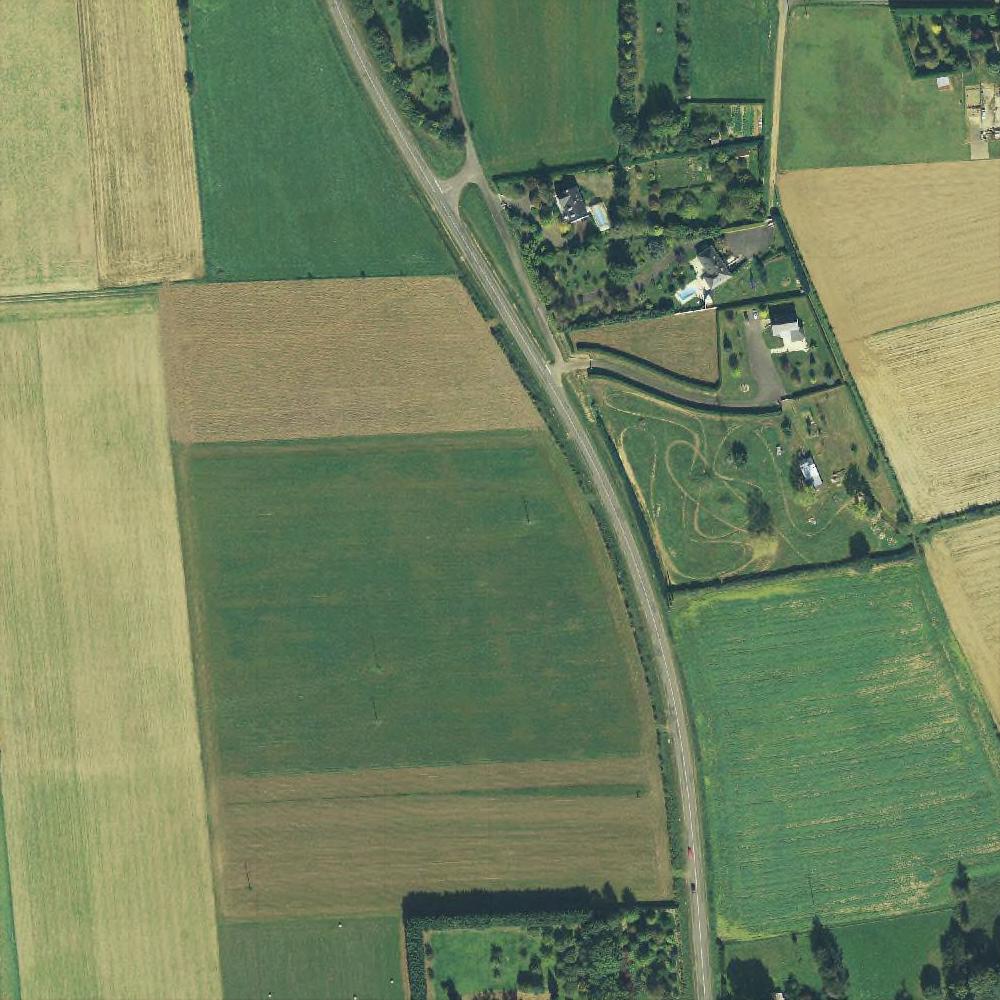} \\
         \includegraphics[width=.09\linewidth]{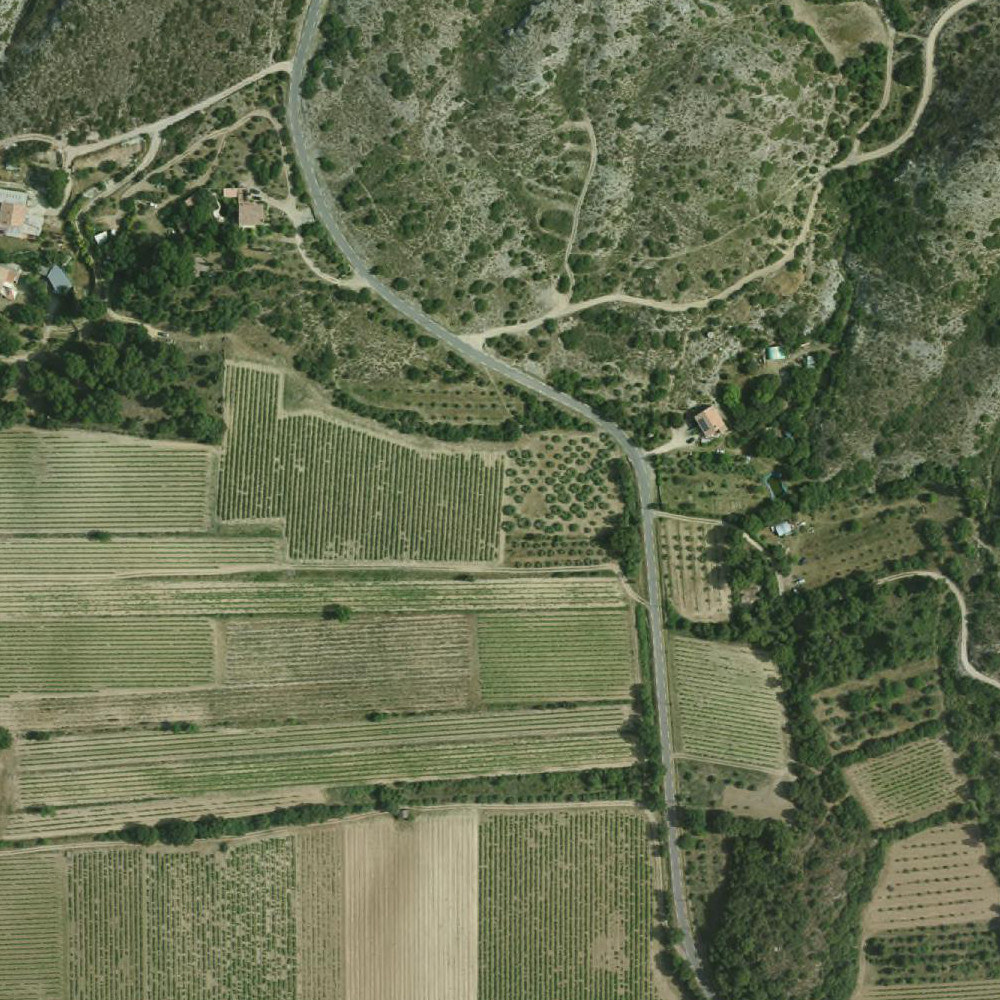} & \includegraphics[width=.09\linewidth]{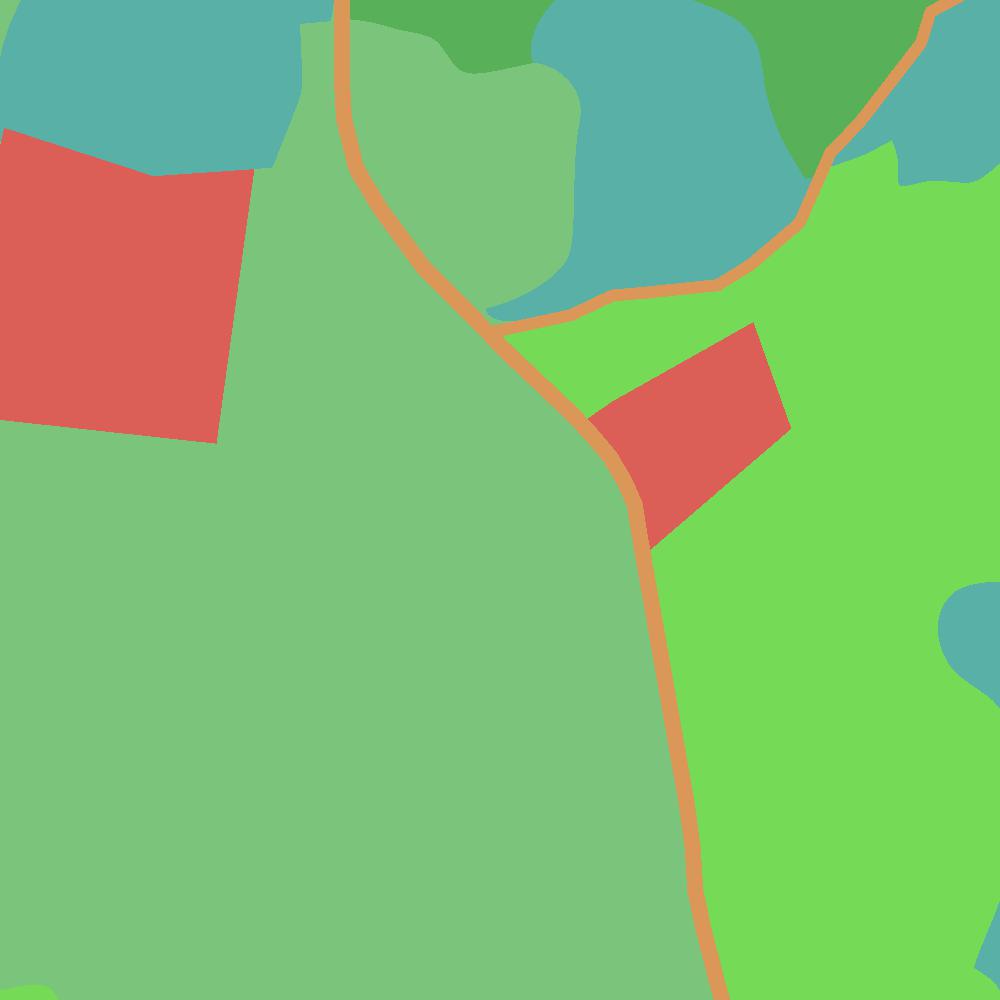} & \includegraphics[width=.09\linewidth]{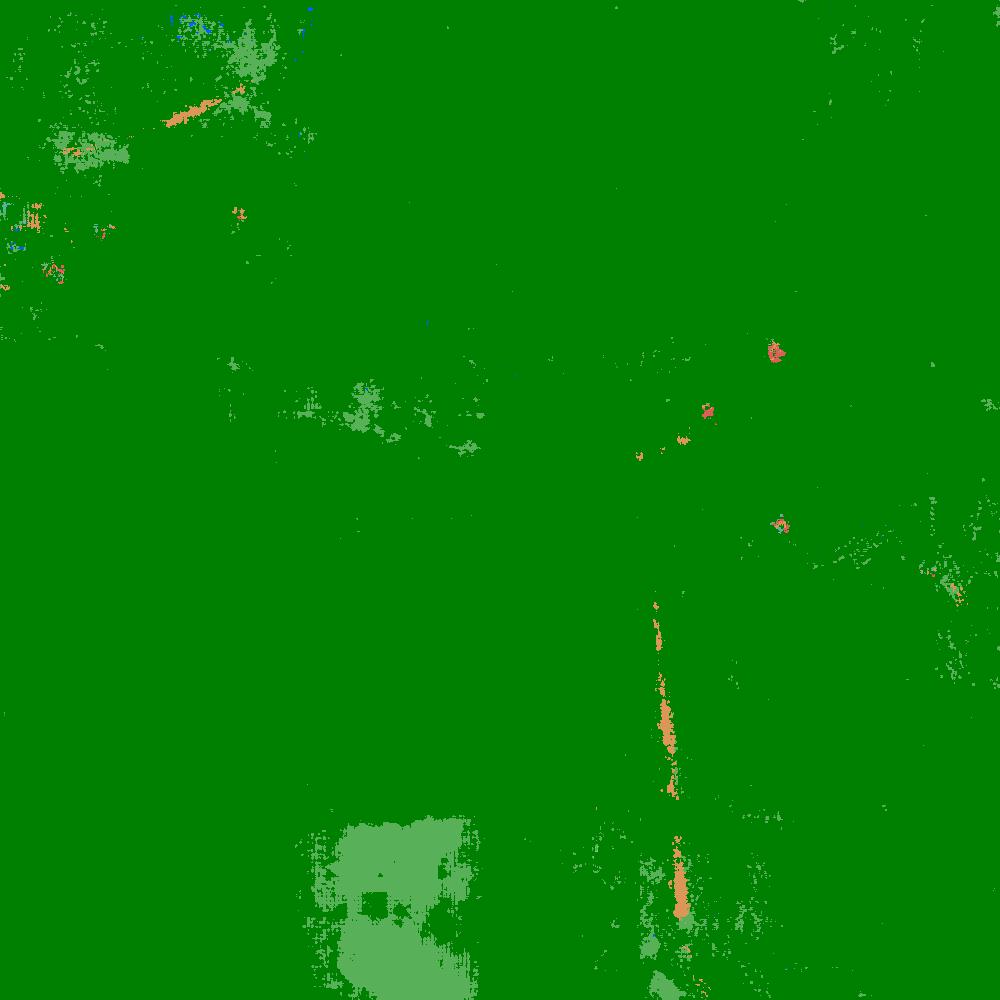} & \includegraphics[width=.09\linewidth]{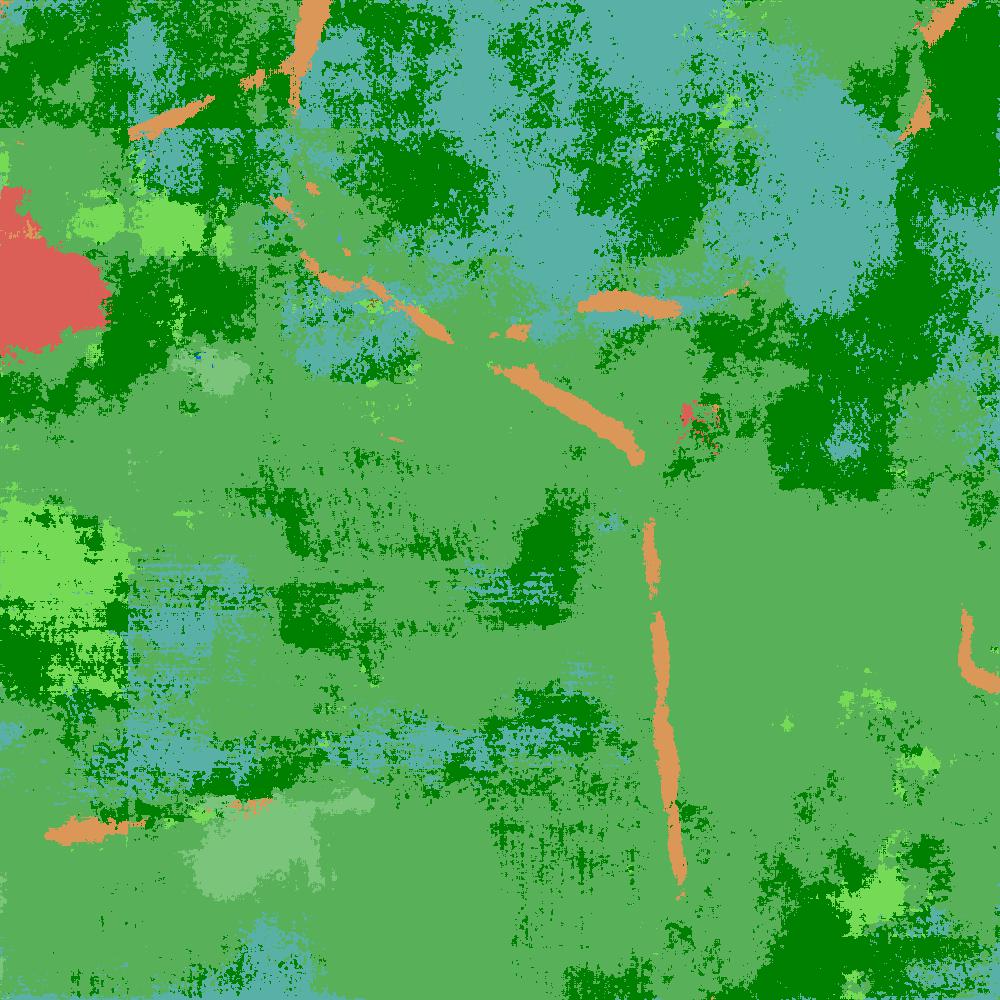} & \includegraphics[width=.09\linewidth]{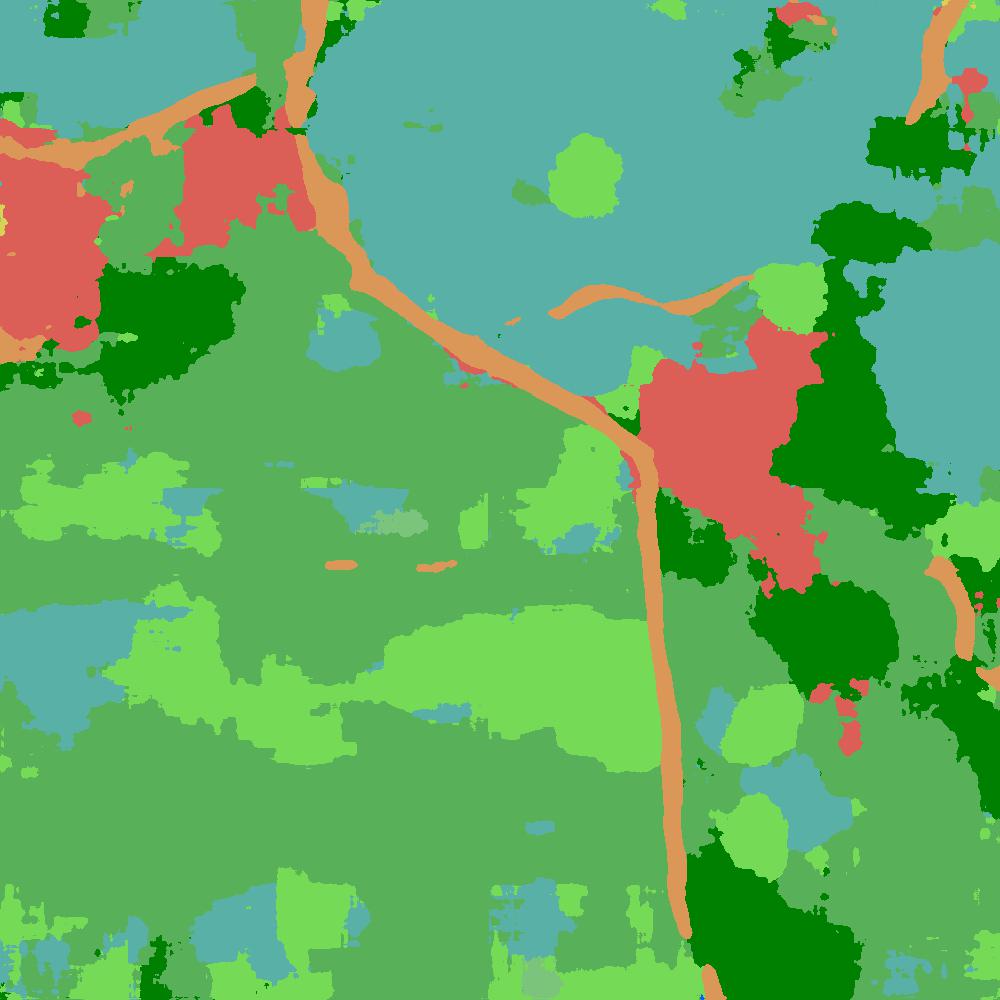} & \includegraphics[width=.09\linewidth]{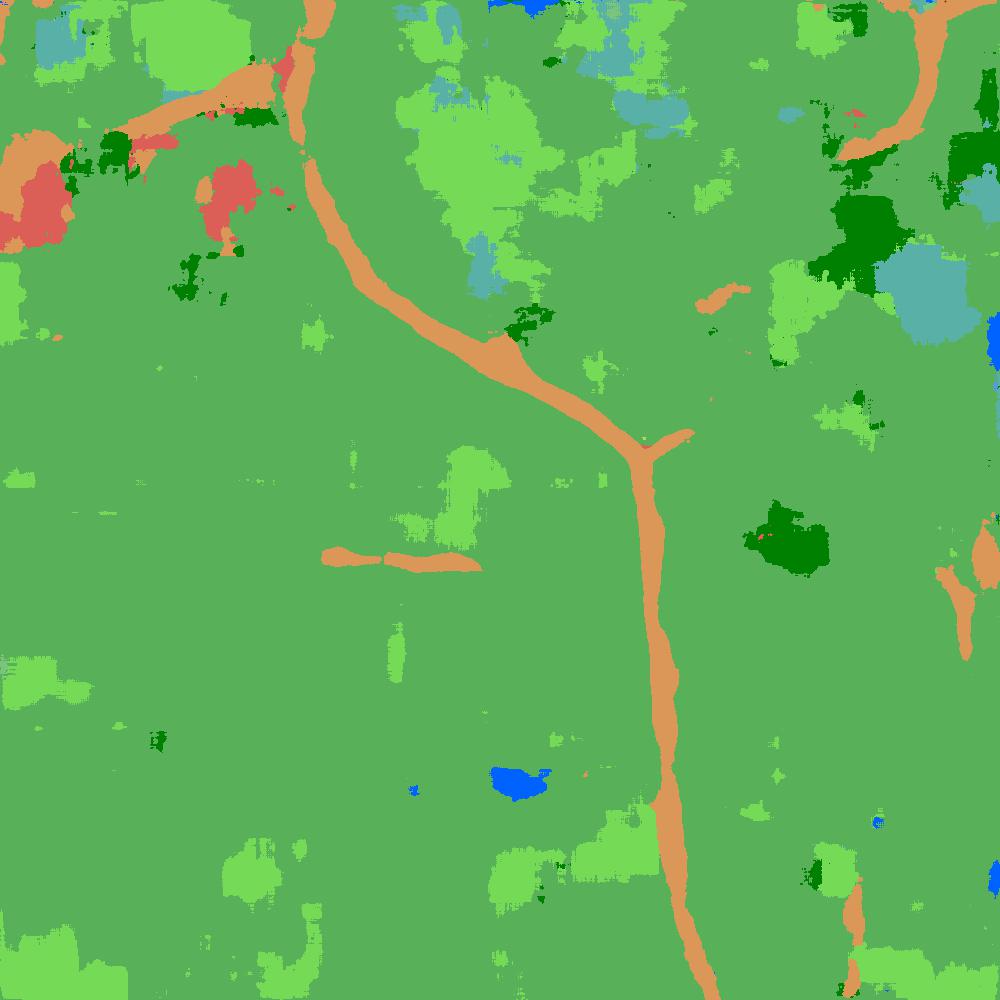} & \includegraphics[width=.09\linewidth]{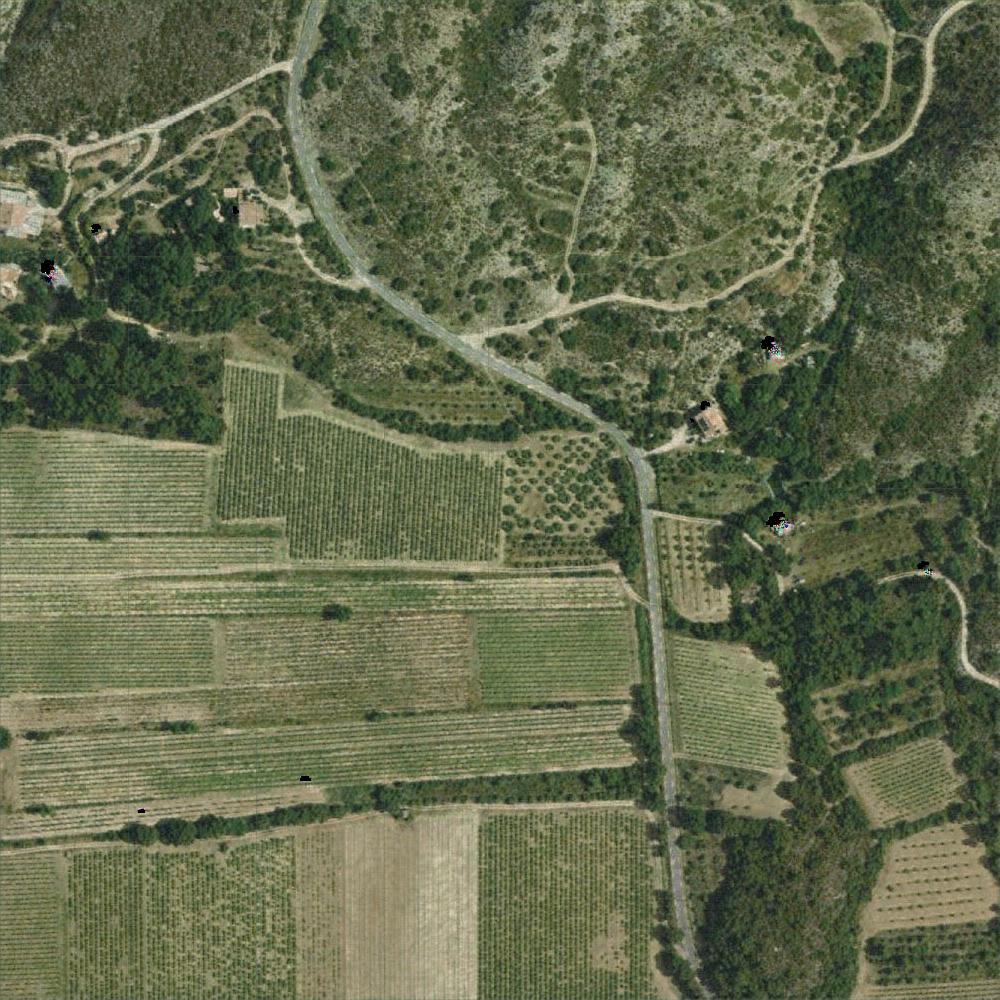} & \includegraphics[width=.09\linewidth]{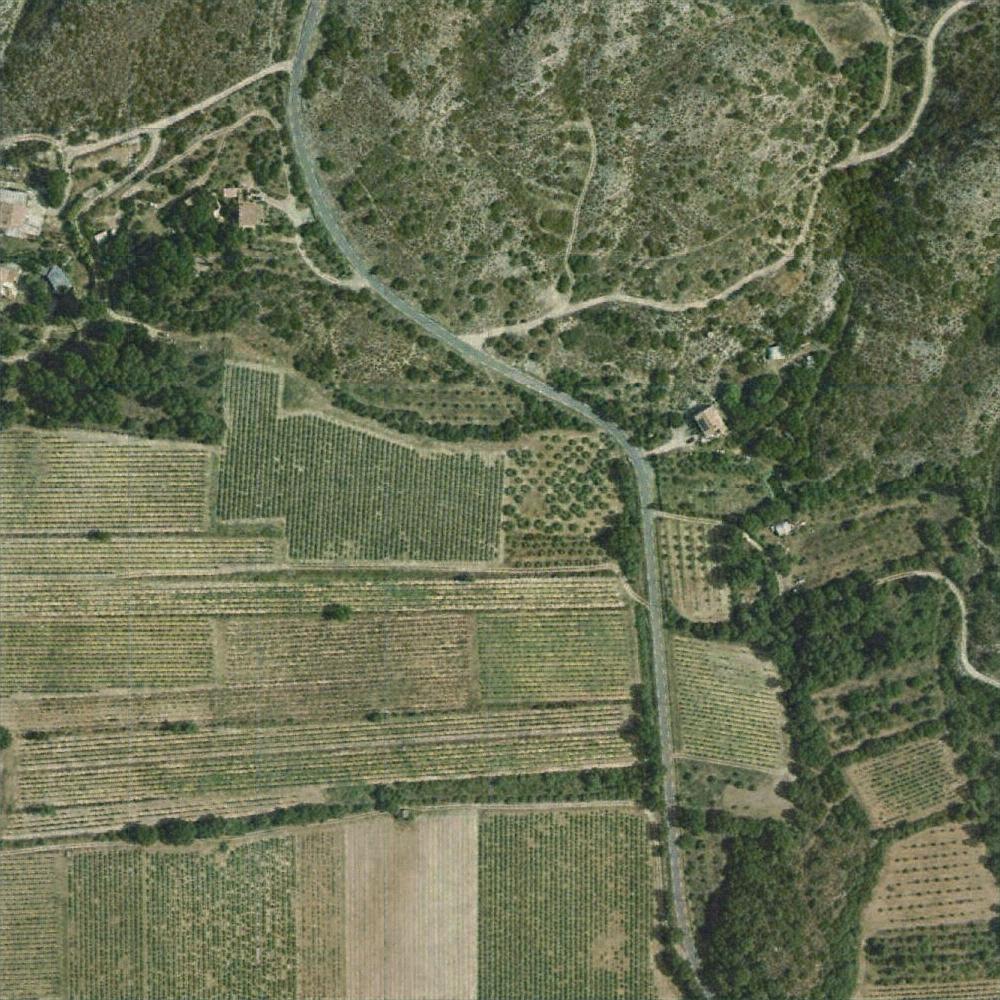} & \includegraphics[width=.09\linewidth]{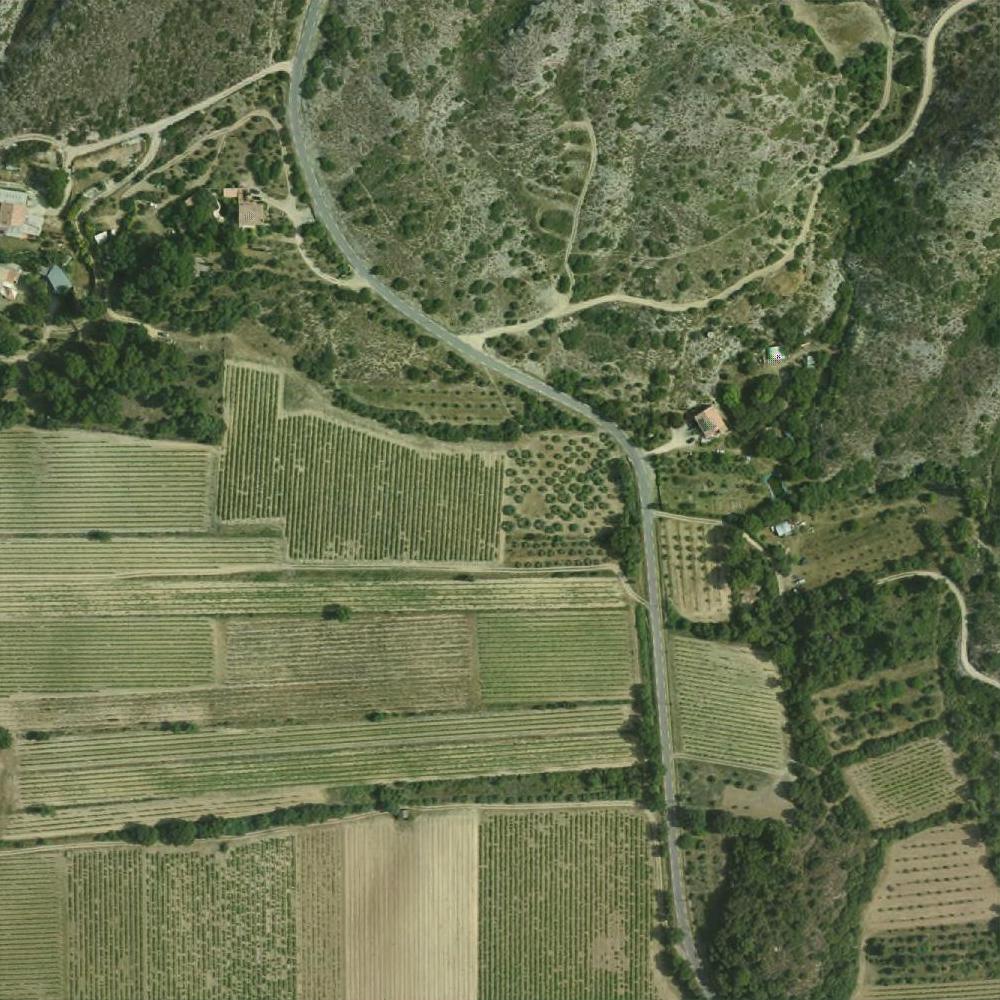} & \includegraphics[width=.09\linewidth]{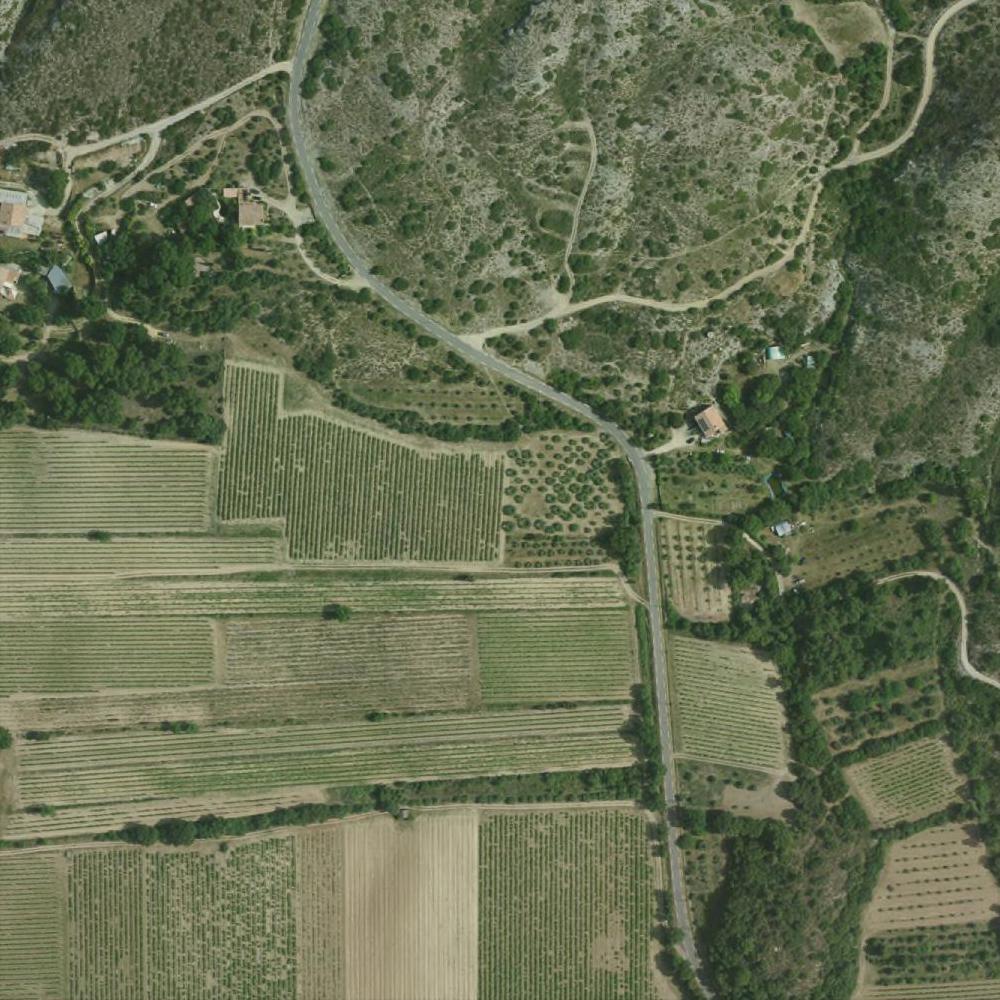} \\
         \includegraphics[width=.09\linewidth]{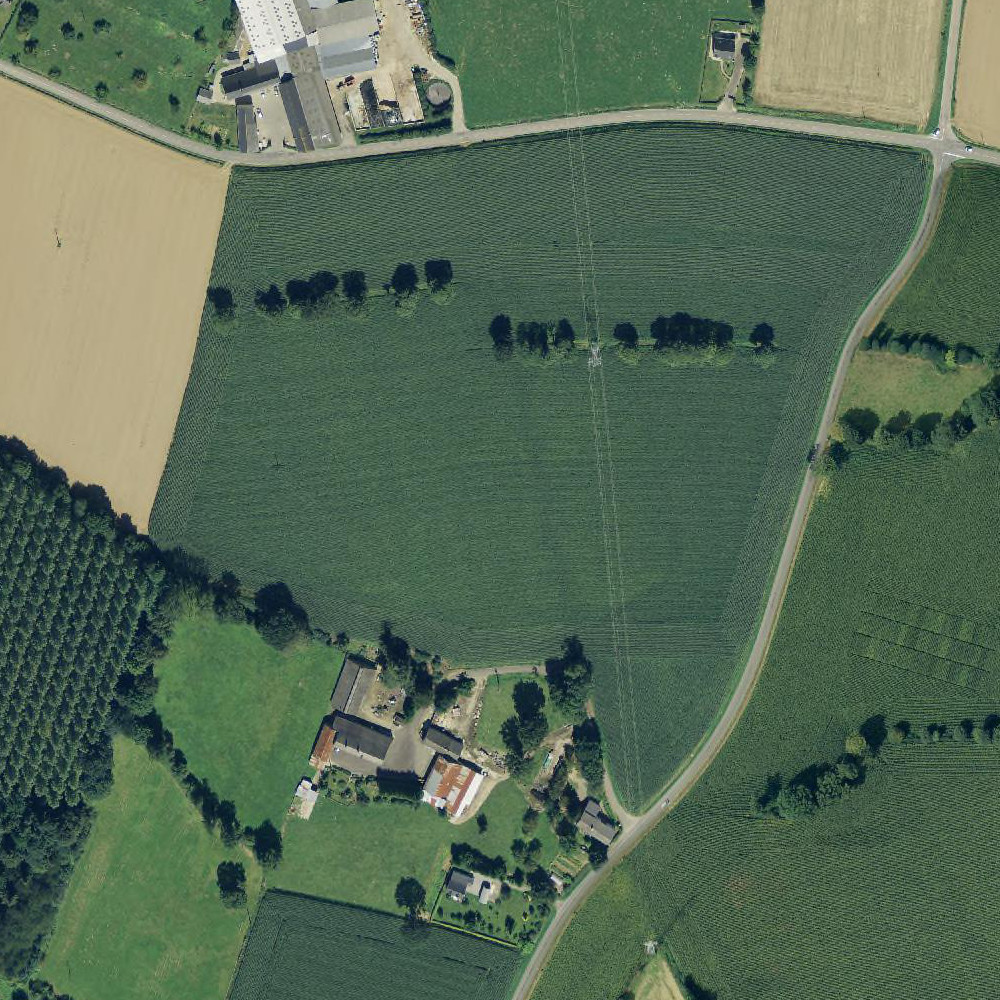} & \includegraphics[width=.09\linewidth]{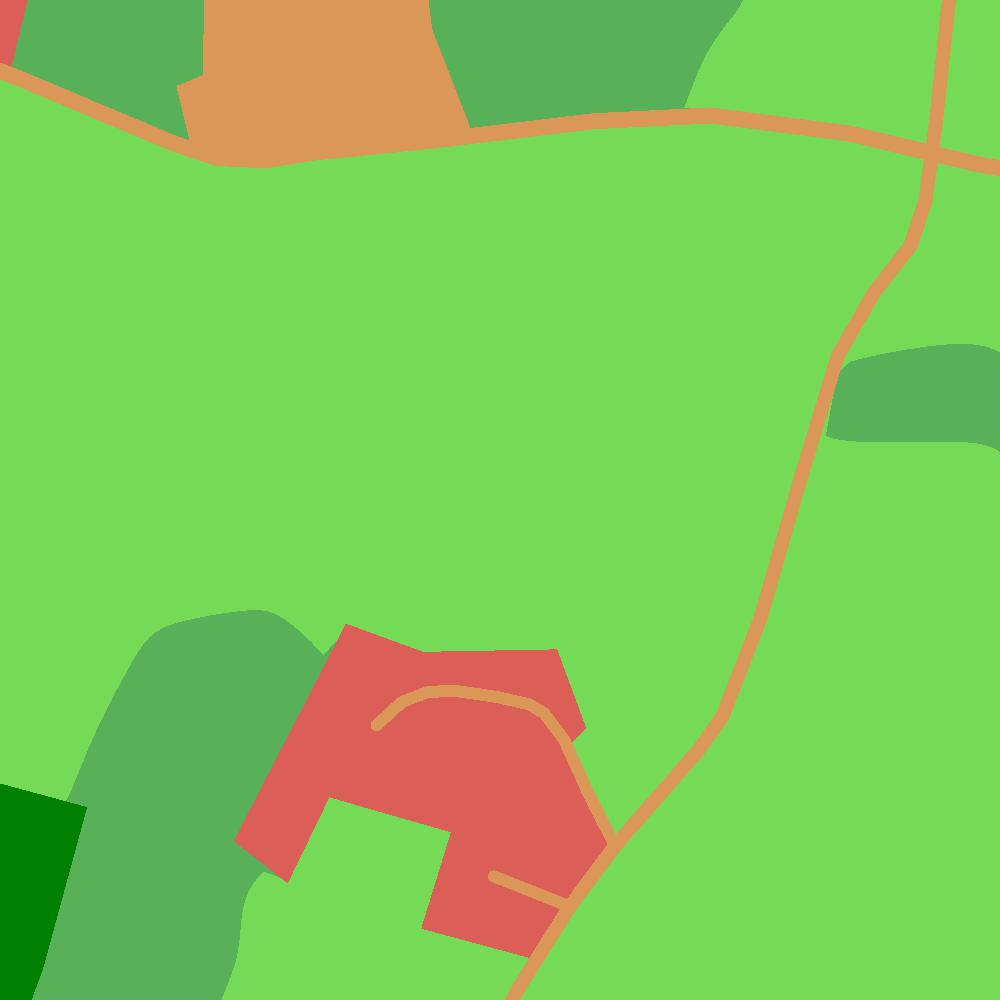} & \includegraphics[width=.09\linewidth]{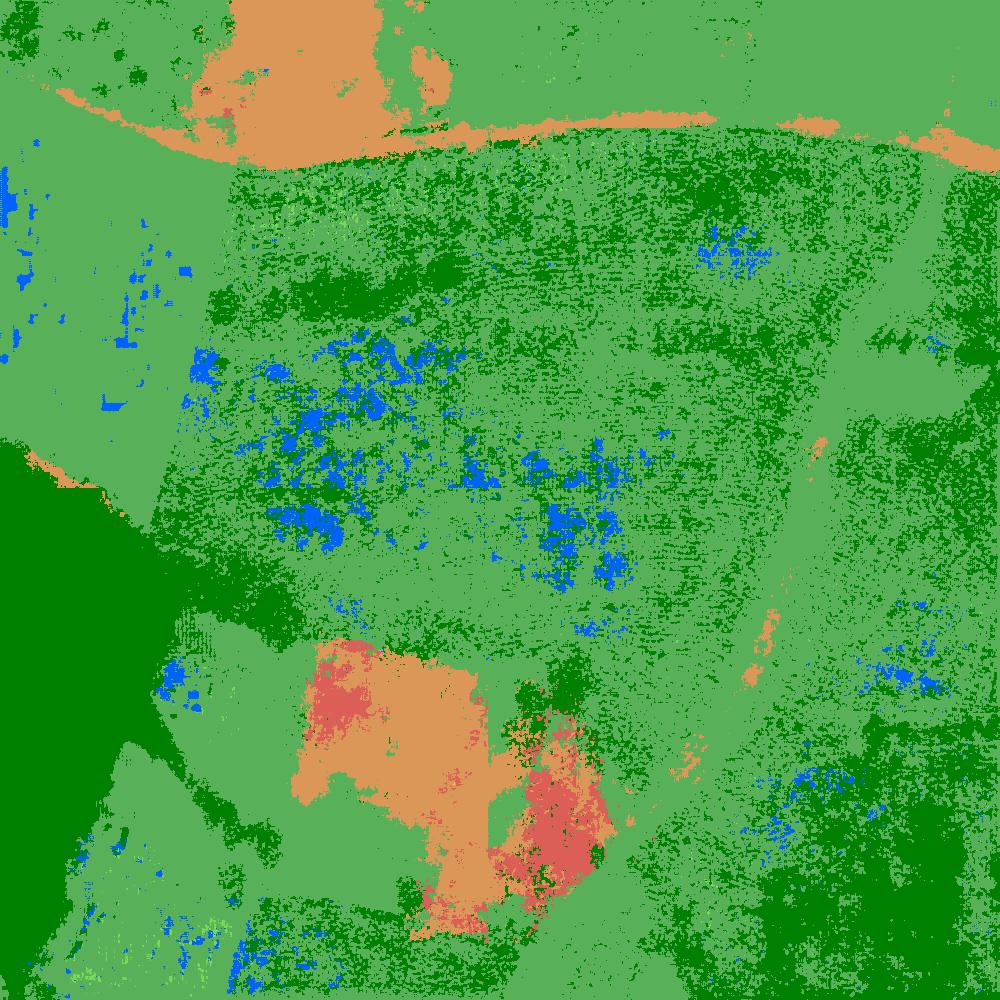} & \includegraphics[width=.09\linewidth]{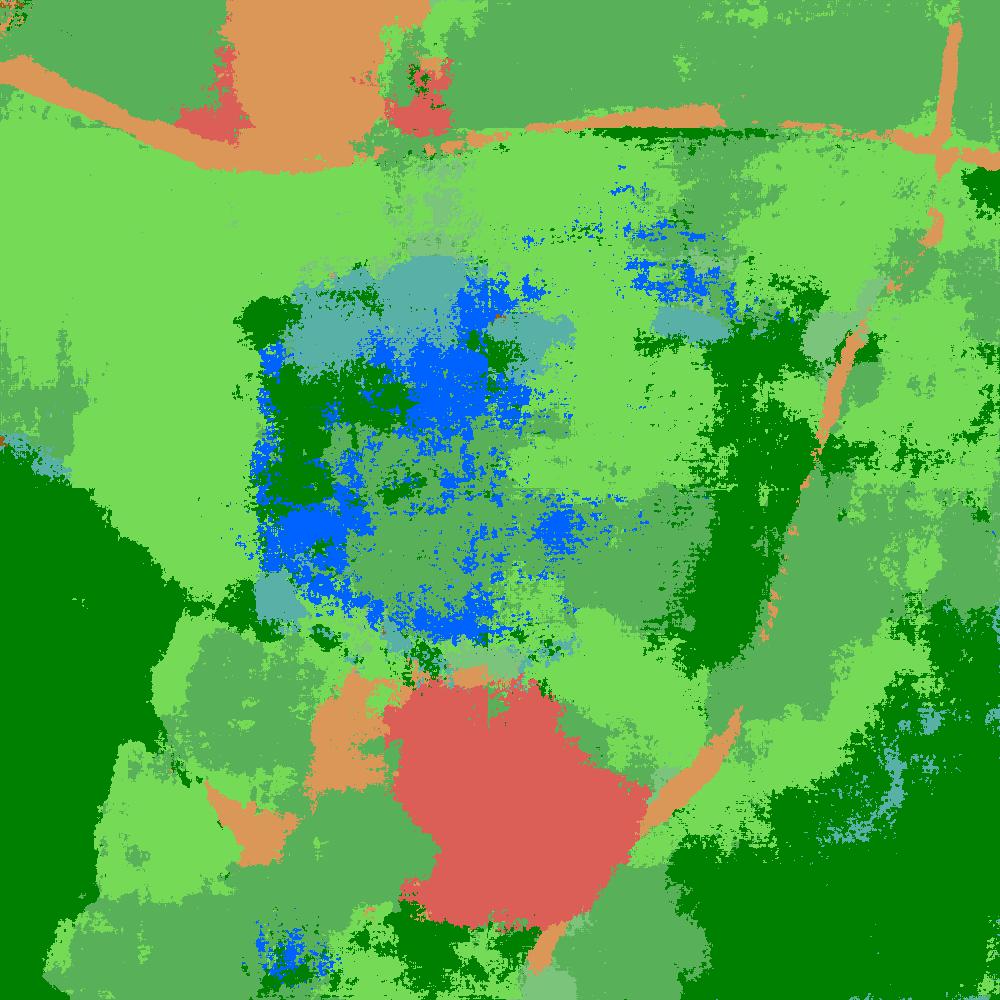} & \includegraphics[width=.09\linewidth]{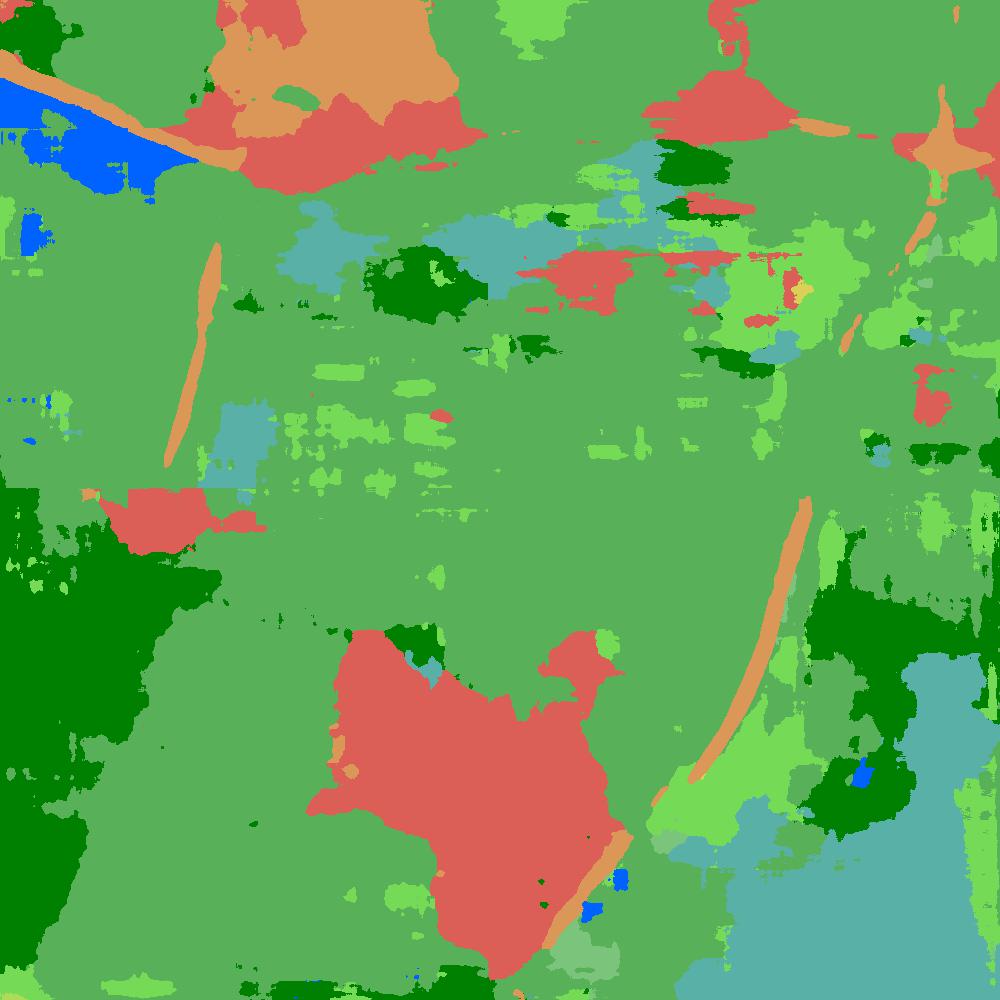} & \includegraphics[width=.09\linewidth]{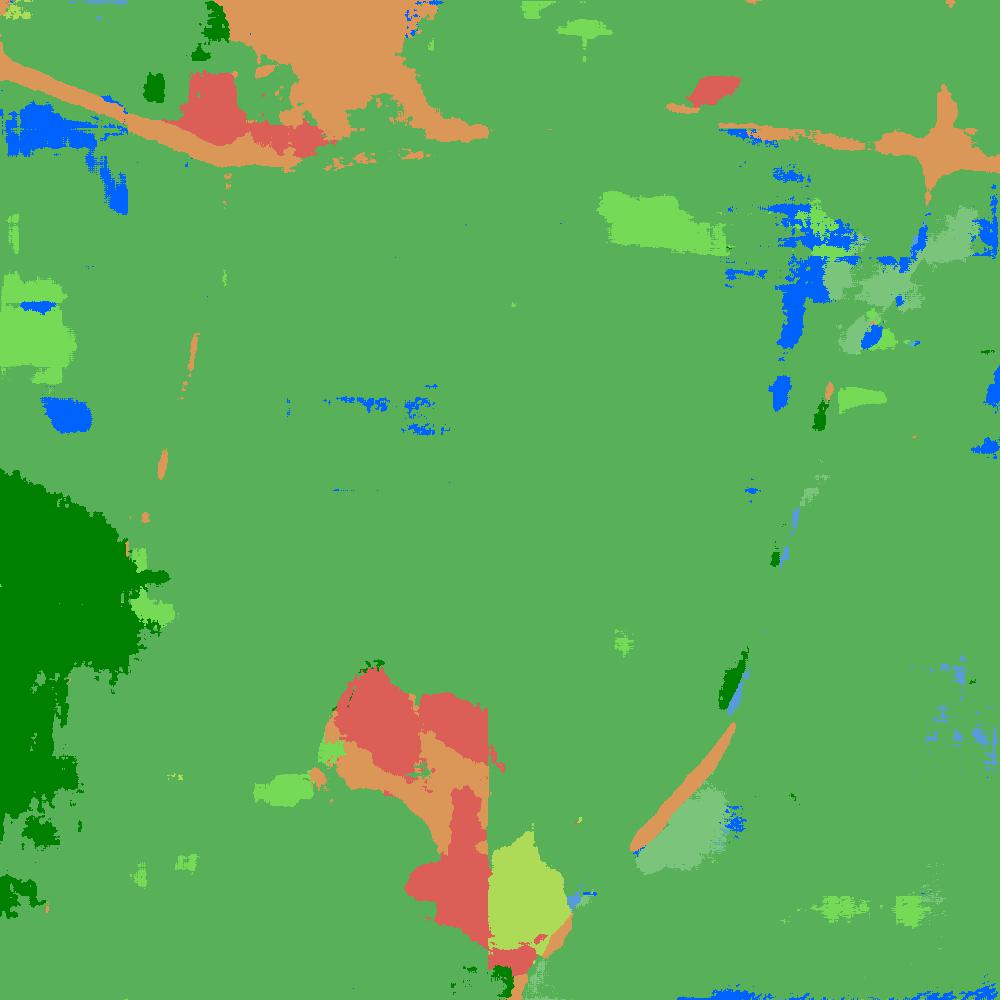} & \includegraphics[width=.09\linewidth]{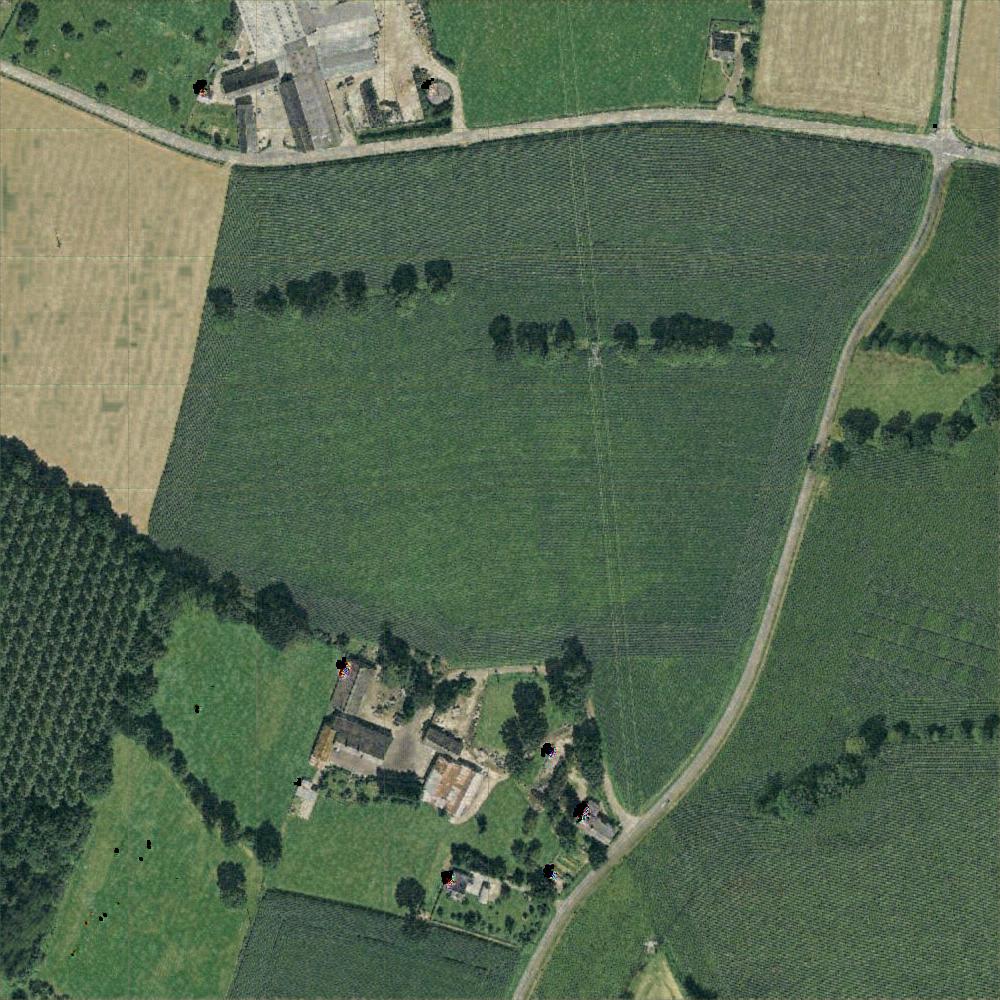} & \includegraphics[width=.09\linewidth]{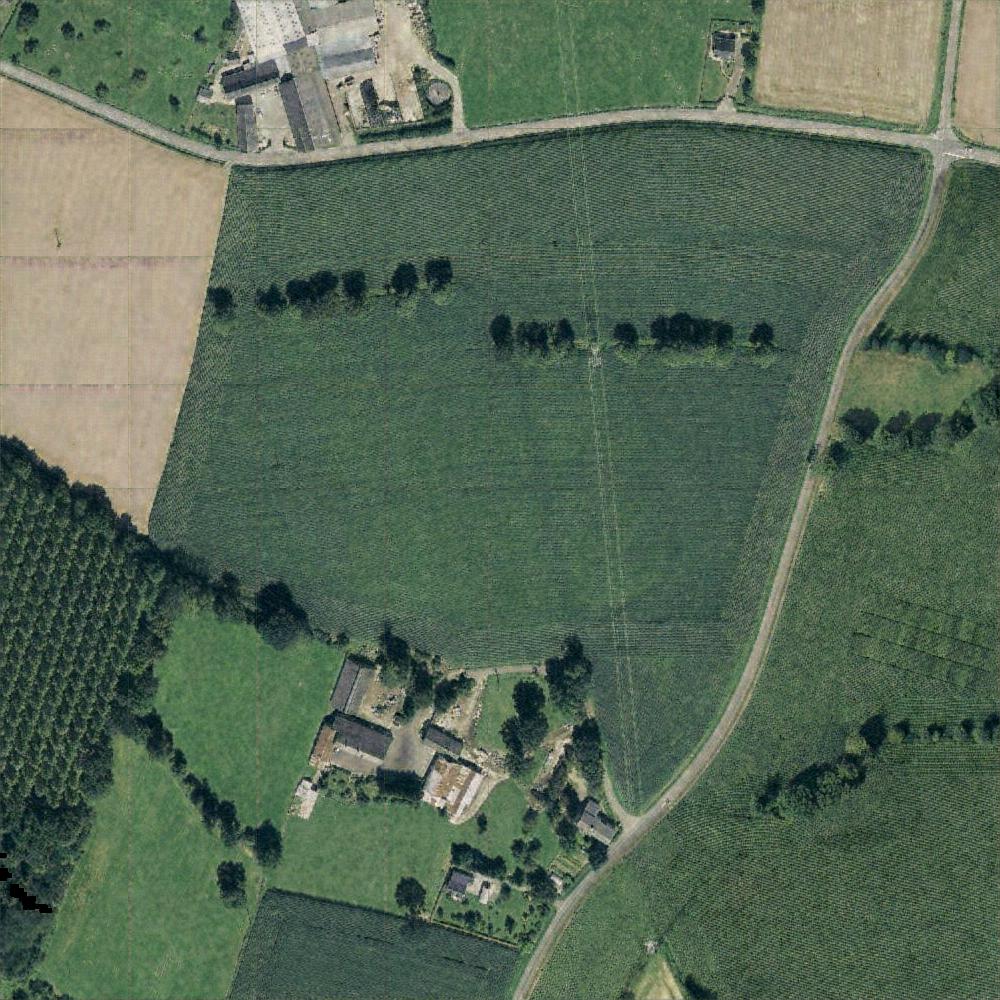} & \includegraphics[width=.09\linewidth]{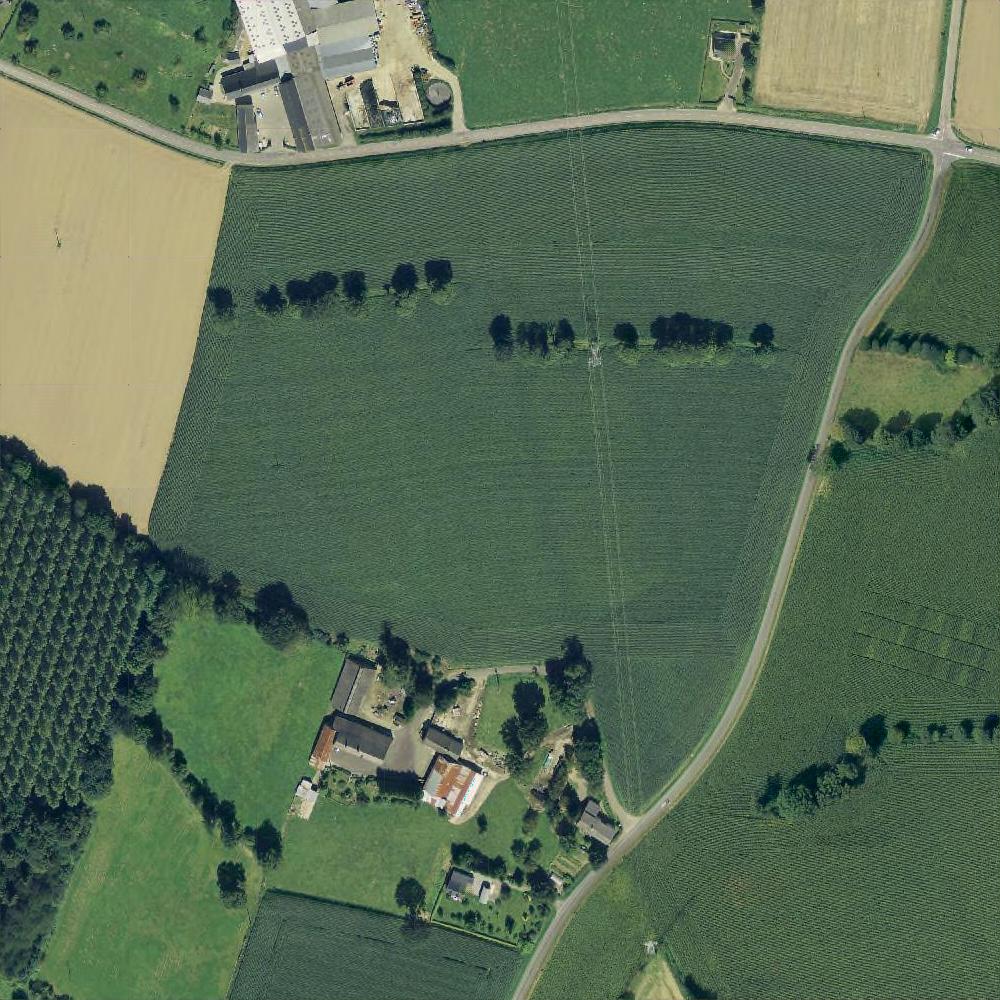} & \includegraphics[width=.09\linewidth]{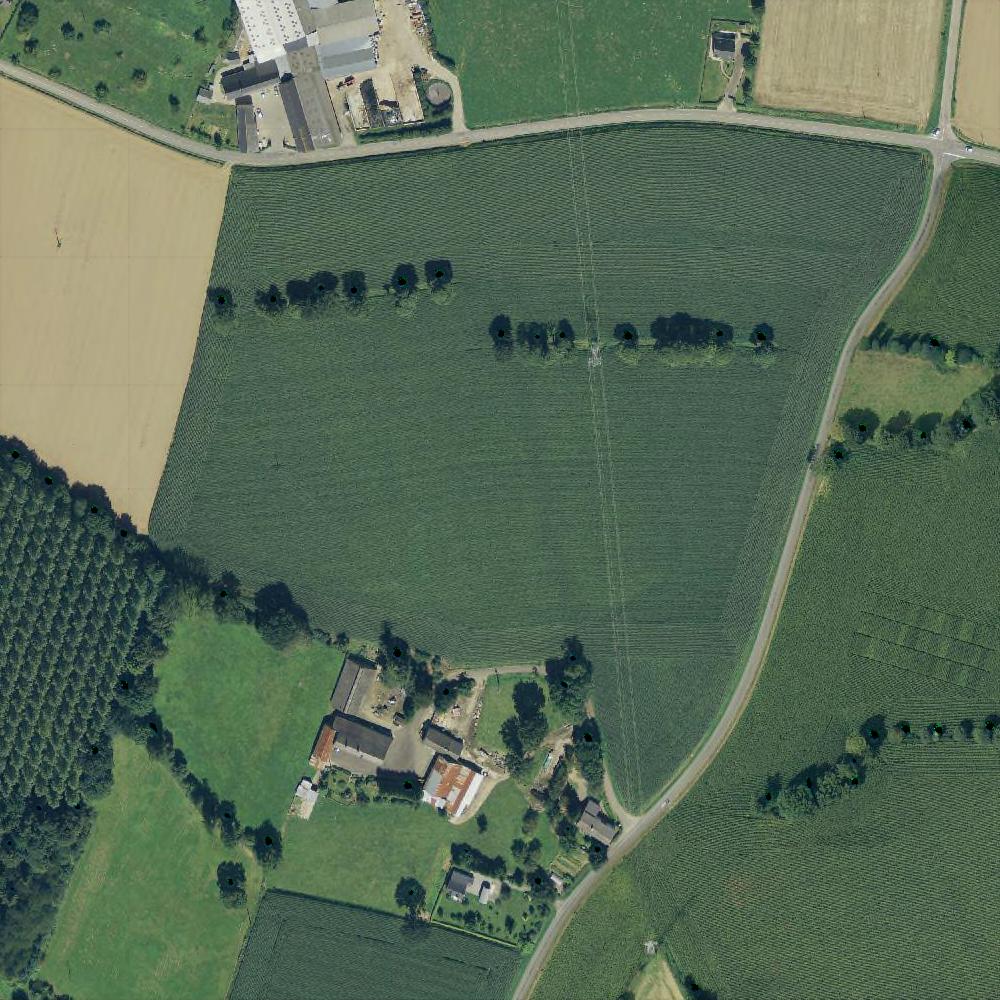} \\
         \includegraphics[width=.09\linewidth]{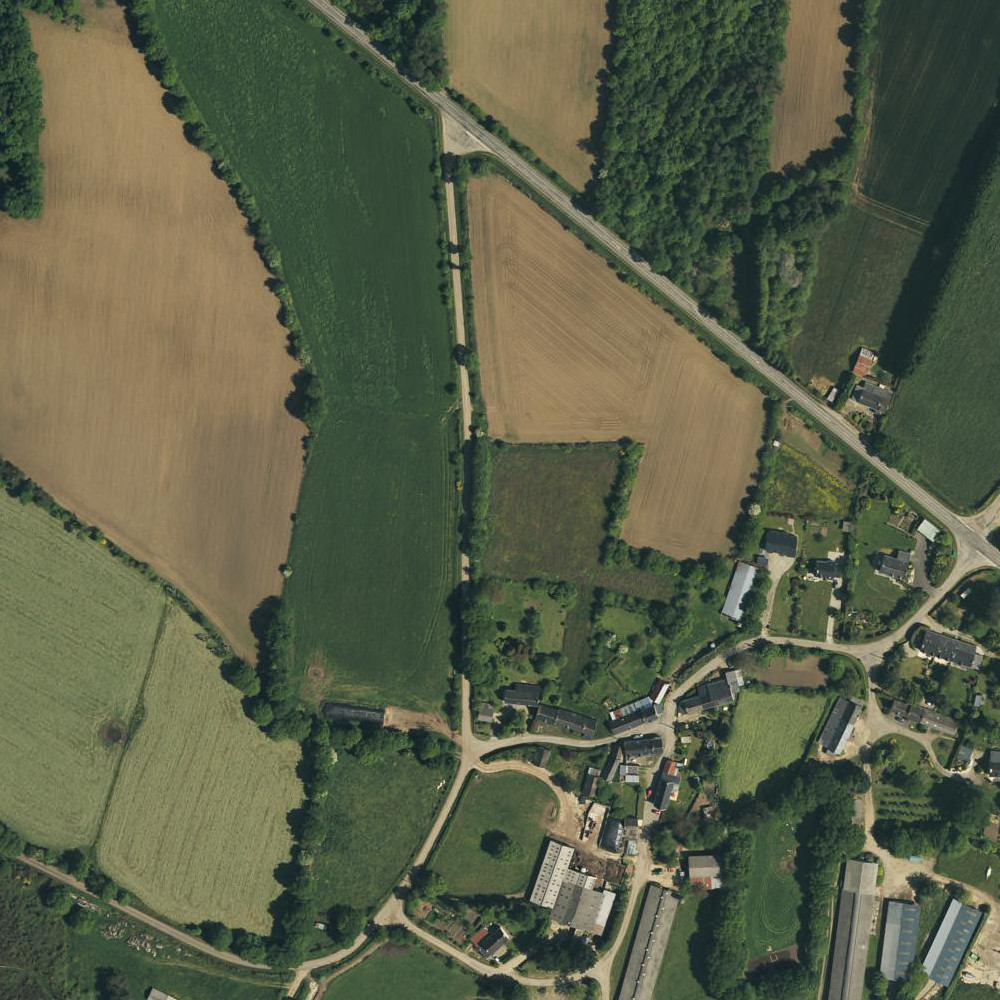} & \includegraphics[width=.09\linewidth]{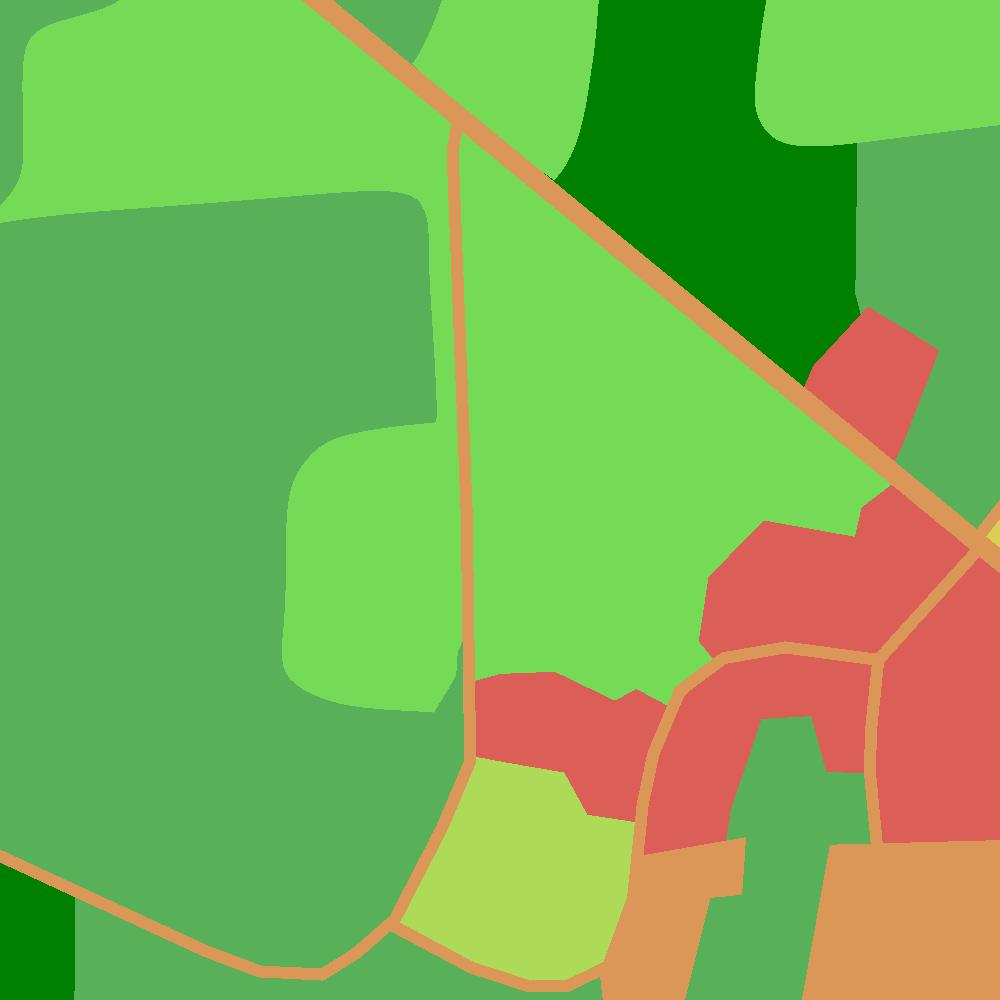} & \includegraphics[width=.09\linewidth]{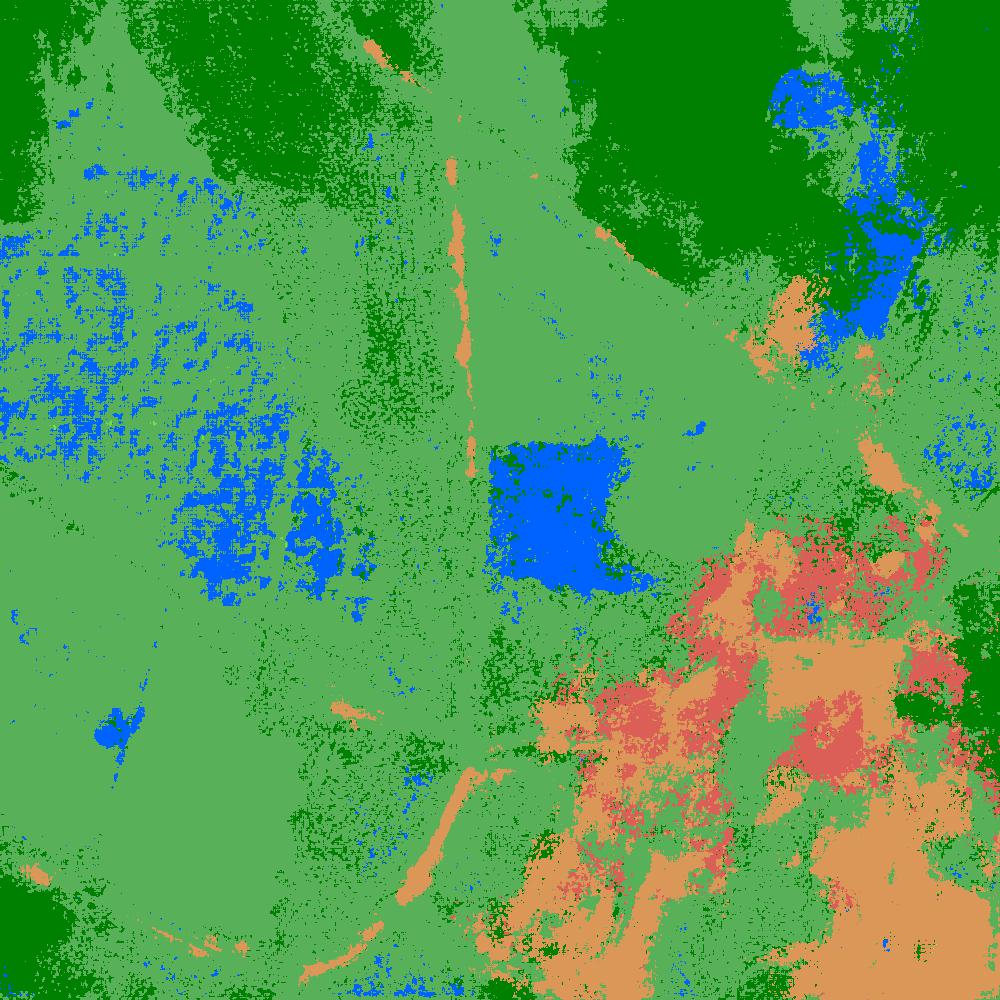} & \includegraphics[width=.09\linewidth]{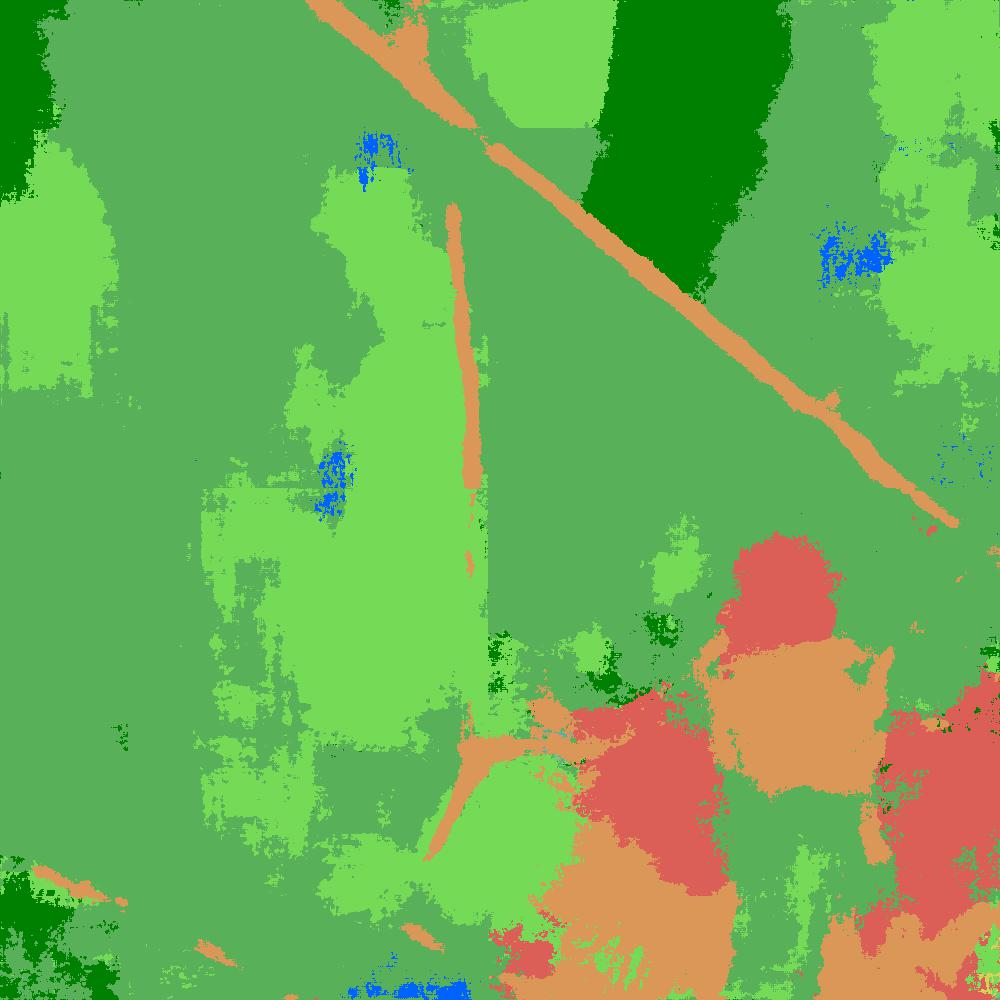} & \includegraphics[width=.09\linewidth]{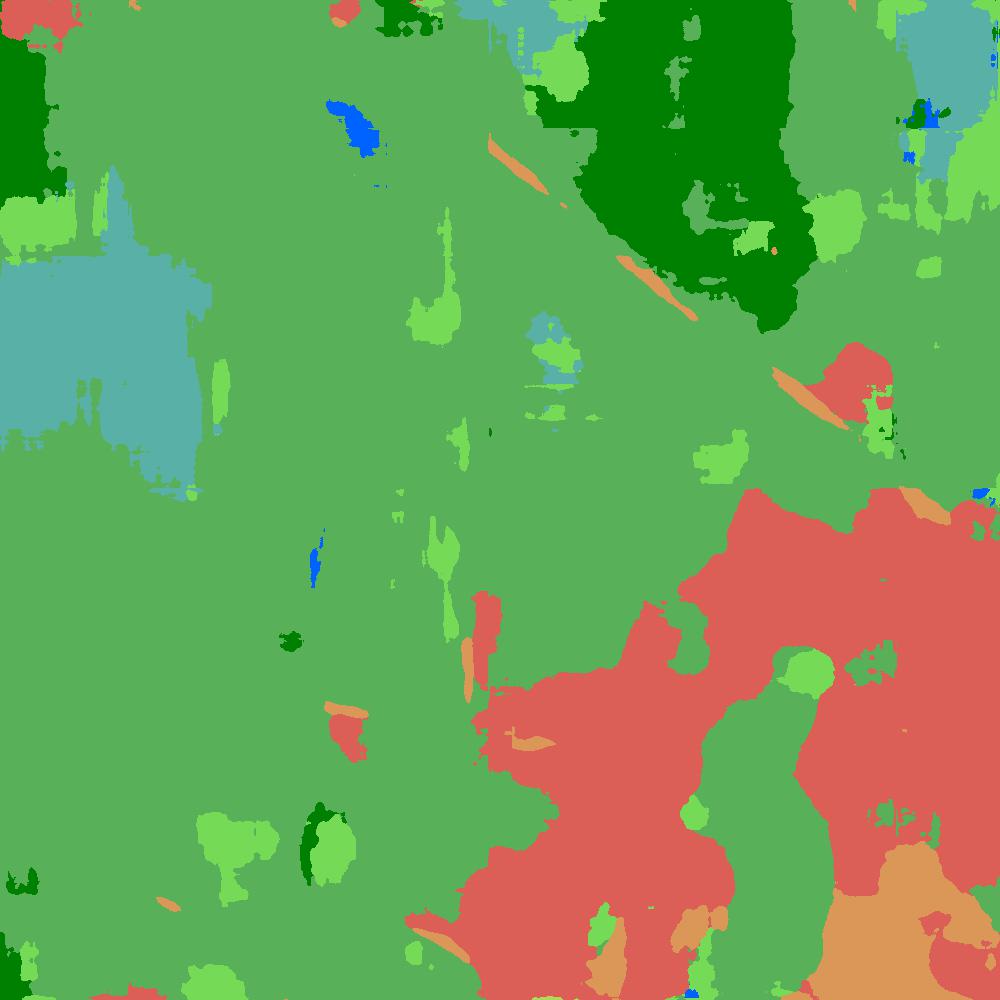} & \includegraphics[width=.09\linewidth]{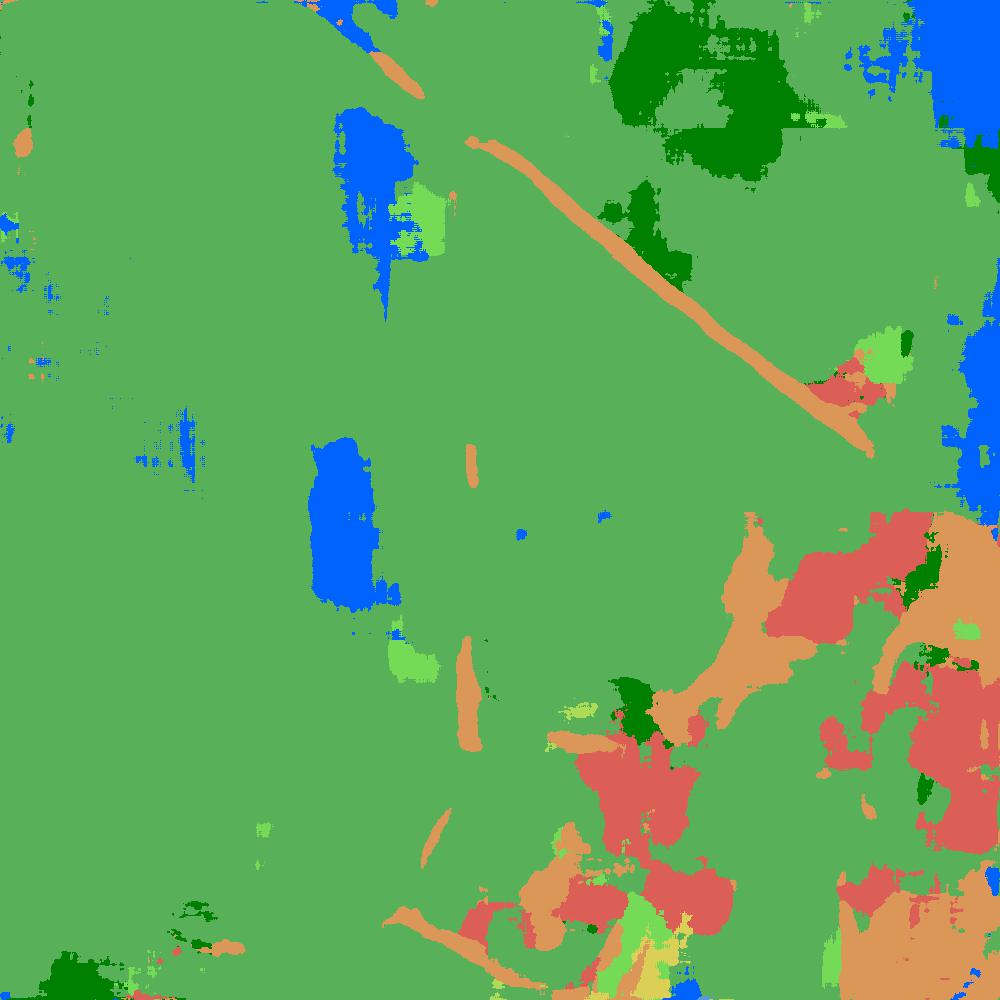} & \includegraphics[width=.09\linewidth]{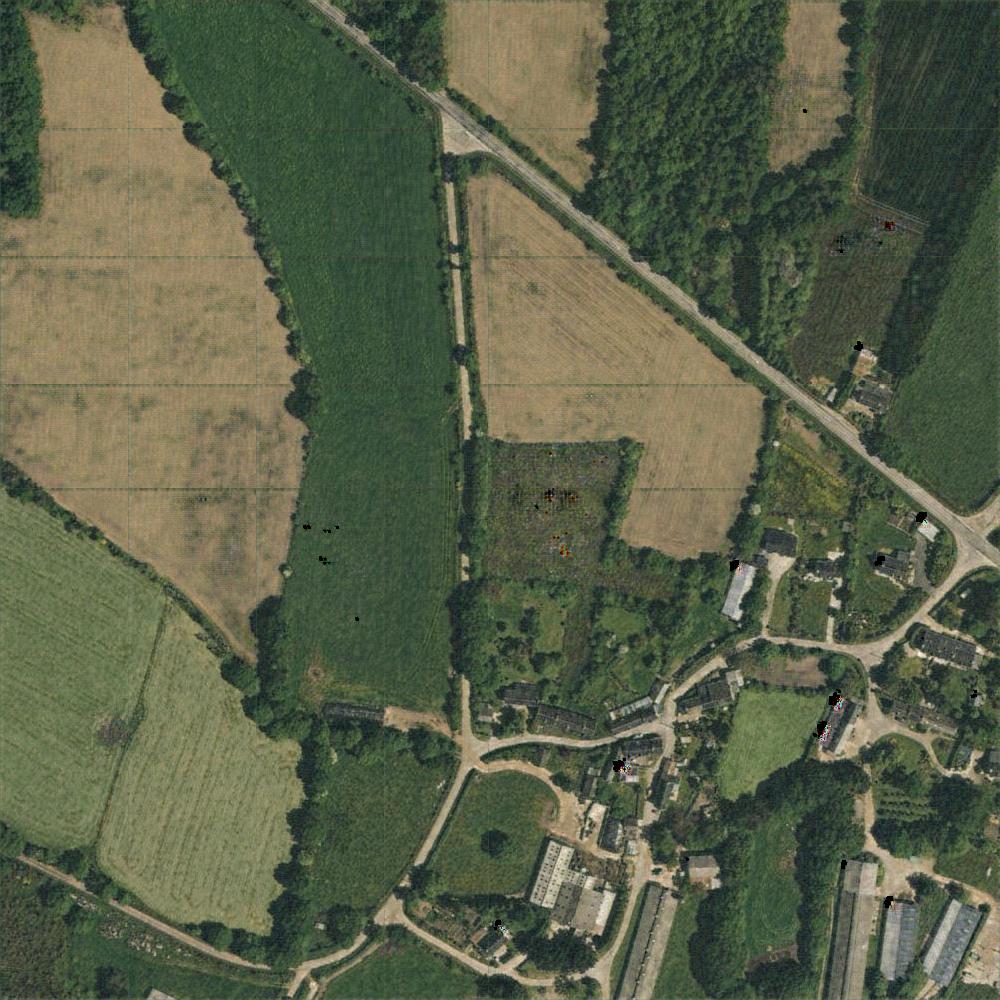} & \includegraphics[width=.09\linewidth]{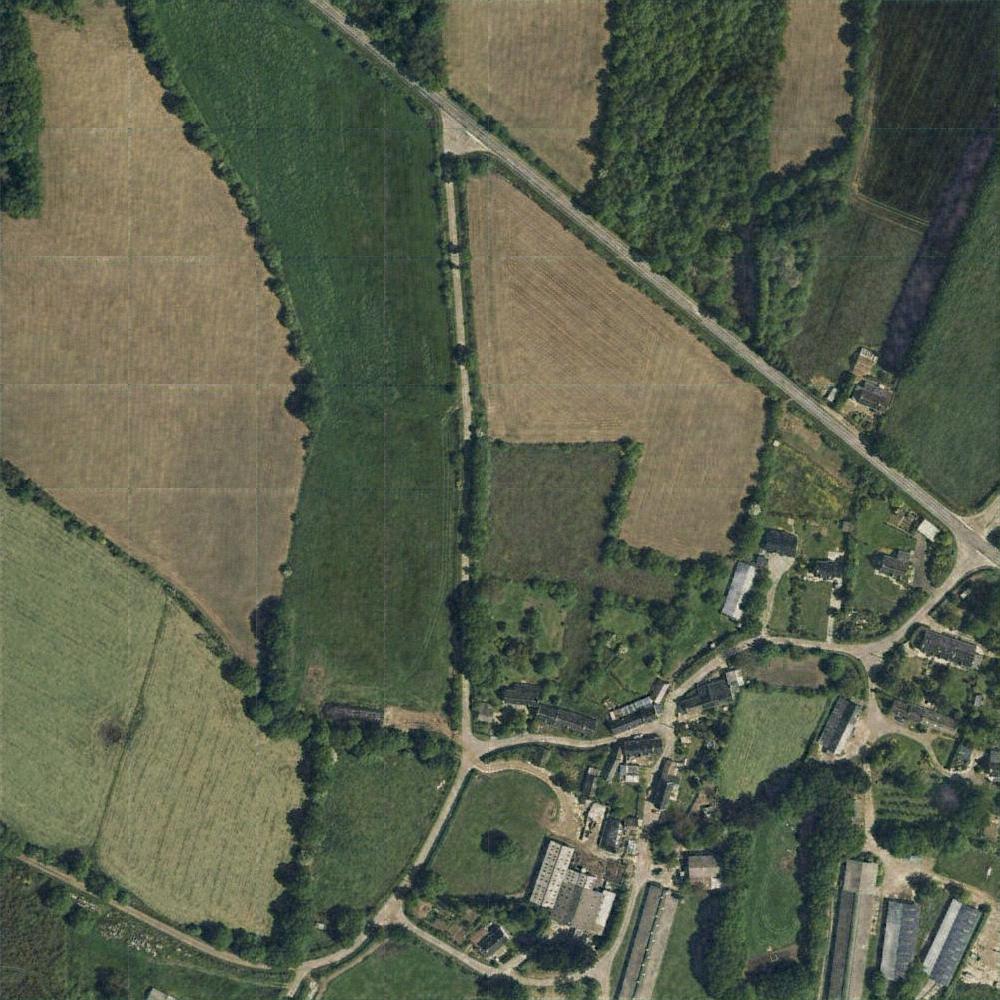} & \includegraphics[width=.09\linewidth]{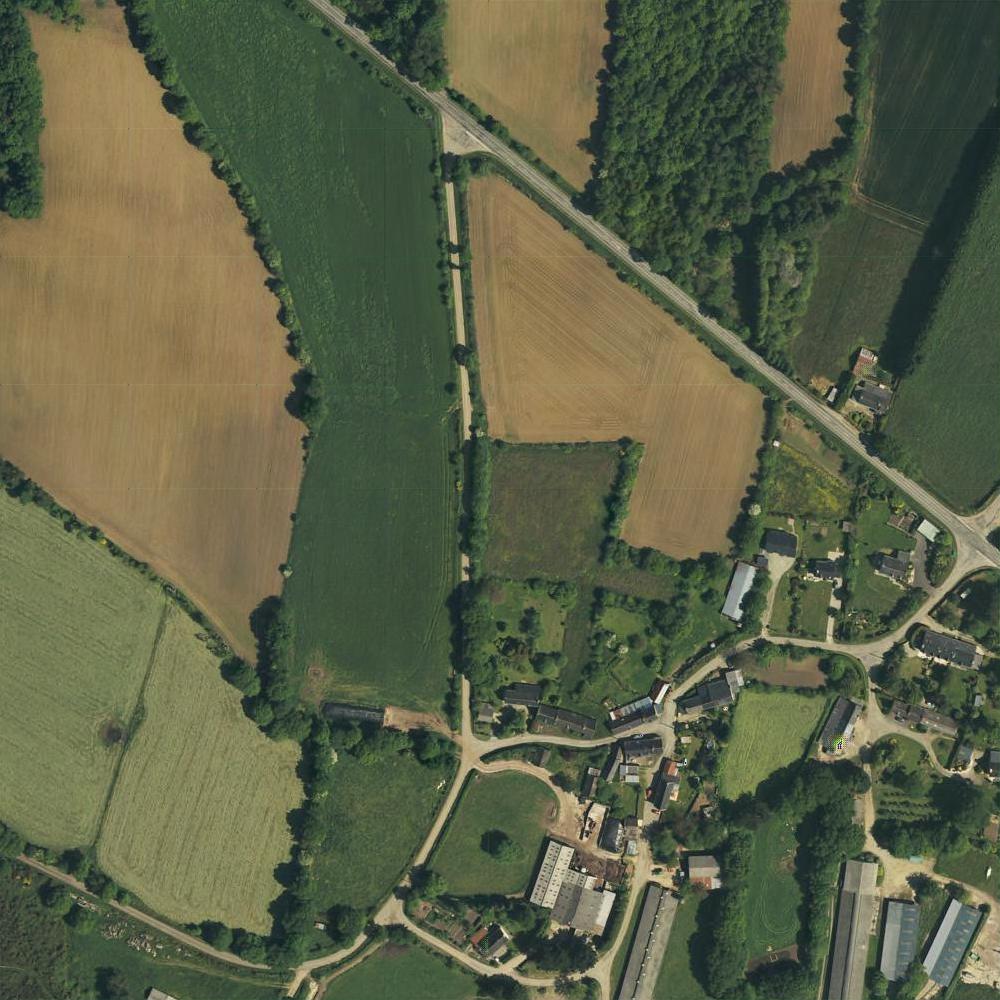} & \includegraphics[width=.09\linewidth]{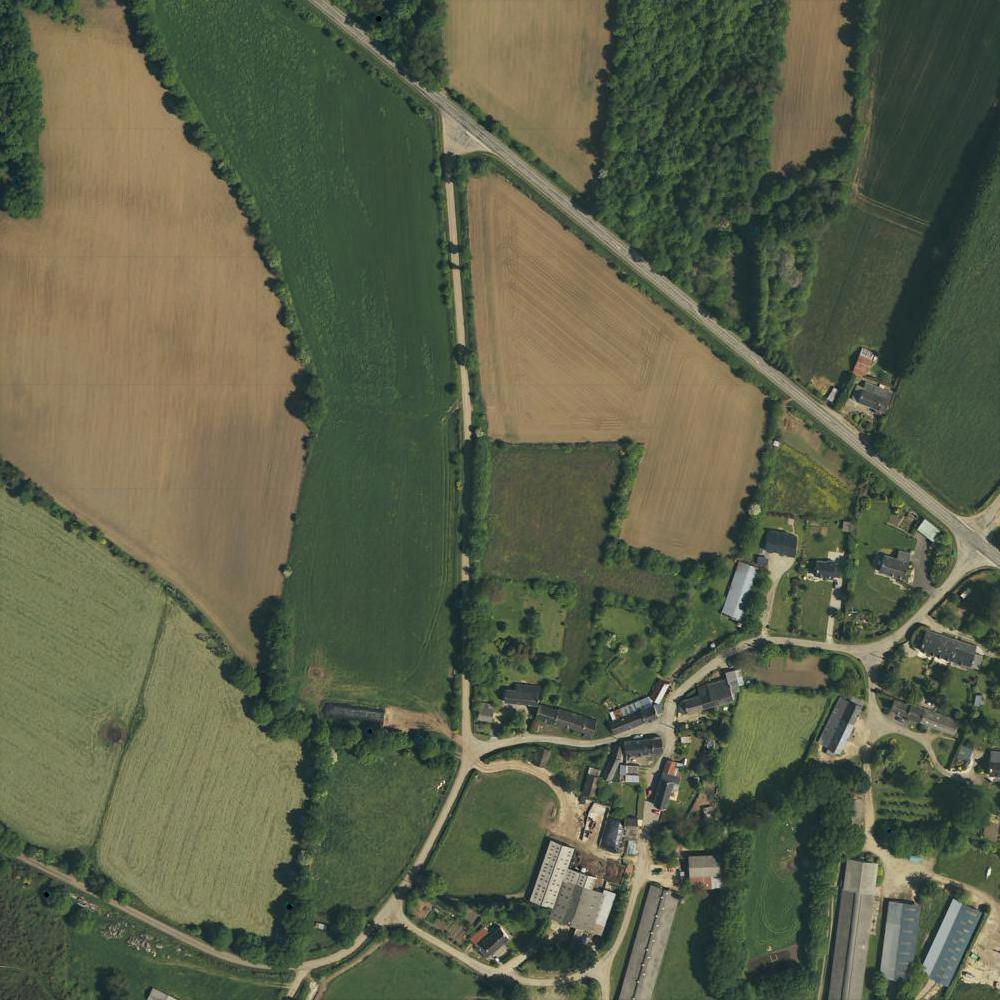}

         \\[3pt]

        && \scriptsize{BN-e} & \scriptsize{BN-l-S} & \scriptsize{BN-l-U} & \scriptsize{W-Net} & \scriptsize{BN-e} & \scriptsize{BN-l-S} & \scriptsize{BN-l-U} & \scriptsize{W-Net}\\
Image & GT & \multicolumn{4}{c}{Semantic Segmentation} & \multicolumn{4}{c}{Reconstruction}  
      \end{tabular}
      \caption{Results comparison for different neural network architectures with reconstruction as auxiliary task ($\LL_1$ auxiliary loss).  BN-e stands for BerundaNet-early, BN-l-S/BN-l-U for BerundaNet-late with SegNet/U-Net backbone, respectively.} \label{fig: results-different-networks-reconstruction}
   \end{center}
\end{figure}

\begin{figure}[!htbp]
   \begin{center}
      \setlength{\tabcolsep}{1pt}
      \begin{tabular}{cc@{\hspace{8pt}}cccc@{\hspace{8pt}}cccc}
         \includegraphics[width=.09\linewidth]{13-2014-0880-6290-LA93-0M50-E080_2786_1082.jpg} & \includegraphics[width=.09\linewidth]{13-2014-0880-6290-LA93-0M50-E080_2786_1082_gt.jpg} & \includegraphics[width=.09\linewidth]{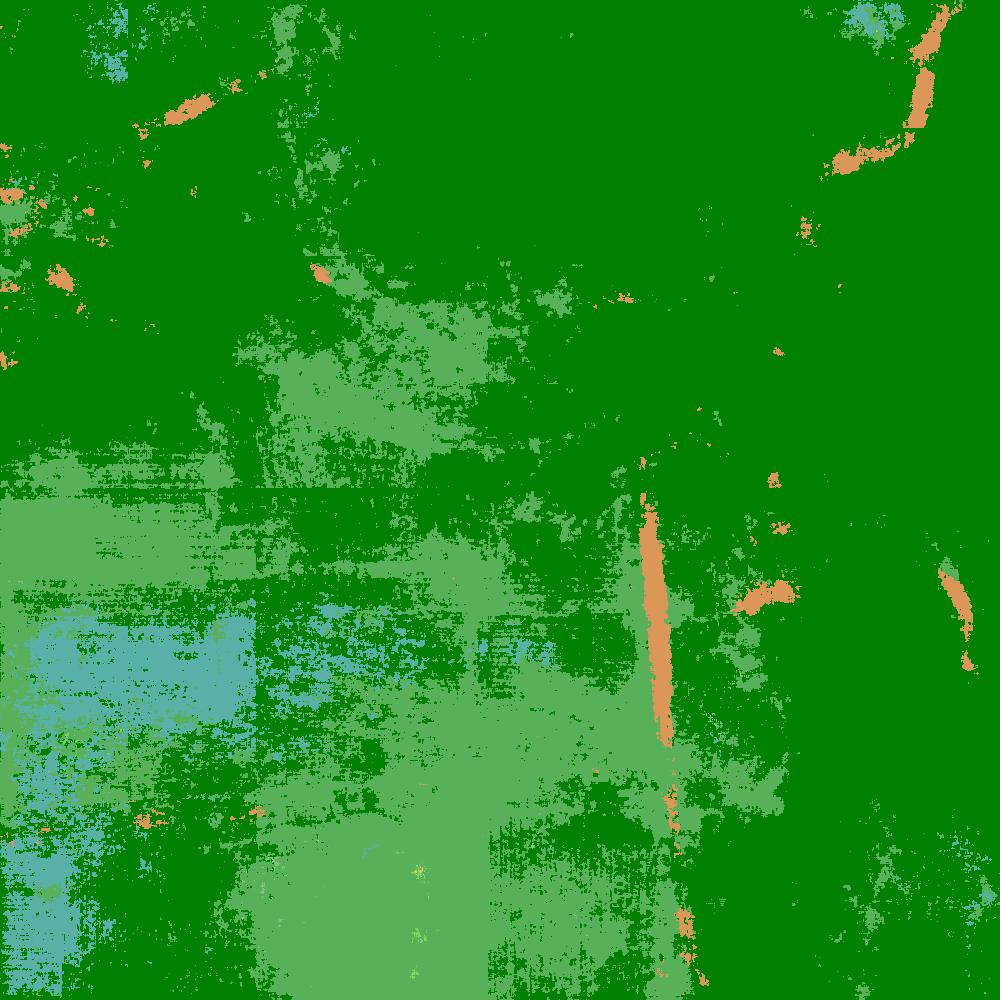} & \includegraphics[width=.09\linewidth]{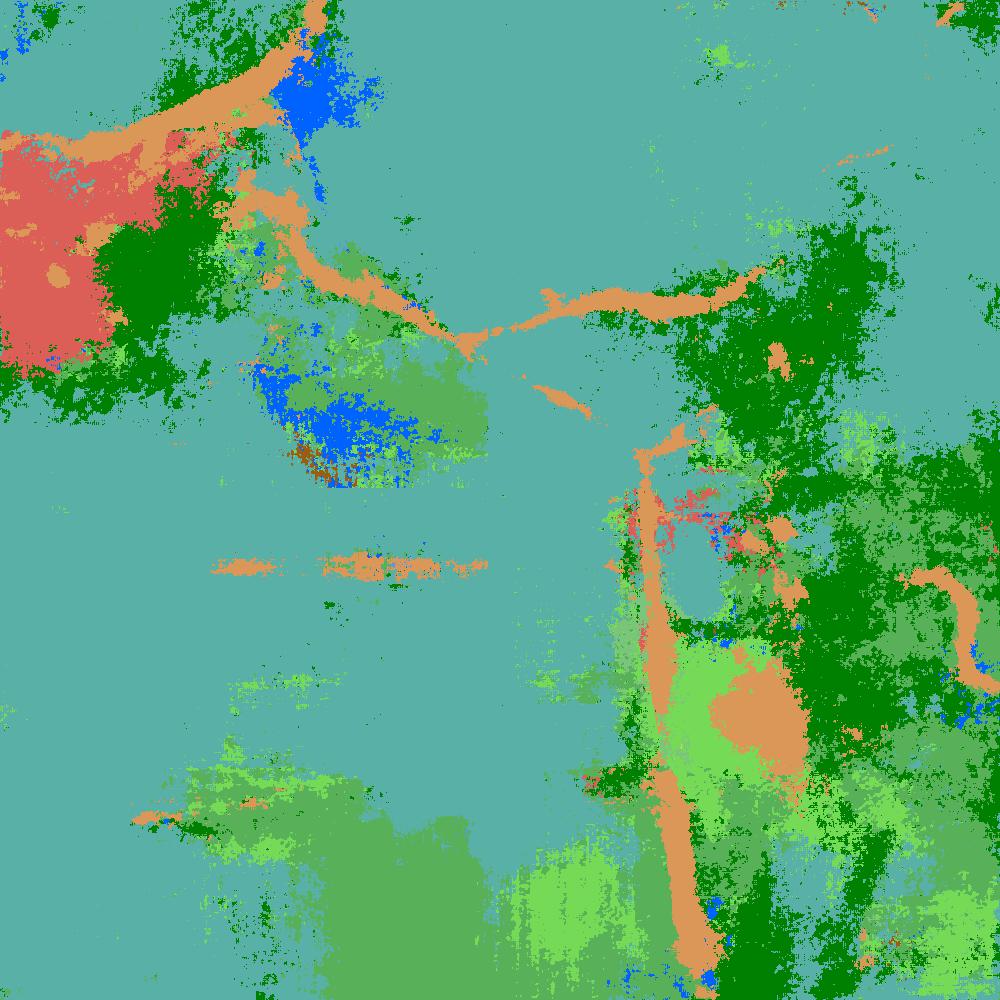}
         & \includegraphics[width=.09\linewidth]{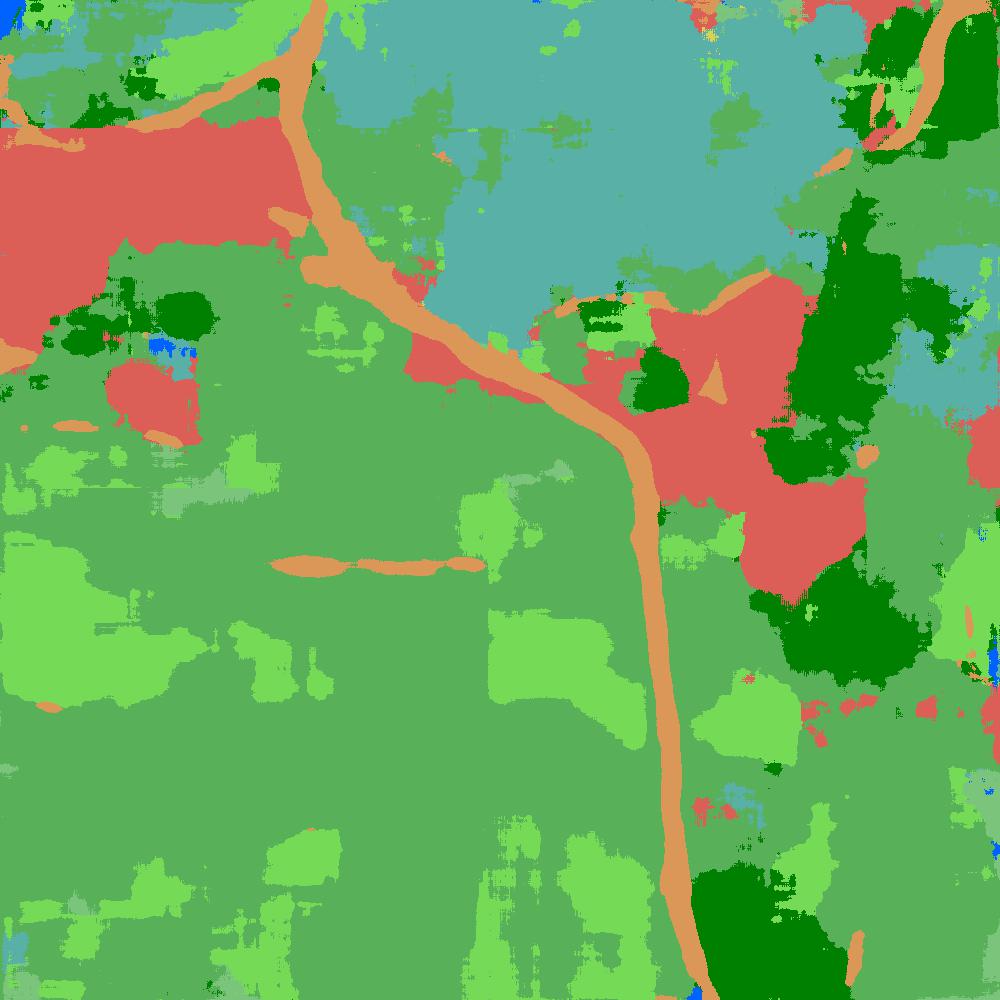} & \includegraphics[width=.09\linewidth]{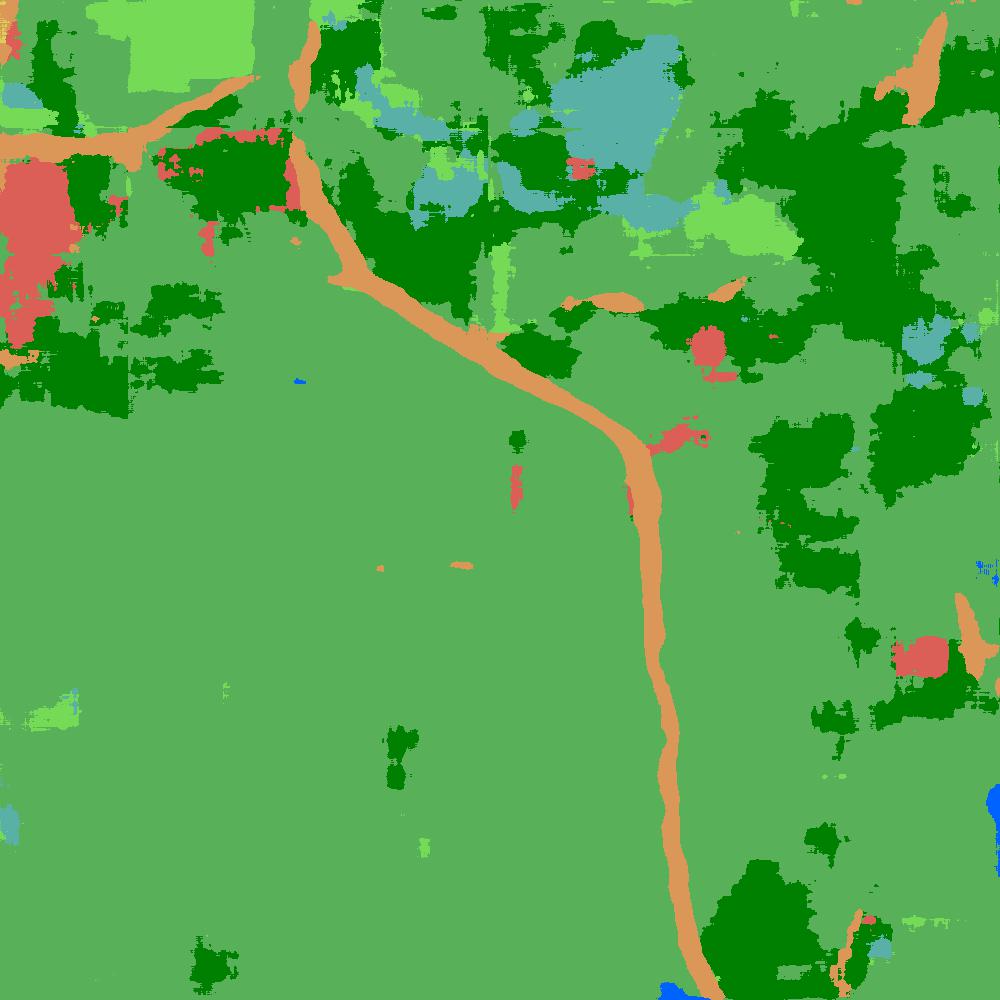} & \includegraphics[width=.09\linewidth]{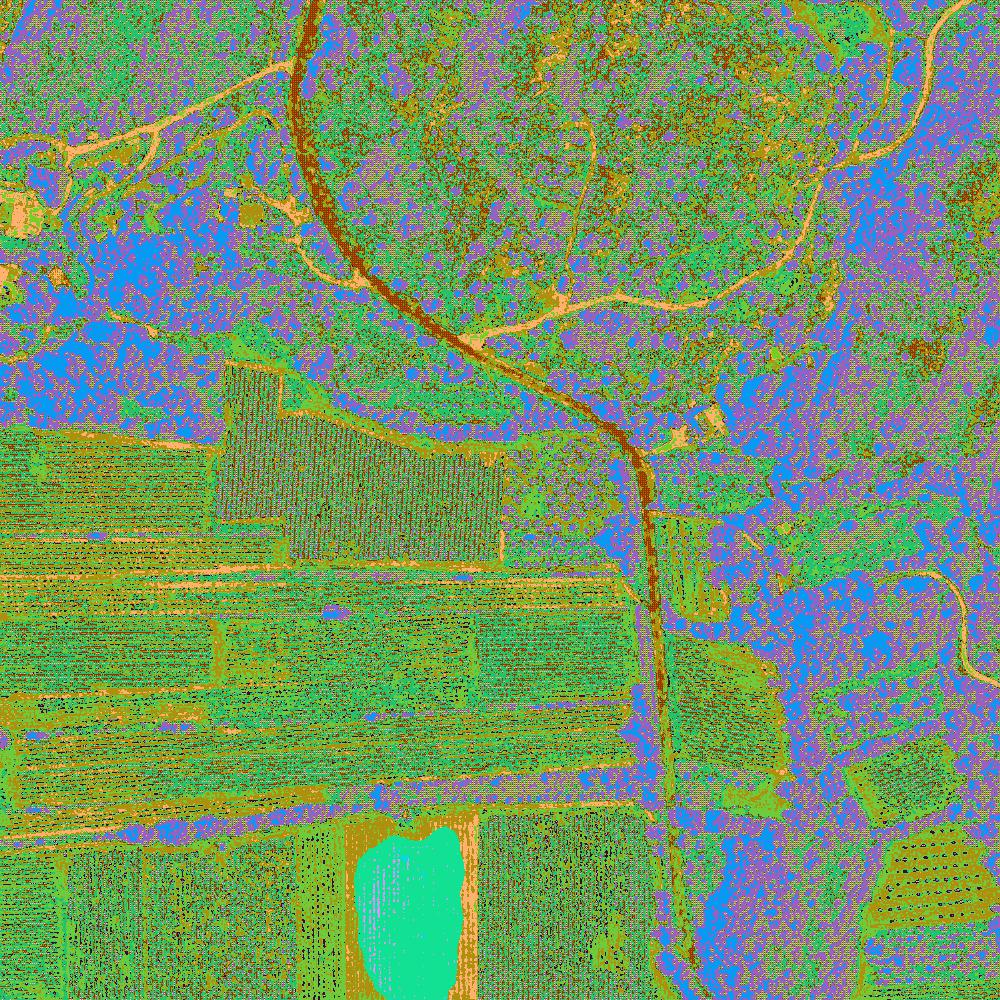} & \includegraphics[width=.09\linewidth]{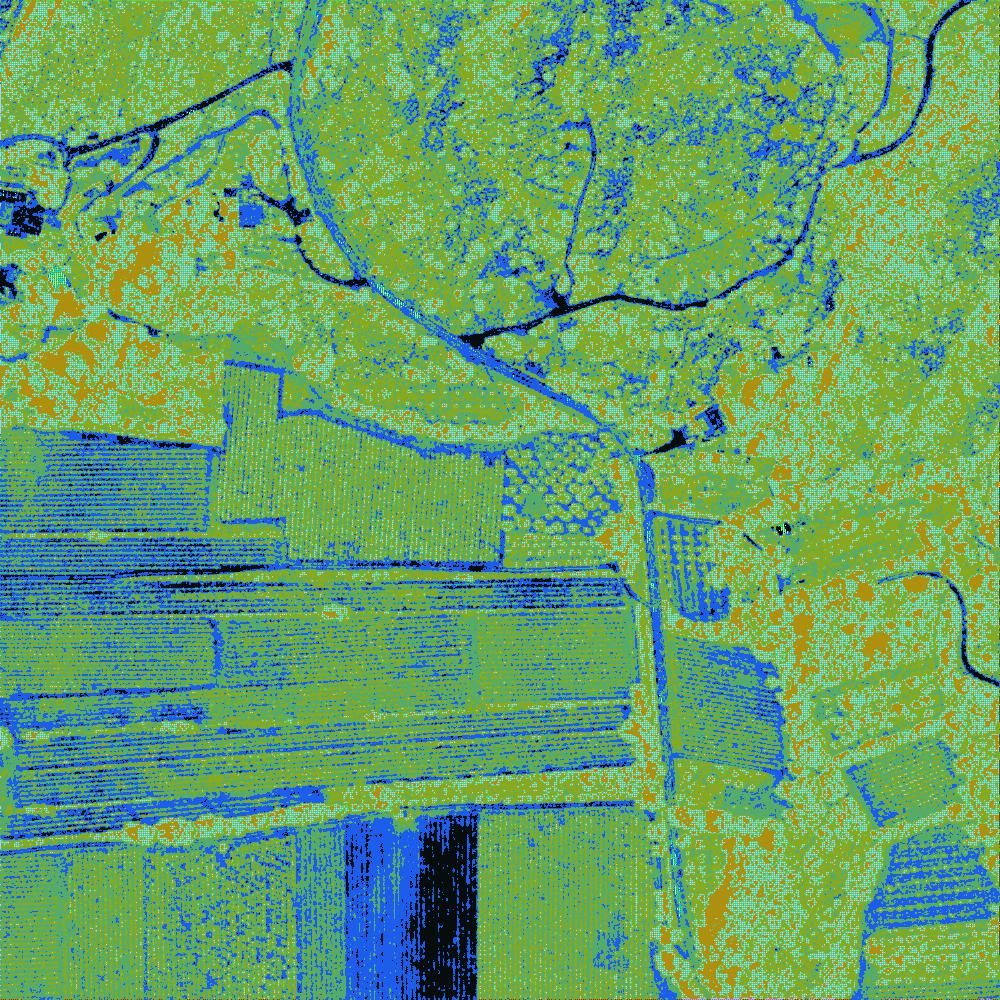} & \includegraphics[width=.09\linewidth]{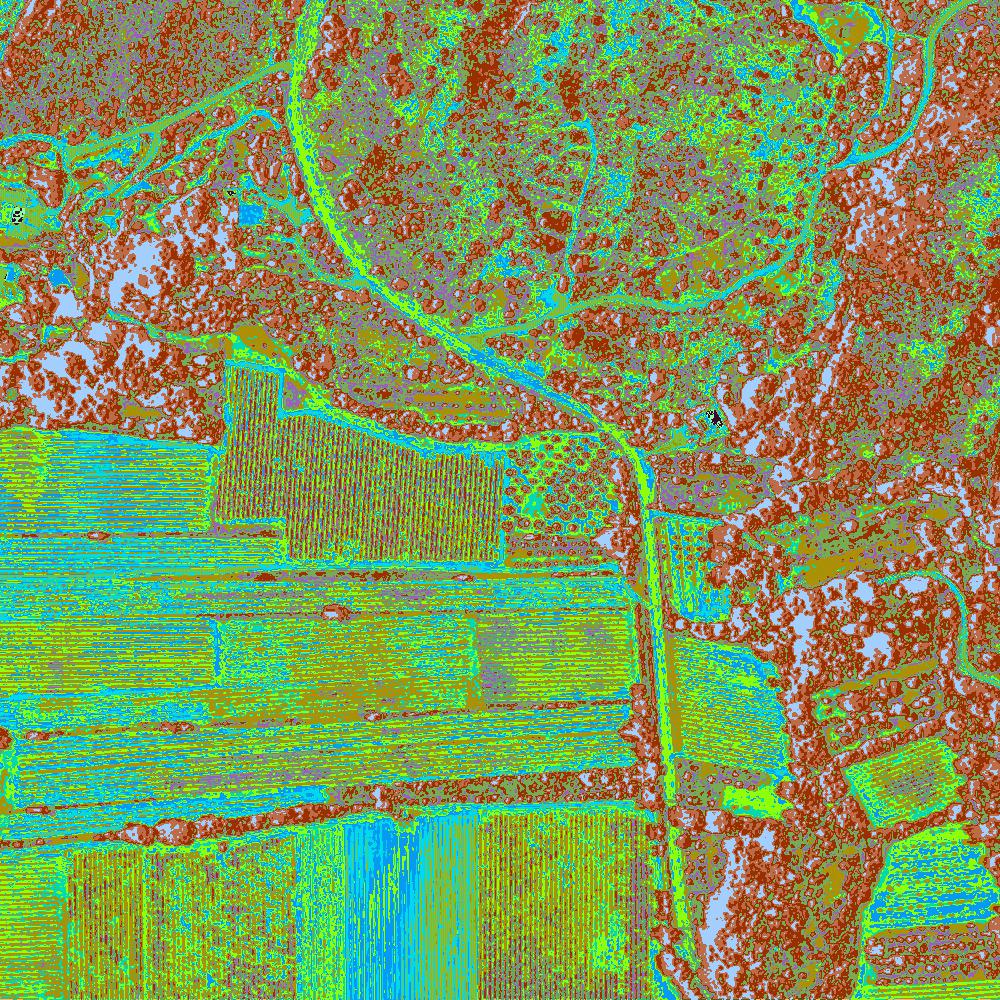}  & \includegraphics[width=.09\linewidth]{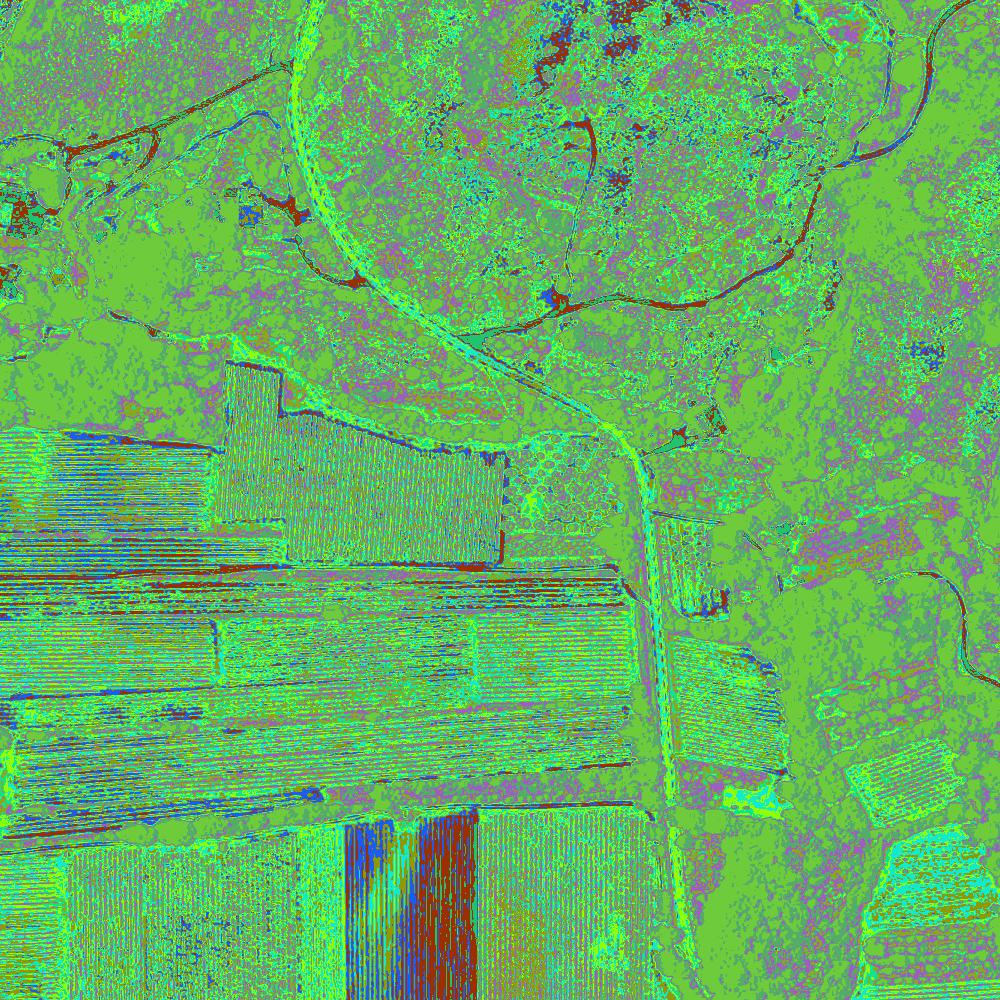} \\
         \includegraphics[width=.09\linewidth]{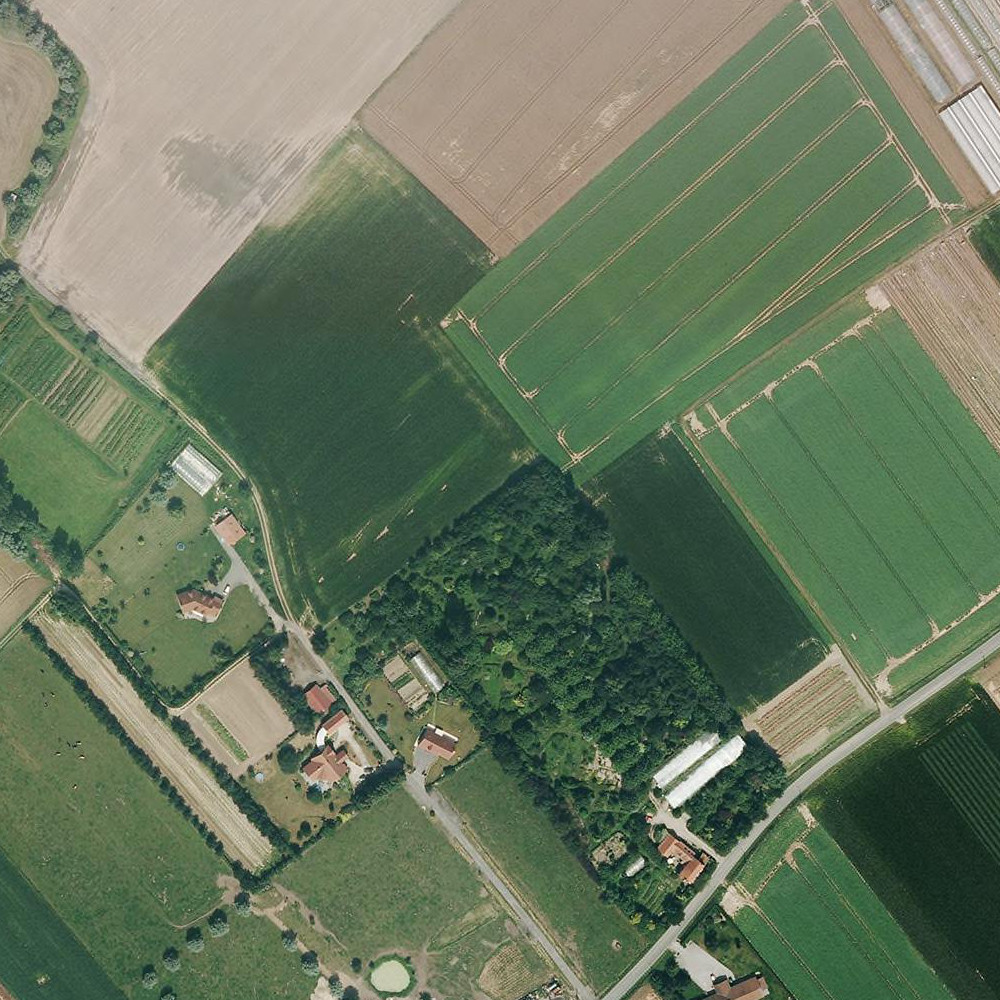} & \includegraphics[width=.09\linewidth]{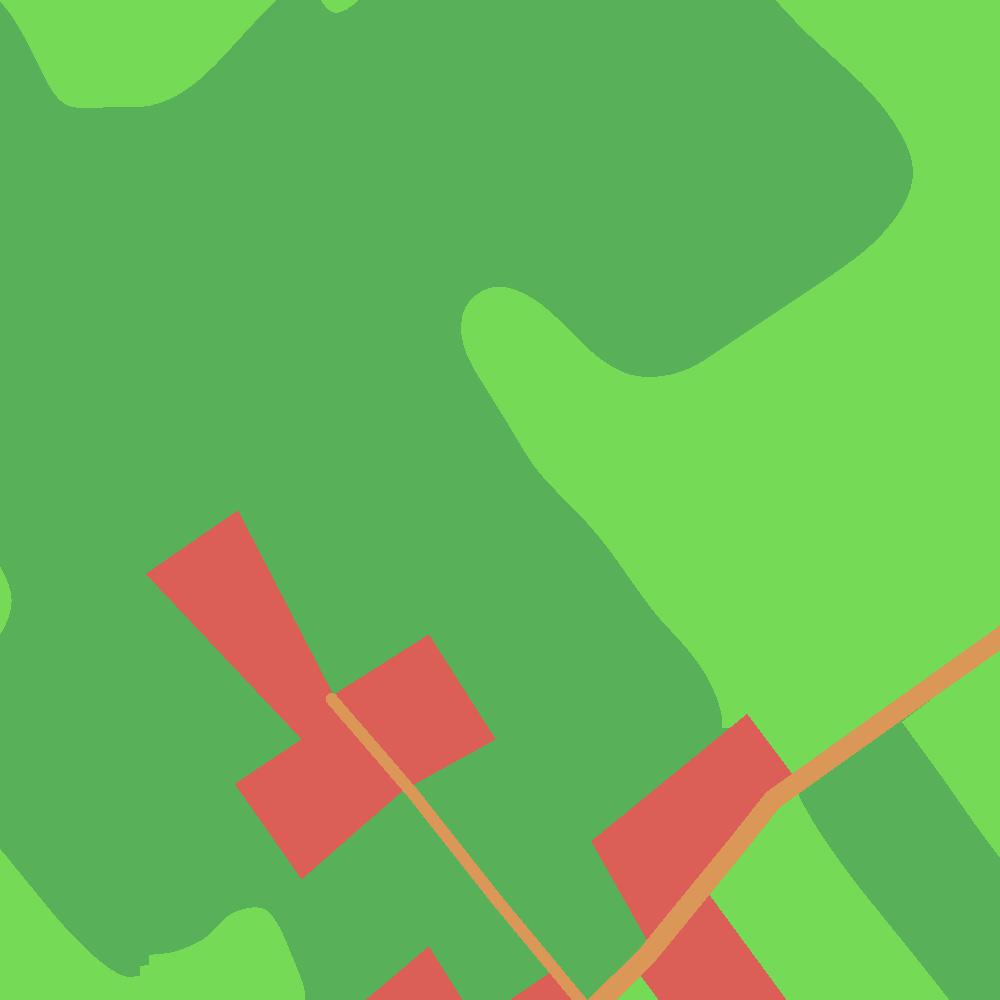} & \includegraphics[width=.09\linewidth]{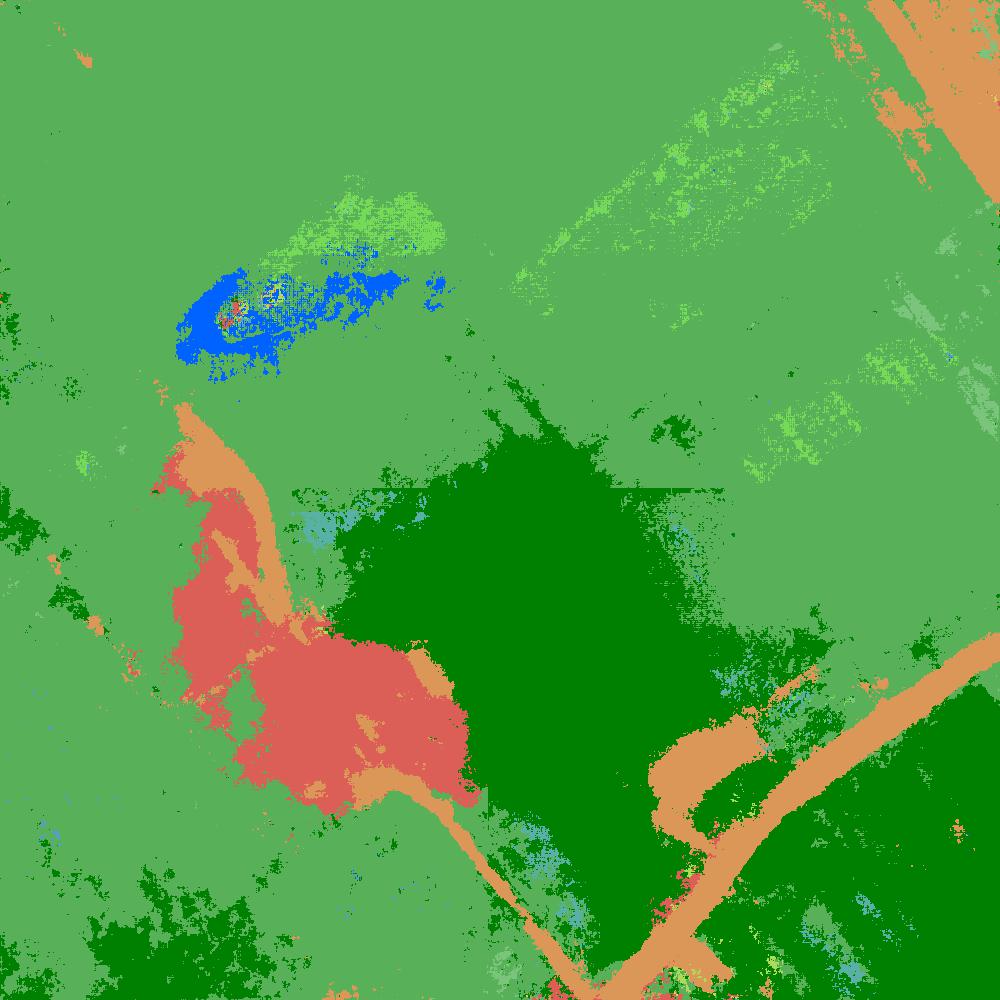} & \includegraphics[width=.09\linewidth]{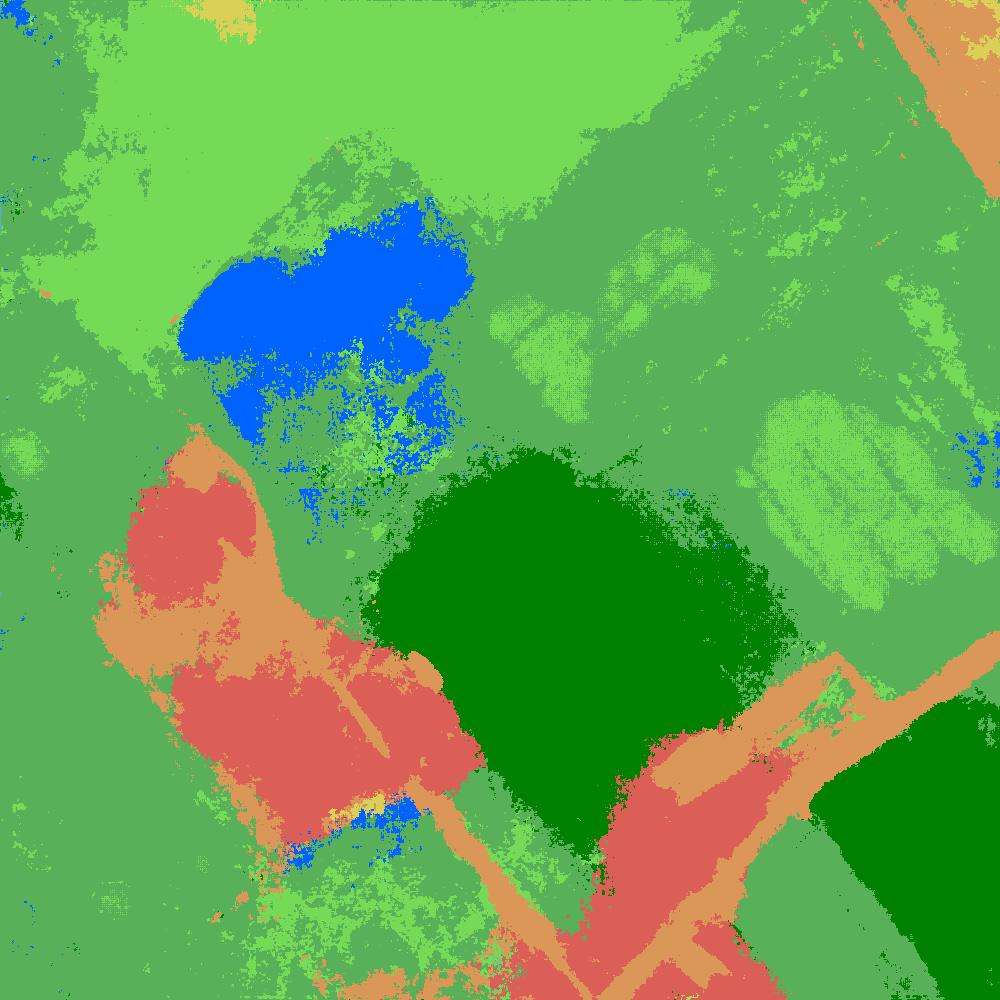} & \includegraphics[width=.09\linewidth]{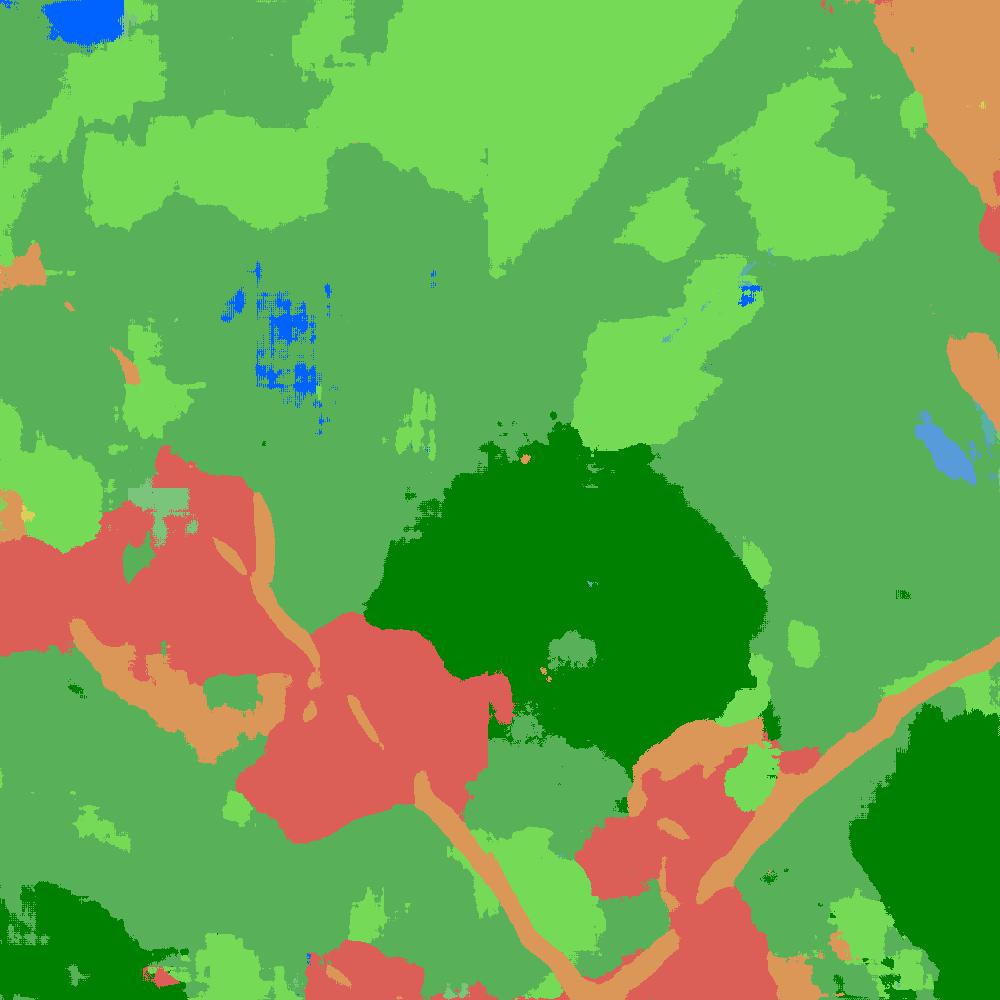} & \includegraphics[width=.09\linewidth]{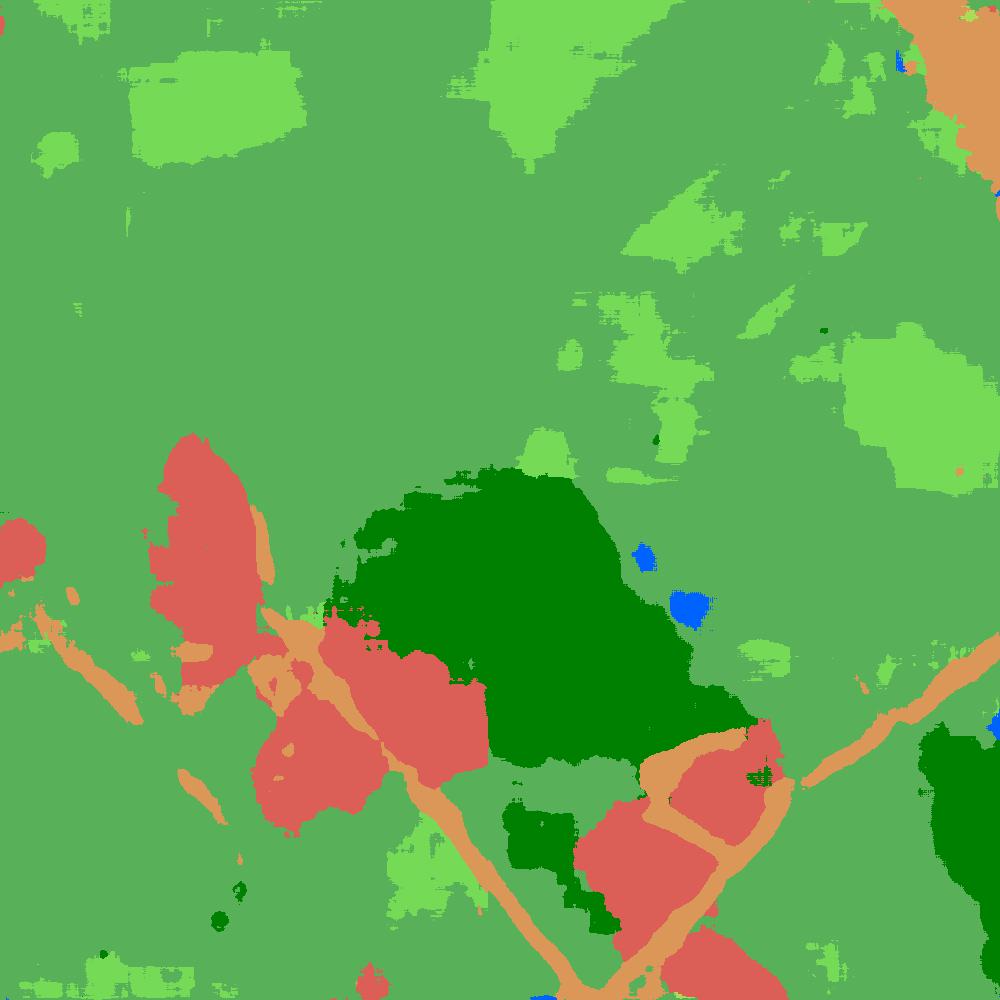} & \includegraphics[width=.09\linewidth]{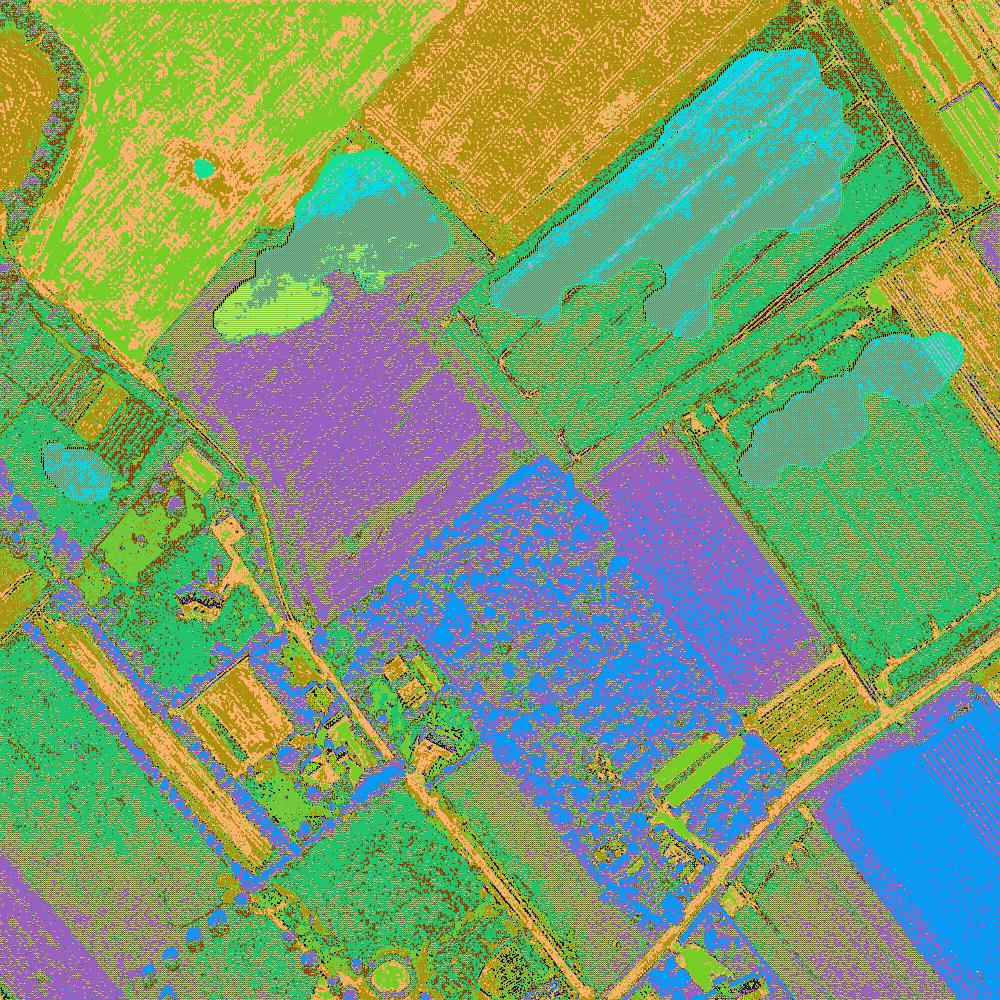} & \includegraphics[width=.09\linewidth]{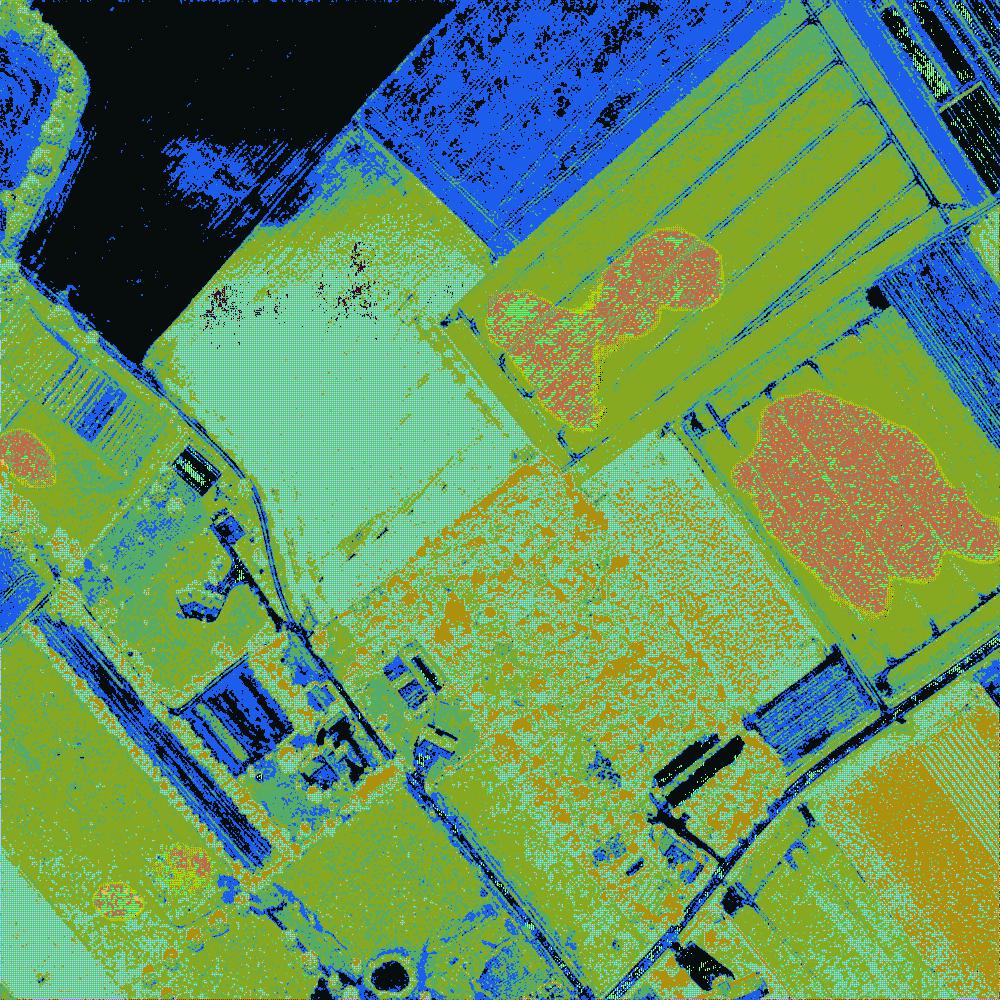} & \includegraphics[width=.09\linewidth]{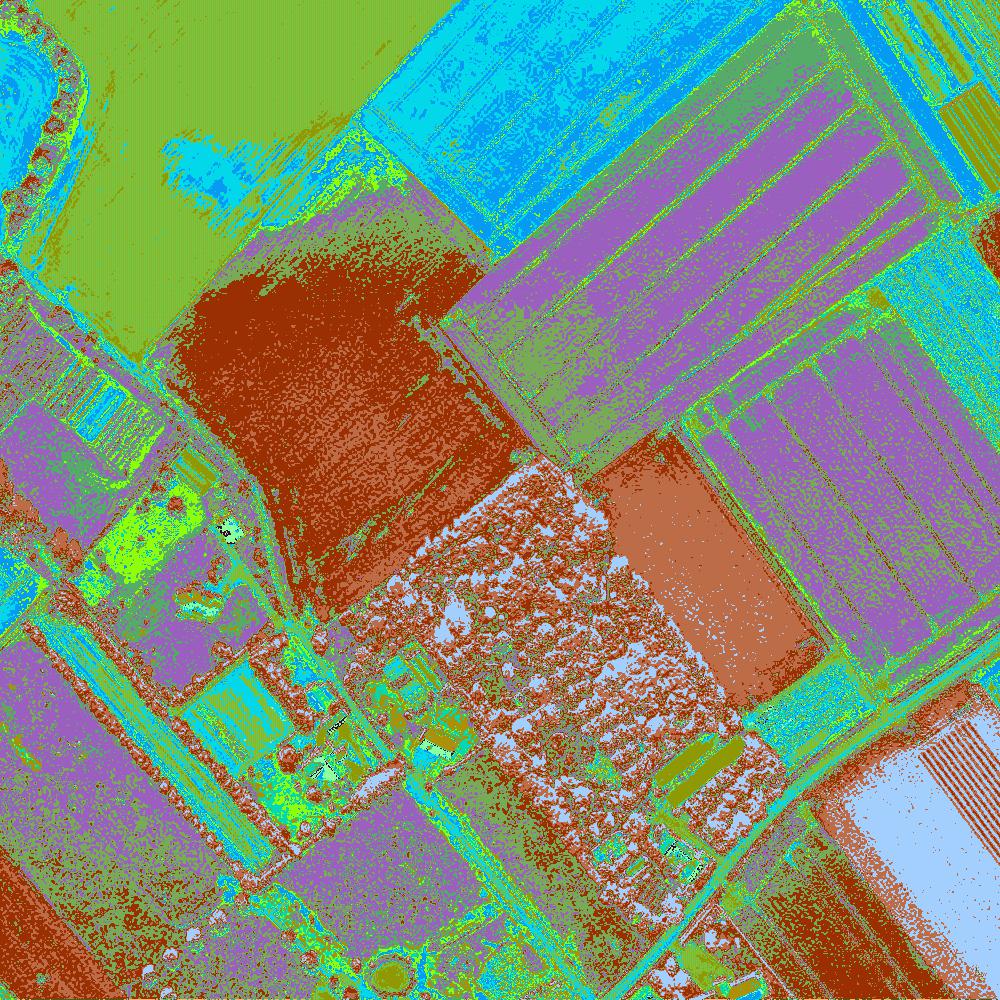} & \includegraphics[width=.09\linewidth]{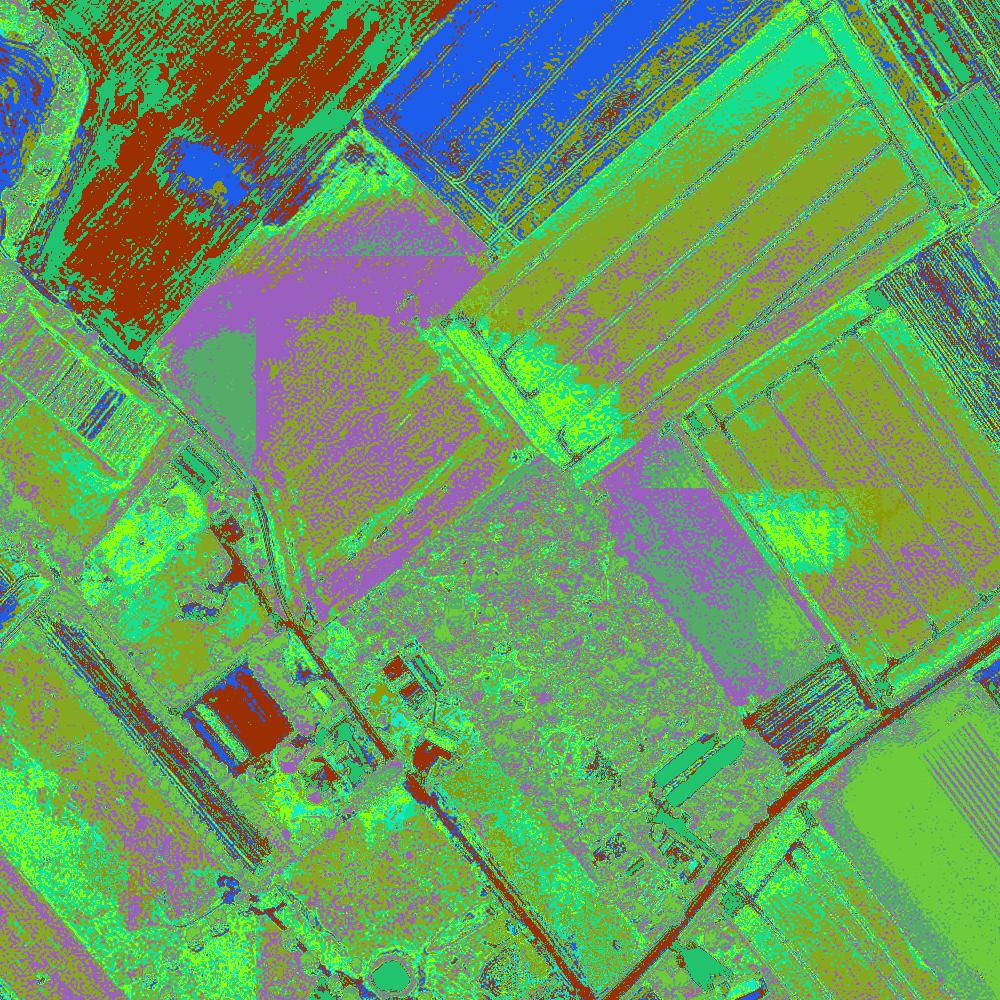} \\
         \includegraphics[width=.09\linewidth]{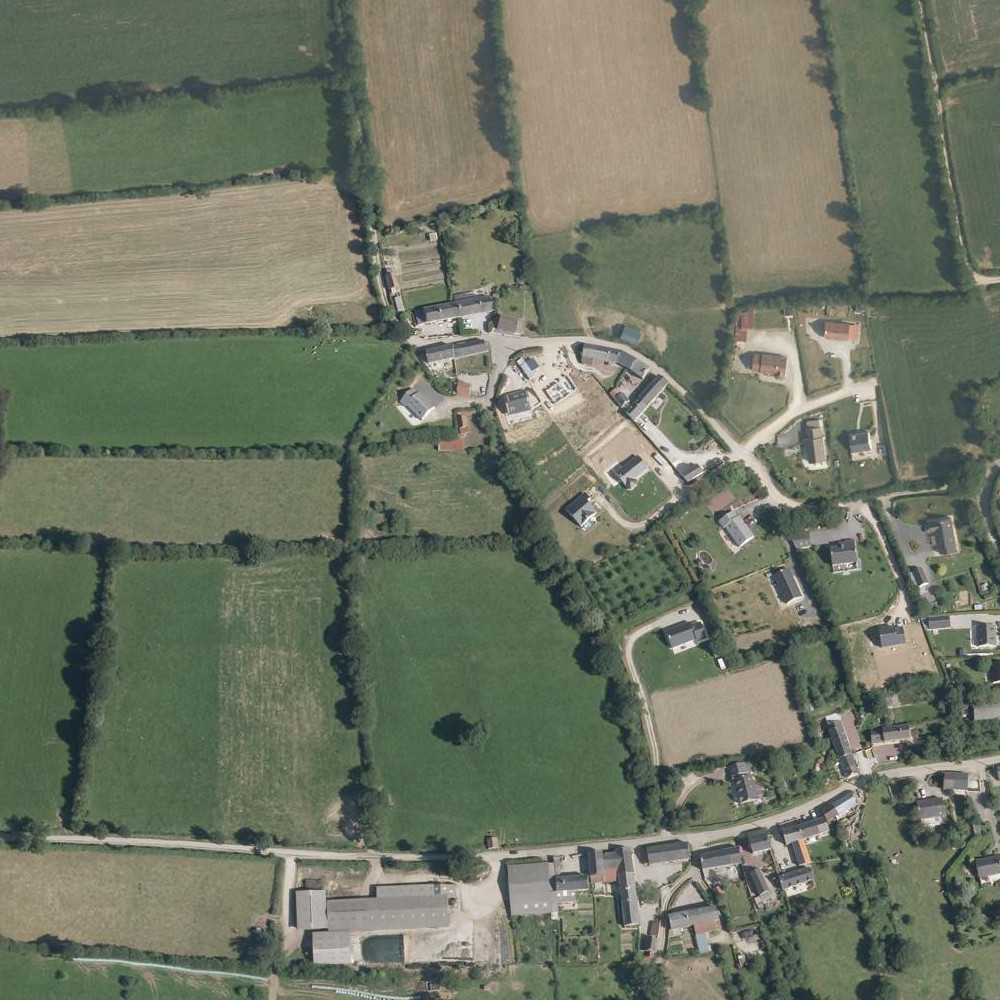} & \includegraphics[width=.09\linewidth]{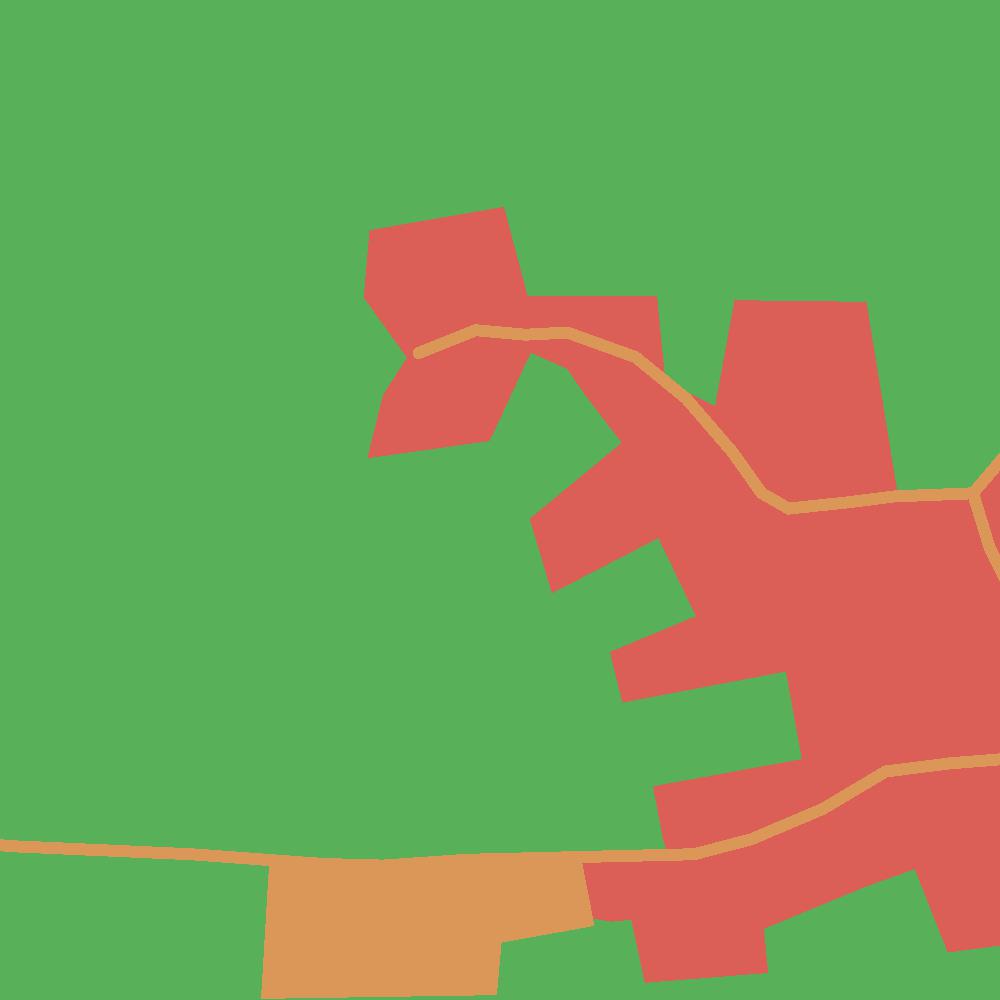} & \includegraphics[width=.09\linewidth]{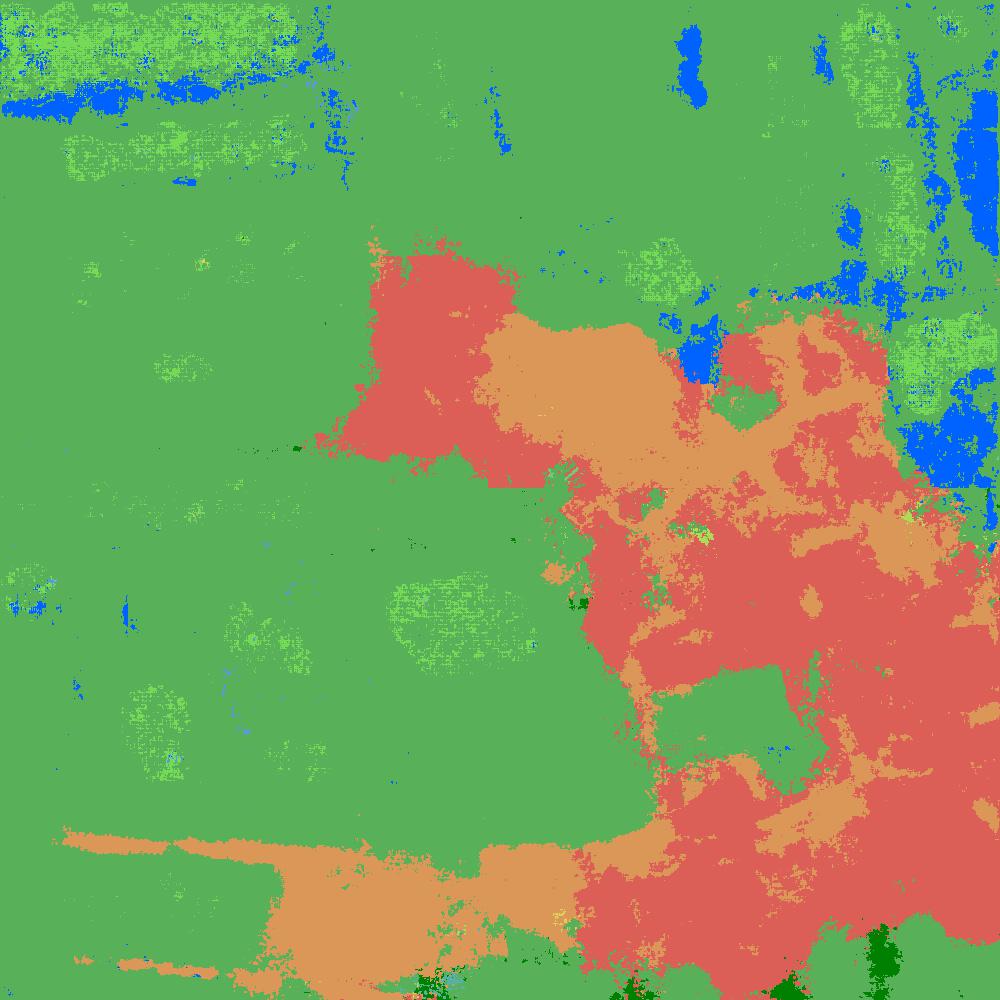} & \includegraphics[width=.09\linewidth]{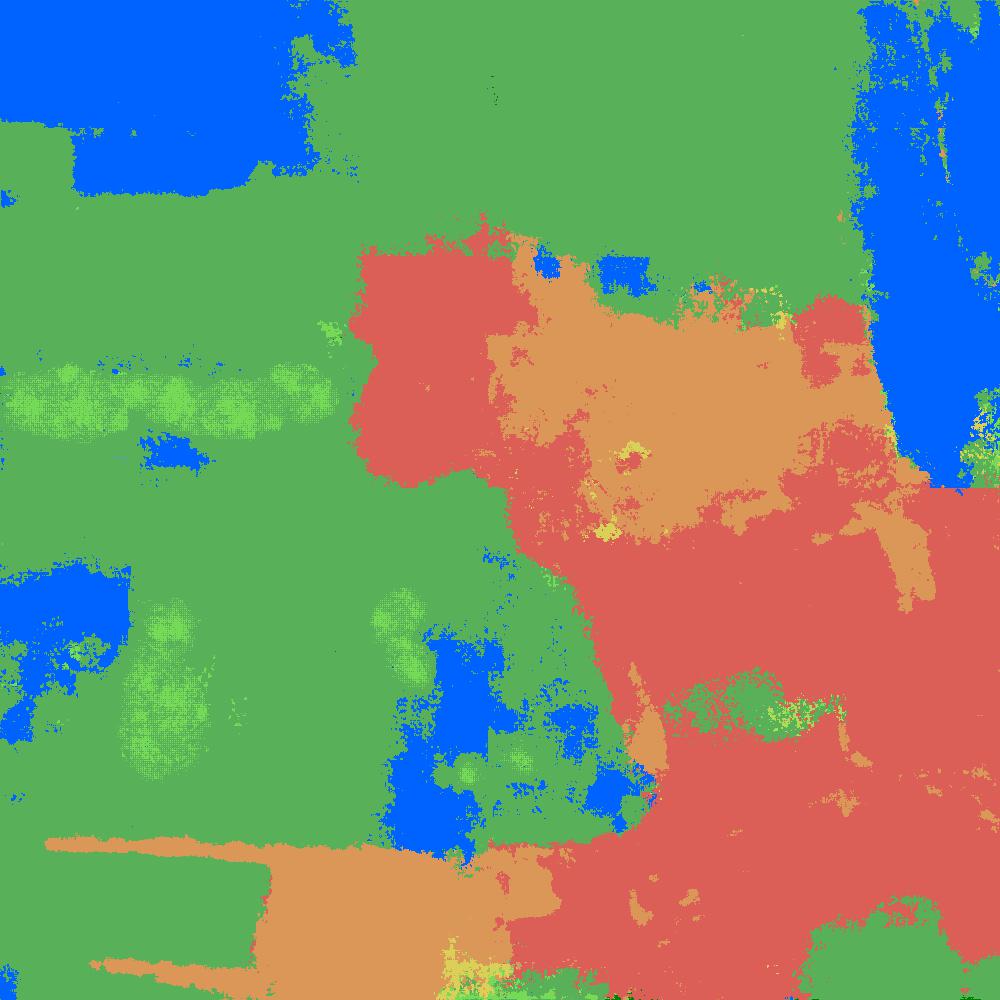} & \includegraphics[width=.09\linewidth]{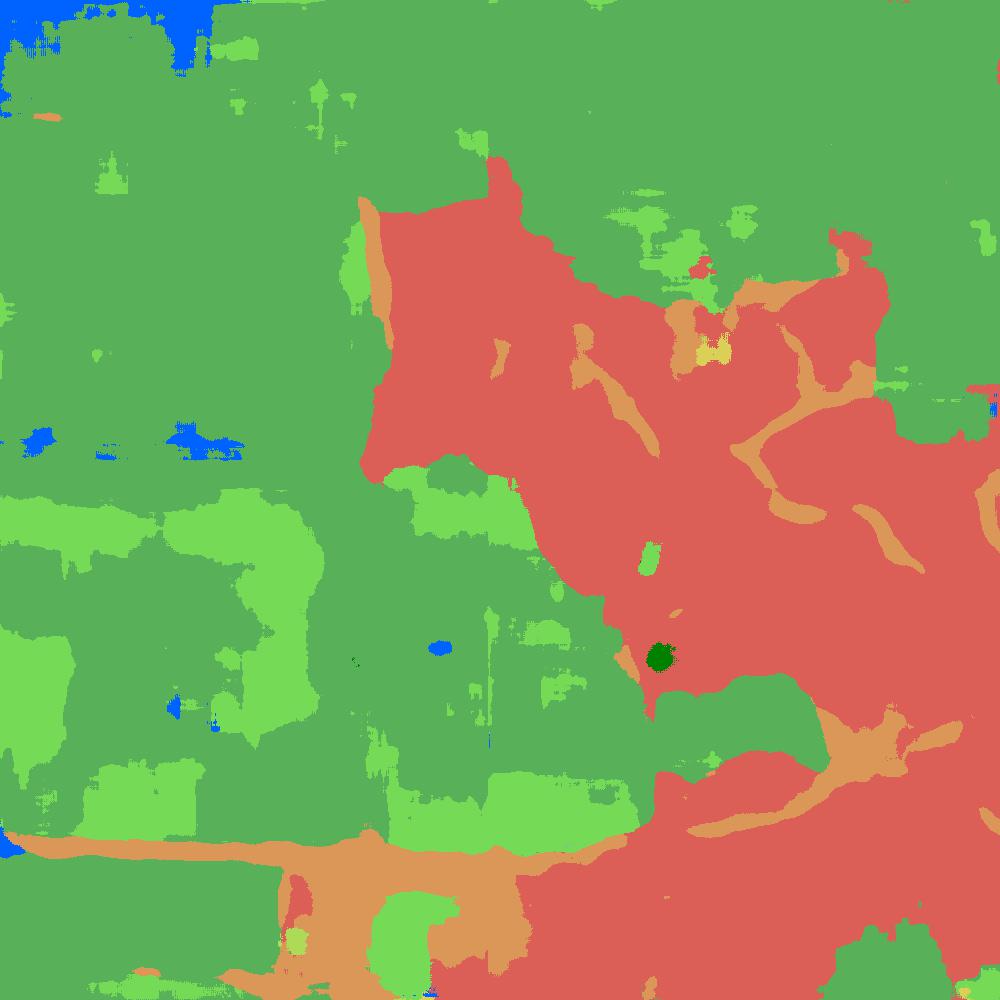} & \includegraphics[width=.09\linewidth]{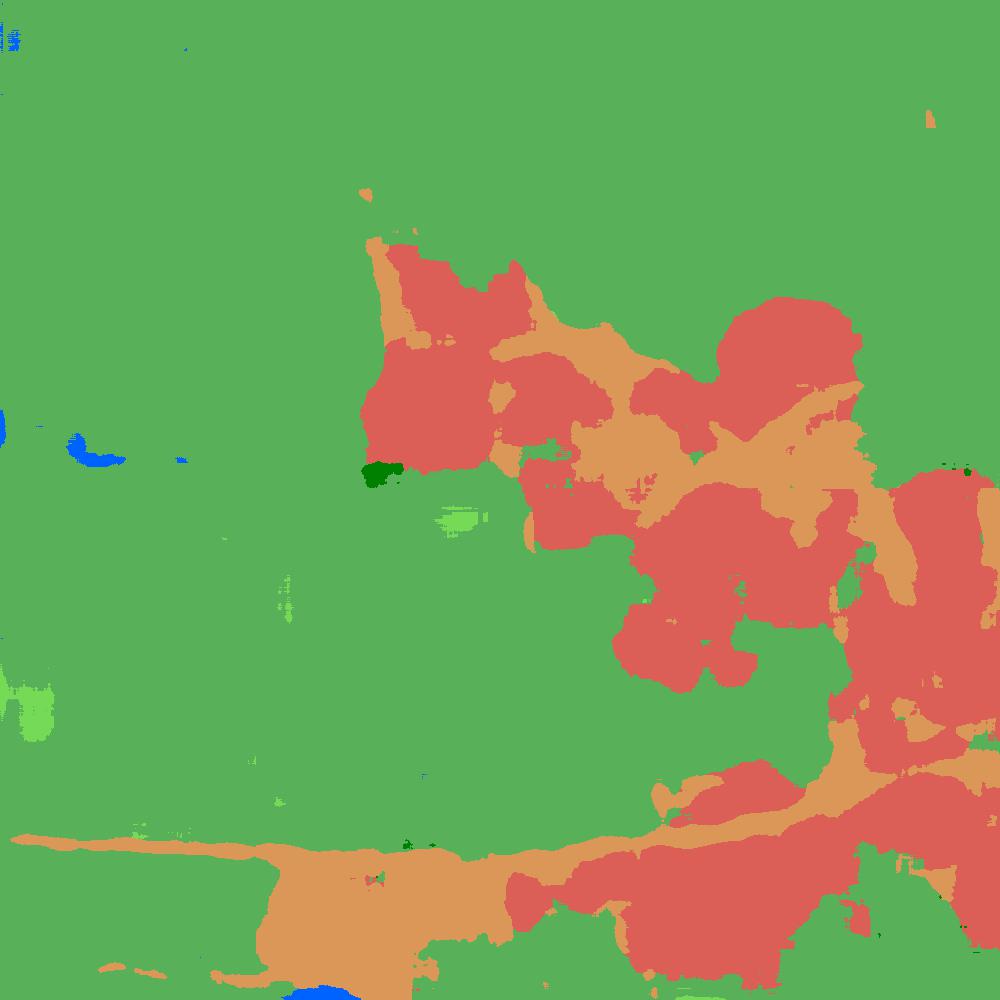} & \includegraphics[width=.09\linewidth]{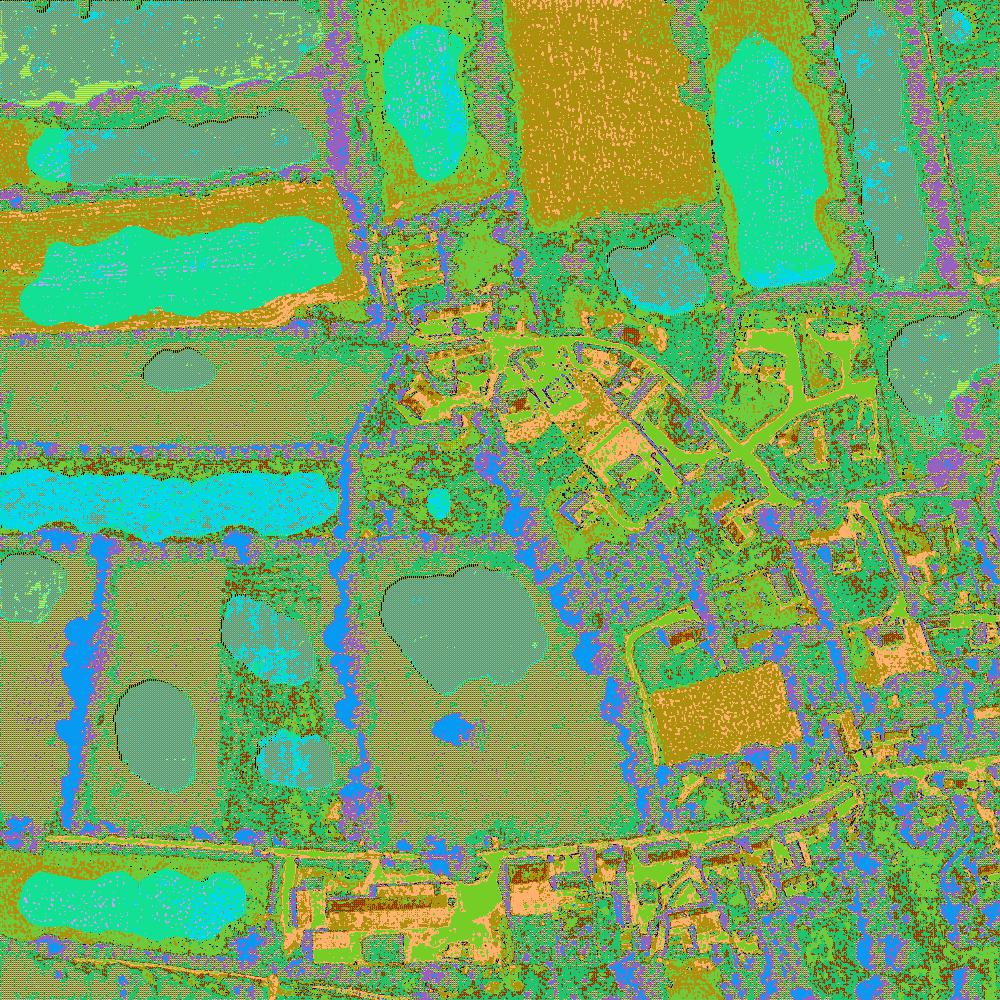} & \includegraphics[width=.09\linewidth]{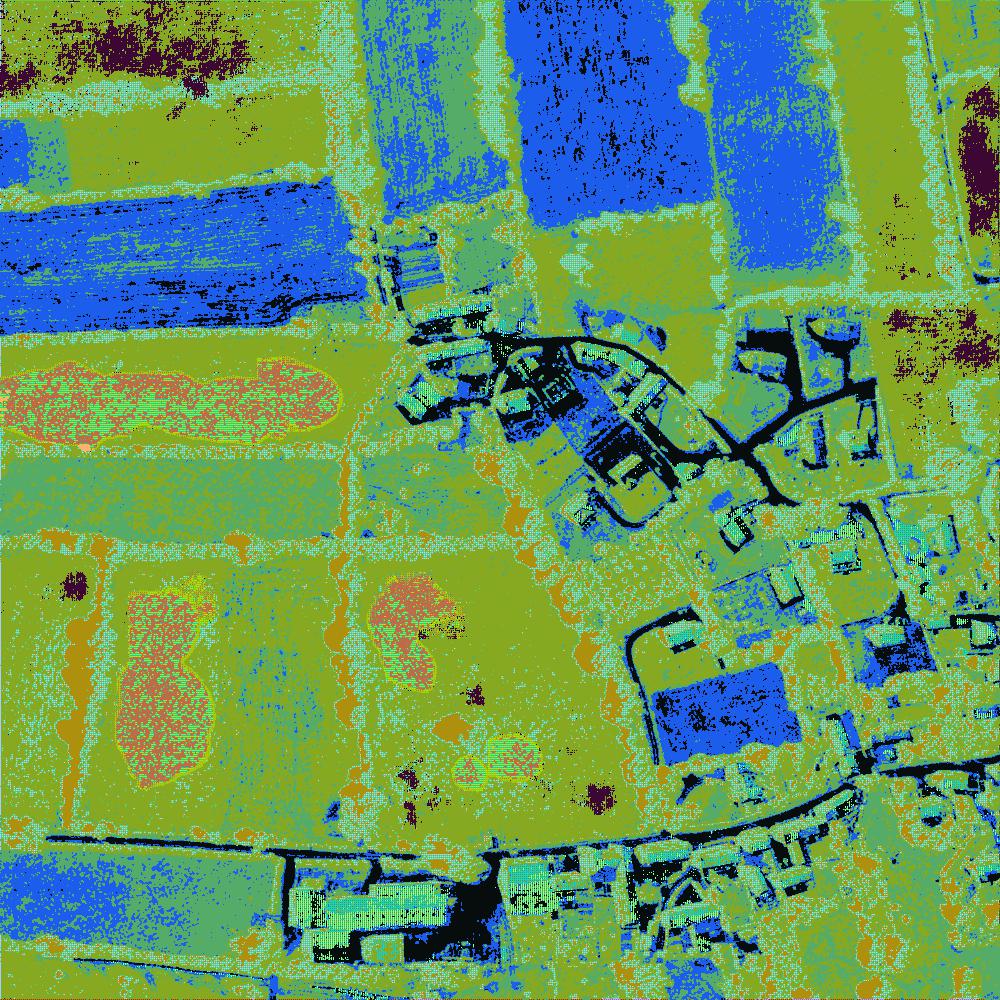} & \includegraphics[width=.09\linewidth]{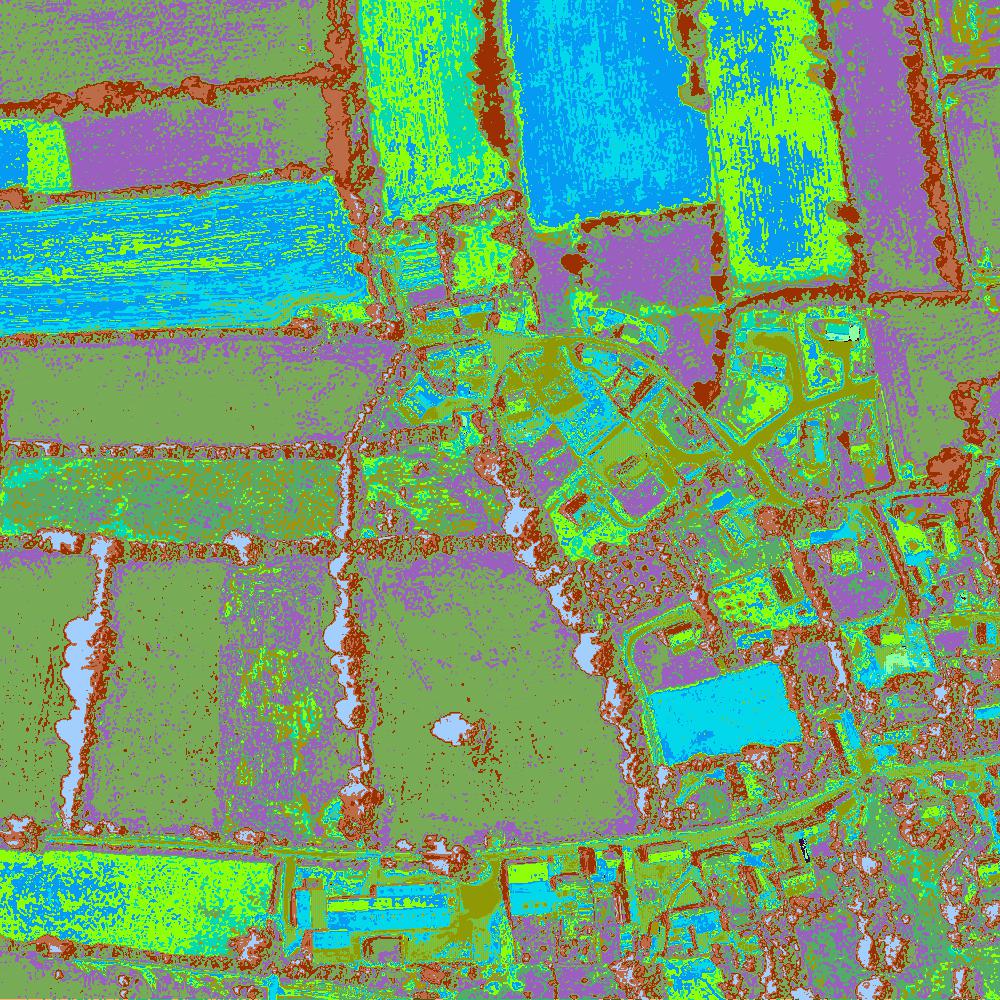} & \includegraphics[width=.09\linewidth]{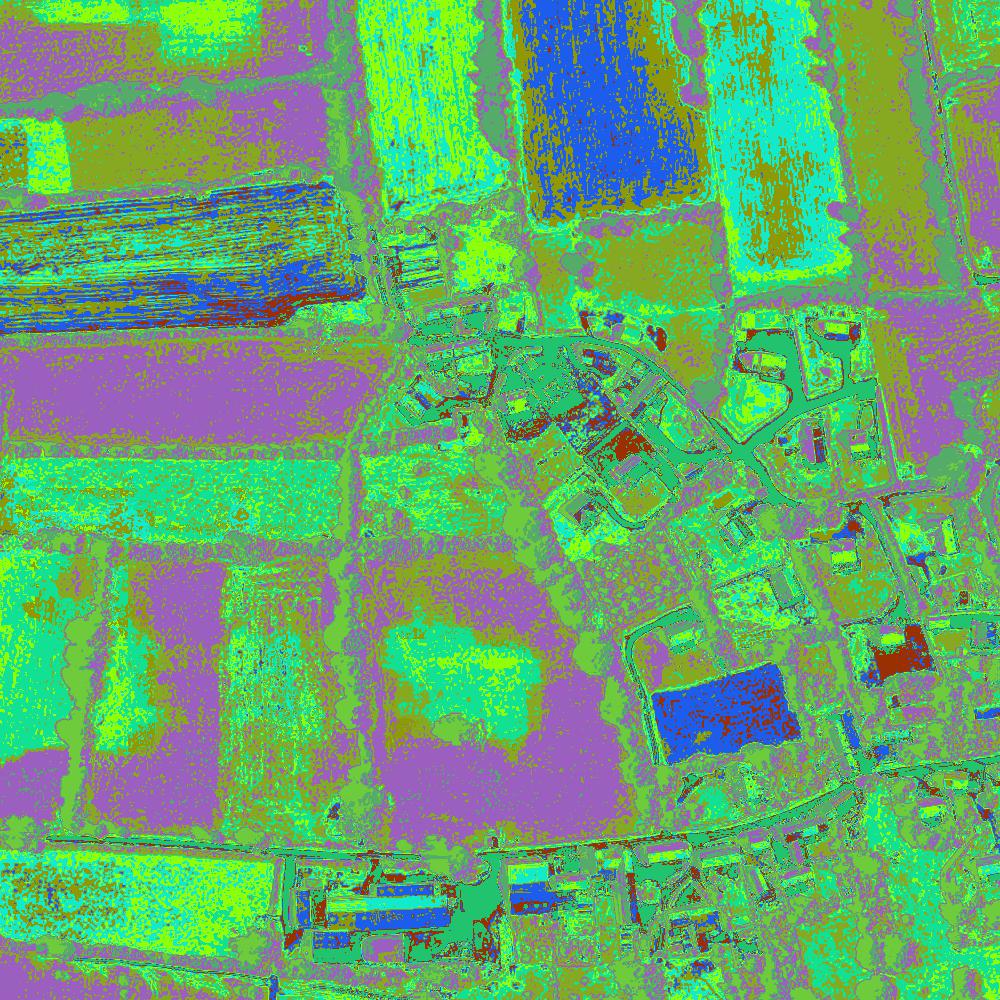} \\
         \includegraphics[width=.09\linewidth]{63-2013-0700-6510-LA93-0M50-E080_5480_4729.jpg} & \includegraphics[width=.09\linewidth]{63-2013-0700-6510-LA93-0M50-E080_5480_4729_gt.jpg} & \includegraphics[width=.09\linewidth]{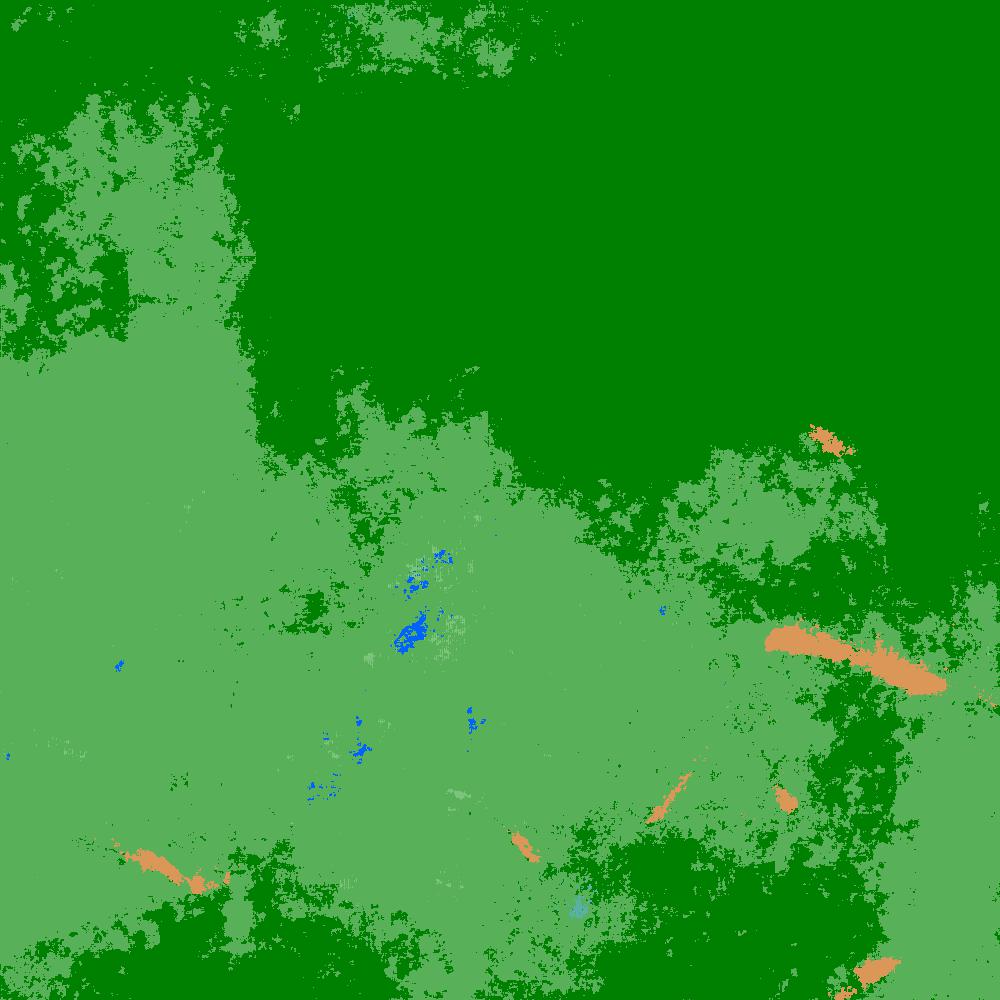} & \includegraphics[width=.09\linewidth]{63-2013-0700-6510-LA93-0M50-E080_5480_4729_segnet_km_pred.jpg} & \includegraphics[width=.09\linewidth]{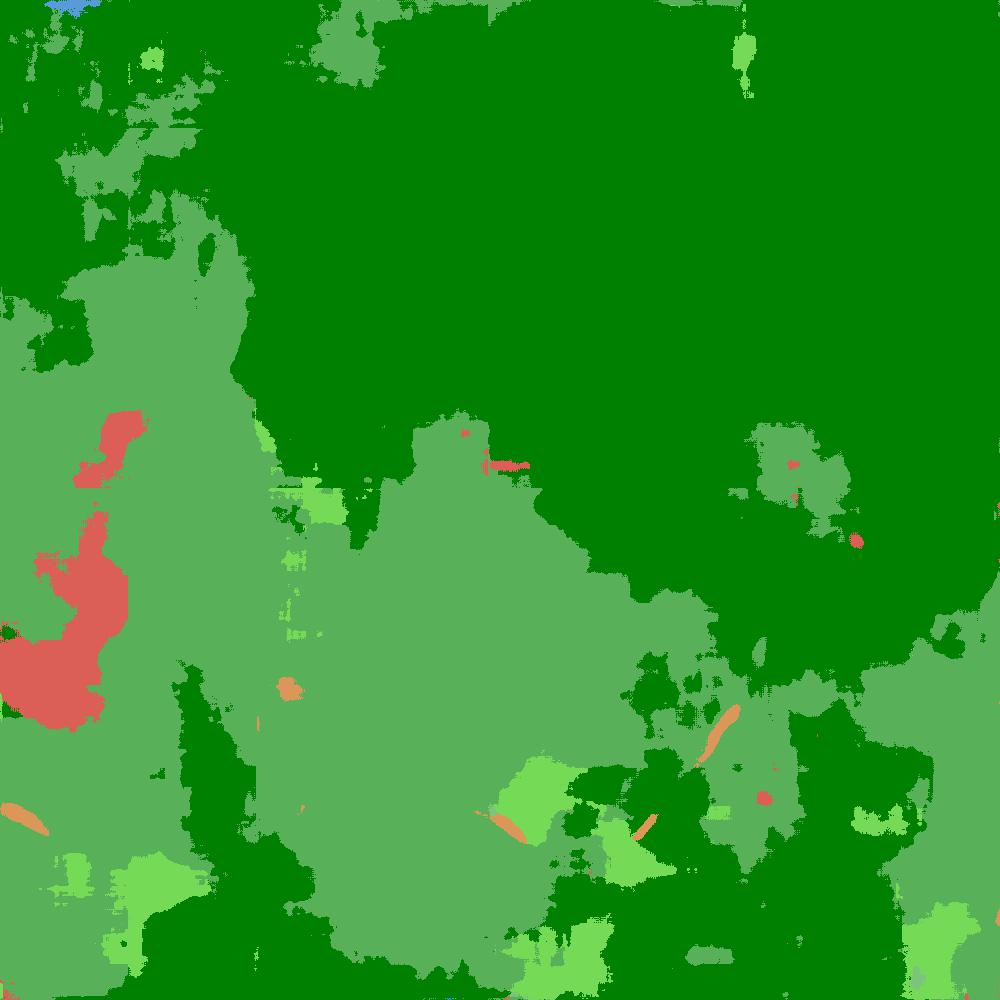} & \includegraphics[width=.09\linewidth]{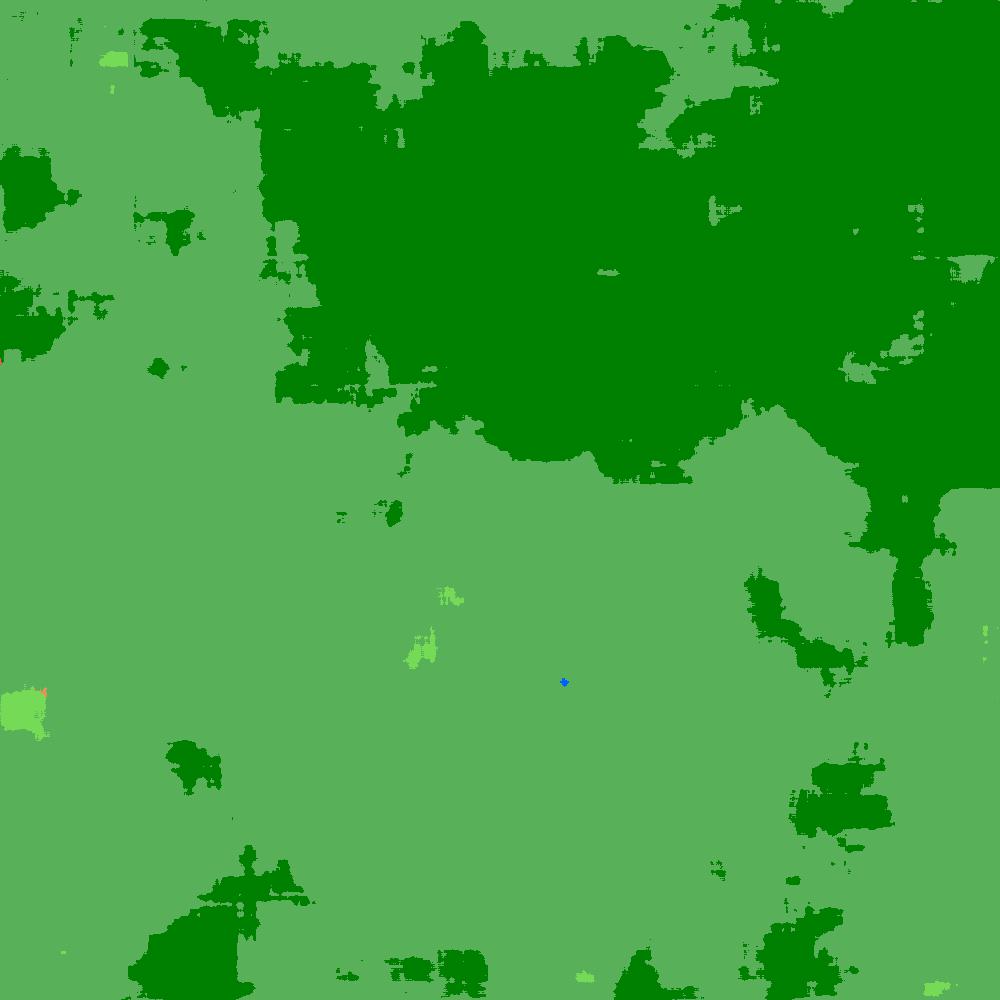} & \includegraphics[width=.09\linewidth]{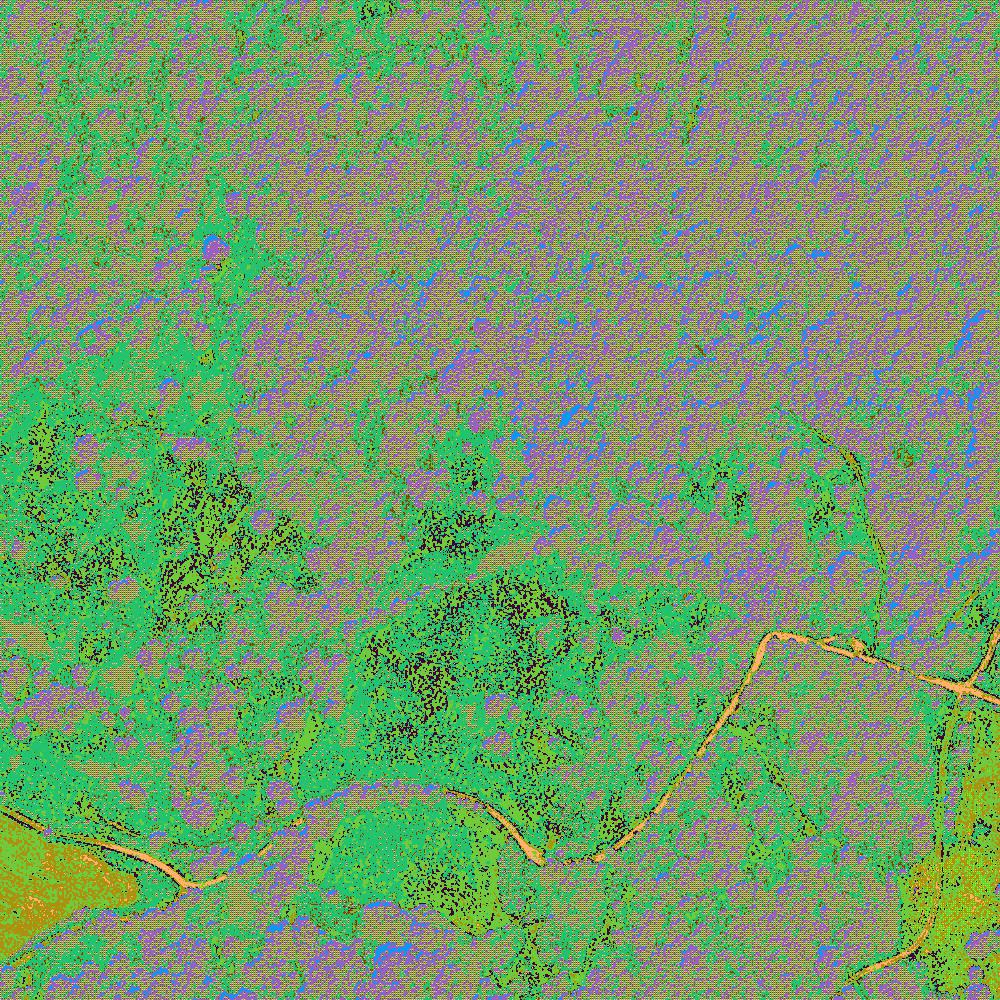} & \includegraphics[width=.09\linewidth]{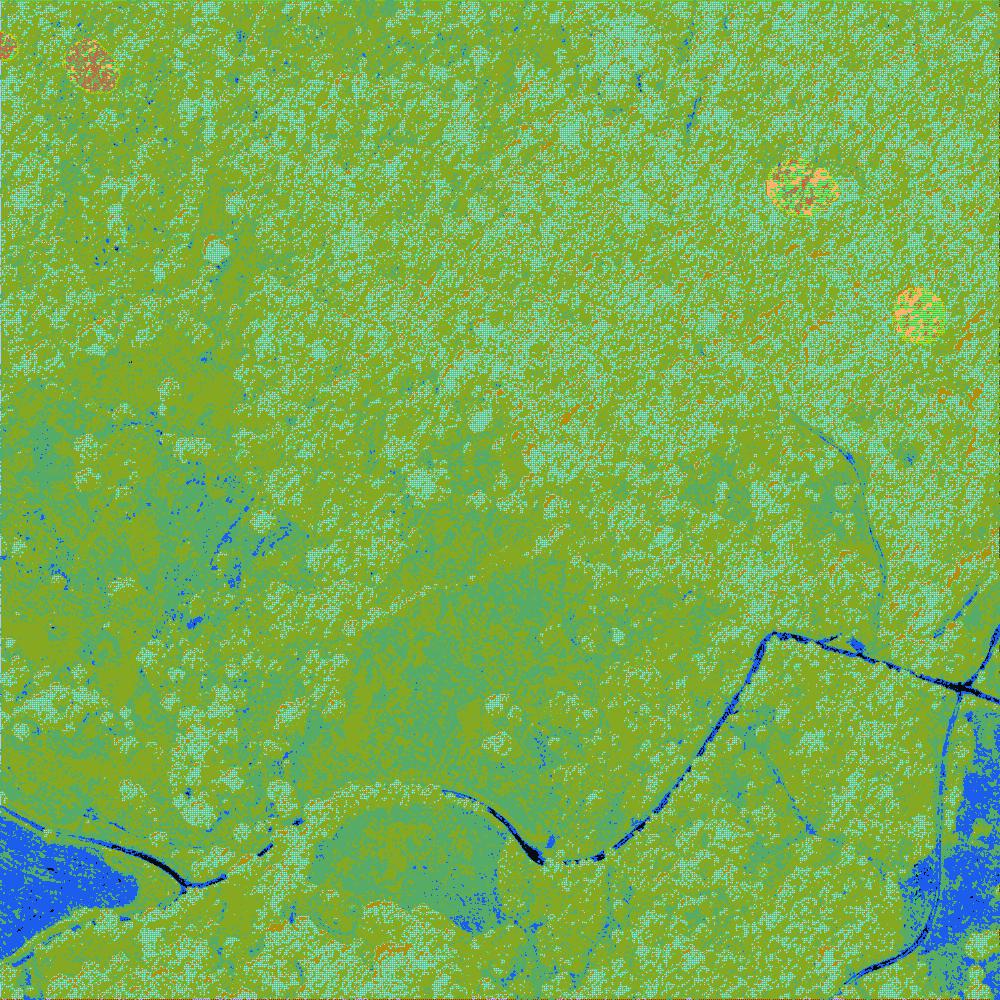} & \includegraphics[width=.09\linewidth]{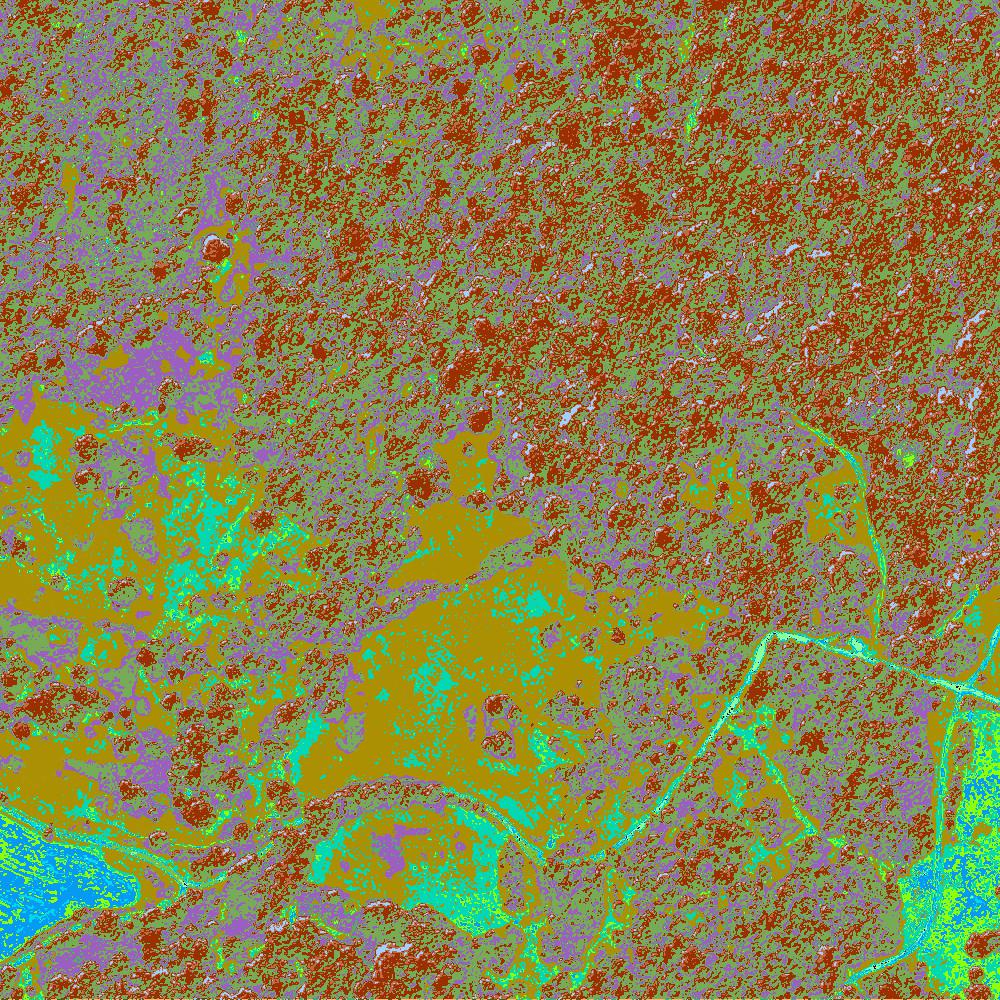} & \includegraphics[width=.09\linewidth]{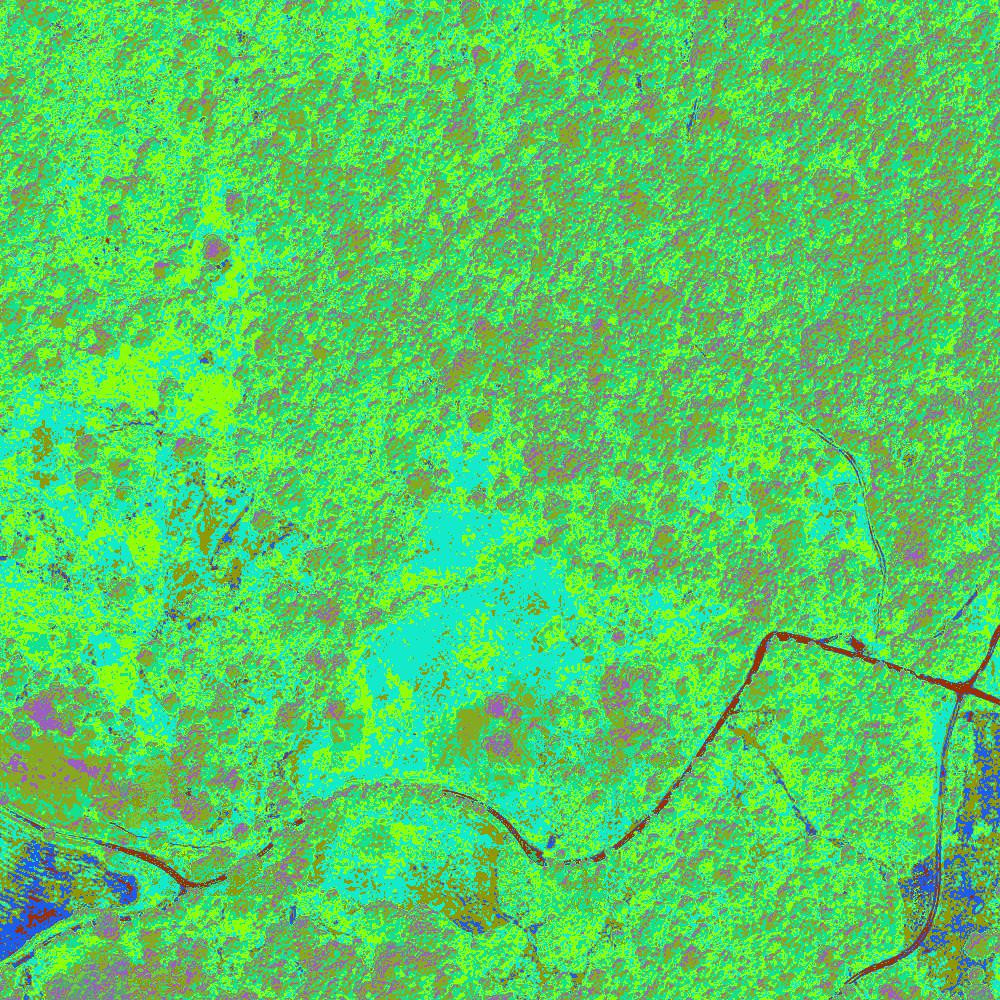} \\
         [3pt]
&& \scriptsize{BN-e} & \scriptsize{BN-l-S} & \scriptsize{BN-l-U} & \scriptsize{W-Net} & \scriptsize{BN-e} & \scriptsize{BN-l-S} & \scriptsize{BN-l-U} & \scriptsize{W-Net}\\
Image & GT & \multicolumn{4}{c}{Semantic Segmentation} & \multicolumn{4}{c}{Unsupervised Segmentation}  
      \end{tabular}
      \caption{Results comparison for different neural networks with unsupervised segmentation as auxiliary task ($\LL_{km}$ auxiliary loss). BN-e stands for BerundaNet-early, BN-l-S/BN-l-U for BerundaNet-late with SegNet/U-Net backbone, respectively.}
      \label{fig: results-different-networks-segmentation}
   \end{center}
\end{figure}

Figures~\ref{fig: results-different-networks-reconstruction} and~\ref{fig: results-different-networks-segmentation} show some examples of semantic maps and unsupervised outputs at inference time for these methods, using reconstruction and unsupervised segmentation as auxiliary task, respectively. From these examples, we confirm that whether we choose reconstruction or segmentation as auxiliary unsupervised task, BerundaNet-late (U-Net backbone) gets the finer and smoother results, especially in the second case. 

Therefore, the choice of the architecture and backbone matters for the semi-supervised task.
BerundaNet-late performs better than BerundaNet-early with same backbone.
Moreover, the U-Net backbone outperforms the SegNet backbone.
Finally, the simple architecture BerundaNet-late presented in this work places it first, before W-Net.

Thus, it seems the choice of architecture is at least as important as the loss design.
This choice does not only rely on the number of parameters (W-Net has about twice the number of parameters of BerundaNet, since it relies on two U-Nets) but also how the supervised and unsupervised information are mixed.

\subsubsection{Influence of the choice of auxiliary loss on semi-supervision}\label{sec: experiments-choice-loss}

In this section, we analyze the effect on the semantic segmentation results of different auxiliary losses presented in section~\ref{sec: unsupervised-losses}. To this end, we train the same network architecture while changing the loss. We choose BerundaNet-late with U-Net backbone, since it was the network with the best scores in the previous sections, regardless of the auxiliary task.

\begin{figure}[!htbp]
   \begin{center}
      \setlength{\tabcolsep}{1.2pt}
      \begin{tabular}{cc@{\hspace{10pt}}cc@{\hspace{10pt}}cc}
         \includegraphics[width=.14\linewidth]{35-2012-0350-6815-LA93-0M50-E080_6893_61.jpg} & \includegraphics[width=.14\linewidth]{35-2012-0350-6815-LA93-0M50-E080_6893_61_gt.jpg} & \includegraphics[width=.14\linewidth]{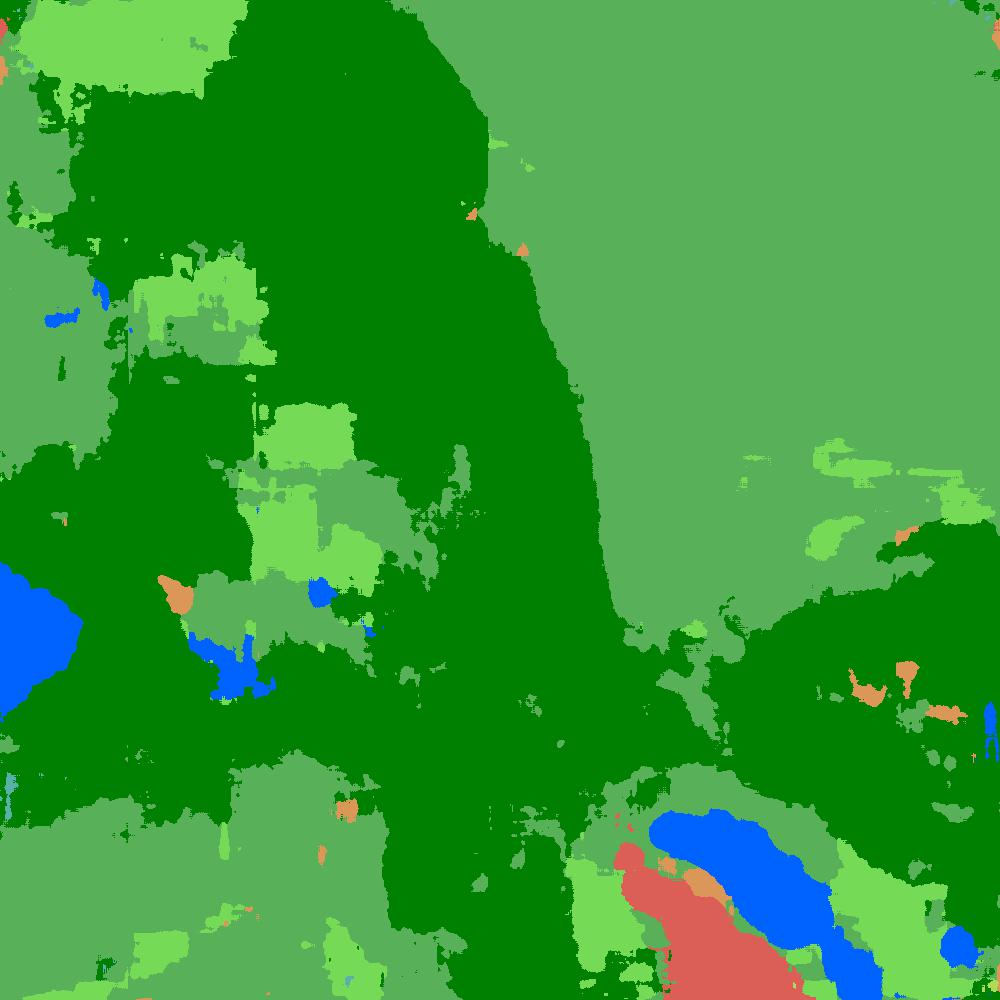} & \includegraphics[width=.14\linewidth]{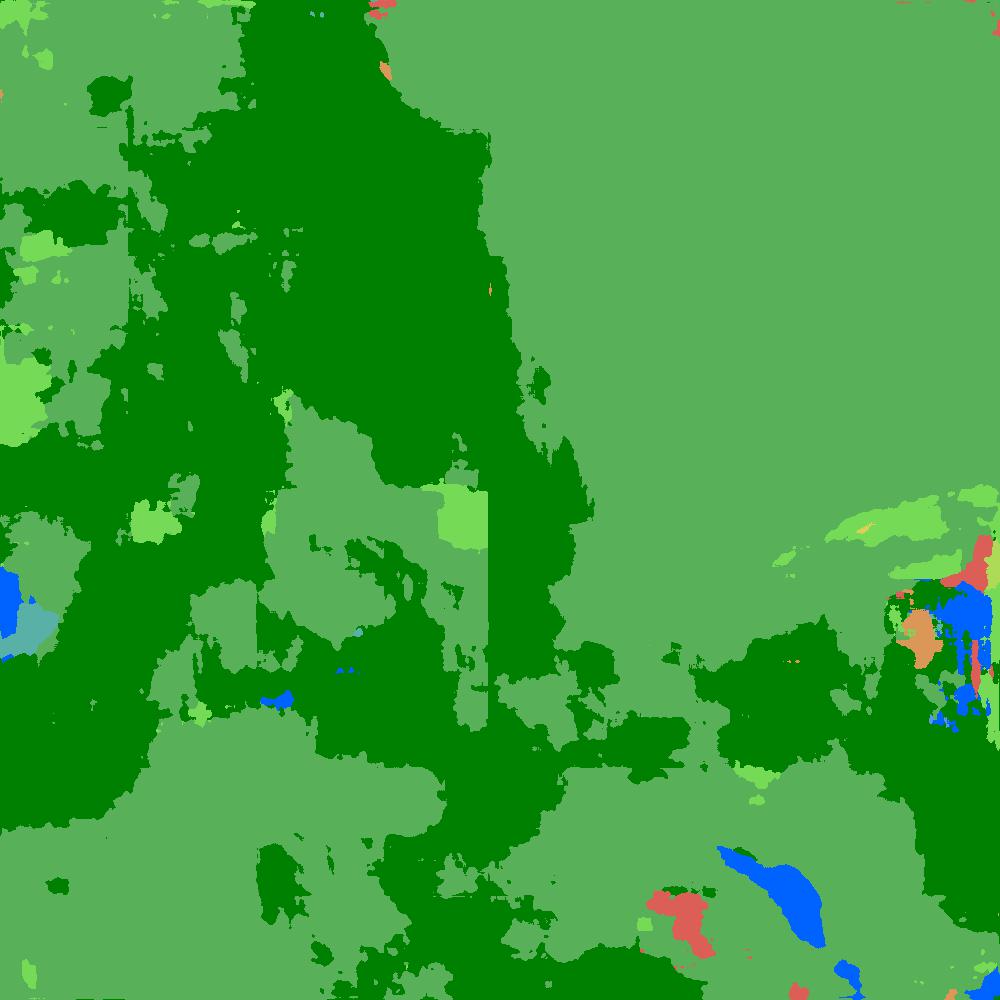} & \includegraphics[width=.14\linewidth]{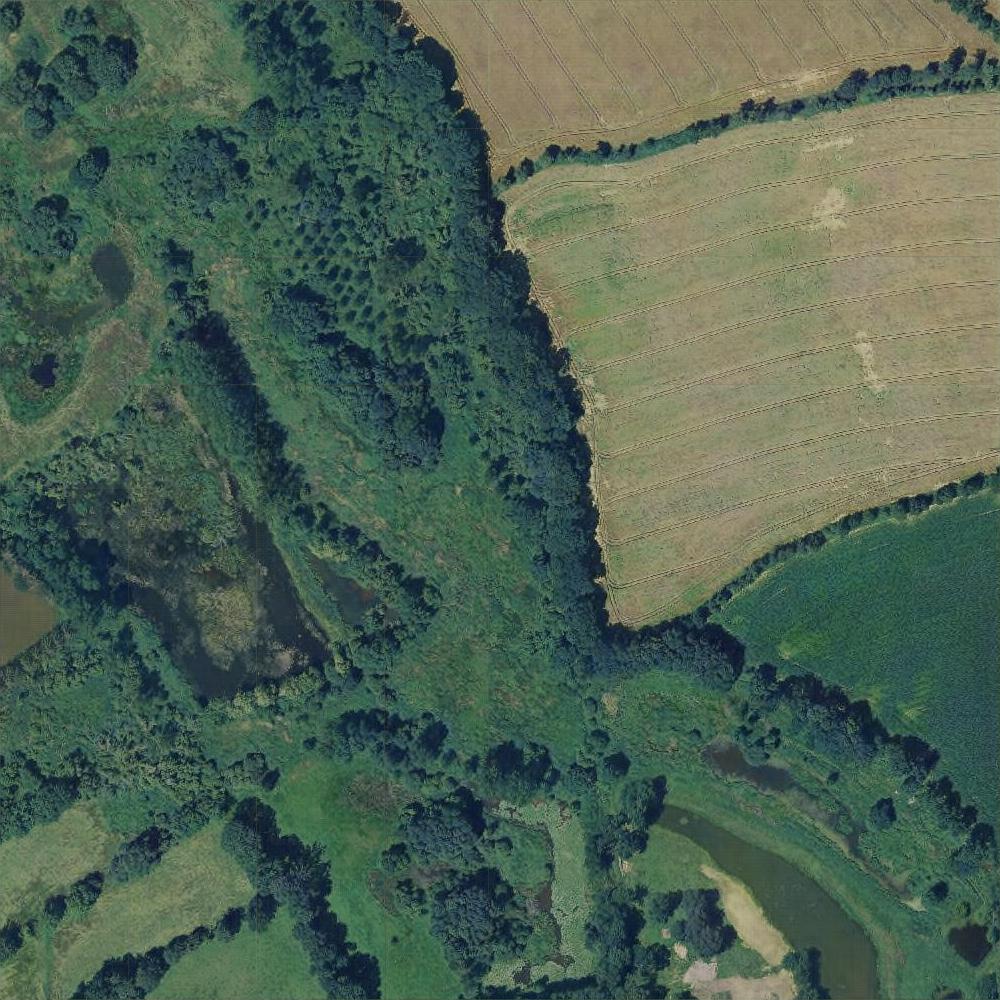} & \includegraphics[width=.14\linewidth]{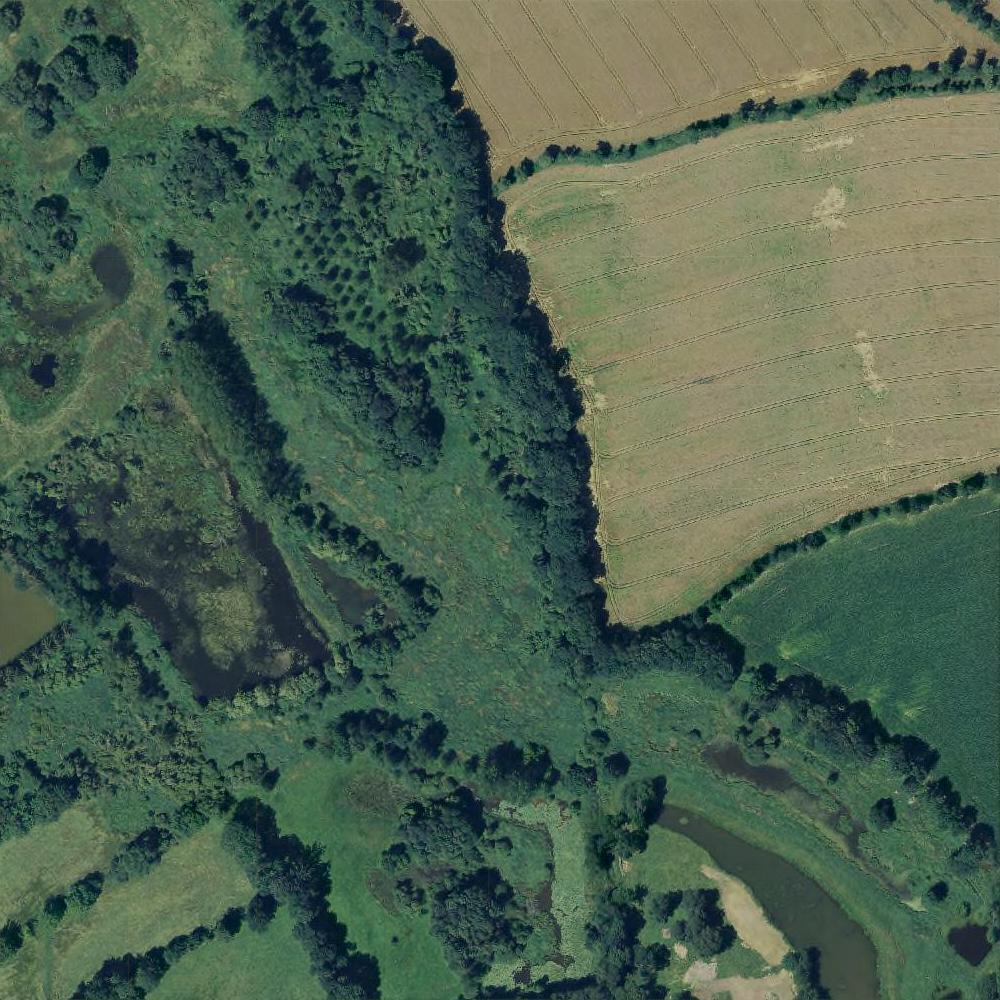} \\
         \includegraphics[width=.14\linewidth]{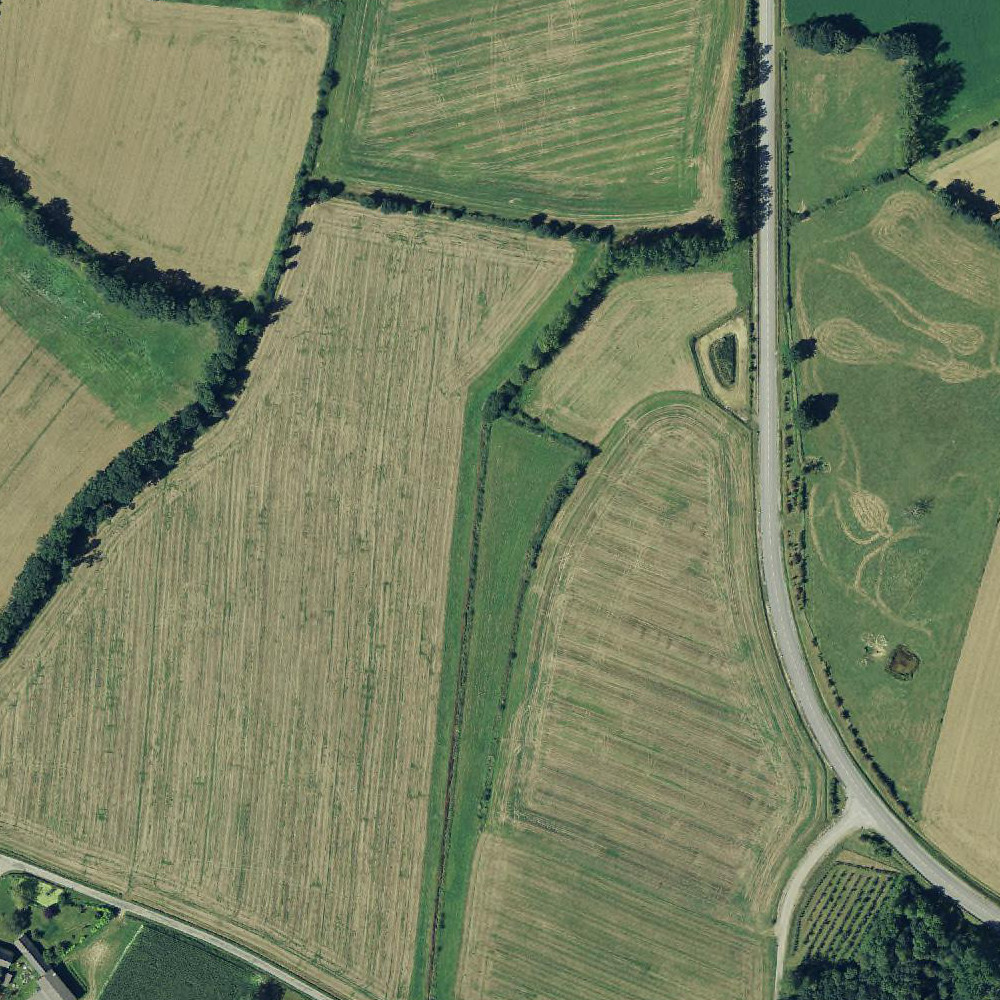} & \includegraphics[width=.14\linewidth]{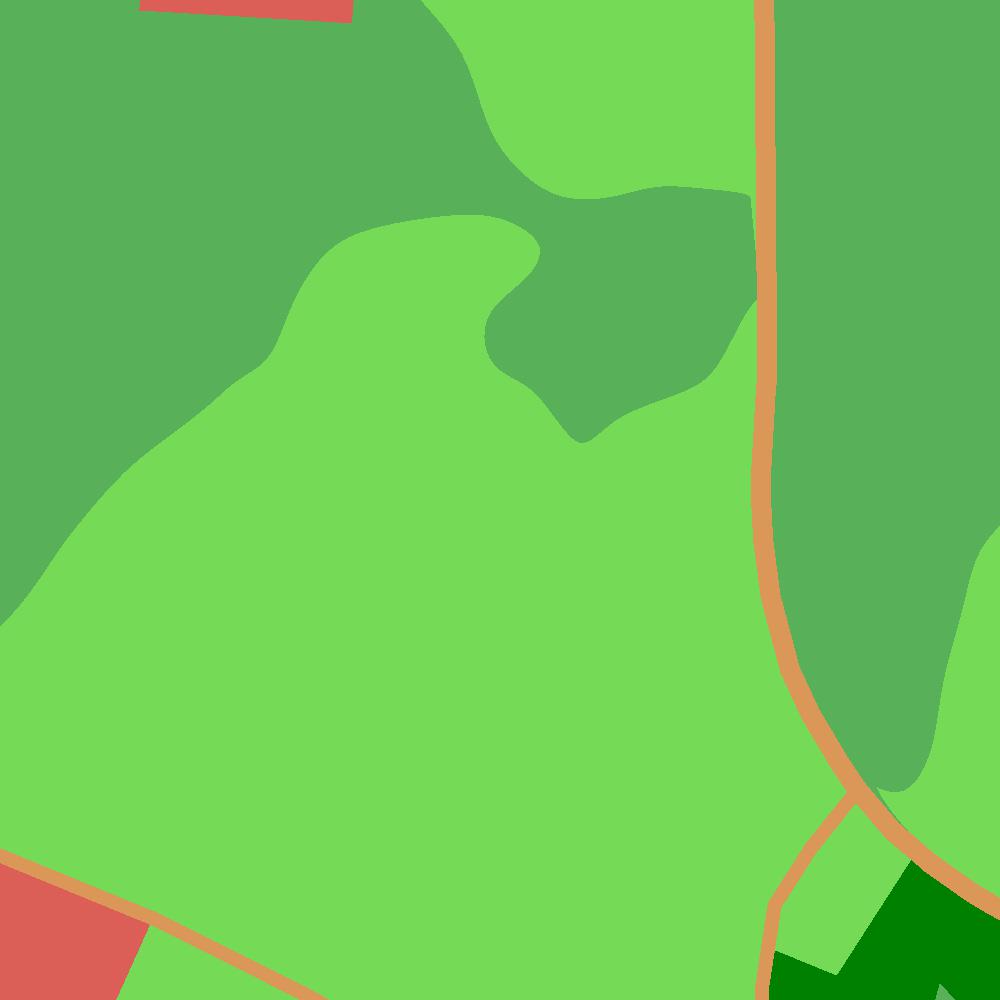} & \includegraphics[width=.14\linewidth]{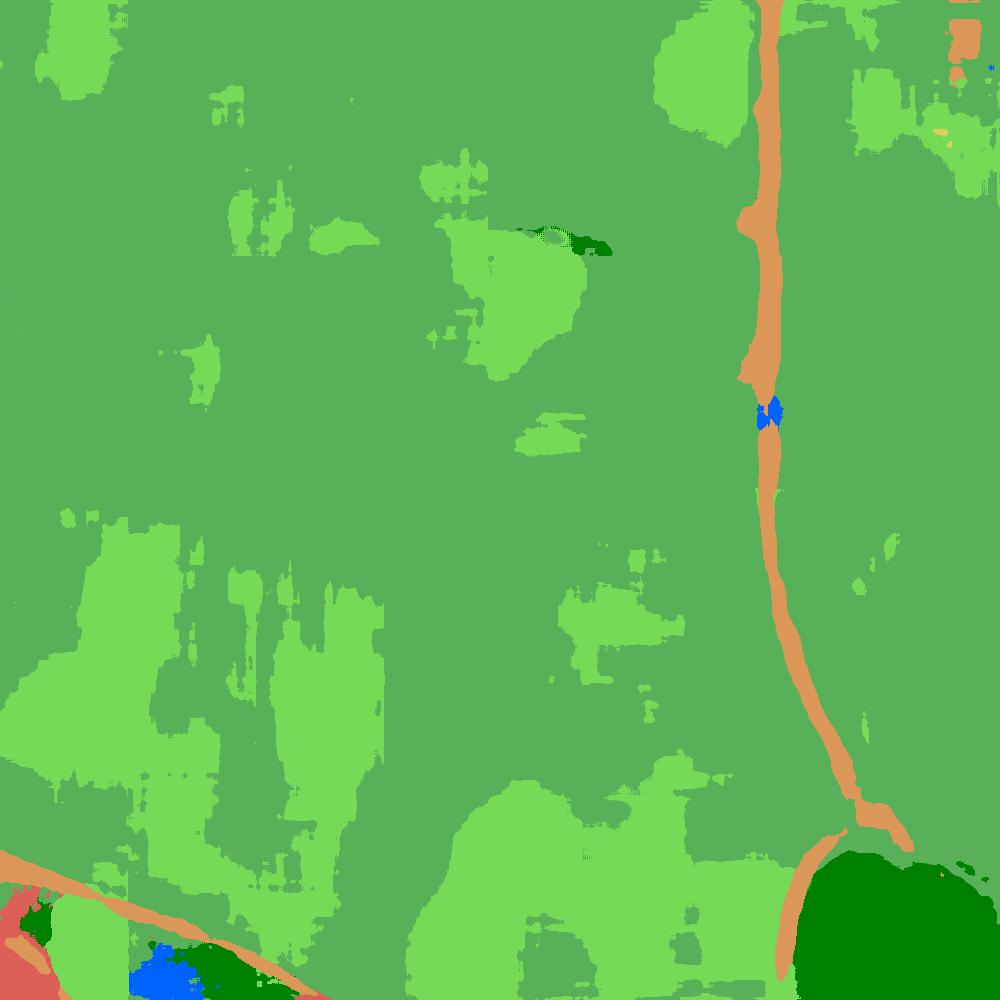} & \includegraphics[width=.14\linewidth]{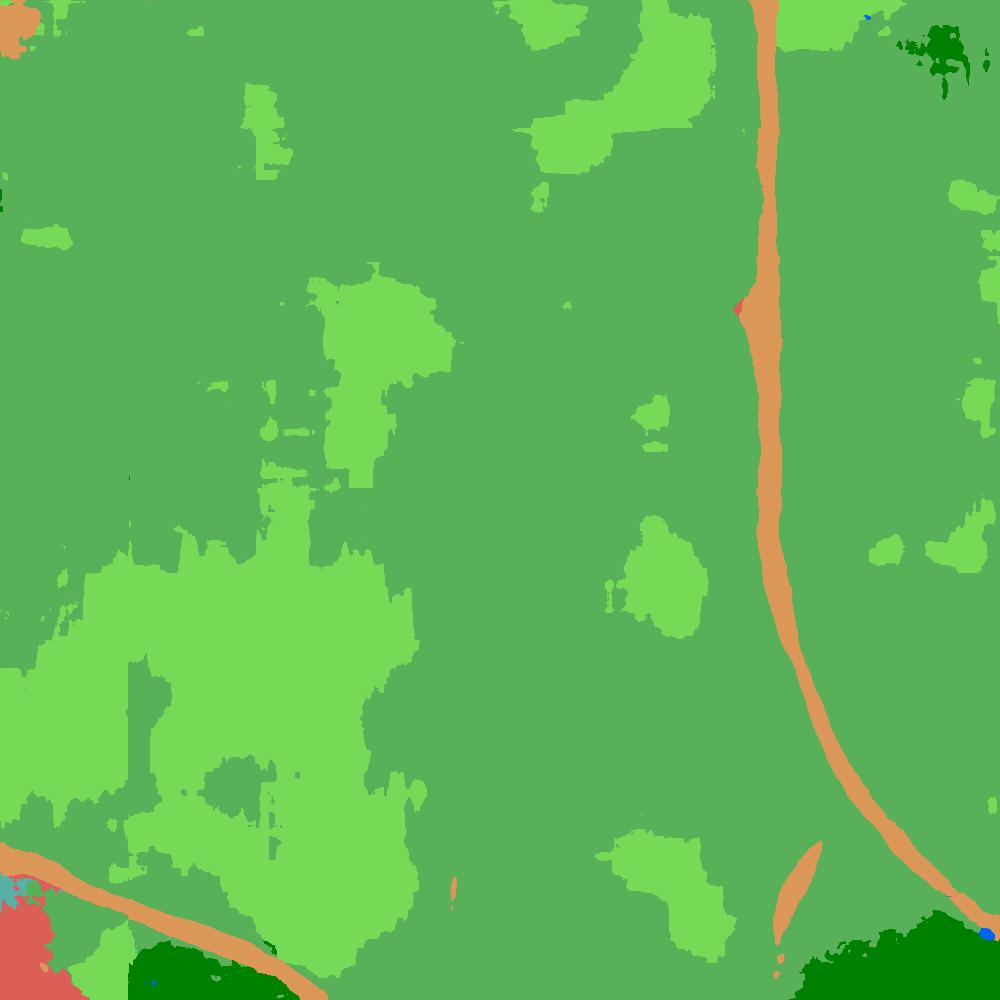} & \includegraphics[width=.14\linewidth]{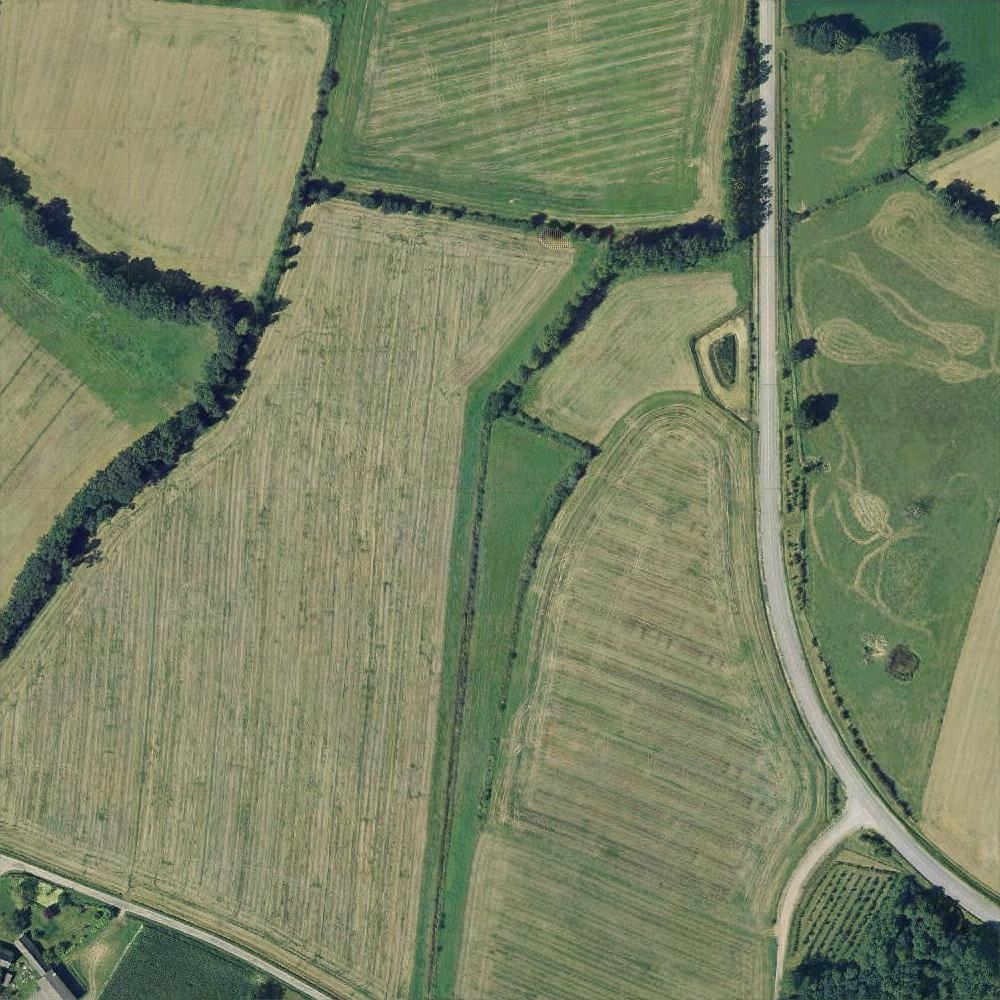} & \includegraphics[width=.14\linewidth]{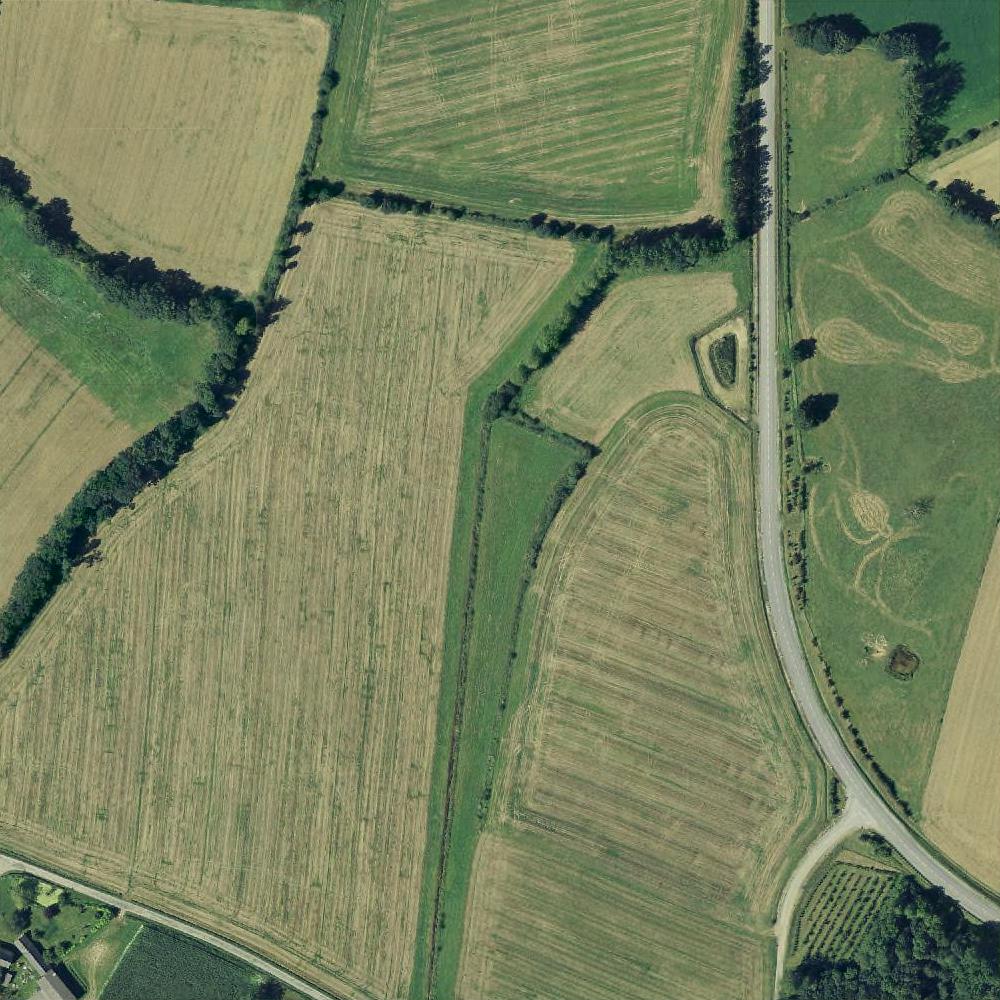} \\
         \includegraphics[width=.14\linewidth]{13-2014-0910-6260-LA93-0M50-E080_663_3753.jpg} & \includegraphics[width=.14\linewidth]{13-2014-0910-6260-LA93-0M50-E080_663_3753_gt.jpg} & \includegraphics[width=.14\linewidth]{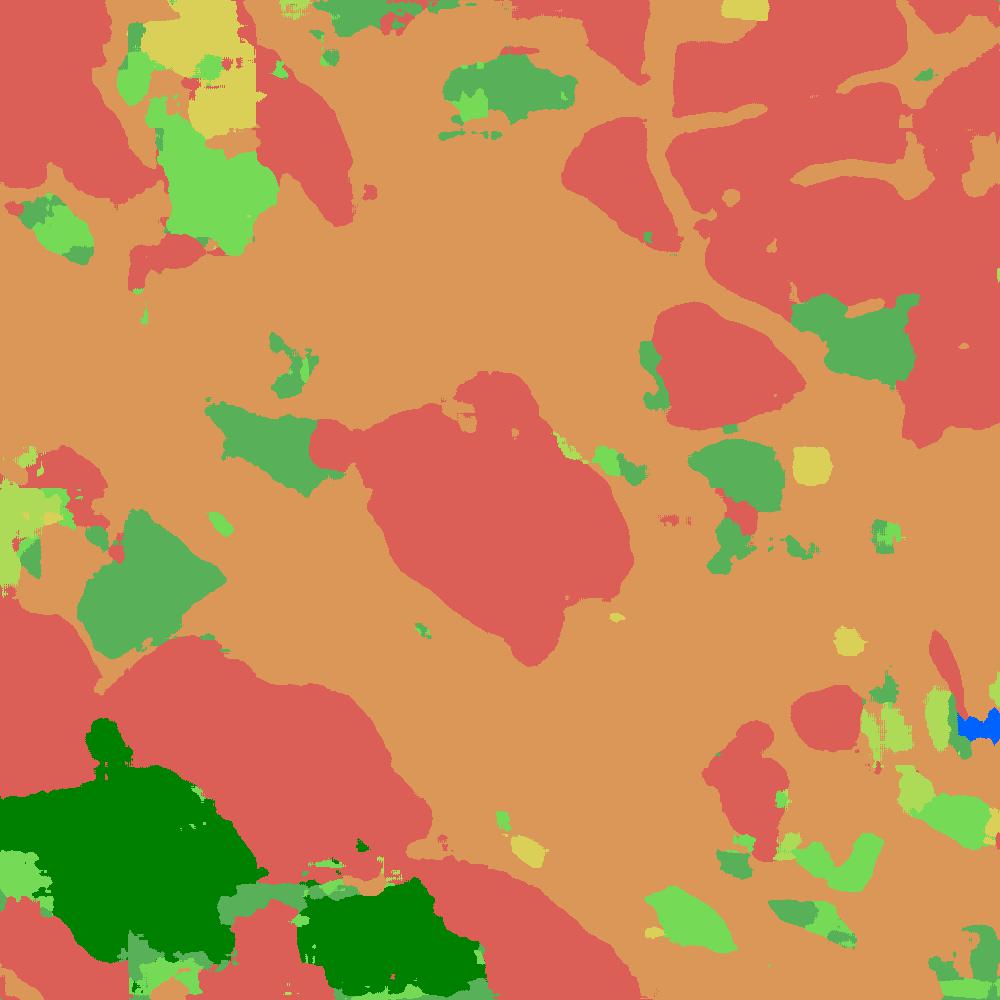} & \includegraphics[width=.14\linewidth]{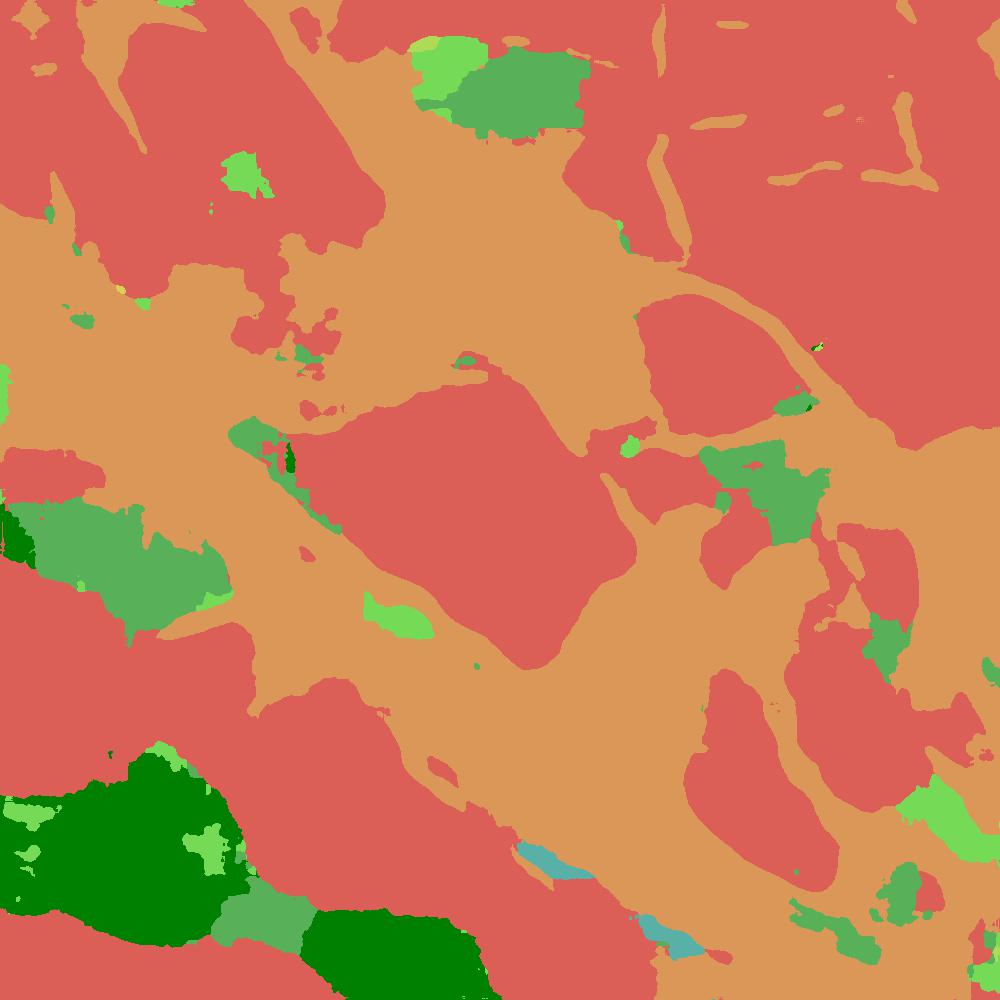} & \includegraphics[width=.14\linewidth]{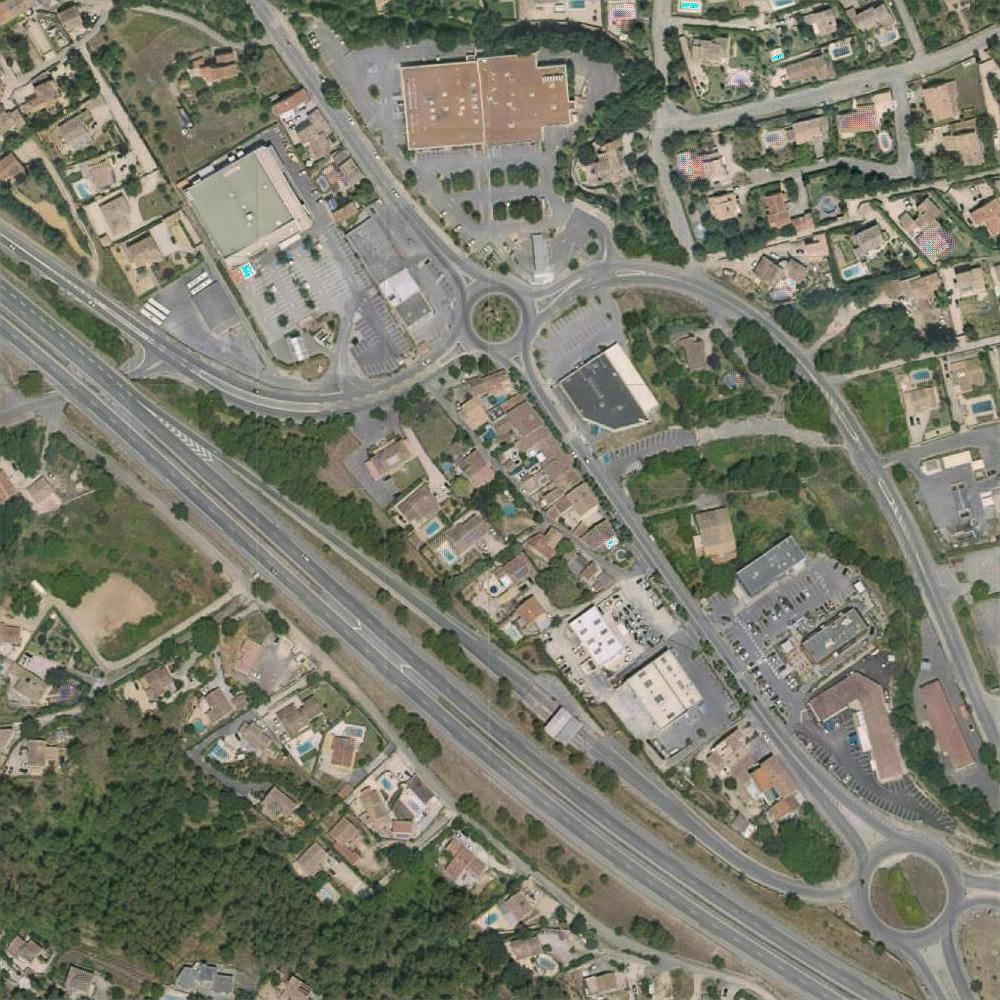} & \includegraphics[width=.14\linewidth]{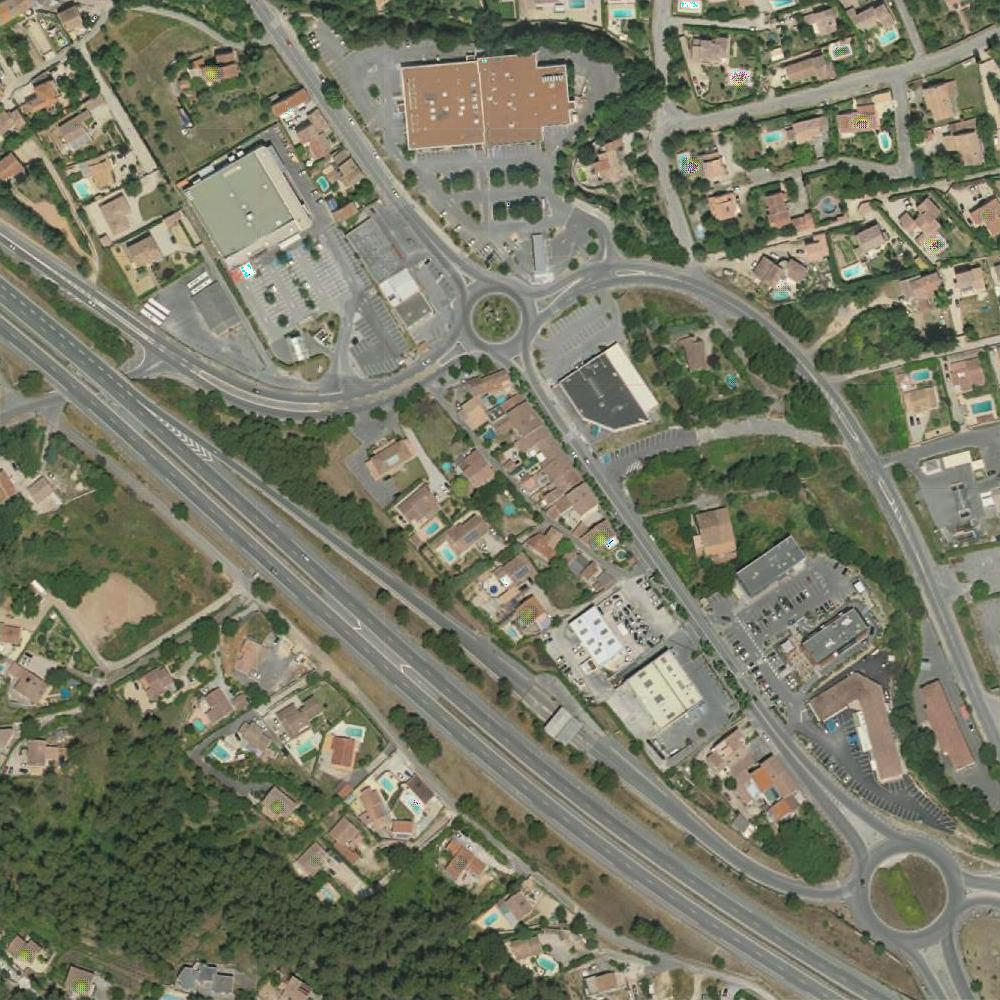} \\

         \includegraphics[width=.14\linewidth]{50-2012-0355-6955-LA93-0M50-E080_6517_222.jpg} & \includegraphics[width=.14\linewidth]{50-2012-0355-6955-LA93-0M50-E080_6517_222_gt.jpg} & \includegraphics[width=.14\linewidth]{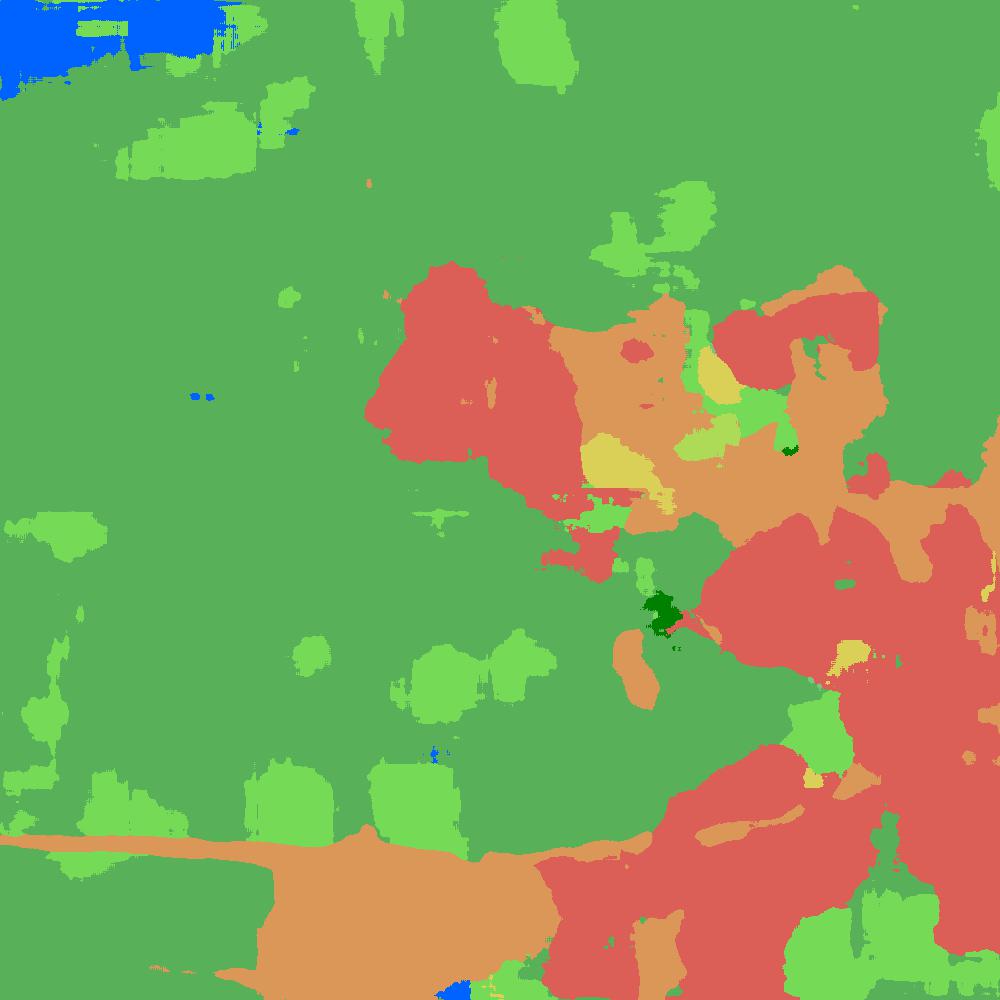} & \includegraphics[width=.14\linewidth]{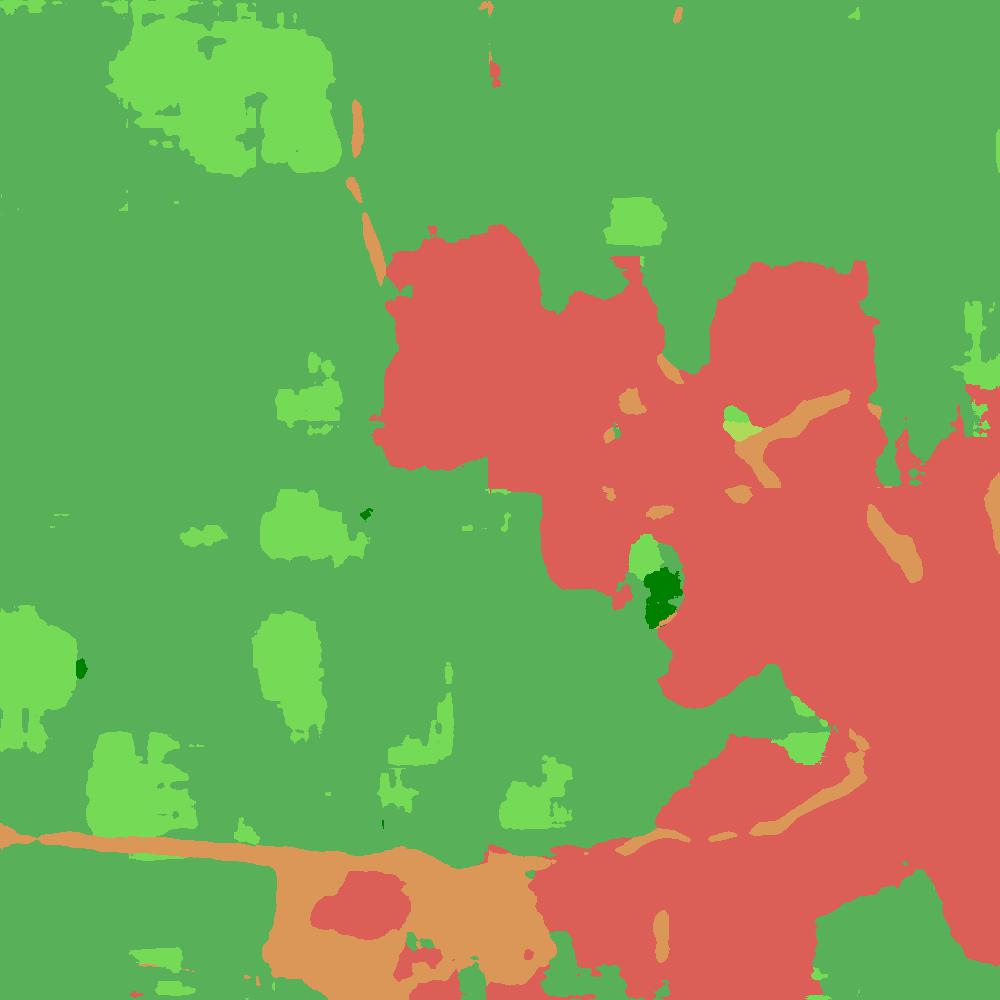} & \includegraphics[width=.14\linewidth]{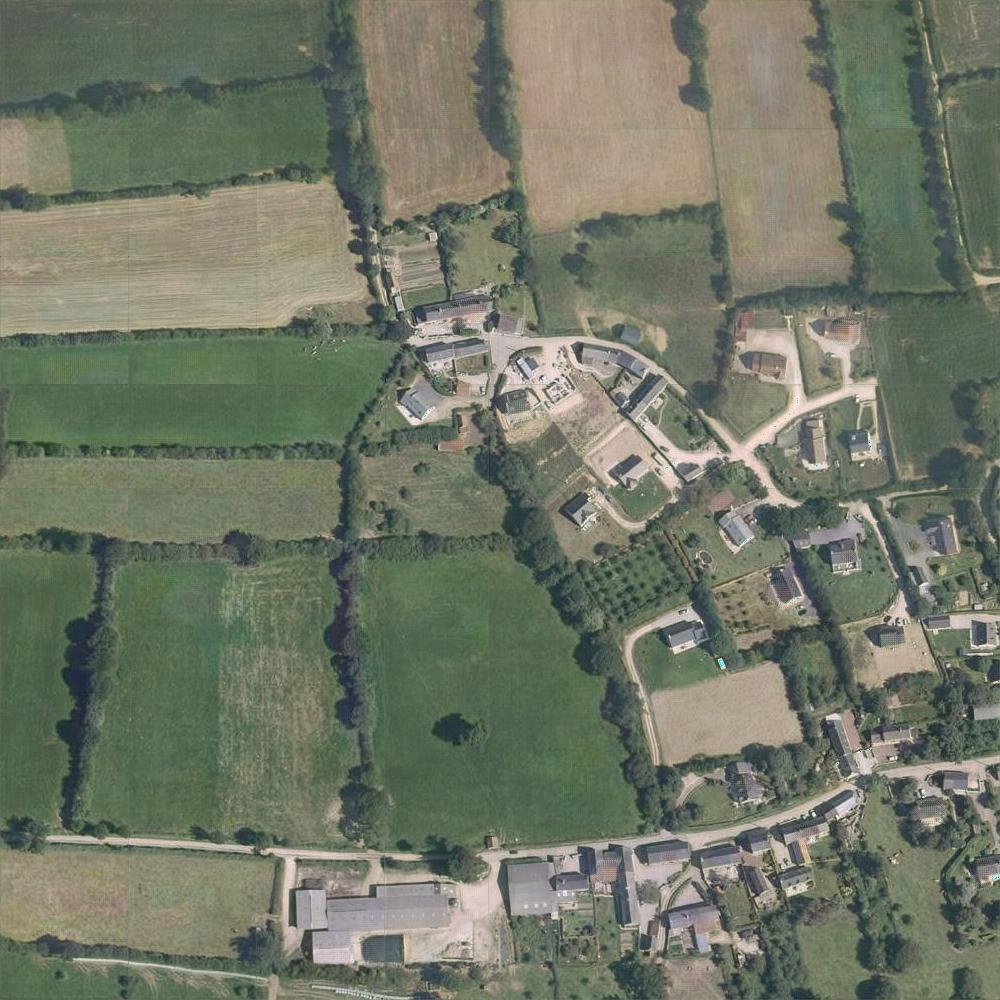} & \includegraphics[width=.14\linewidth]{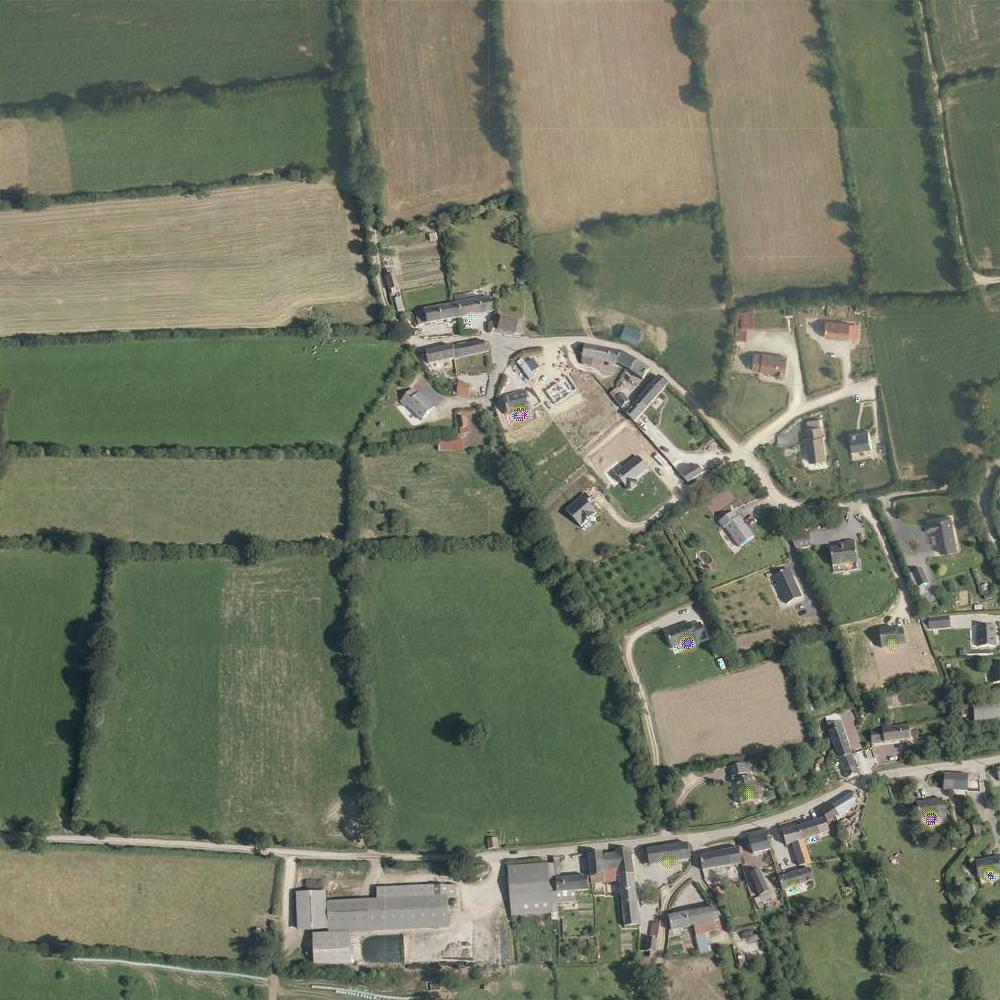} \\[3pt]

         Image & GT & $\LL_2$ & $\LL_1$ & $\LL_2$ & $\LL_1$ \\[4pt]
         \multicolumn{2}{c}{}& \multicolumn{2}{c}{Semantic Segmentation\quad\hfill} & \multicolumn{2}{c}{Reconstruction} 
      \end{tabular}
      \caption{Segmentation maps and reconstruction outputs for BerundaNet-late (U-Net backbone), using different unsupervised reconstruction losses for the auxiliary task.} \label{fig: results-different-losses-reconstruction}
   \end{center}
\end{figure}

\begin{figure}[!htbp]
   \begin{center}
      \setlength{\tabcolsep}{1.2pt}
      \begin{tabular}{cc@{\hspace{10pt}}cc@{\hspace{10pt}}cc}
         \includegraphics[width=.14\linewidth]{35-2012-0350-6815-LA93-0M50-E080_6893_61.jpg} & \includegraphics[width=.14\linewidth]{35-2012-0350-6815-LA93-0M50-E080_6893_61_gt.jpg} & \includegraphics[width=.14\linewidth]{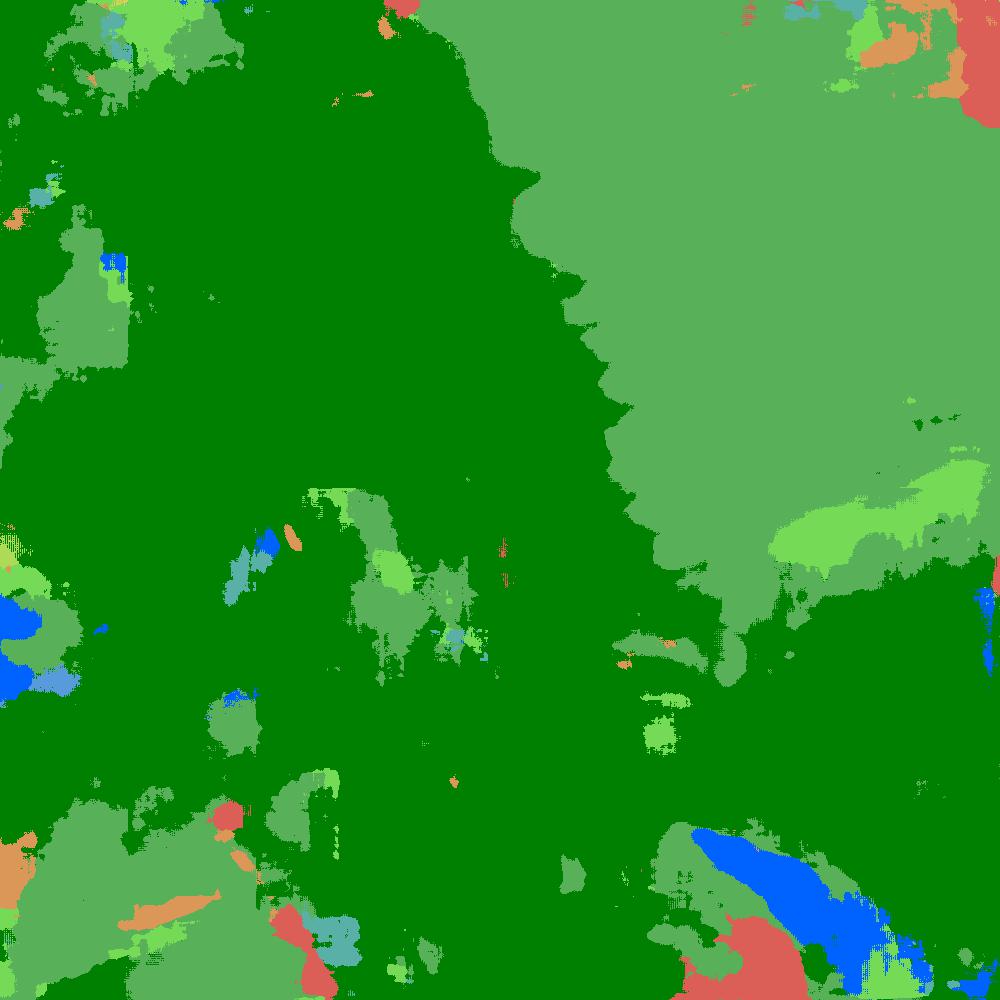} & \includegraphics[width=.14\linewidth]{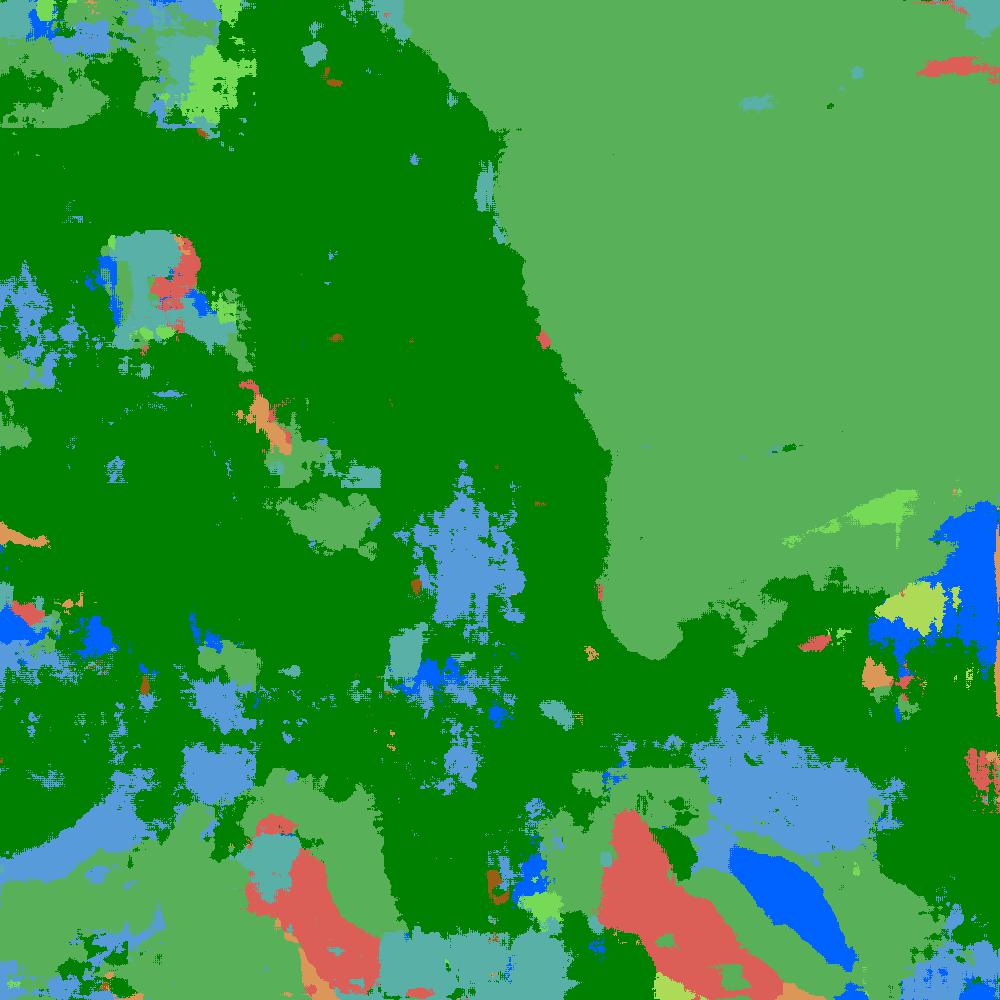} & \includegraphics[width=.14\linewidth]{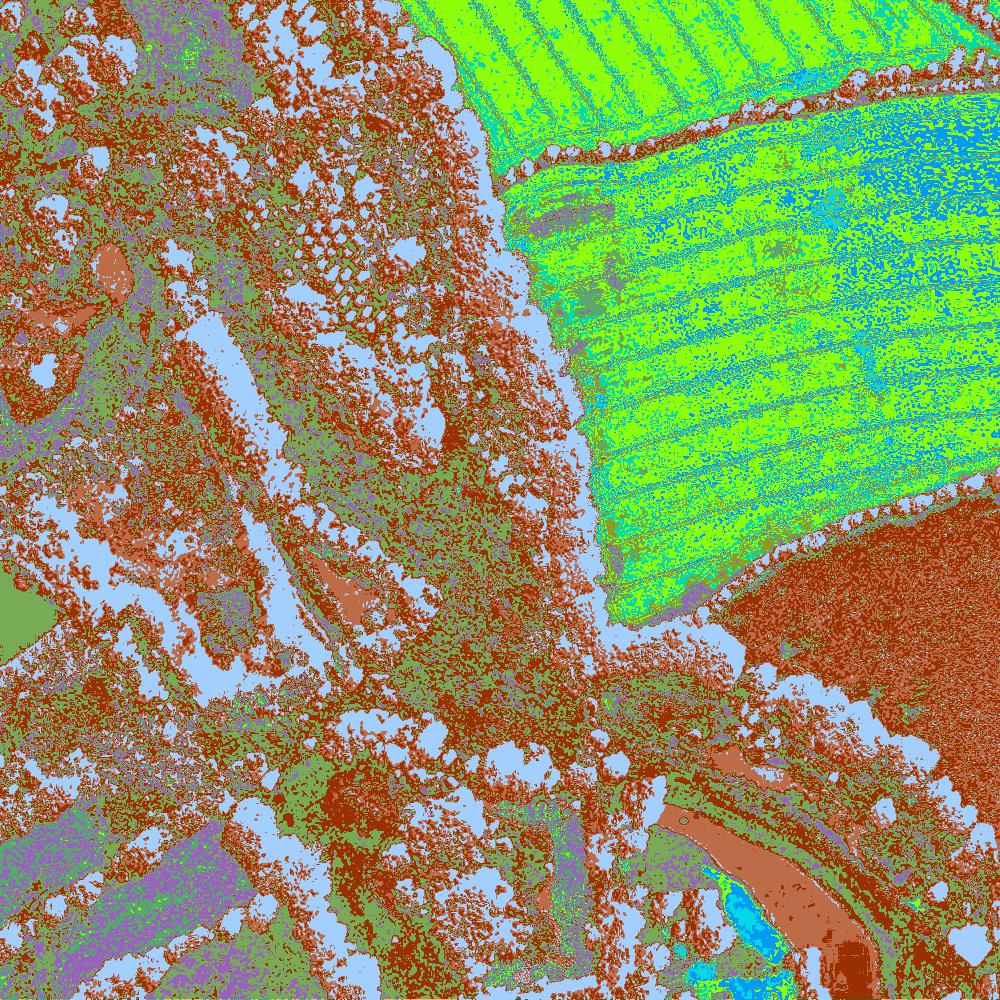} & \includegraphics[width=.14\linewidth]{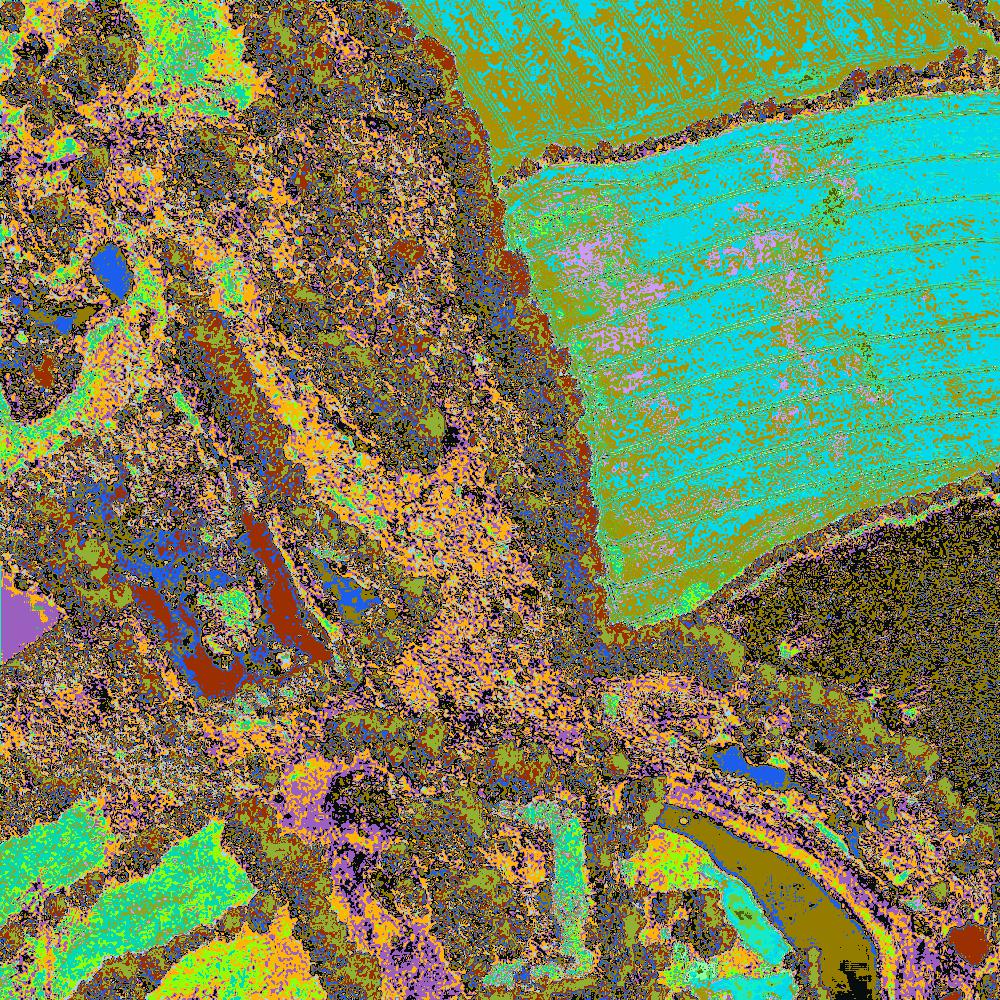} \\
         \includegraphics[width=.14\linewidth]{35-2012-0360-6770-LA93-0M50-E080_4563_8367.jpg} & \includegraphics[width=.14\linewidth]{35-2012-0360-6770-LA93-0M50-E080_4563_8367_gt.jpg} & \includegraphics[width=.14\linewidth]{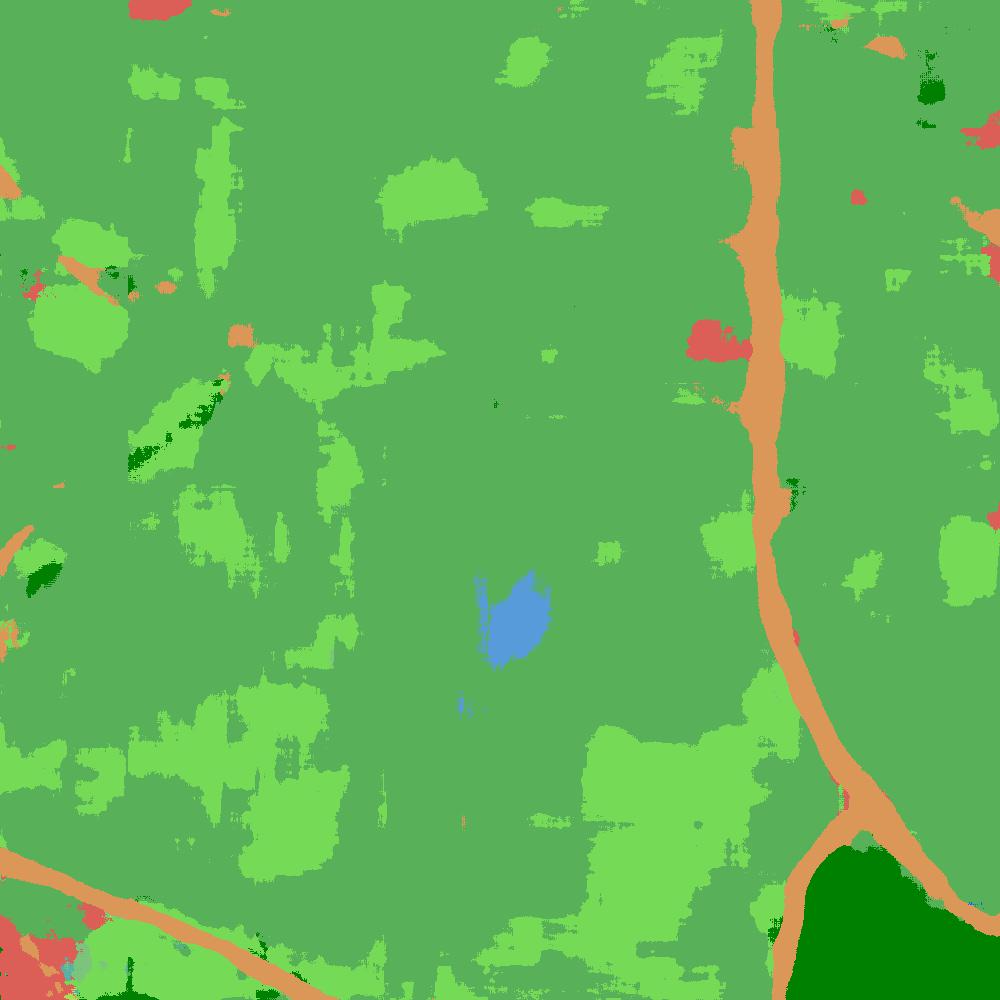} & \includegraphics[width=.14\linewidth]{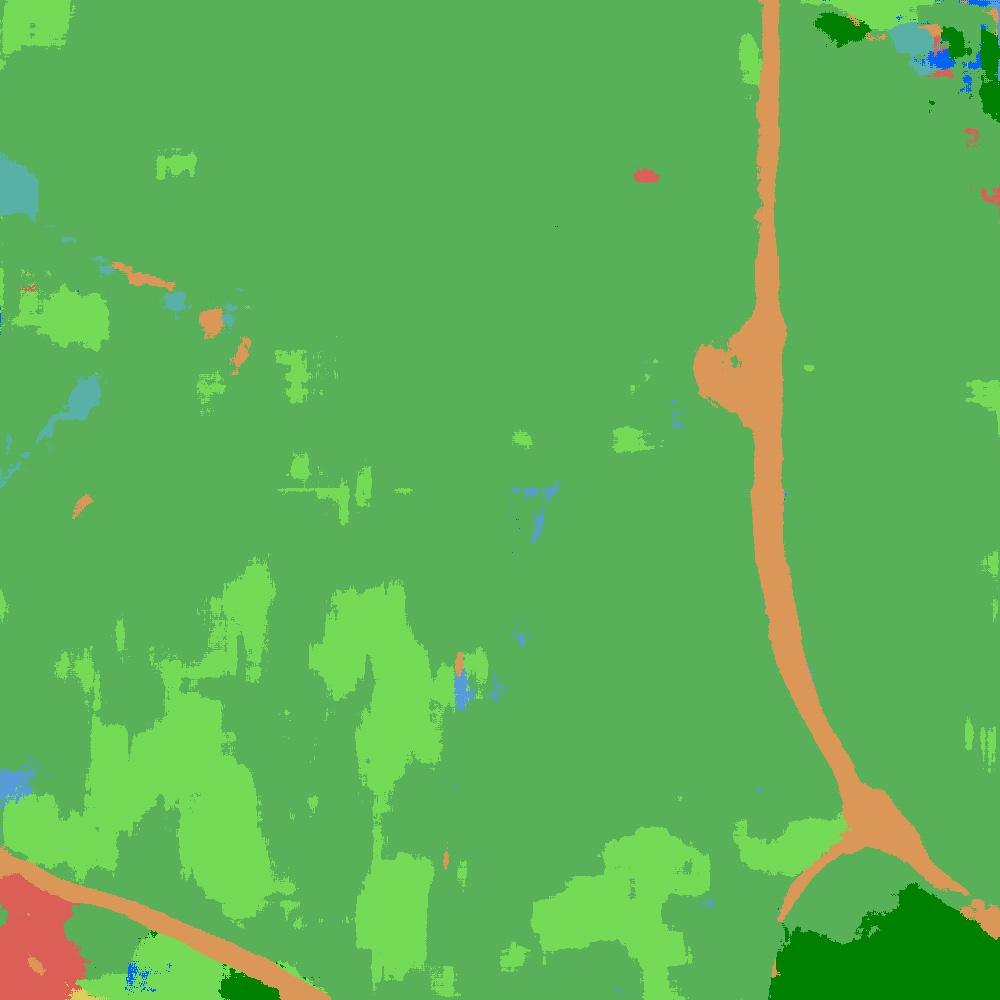} & \includegraphics[width=.14\linewidth]{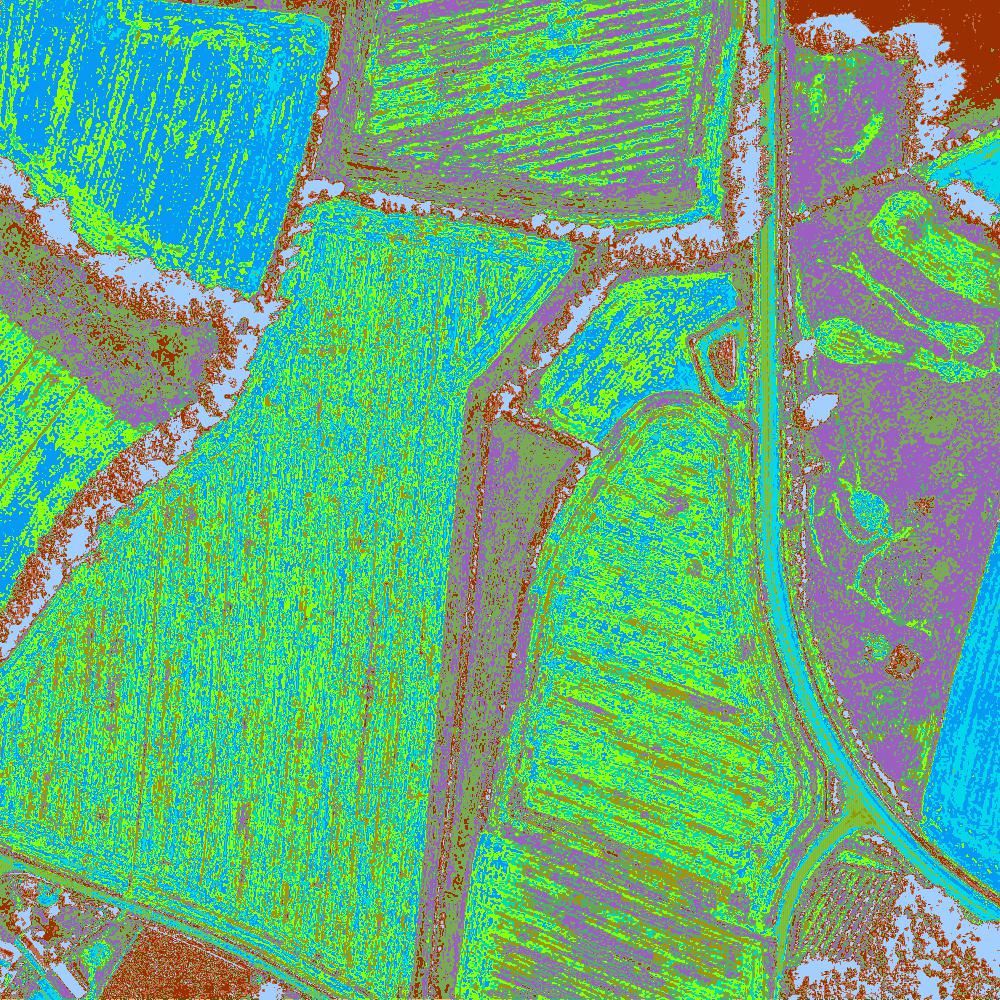} & \includegraphics[width=.14\linewidth]{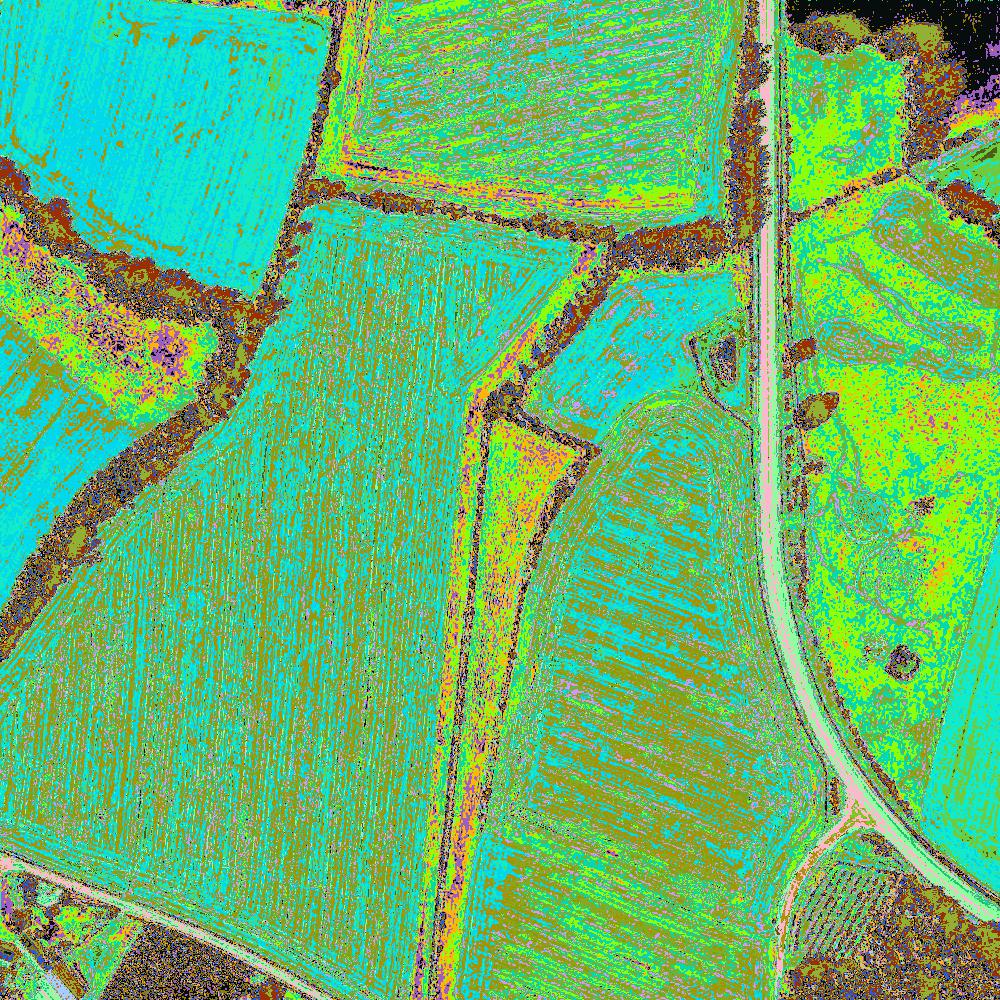} \\

         \includegraphics[width=.14\linewidth]{13-2014-0910-6260-LA93-0M50-E080_663_3753.jpg} & \includegraphics[width=.14\linewidth]{13-2014-0910-6260-LA93-0M50-E080_663_3753_gt.jpg} & \includegraphics[width=.14\linewidth]{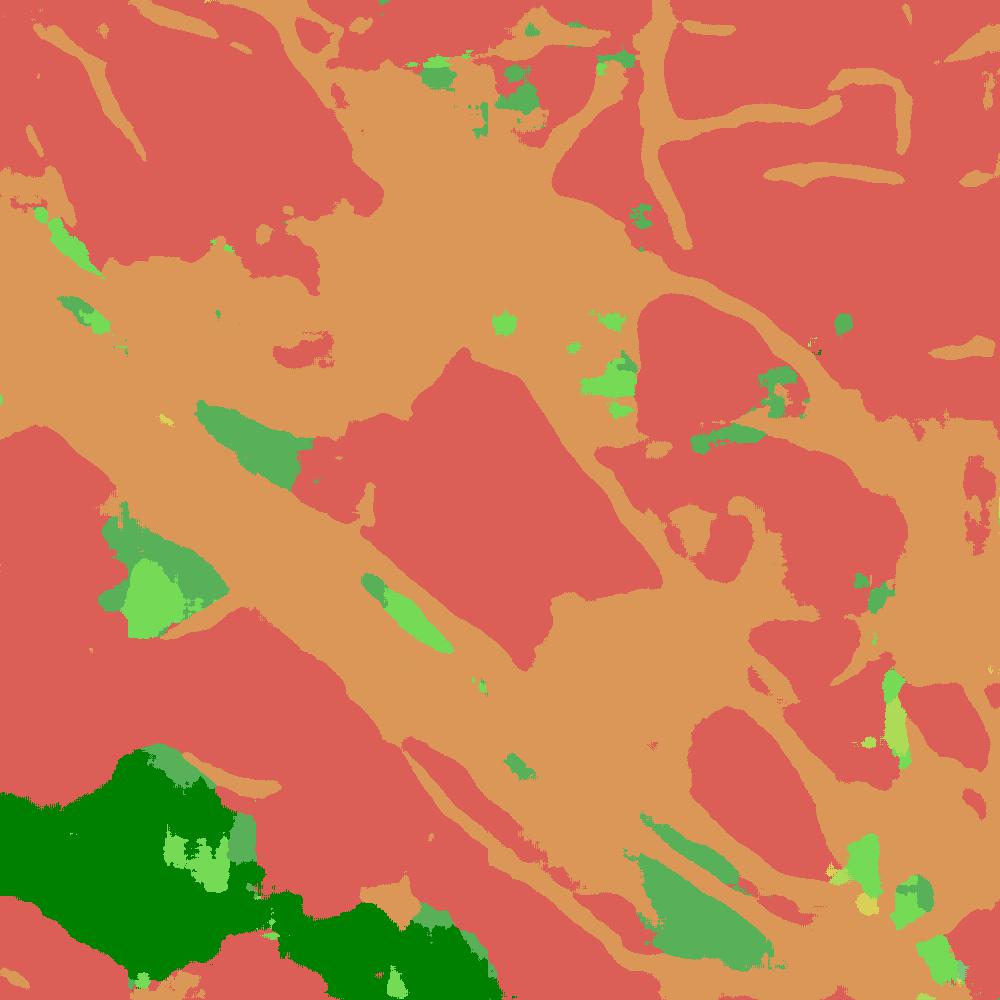} & \includegraphics[width=.14\linewidth]{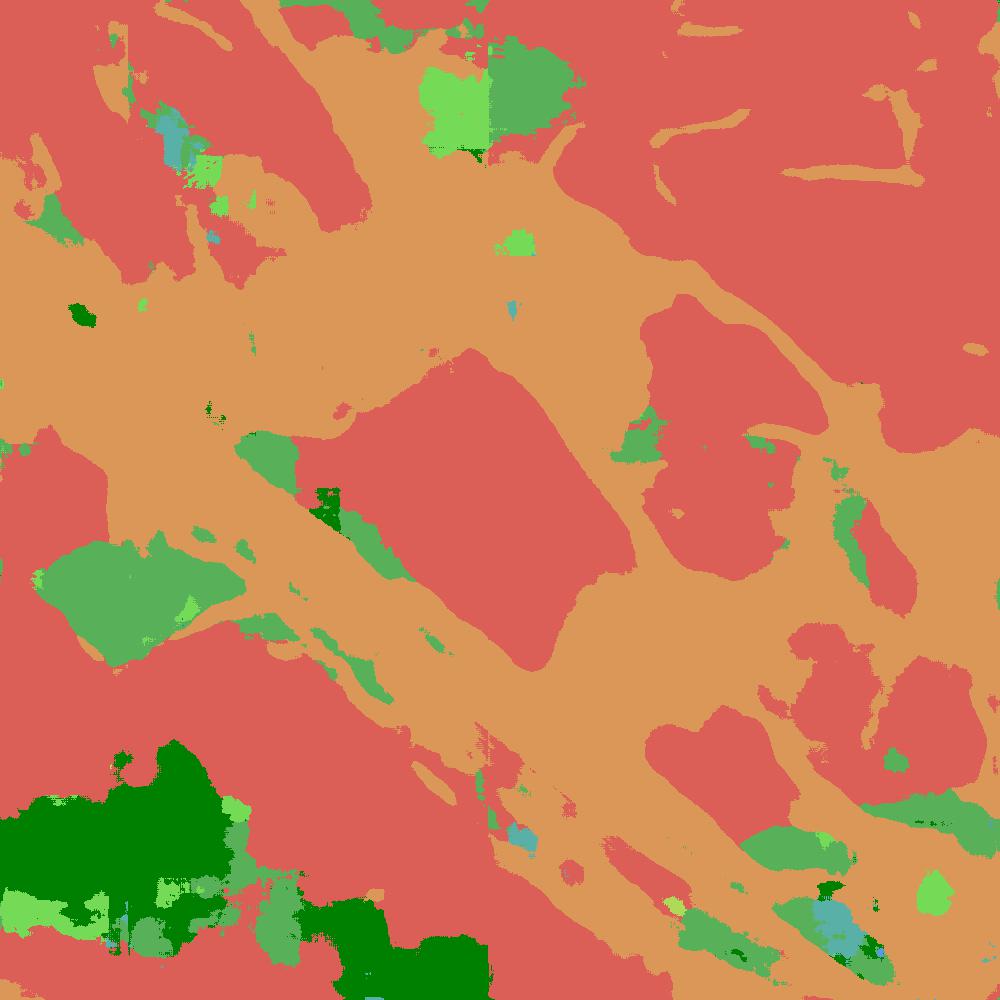} & \includegraphics[width=.14\linewidth]{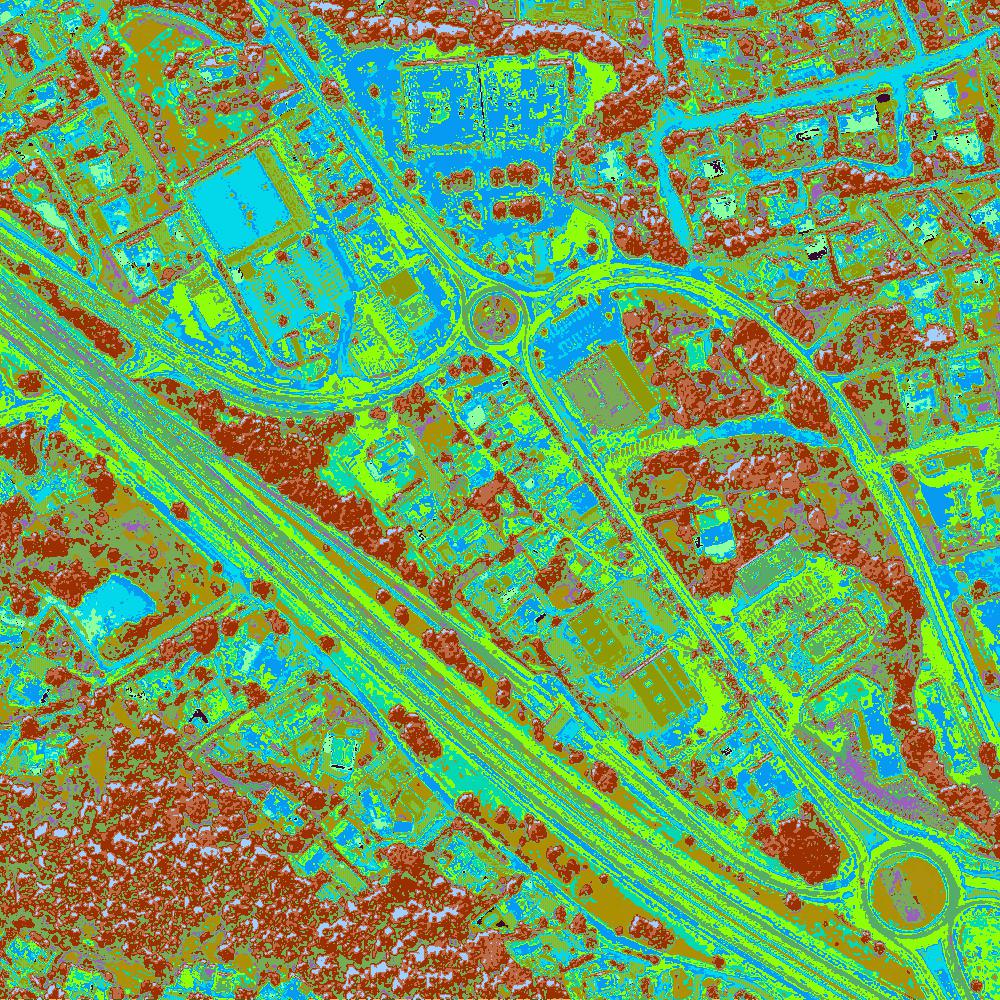} & \includegraphics[width=.14\linewidth]{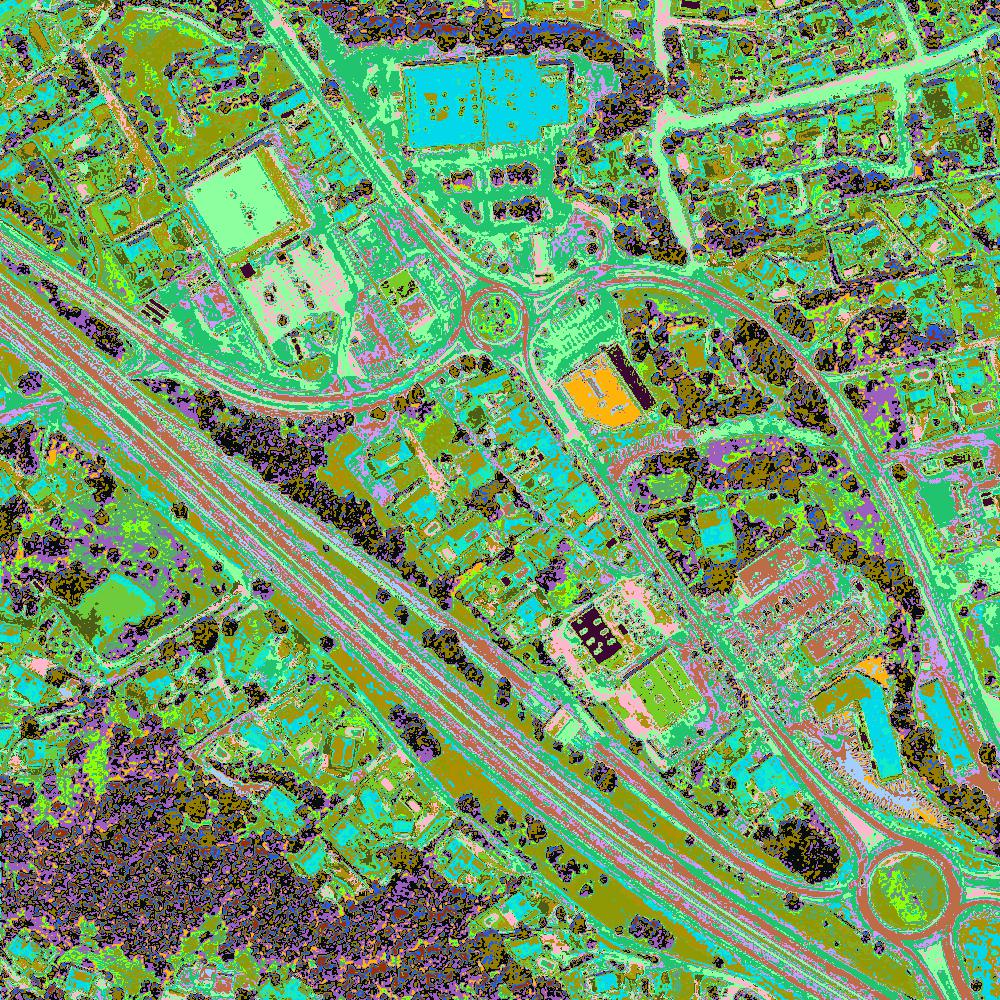} \\

         \includegraphics[width=.14\linewidth]{50-2012-0355-6955-LA93-0M50-E080_6517_222.jpg} & \includegraphics[width=.14\linewidth]{50-2012-0355-6955-LA93-0M50-E080_6517_222_gt.jpg} & \includegraphics[width=.14\linewidth]{50-2012-0355-6955-LA93-0M50-E080_6517_222_unet_km_pred.jpg} & \includegraphics[width=.14\linewidth]{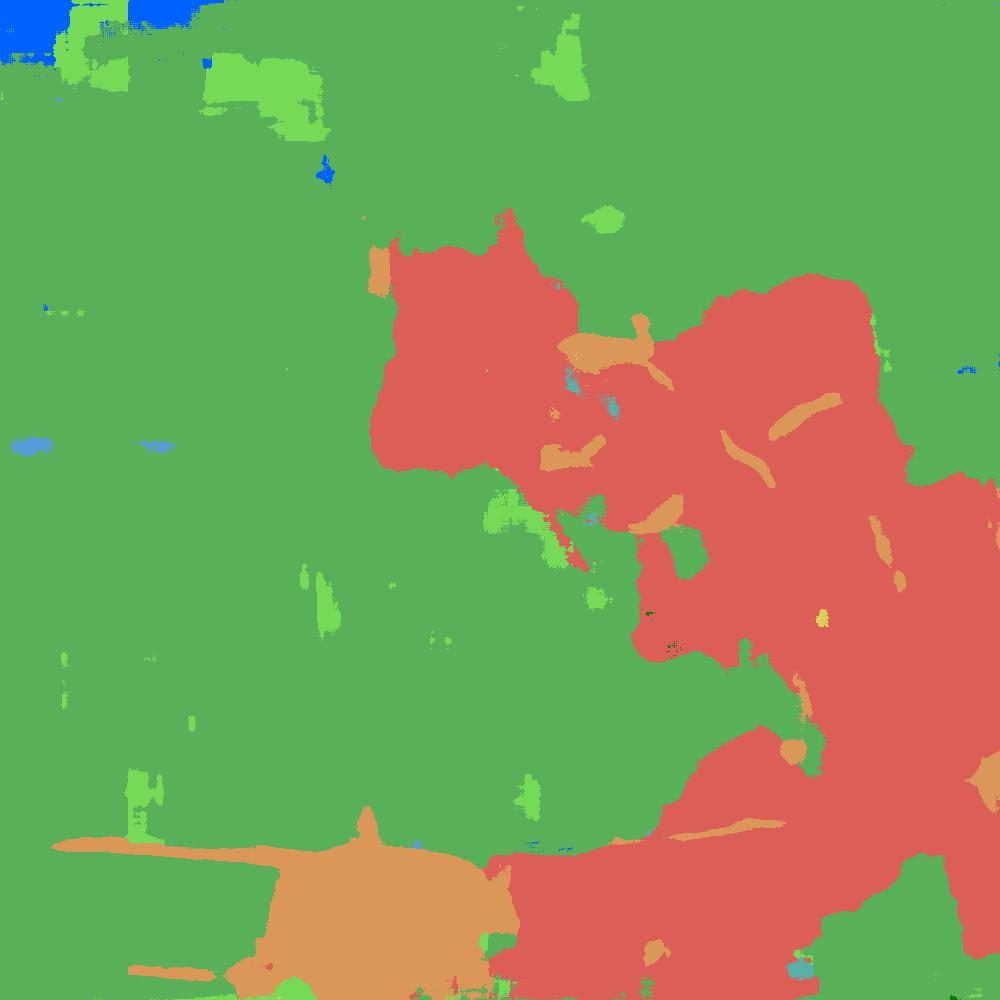} & \includegraphics[width=.14\linewidth]{50-2012-0355-6955-LA93-0M50-E080_6517_222_unet_km_unsup.jpg} & \includegraphics[width=.14\linewidth]{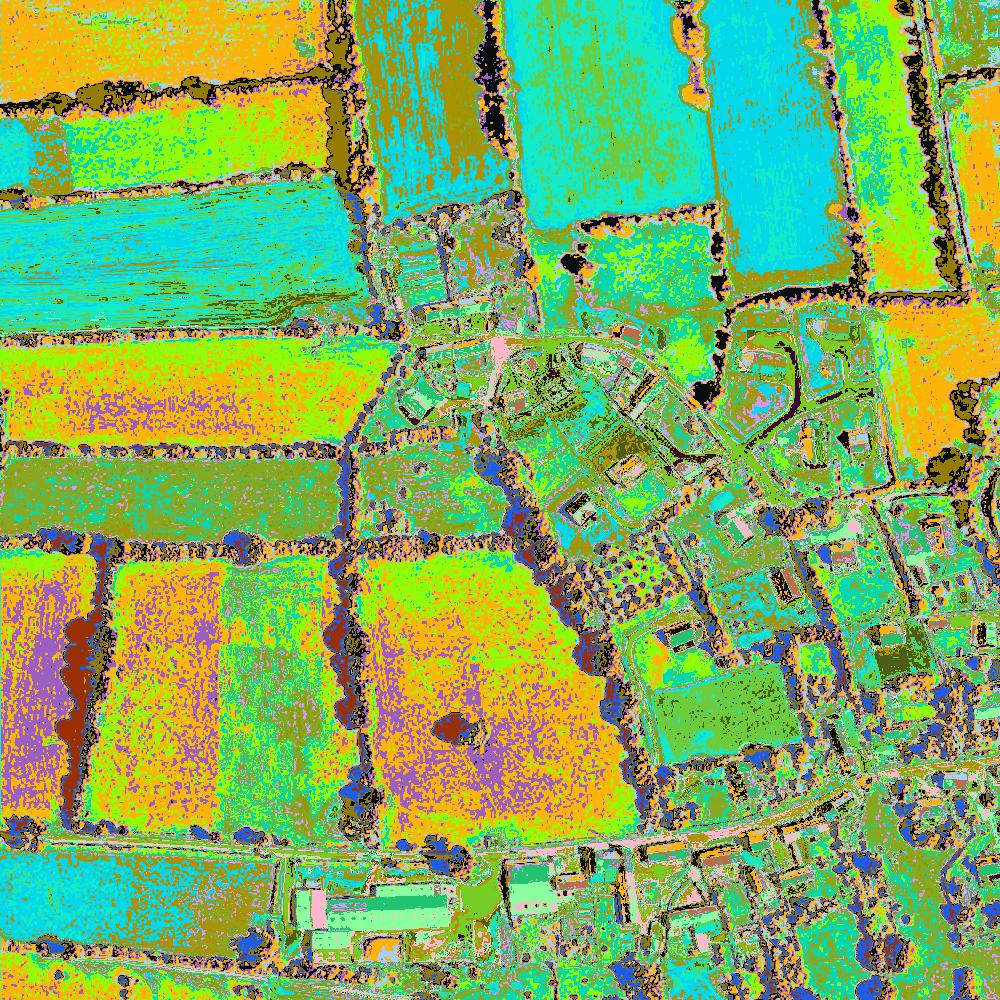} \\[3pt]

         Image & GT & $\LL_{km}$ & $\LL_{MS}$ & $\LL_{km}$ & $\LL_{MS}$ \\[4pt]
         \multicolumn{2}{c}{}& \multicolumn{2}{c}{Semantic Segmentation\quad\hfill} & \multicolumn{2}{c}{Unsup. Segmentation} 
      \end{tabular}
      \caption{ Semantic segmentation maps and unsupervised segmentation outputs for BerundaNet-late (U-Net backbone), using different unsupervised segmentation losses for auxiliary task.} \label{fig: results-different-losses-segmentation}
   \end{center}
\end{figure}

Table~\ref{tab:tmf-losses-comparison} reports the results obtained through these experiments. Figure~\ref{fig: results-different-losses-reconstruction} exhibits some examples of segmentation maps and unsupervised outputs obtained by BerundaNet-late with reconstructions losses ($\LL_1$ and $\LL_2$) at inference time, while Figure~\ref{fig: results-different-losses-segmentation} shows examples using unsupervised segmentation as auxiliary task.

For the reconstruction task, $\LL_1$ loss outperforms the $\LL_2$ approach, this is confirmed by visual examples in Fig.~\ref{fig: results-different-losses-reconstruction} where we perceive that results are marginally better for $\LL_1$ than for $\LL_2$ in terms of smoothness, especially in urban areas like the third and fourth row examples.

In the case of segmentation, $\LL_{km}$ and $\LL_{MS}$ are somehow equivalent. However, from Figure~\ref{fig: results-different-losses-segmentation} the $\LL_{km}$ loss seems to be superior to $\LL_{MS}$ in most cases, especially when it comes to road detection.

\begin{table}[H]
   \begin{center}
      \caption{Auxiliary unsupervised loss  effect comparison using BerundaNet-late with U-Net backbone.}\label{tab:tmf-losses-comparison}
      \begin{tabular}{cccc} \toprule
      \emph{Auxiliary Task} & \emph{Aux. Loss} & \emph{OA (\%)} &\emph{ mIoU (\%)} \\ \midrule
      \multirow{2}{*}{Reconstruction} & $\LL_1$ & \bf{47.90} & \bf{18.70} \\ 
       & $\LL_2$ & 44.55 & 16.27 \\\midrule
      \multirow{2}{*}{Segmentation} & $\LL_{km}$ & 46.92 & 18.26 \\
      & $\LL_{MS}$~\cite{kim2019mumford} & 46.88 & 18.57\\
      \bottomrule
      \end{tabular}
   \end{center}   
\end{table}

\subsection{\added{Experiments on the Christchurch Aerial Semantic Dataset}}

\added{We also perform experiments on the Christchurch Aerial Semantic Dataset (CASD)}\footnote{Available at \url{https://doi.org/10.5281/zenodo.3566005}} \added{to test the reliability of our framework~\cite{castillo2020onauxiliarylosses}}.

\added{CASD comprises aerial imagery at 10~cm/px resolution over Christchurch, New Zealand. Dense semantic annotations were produced by ONERA/DTIS on 4 images, considering 4 classes: buildings, cars, vegetation and background~\cite{audebert2017segment,randrianarivo2013urban}. The dataset also includes 20 aerial images without annotations, which makes it suitable for semi-supervised learning algorithms.}

\added{For these experiments, we use a training partition containing labeled and unlabeled data --2 annotated tiles and 20 non-annotated tiles--, and keep 2 annotated tiles for validation. We train a BerundaNet-late architecture with U-Net backbone, because of its simplicity and efficiency. The network is trained during 50 pseudo-epochs with 5000 labeled iterations and 5000 unlabeled iterations. Since the dataset allows it (training only takes a few hours), we also evaluate different values of the hyperparameter $\lambda$ (in Eq.~\eqref{eq: semisupervised-loss})}.

\added{Results are reported in Table~\ref{tab:christchurch-semi-sup}. Mean and variance are obtained over 4 runs of each experiment. We note that semi-supervised methods outperform the supervised setting. Moreover, best scores are obtained with unsupervised segmentation losses, and especially our relaxed K-means loss allows to improve the mIoU score by +3.39\% and overall accuracy by +1.97\%, with respect to the supervised setting.   }

\begin{table}[htpb]
   \caption{\added{Results comparison for supervised and semi-supervised methods over the Christchurch Aerial Semantic Dataset.}} 
   \centering
   \setlength{\tabcolsep}{10pt}
   \begin{tabular}{>{\centering\arraybackslash}m{1.4cm}>{\centering\arraybackslash}m{.8cm}>{\centering\arraybackslash}m{.8cm}>{\centering\arraybackslash}m{.5cm}cc}
   \toprule
   \emph{Mode} & \emph{Aux. Task} & \emph{Aux. Loss} & $\lambda$ & \emph{OA (\%)} & \emph{mIoU (\%)} \\ \midrule
   \multirow{1}{*}{Sup} & - & -  & - & $81.06 \pm 0.46$ & $67.43 \pm 0.49$ \\ \midrule
   \multirow{4}{*}{Semi-sup} &\multirow{2}{*}{Rec} & $\mathcal{L}_1$ & 0.5 & $82.28 \pm 0.55 $ & $68.78 \pm 1.27$ \\
        & & $\mathcal{L}_2$ & 5 & $82.36 \pm 0.42 $ & $ 68.99 \pm 0.85 $ \\ \cmidrule{2-6}
        & \multirow{2}{*}{Seg} & $\mathcal{L}_{km}$ & 1 & $\mathbf{83.03\pm 0.42}$ & $ \mathbf{70.82\pm 0.35}$\\ 
        & & $\mathcal{L}_{MS}$ & 1 & $82.94\pm0.26$ & $70.24\pm0.84$\\ \bottomrule
   \end{tabular}
   
   \label{tab:christchurch-semi-sup}
\end{table}

\begin{figure}[htbp]
   \begin{center}
      \setlength{\tabcolsep}{1pt}
       \begin{tabular}{ccccccc}
       \small
       \includegraphics[width=.14\linewidth]{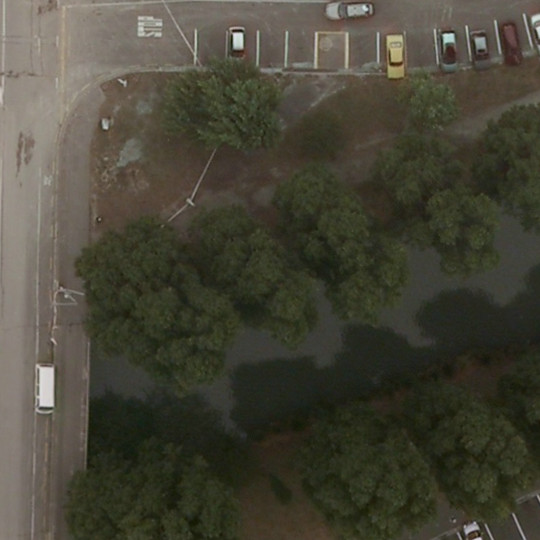} & \includegraphics[width=.14\linewidth]{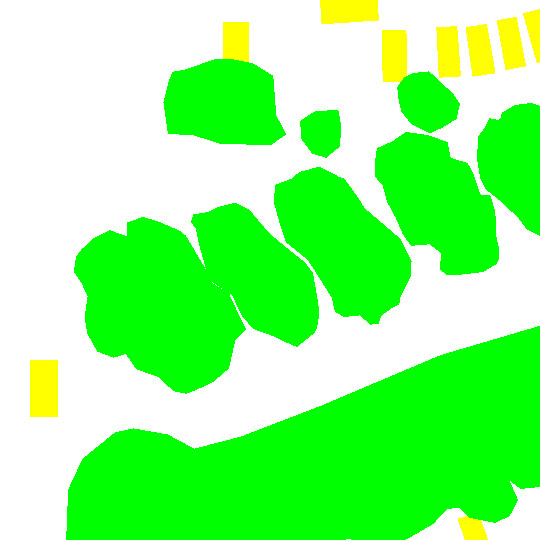} & \includegraphics[width=.14\linewidth]{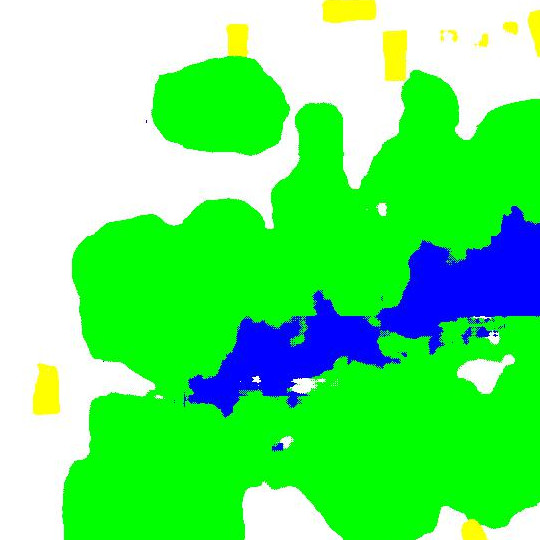}& \includegraphics[width=.14\linewidth]{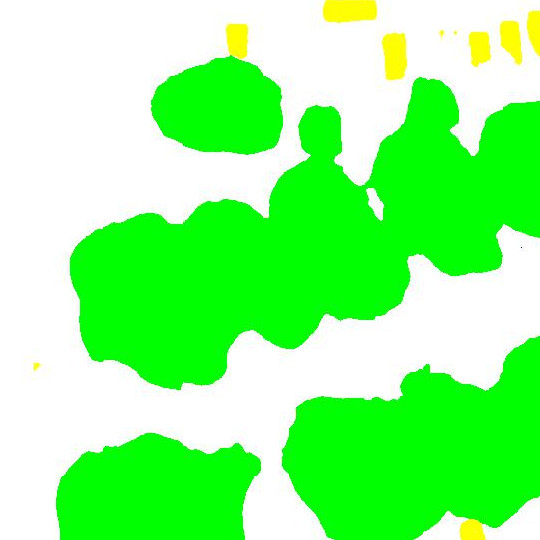} & \includegraphics[width=.14\linewidth]{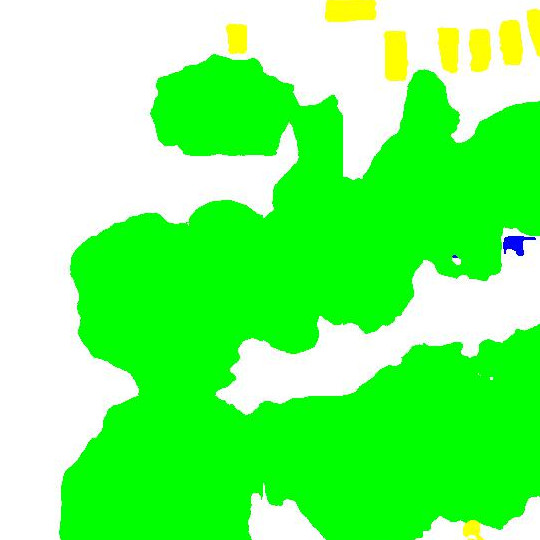} 
       & \includegraphics[width=.14\linewidth]{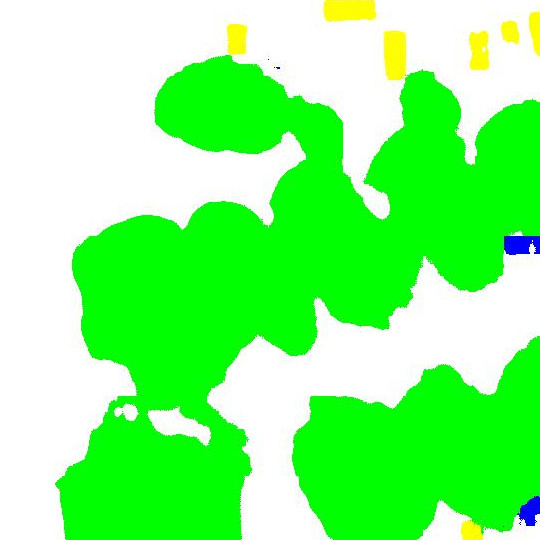} & \includegraphics[width=.14\linewidth]{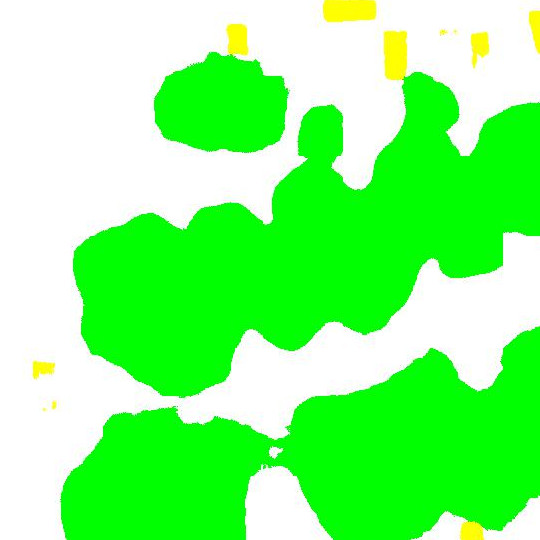} \\
       \includegraphics[width=.14\linewidth]{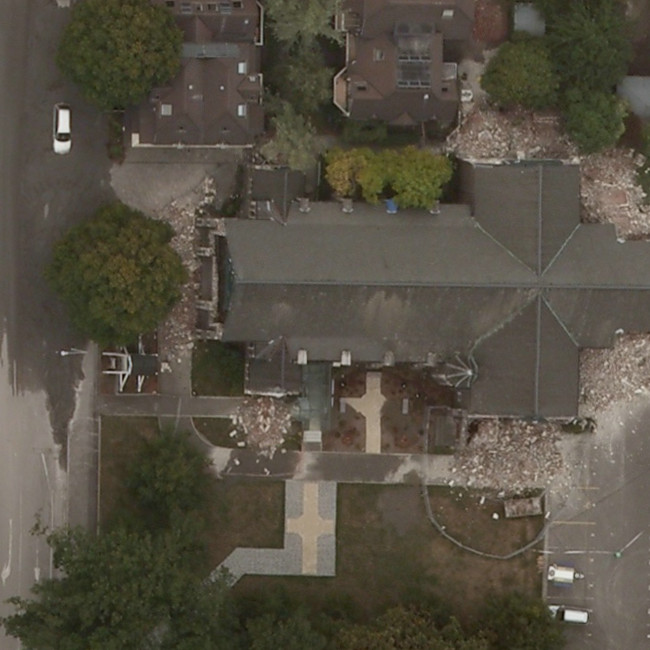} & \includegraphics[width=.14\linewidth]{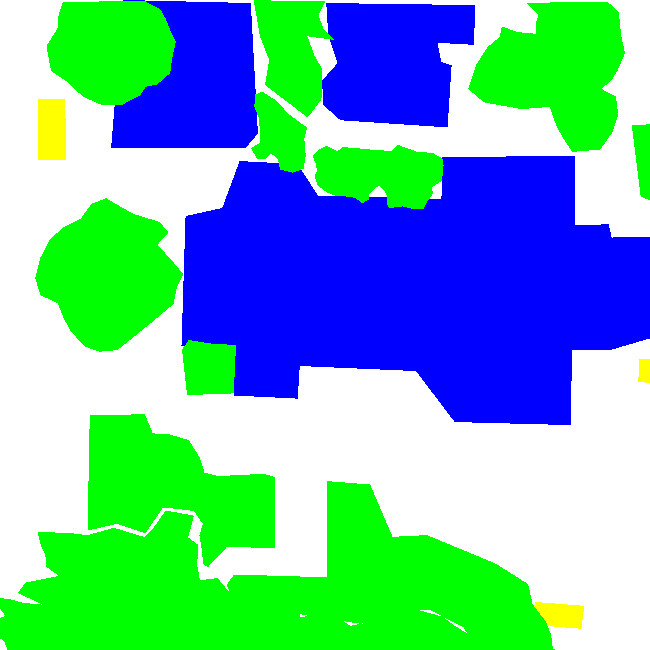} & \includegraphics[width=.14\linewidth]{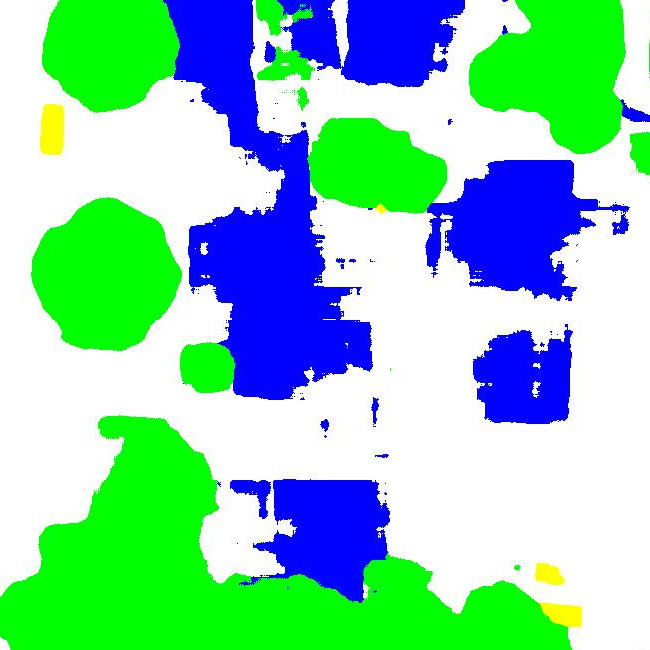}& \includegraphics[width=.14\linewidth]{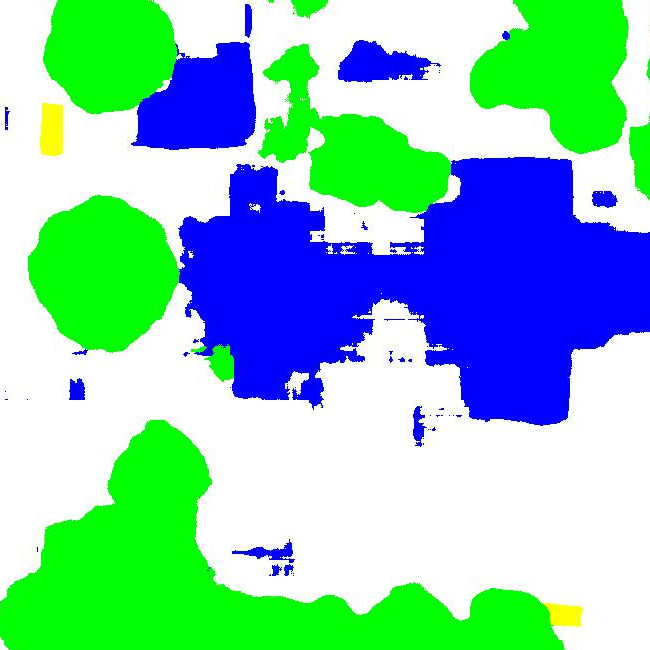} & \includegraphics[width=.14\linewidth]{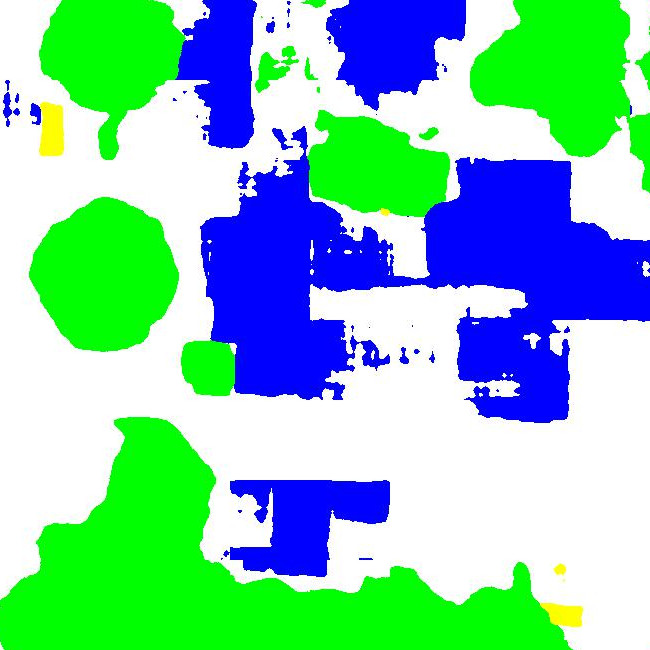}
       & \includegraphics[width=.14\linewidth]{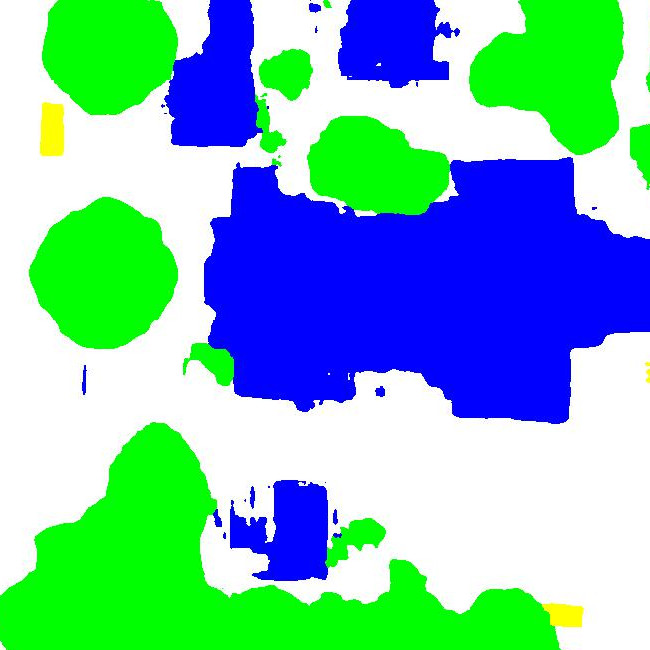} & \includegraphics[width=.14\linewidth]{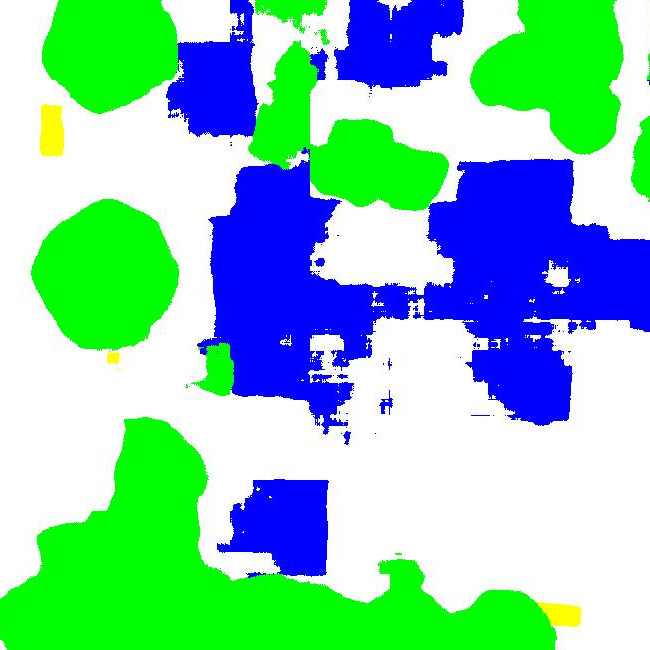} \\
       \emph{Image} & \emph{GT} & \emph{Supervised} & $\LL_1$ & $\LL_2$ & $\LL_{km}$ & $\LL_{MS}$\\

      \end{tabular}
      \caption{\added{Two examples of inference over the CASD dataset.} 
      \mysmallbox{blue}{blue} buildings, \mysmallbox{yellow}{yellow} cars, \mysmallbox{green}{green} vegetation and \mysmallbox{black}{white} background.} \label{fig: results-christchurch}
   \end{center}
\end{figure}

\added{Figure~\ref{fig: results-christchurch} shows two examples of segmentation maps obtained by the different methods. In the first row example, the supervised approach is the only one that mistakes the river as a building; the supplementary information provided by unlabeled images to the semi-supervised methods allows to prevent this error. In the second row, the $\LL_{km}$ loss is the only one that correctly segments the central building, likely due to its color clustering capacity.}

\added{In general, we observe from the experiments over CASD that including unlabeled data during training helps to improve the segmentation maps with respect to the case where we only use our limited labeled data.}

\subsection{Experiments on \MF}\label{sec: experiments-MF}

All the results and analysis exposed above were conducted using the \tmf\ dataset, due to computing capacity and processing time.
In this section we present the first semi-supervised results over the entire \MF\ dataset. 

\begin{figure}[!htbp]
   \begin{center}
      \setlength{\tabcolsep}{1.2pt}
      \begin{tabular}{cc@{\hspace{10pt}}cc}
         \includegraphics[width=.2\linewidth]{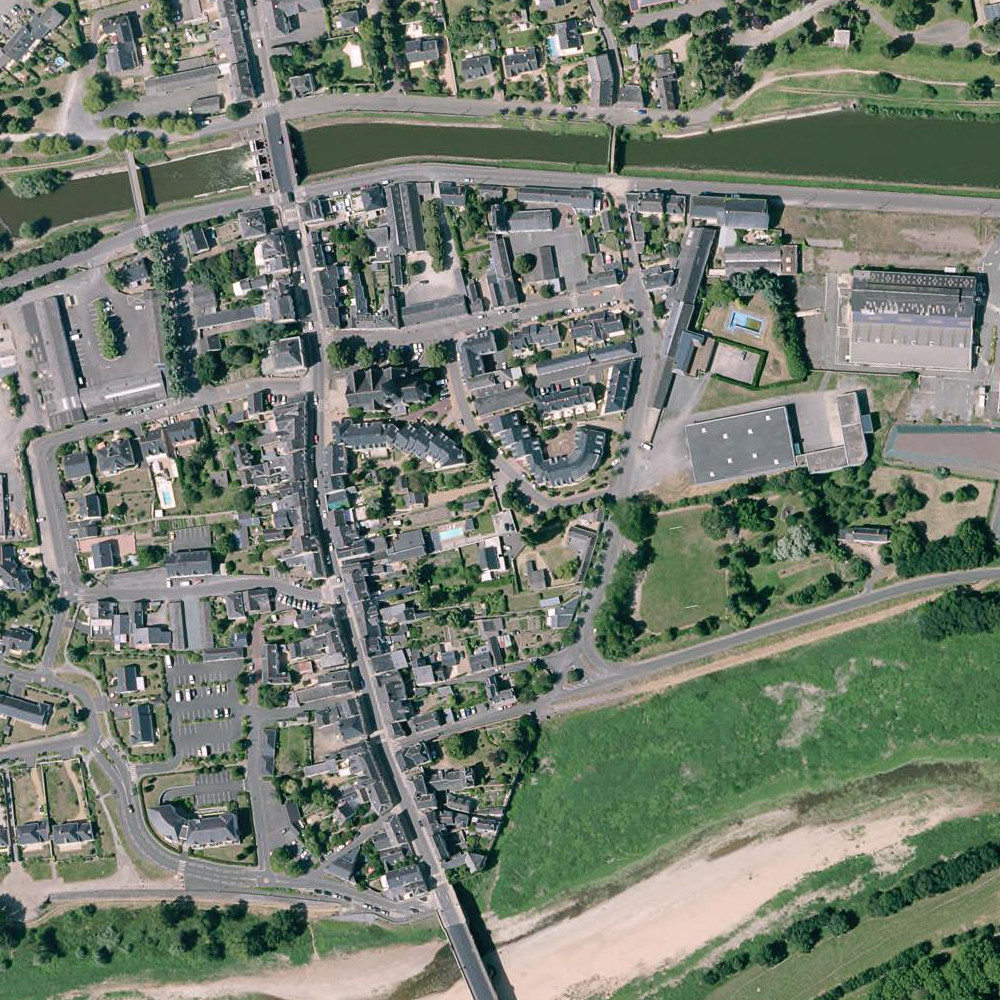} & \includegraphics[width=.2\linewidth]{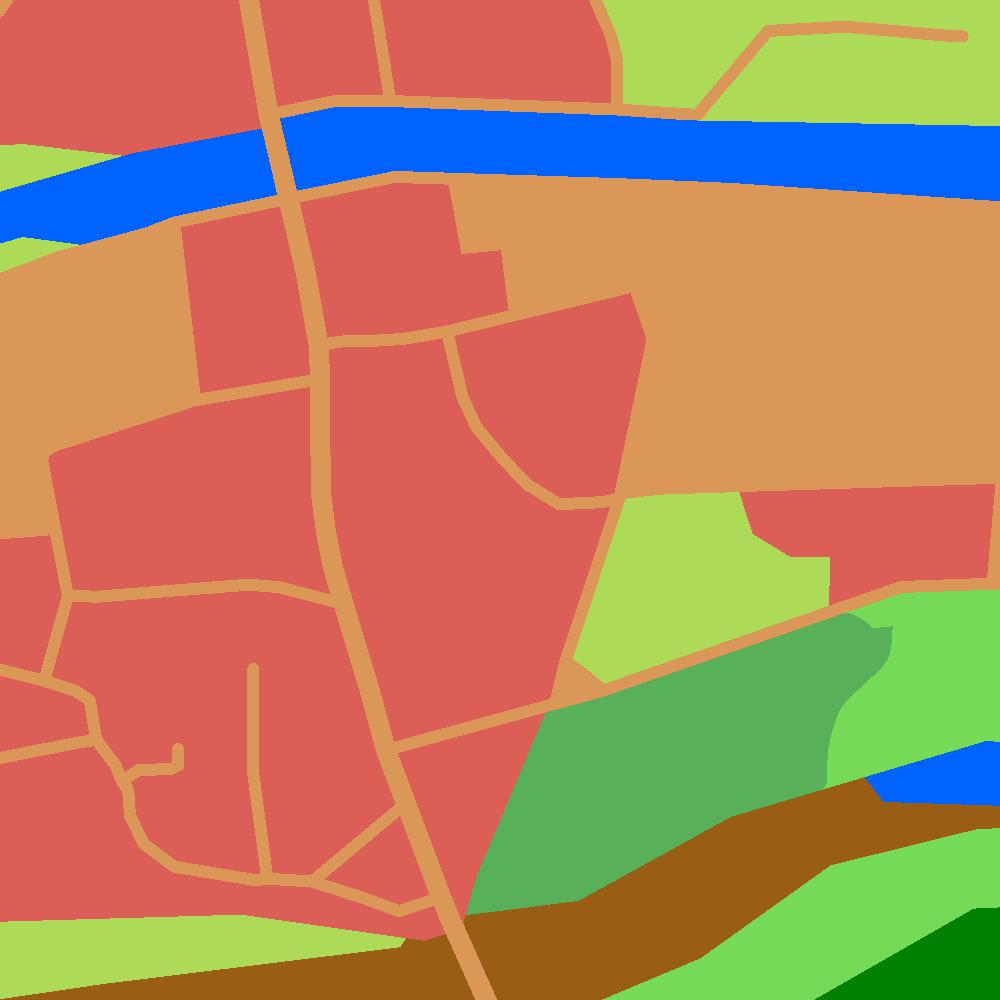} & \includegraphics[width=.2\linewidth]{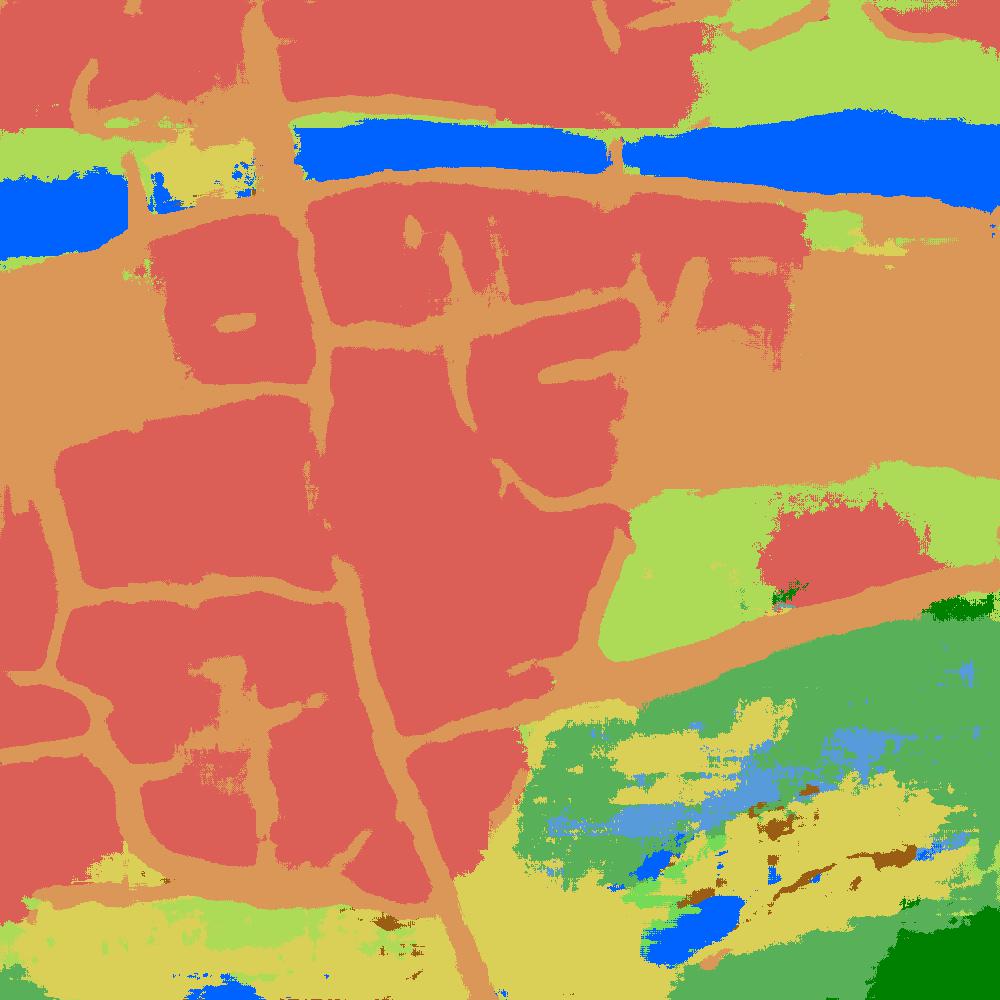} & \includegraphics[width=.2\linewidth]{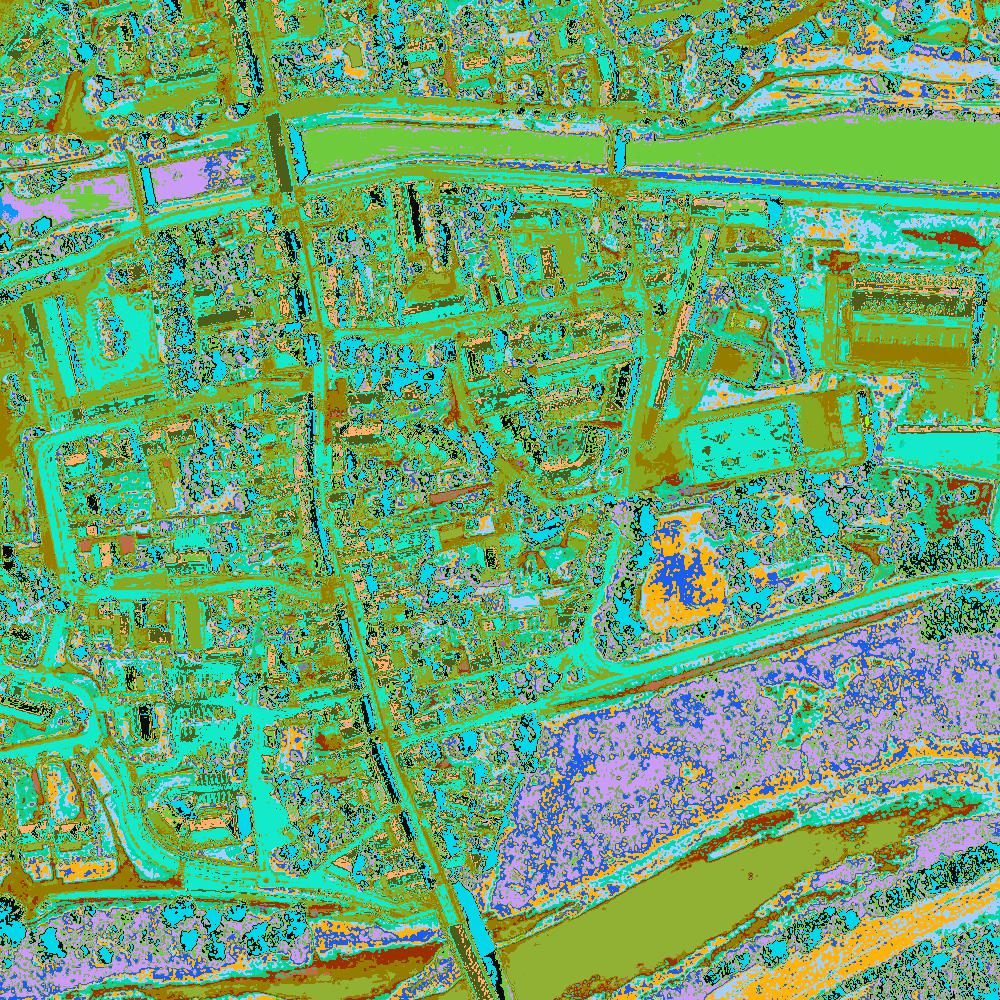} \\
         \includegraphics[width=.2\linewidth]{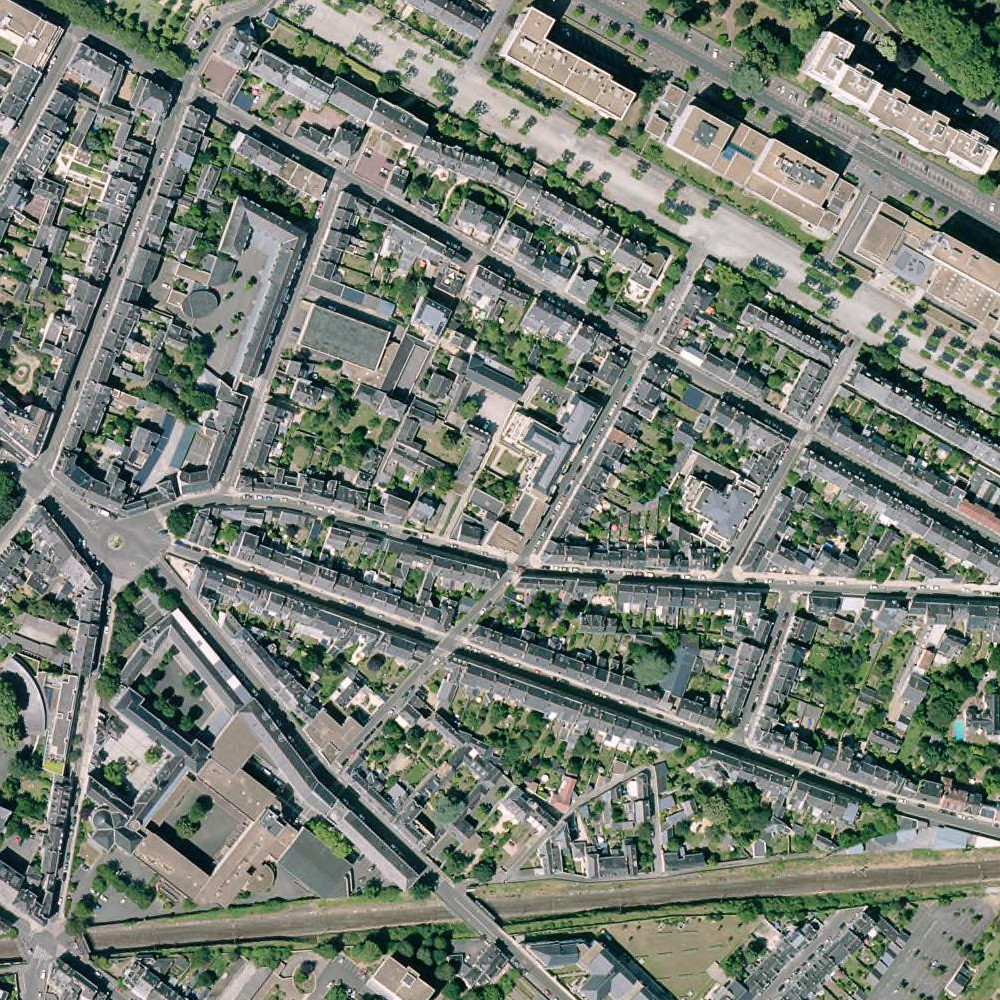} & \includegraphics[width=.2\linewidth]{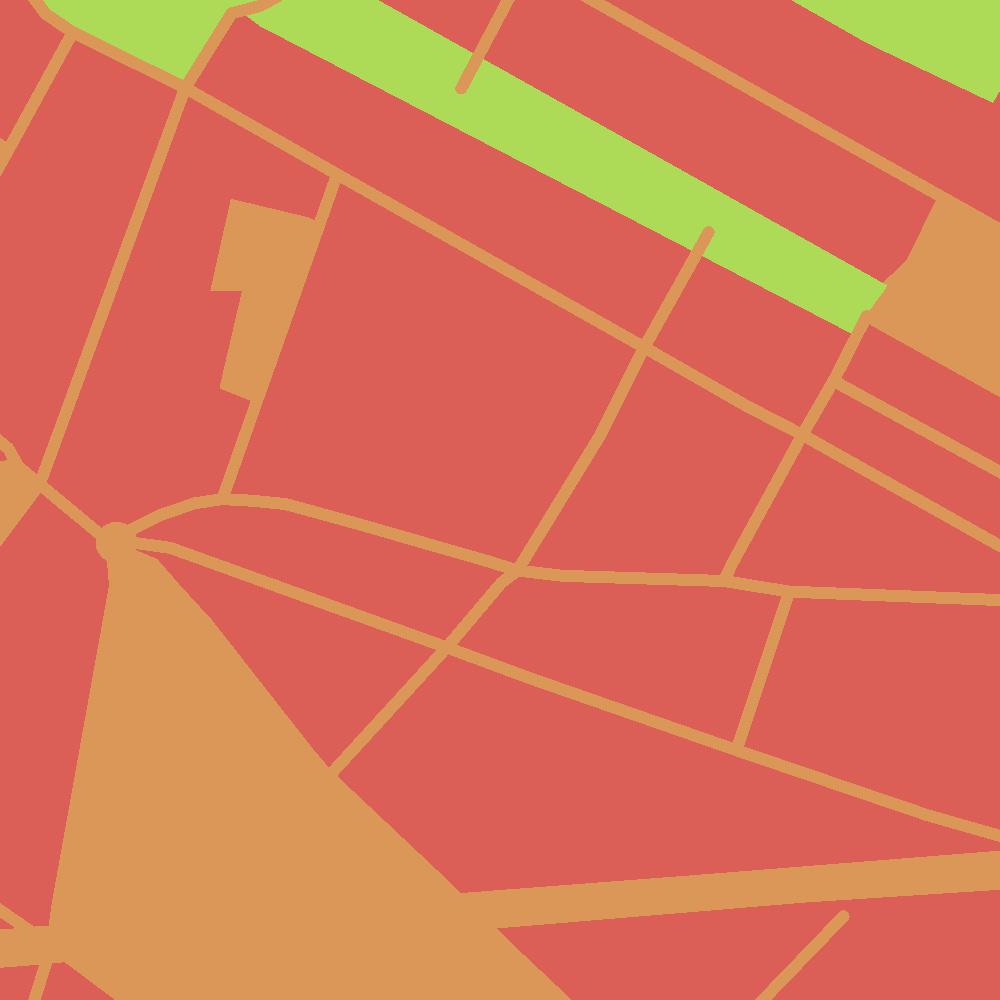} & \includegraphics[width=.2\linewidth]{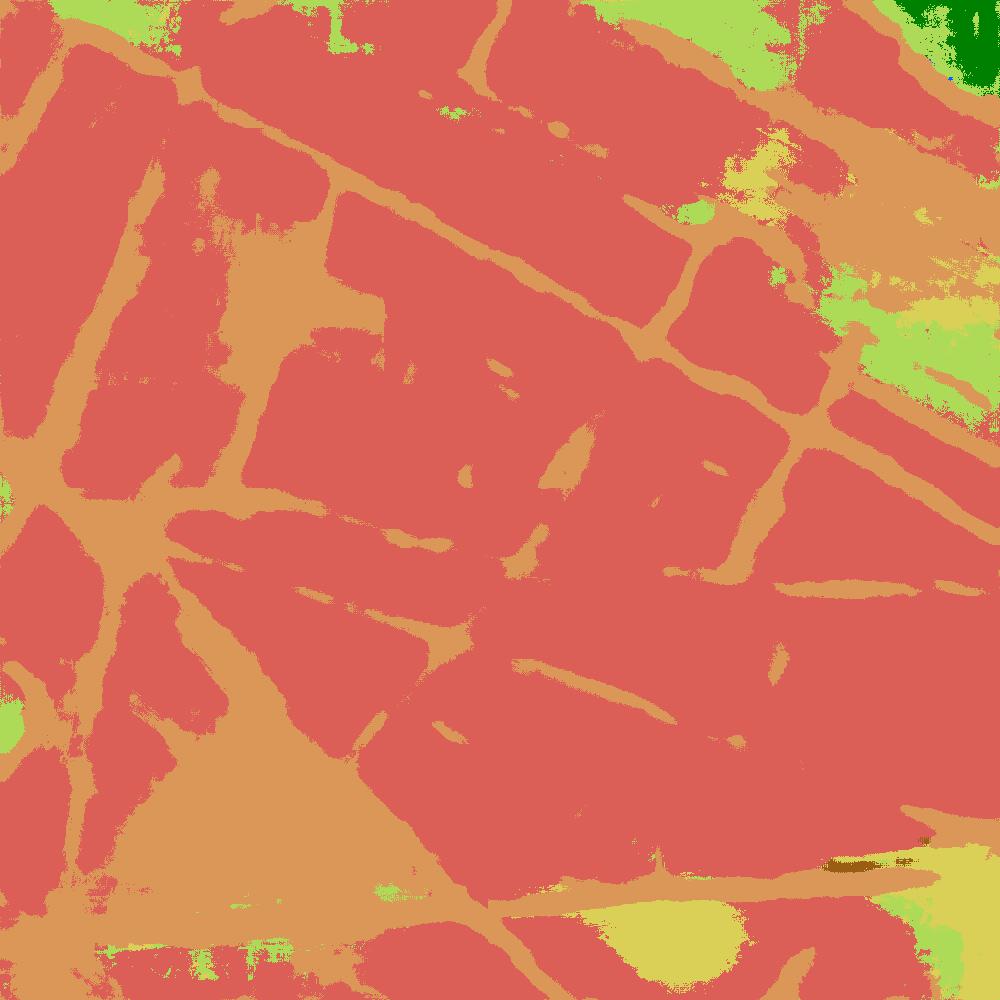} & \includegraphics[width=.2\linewidth]{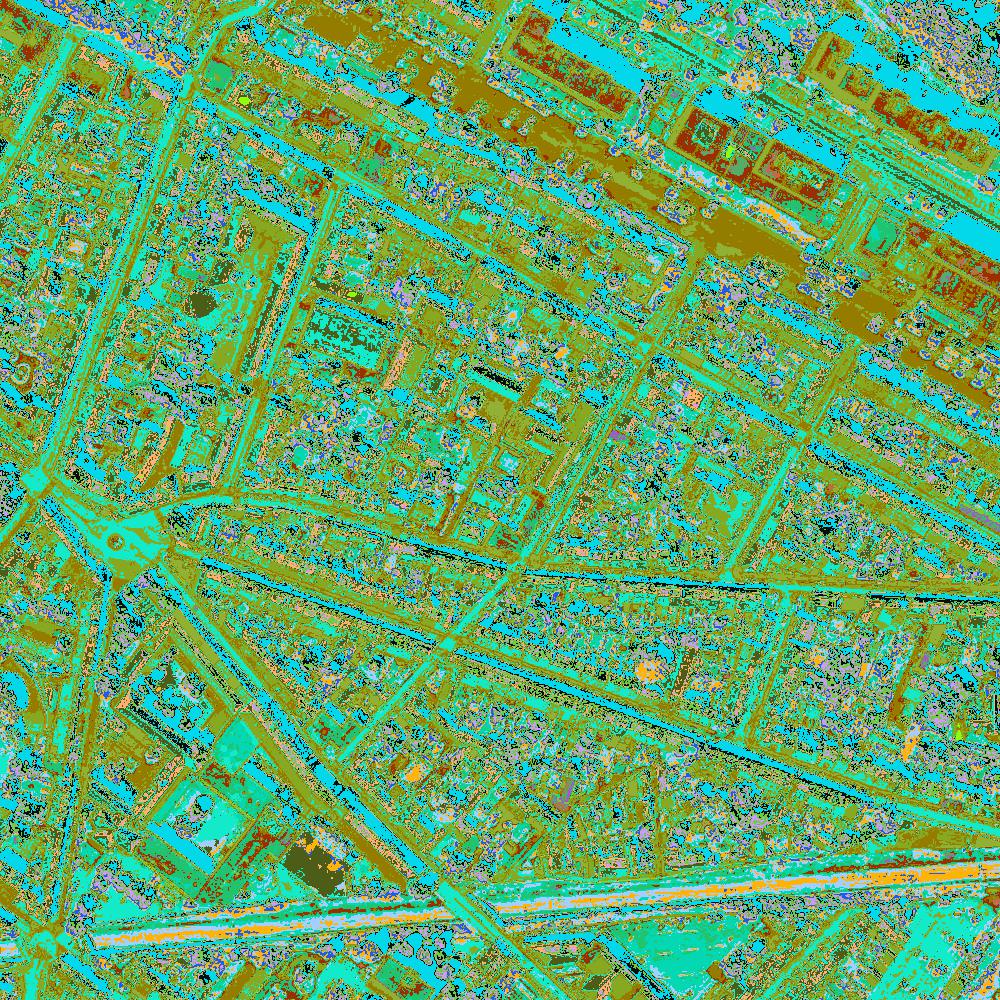} \\
         \includegraphics[width=.2\linewidth]{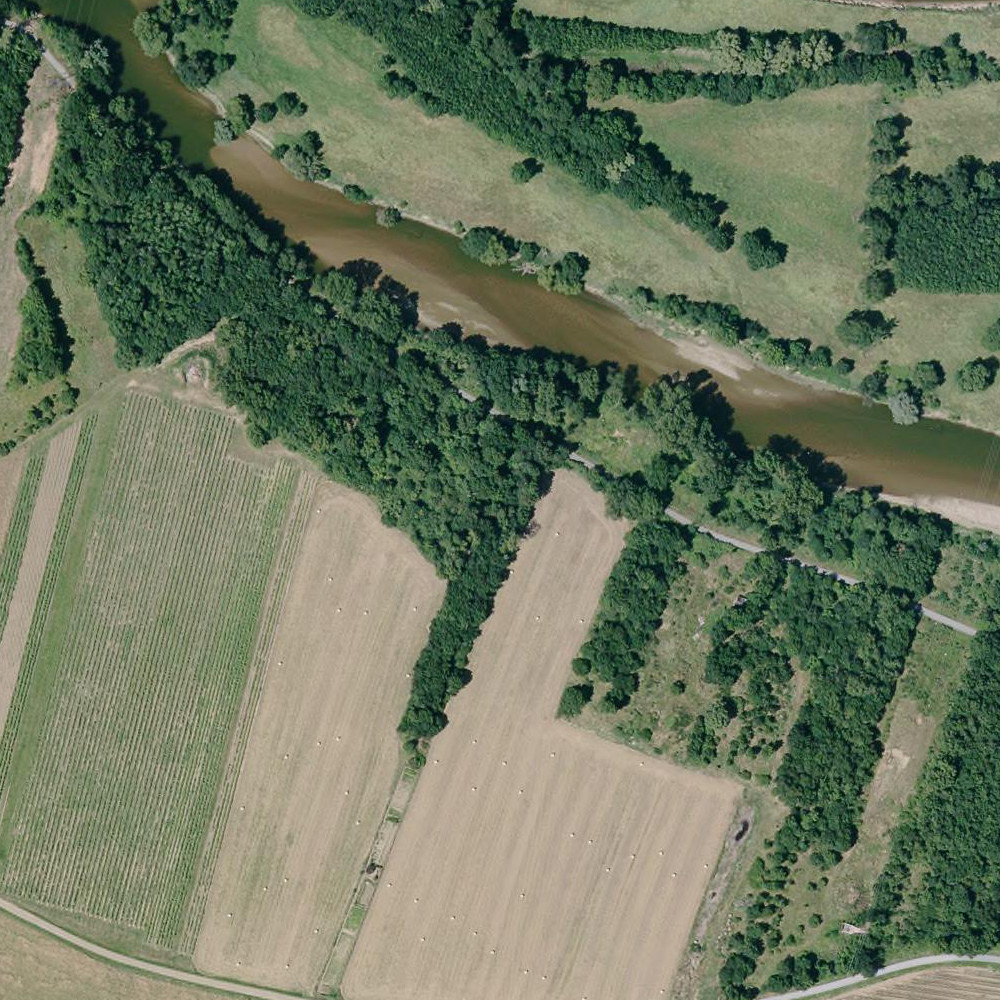} & \includegraphics[width=.2\linewidth]{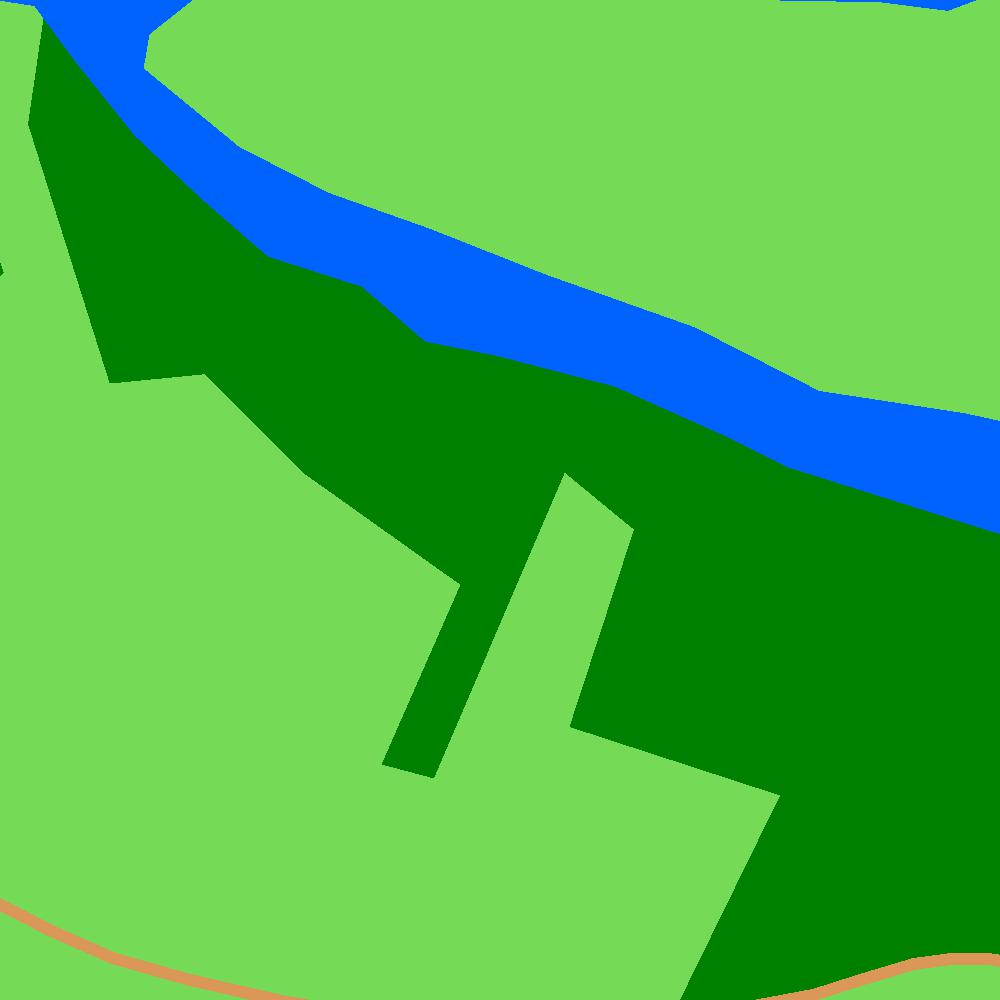} & \includegraphics[width=.2\linewidth]{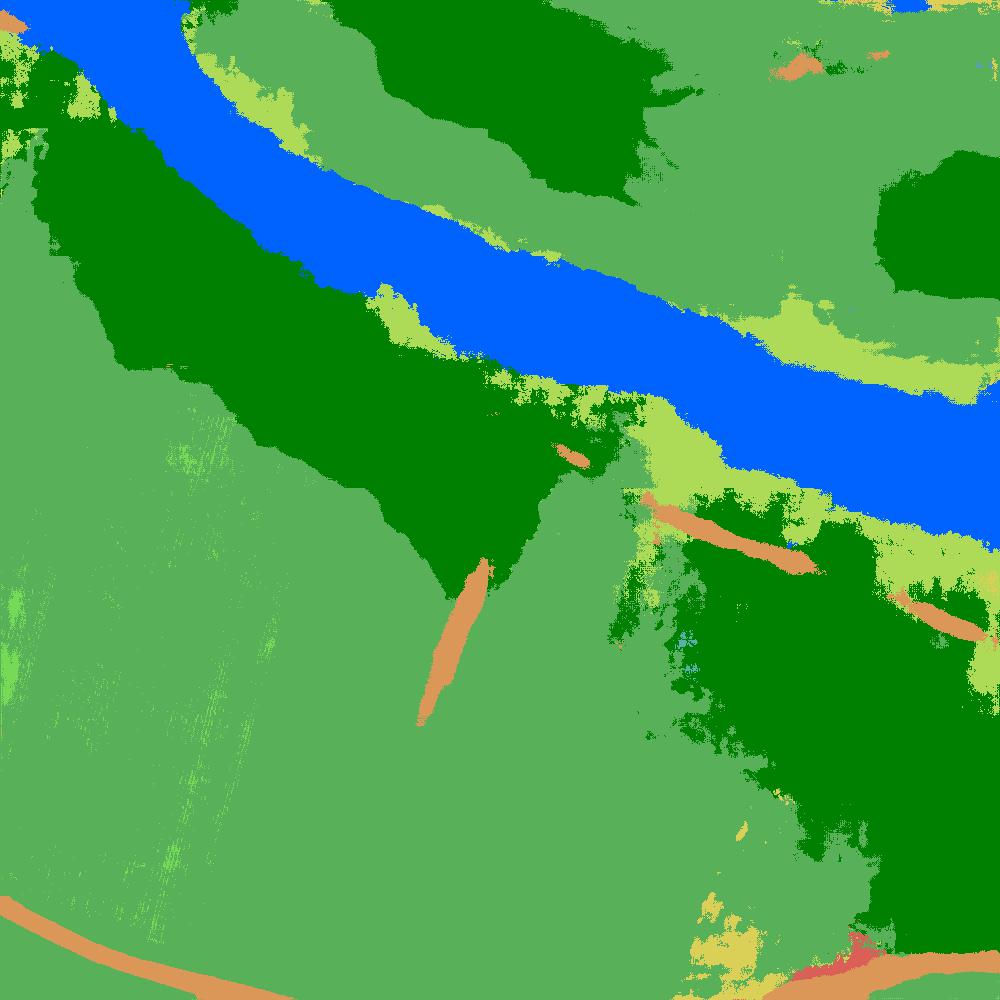} & \includegraphics[width=.2\linewidth]{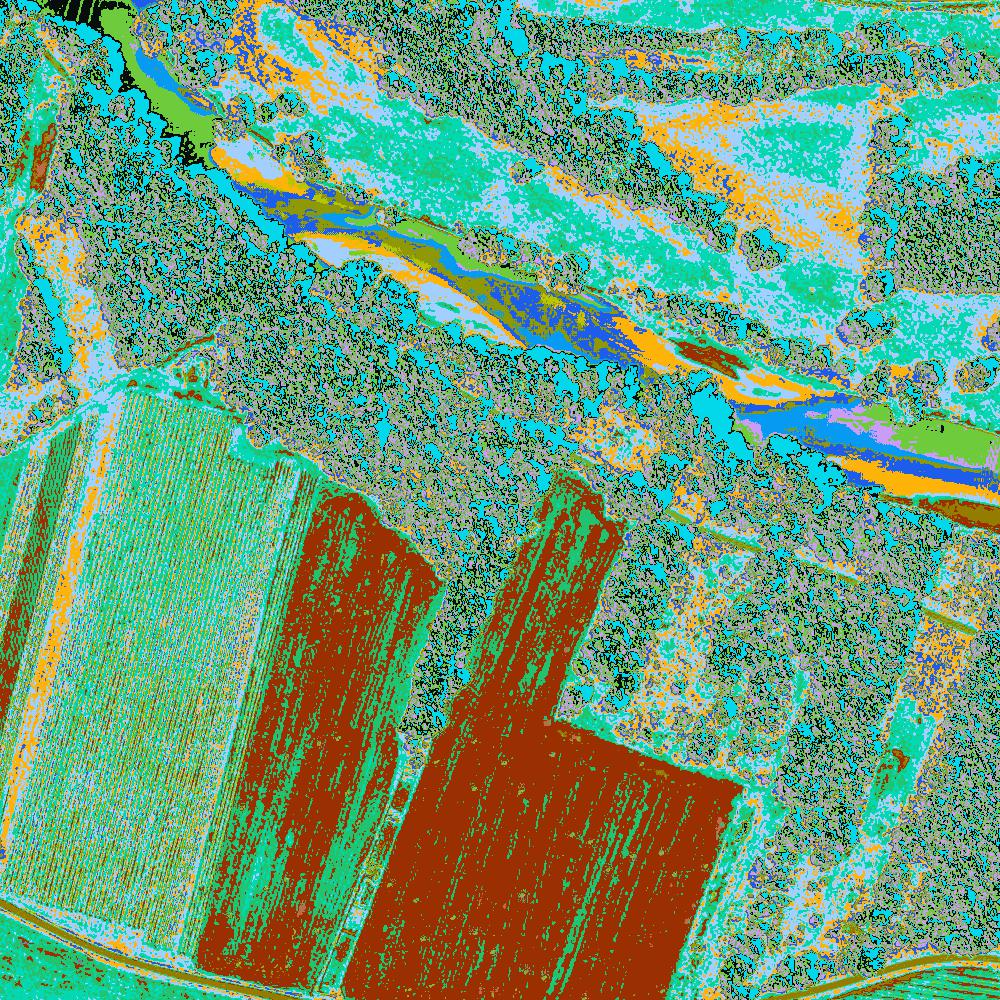}\\
         \includegraphics[width=.2\linewidth]{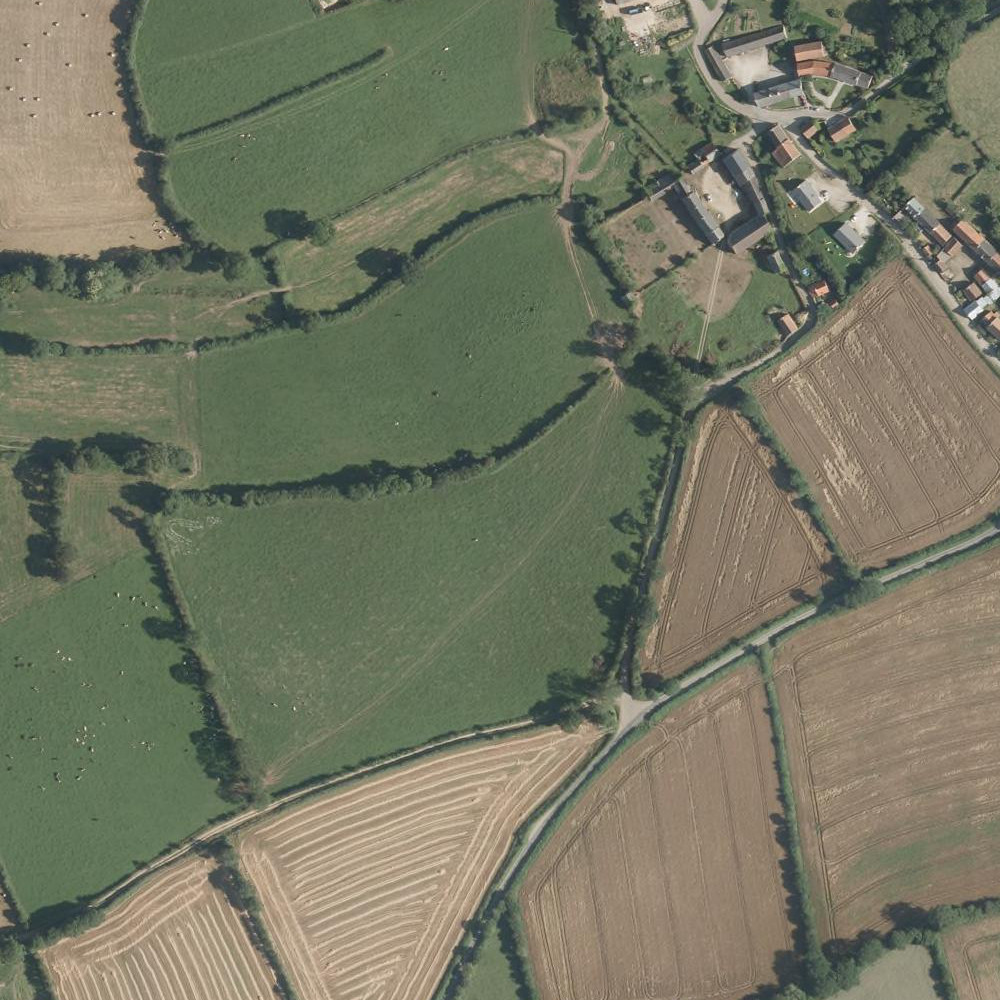} & \includegraphics[width=.2\linewidth]{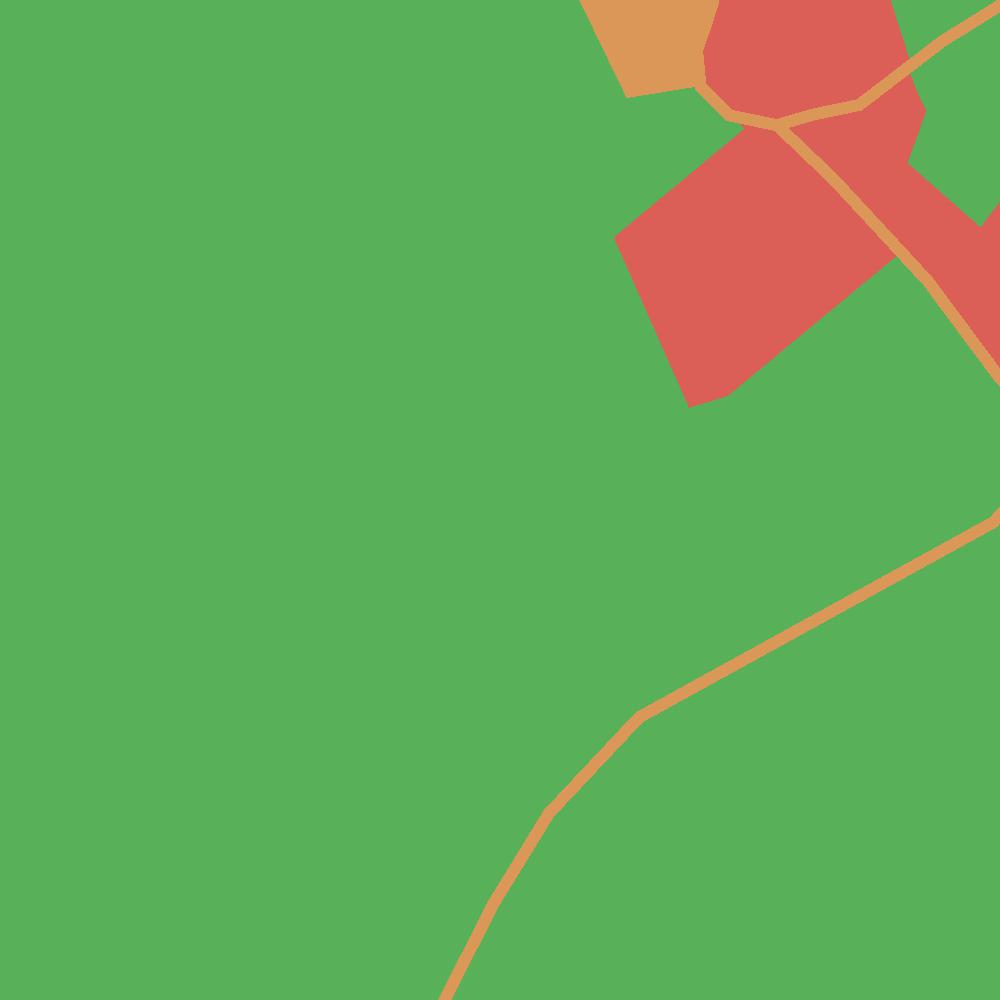} & \includegraphics[width=.2\linewidth]{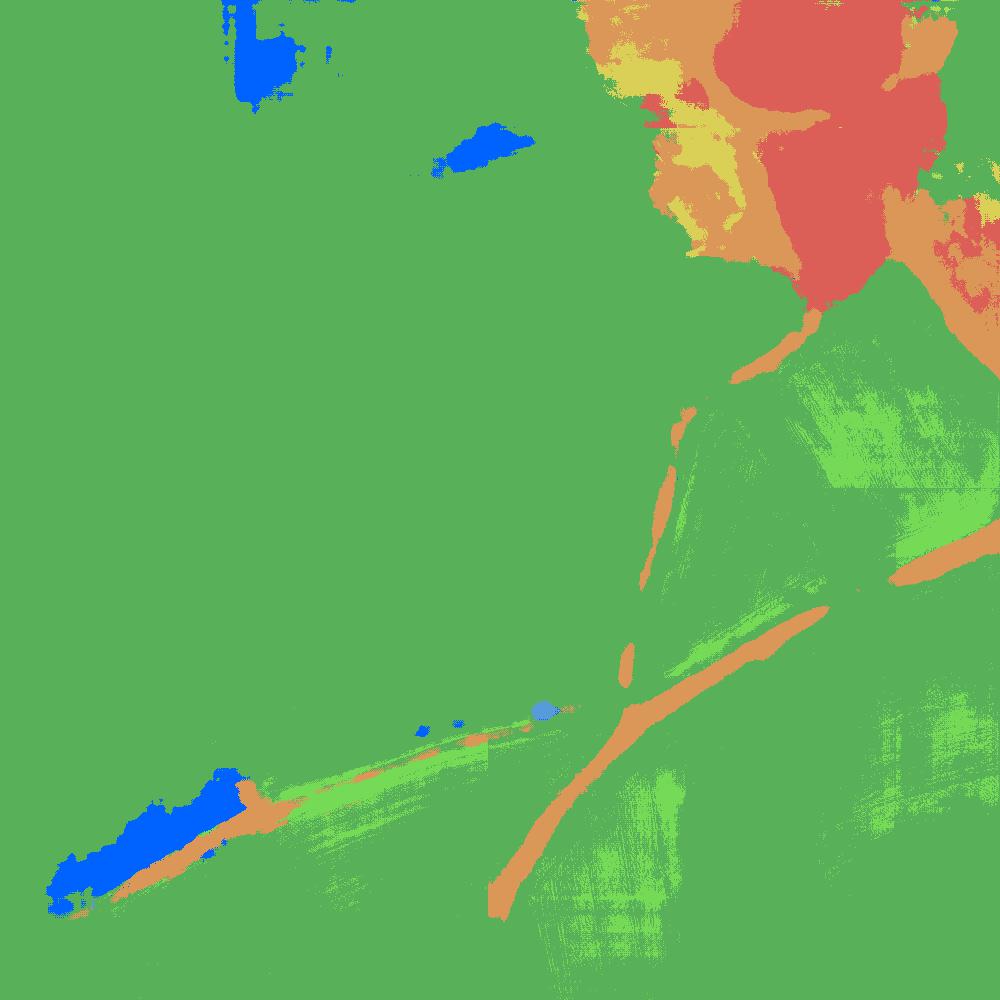} & \includegraphics[width=.2\linewidth]{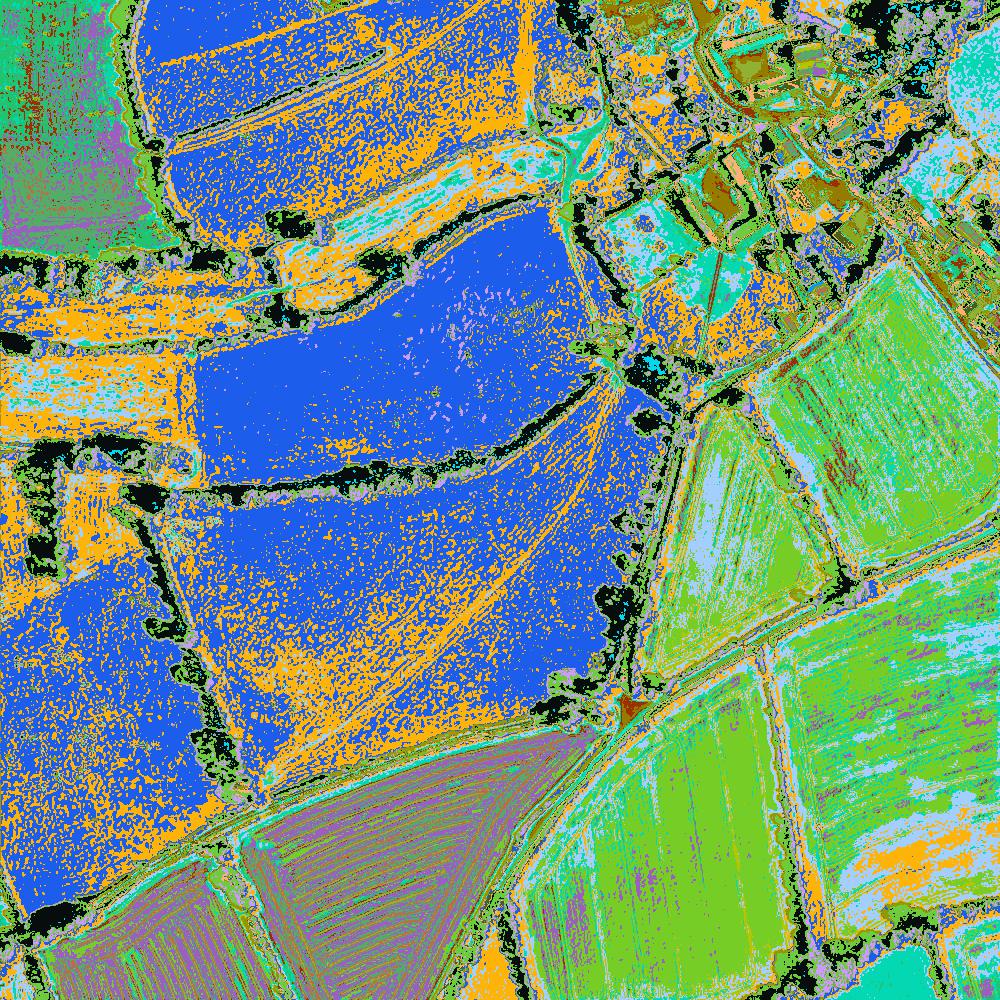} \\
         \includegraphics[width=.2\linewidth]{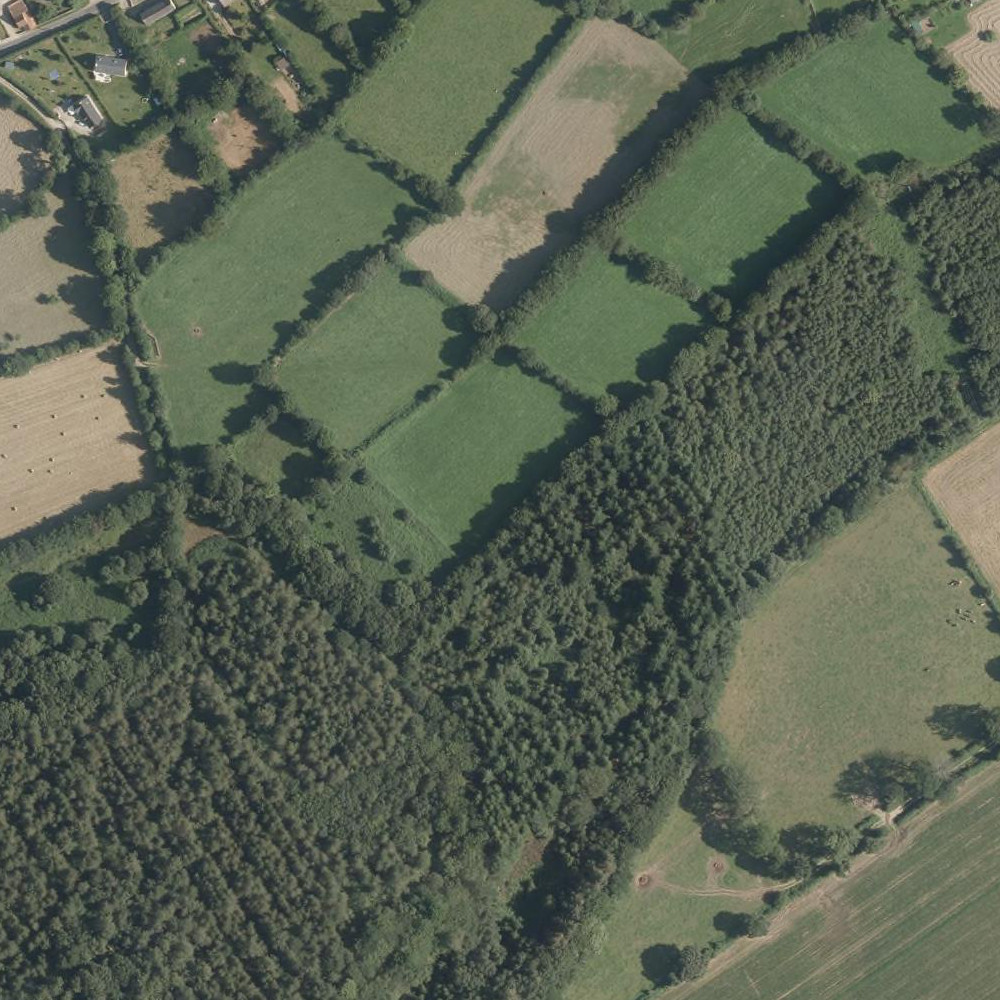} & \includegraphics[width=.2\linewidth]{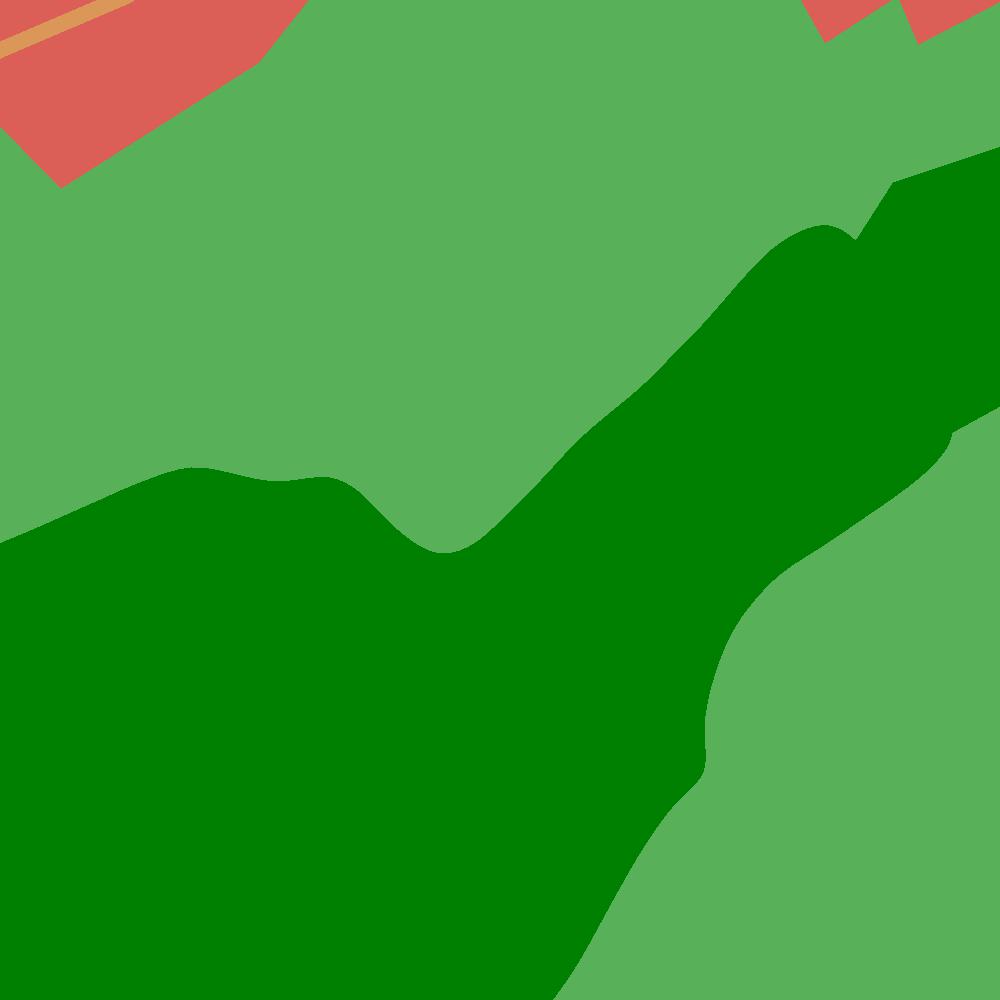} & \includegraphics[width=.2\linewidth]{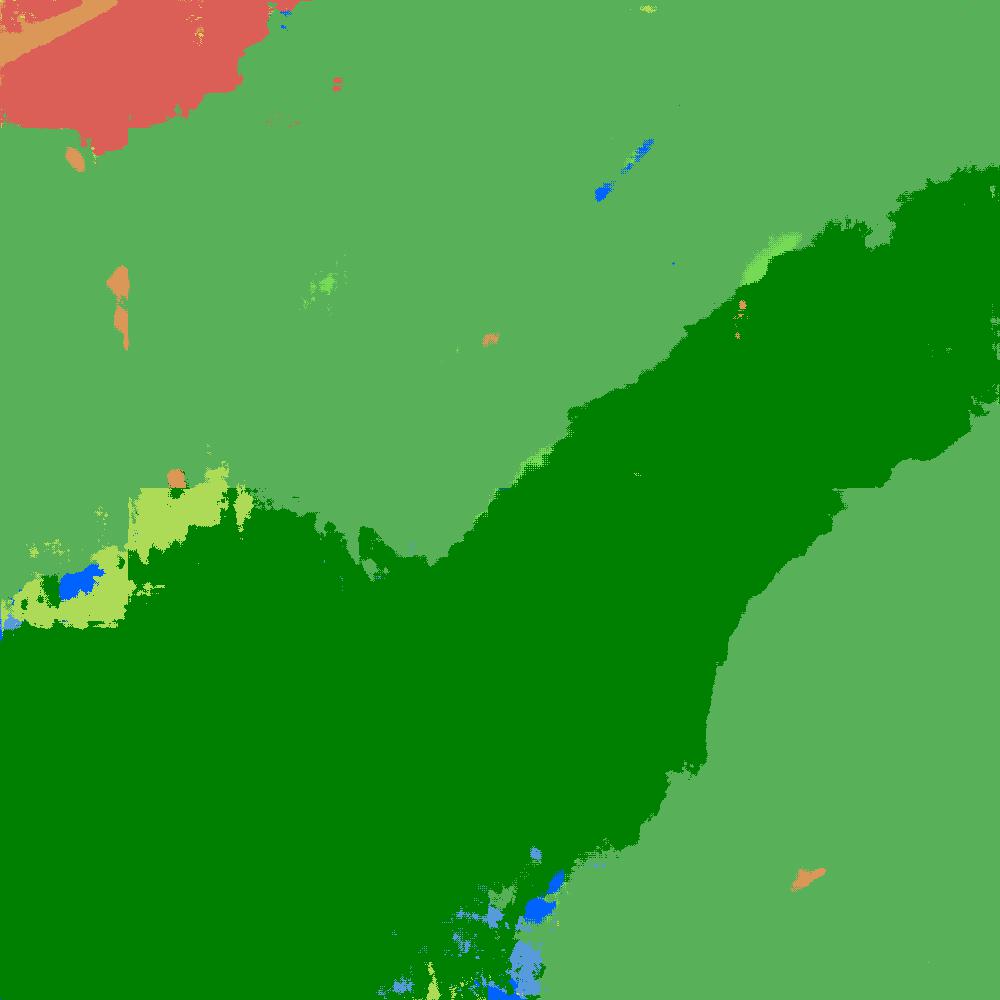} & \includegraphics[width=.2\linewidth]{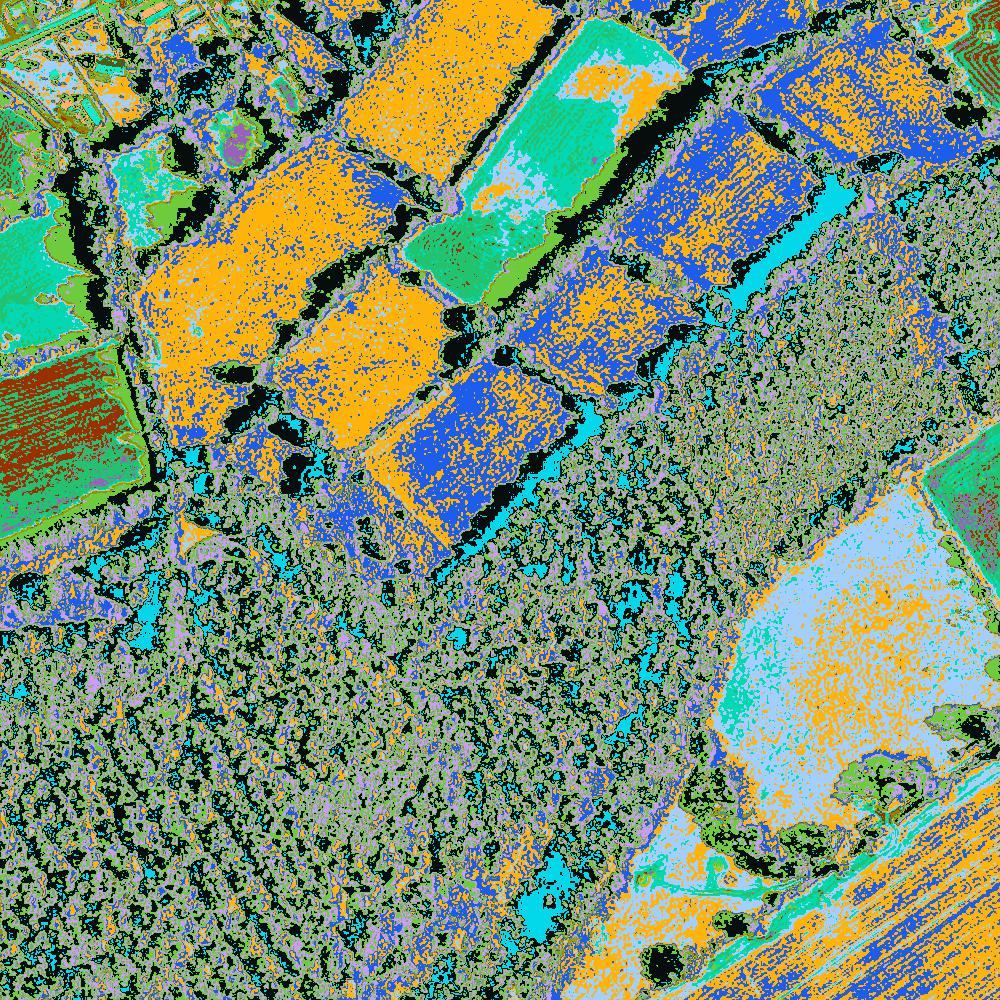} \\
         \includegraphics[width=.2\linewidth]{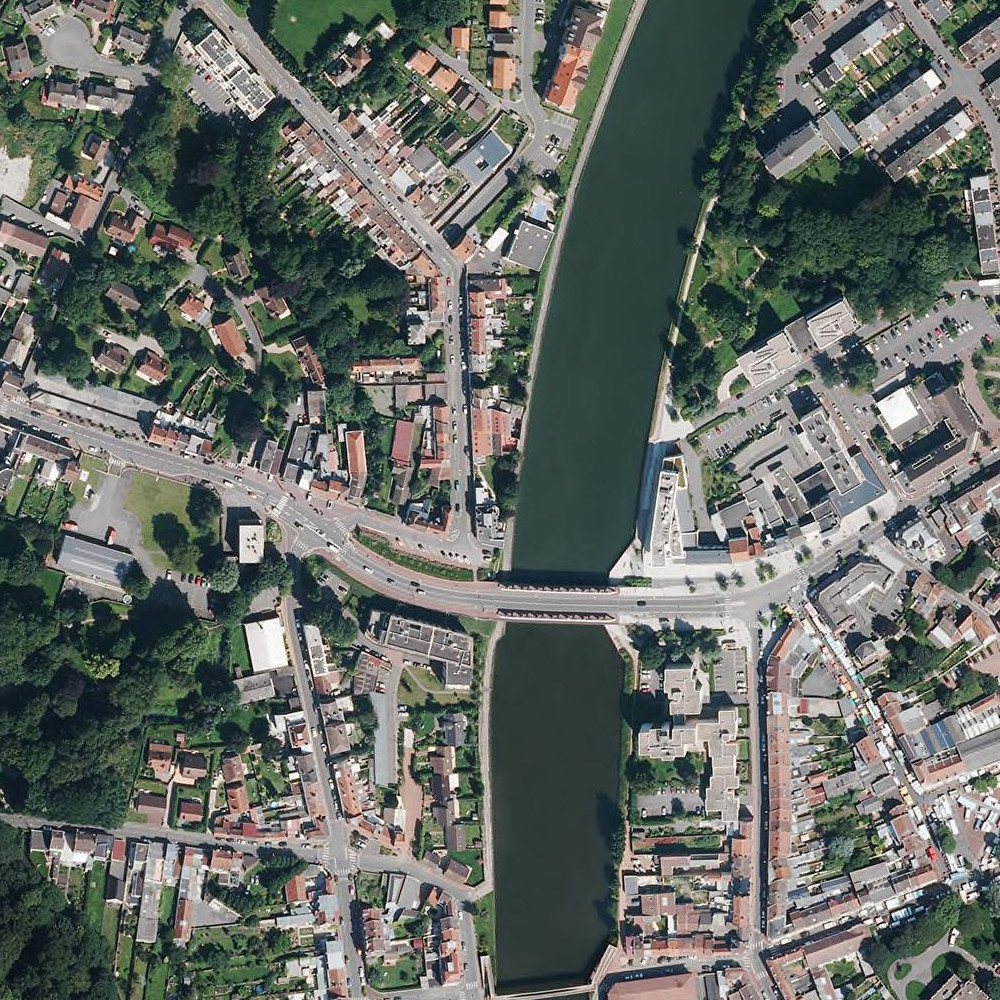} & \includegraphics[width=.2\linewidth]{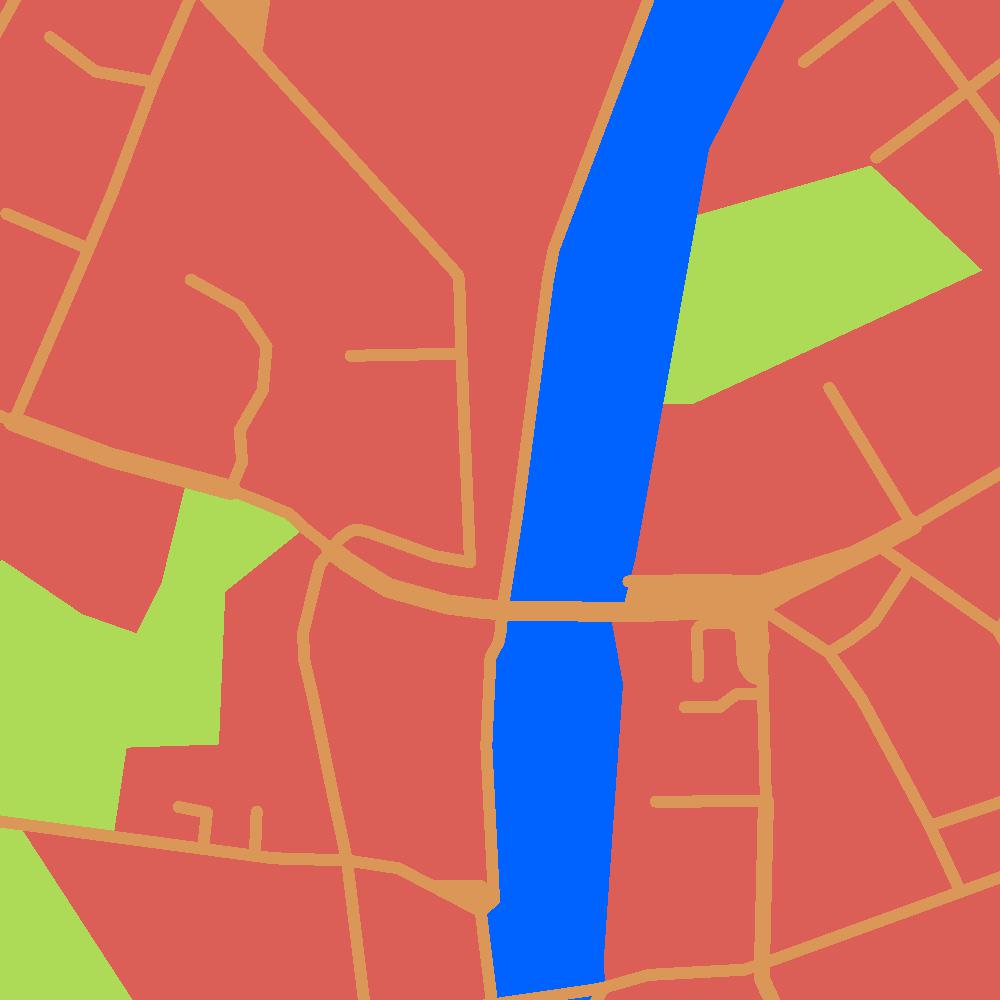} & \includegraphics[width=.2\linewidth]{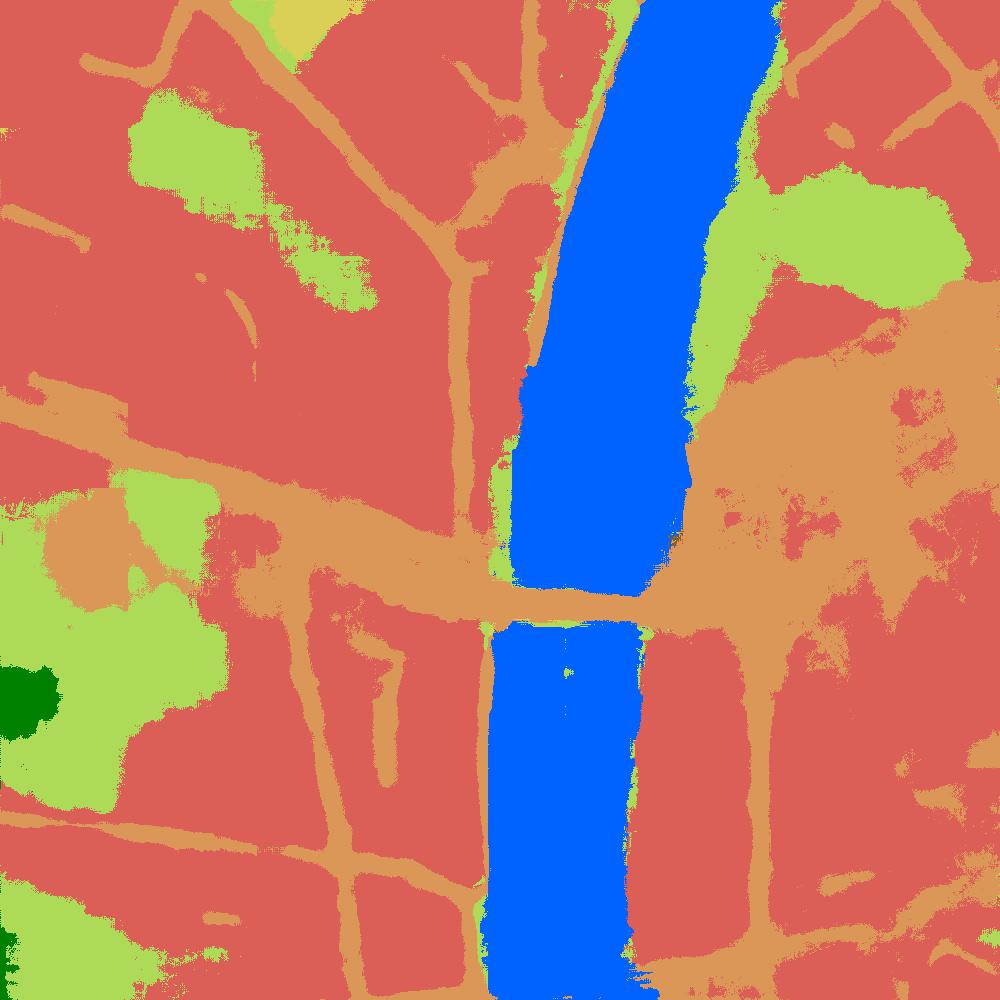} & \includegraphics[width=.2\linewidth]{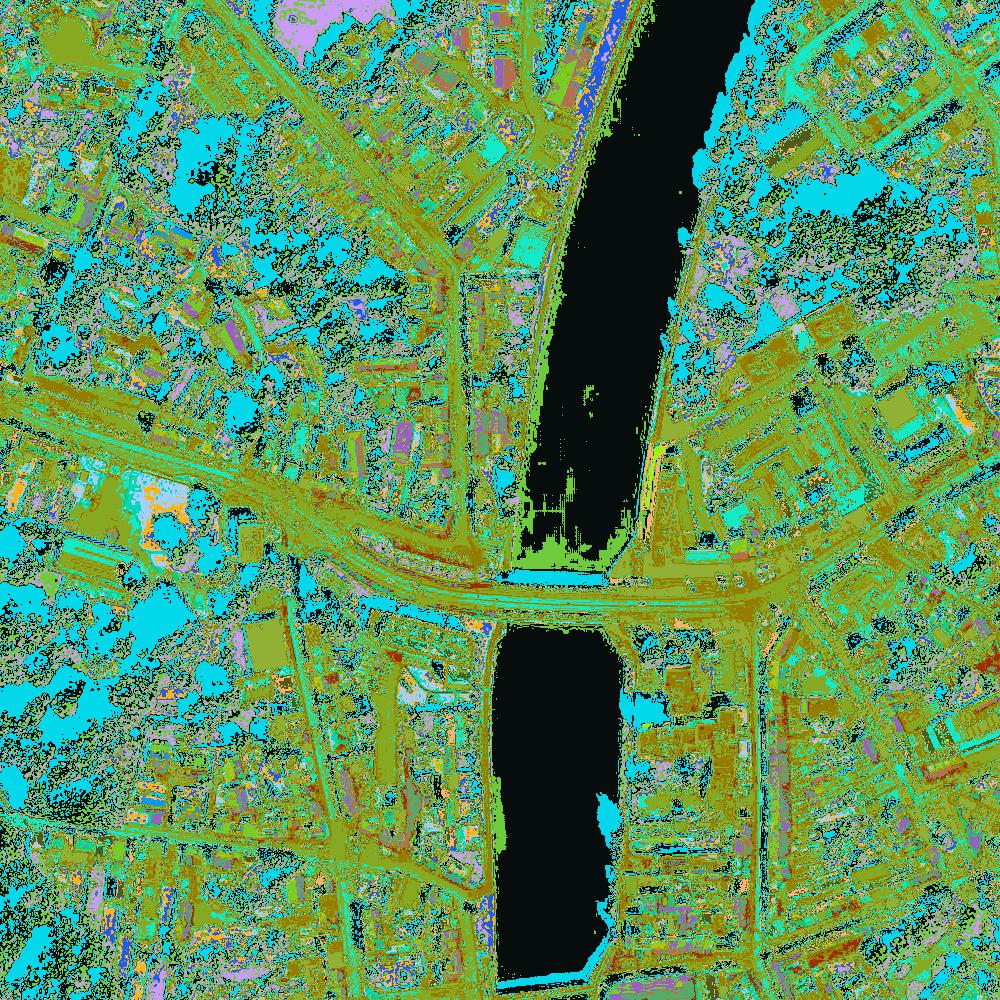} \\[3pt]

         Image & GT & \makecell{Semantic \\ segmentation} & \makecell{Unsupervised \\ segmentation}\\[4pt]

      \end{tabular}
      \caption{Semi-supervised results over MiniFrance. BerundaNet-late with U-Net backbone and $\LL_{km}$ as auxiliary loss.} \label{fig: results-minifrance}
   \end{center}
\end{figure}

To this end, we train a BerundaNet-late with U-Net backbone as it is the best result we got in a semi-supervised setting (see Table~\ref{tab: tmf-supervised-vs-semisupervised}). We use our regularized k-means loss ($\LL_{km}$) as auxiliary unsupervised loss.\added{
We also train a U-Net network on the labeled partition of \MF\ in a classic supervised way for comparison with the semi-supervised setting. Results are reported in Table~\ref{tab: mf-semisupervised} and some visual results of the semi-supervised experiment are shown in Figure~\ref{fig: results-minifrance}.}  

These results on \MF\ are coherent with previous ones reported with \tmf. They confirm our hypothesis that \tmf\ is a good representation of the entire \MF\ dataset. \added{Moreover, they confirm that including unlabeled data during the learning process helps to improve the results on semantic segmentation.}

It is worth to mention that training these models over the entire \MF\ dataset for 450 pseudo-epochs takes roughly 3 weeks. While inference time -- processing all the tiles on the testing partition -- takes about 6 days (with a single GPU).

\begin{table}[H]
   \begin{center}
      \caption{\added{First semi-supervised results over \MF.}}\label{tab: mf-semisupervised}
      \begin{tabular}{cccccc} \toprule
      \emph{Method} & \emph{Network} & \emph{Backbone} & \emph{Aux. Loss} & \emph{OA} & \emph{mIoU} \\ \midrule \midrule
      
      Supervised & U-Net & U-Net & - & 44.28 & 20.77 \\[2pt]
      Semi-Supervised & BerundaNet-late & U-Net & $\LL_{km}$ & \bf{45.16} & \bf{21.20} \\ 
      \bottomrule
      \end{tabular}
   \end{center}   
\end{table}

\section{Conclusions} 

We have introduced the \MF\ suite, a new large-scale dataset designed for semi-supervised semantic segmentation in Earth Observation. \MF\ has unprecedented properties, the diversity of landscapes and scenes reflects the complexity of reality. Above all, it was thoroughly designed for semi-supervised learning, including labeled and unlabeled data in its training partition and recreating a life-like application setting, which makes \MF\ unique. In addition to the dataset, we presented a comprehensive analysis of the data in terms of appearance similarity and representativeness, showing that \MF\ is well-suited to address the semi-supervised problem.

We also introduced deep neural networks, based on multi-task learning, to perform semi-supervised semantic segmentation. In particular, we presented BerundaNet -- a simple extension of classic encoder-decoder architectures -- which proves to be very effective in the semi-supervised task. Together with these architectures, we explored unsupervised auxiliary losses to use alongside with semantic segmentation. Especially, we introduced the relaxed k-means loss to perform unsupervised image  segmentation.

Our experiments have shown that we can benefit from unlabeled data during the learning process to improve semantic segmentation maps. Indeed, semi-supervised approaches allow to generate finer and more homogeneous predictions. We also observed that a simple architecture like BerundaNet-late with a suitable backbone such as U-Net is enough to enhance the segmentation performances. These results are very encouraging and will serve as baselines for future works on semi-supervised semantic segmentation over the \MF\ dataset.

Nevertheless, the problem of semi-supervised learning is not solved. We have seen that our approaches can improve semantic segmentation results, but it is not always the case. In a multi-task approach as ours, we must be careful on the choice of architecture and the auxiliary task to perform along. Furthermore, there exist other possible ways to solve the semi-supervised problem. For instance, one could develop generative models to learn the intrinsic distribution of data from labeled and unlabeled examples and use this information together with labels to improve the segmentation. Another possibility is the use of pseudo-label methods that propagate labels from annotated examples through non-annotated ones, based on a confidence criterion, to enlarge available training data. These methods were not explored in this work, but they should be considered in future research.

\begin{acknowledgements}
Javiera Castillo-Navarro's work is partially funded by a grant from CNES (Centre National d'Études Spatiales). The authors acknowledge the IGN for providing the BD ORTHO database under Open Licence v1.0 (\url{https://www.etalab.gouv.fr/licence-ouverte-open-licence}) and the European Copernicus Program for providing the Urban Atlas data (\url{https://land.copernicus.eu/local/urban-atlas}). This work contains modified Copernicus Urban Atlas data.
\end{acknowledgements}

\clearpage

    %
    %

    \bibliographystyle{spmpsci}      
    \bibliography{biblio}   
    
    
    \end{document}